\documentclass[11pt]{article}

\usepackage{amsmath}
\usepackage{amssymb}
\usepackage{amsthm}
\usepackage{algorithm}
\usepackage{algorithmic}
\usepackage{hyperref}
\usepackage{multirow}
\usepackage{microtype}
\usepackage{natbib}
\usepackage{nameref}
\usepackage[margin=1.25in]{geometry}

\usepackage{subfig}
\usepackage{tikz}
\usetikzlibrary{backgrounds,fit}

\newcommand{\condcommand}[3]{
  \ifdefined #2
    \PackageWarning{condcommand}{#2 is being redefined}
    \renewcommand{#2}[#1]{#3}
  \else
    \newcommand{#2}[#1]{#3}
  \fi
}

\condcommand{1}{\todo}{\textcolor{red}{TODO: #1}}
\mathchardef\mhyphen="2D

\condcommand{0}{\E}{\mathop{\mathbb{E}}}
\condcommand{0}{\I}{\mathbb{I}}
\condcommand{0}{\ind}{\boldsymbol{1}}
\condcommand{0}{\reals}{\mathbb{R}}
\condcommand{0}{\ints}{\mathbb{Z}}

\condcommand{0}{\C}{\mathcal{C}}
\condcommand{0}{\K}{\mathcal{K}}
\condcommand{0}{\L}{\mathcal{L}}
\condcommand{0}{\N}{\mathcal{N}}
\condcommand{0}{\P}{\mathcal{P}}
\condcommand{0}{\T}{\mathcal{T}}
\condcommand{0}{\W}{\mathcal{W}}
\condcommand{0}{\X}{\mathcal{X}}
\condcommand{0}{\Y}{\mathcal{Y}}

\condcommand{0}{\bpsi}{\boldsymbol{\psi}}
\condcommand{0}{\bzero}{\boldsymbol{0}}
\condcommand{0}{\bone}{\boldsymbol{1}}
\condcommand{0}{\bepsilon}{\boldsymbol{\epsilon}}
\condcommand{0}{\bmu}{\boldsymbol{\mu}}
\condcommand{0}{\bpi}{\boldsymbol{\pi}}
\condcommand{0}{\btau}{\boldsymbol{\tau}}
\condcommand{0}{\btheta}{\boldsymbol{\theta}}
\condcommand{0}{\b}{\boldsymbol{b}}
\condcommand{0}{\e}{\boldsymbol{e}}
\condcommand{0}{\f}{\boldsymbol{f}}
\condcommand{0}{\g}{\boldsymbol{g}}
\condcommand{0}{\r}{\boldsymbol{r}}
\condcommand{0}{\v}{\boldsymbol{v}}
\condcommand{0}{\w}{\boldsymbol{w}}
\condcommand{0}{\p}{\boldsymbol{p}}
\condcommand{0}{\x}{\boldsymbol{x}}
\condcommand{0}{\y}{\boldsymbol{y}}
\condcommand{0}{\m}{\boldsymbol{m}}
\condcommand{0}{\z}{\boldsymbol{z}}
\condcommand{0}{\bY}{\boldsymbol{Y}}
\condcommand{0}{\bA}{\boldsymbol{A}}
\condcommand{0}{\bJ}{\boldsymbol{J}}
\condcommand{0}{\bD}{\boldsymbol{D}}

\condcommand{0}{\cv}{\boldsymbol{\hat v}}
\condcommand{0}{\cV}{\hat V}
\condcommand{0}{\M}{M}
\condcommand{0}{\word}{w}
\condcommand{0}{\allwords}{W}

\condcommand{0}{\argmax}{\mathop{\arg\max}}
\condcommand{0}{\argmin}{\mathop{\arg\min}}
\condcommand{0}{\edge}{\mathrm{edge}}
\condcommand{0}{\local}{\mathrm{LOCAL}}
\condcommand{0}{\map}{\mathrm{MAP}}
\condcommand{0}{\mbr}{\mathrm{MBR}}
\condcommand{0}{\marg}{\mathrm{MARG}}
\condcommand{0}{\node}{\mathrm{node}}
\condcommand{0}{\proj}{\mathrm{Proj}}
\condcommand{0}{\st}{\mathrm{s.t.}}
\condcommand{0}{\trans}{\top}
\condcommand{0}{\dist}{\mathrm{dist}}
\condcommand{0}{\rank}{\mathrm{rank}}
\condcommand{0}{\diag}{\mathrm{diag}}
\condcommand{0}{\tr}{\mathrm{tr}}
\condcommand{0}{\length}{\mathrm{length}}
\condcommand{0}{\factor}{\alpha}
\condcommand{0}{\ftr}{\phi}
\condcommand{0}{\var}{\mathrm{Var}}
\condcommand{0}{\vol}{\mathrm{Vol}}
\condcommand{0}{\inc}{\mathrm{in}}
\condcommand{0}{\exc}{\mathrm{out}}
\condcommand{0}{\distinct}{\mathrm{distinct}}
\condcommand{0}{\mrf}{\mathrm{MRF}}
\condcommand{0}{\dpp}{\mathrm{DPP}}
\condcommand{0}{\mmr}{\mathrm{MMR}}
\condcommand{0}{\words}{\mathrm{words}}
\condcommand{0}{\tf}{\mathrm{tf}}
\condcommand{0}{\idf}{\mathrm{idf}}
\condcommand{0}{\tfidf}{\mathrm{tfidf}}
\condcommand{0}{\dsim}{\overset{\scriptstyle D}{\sim}}
\condcommand{0}{\comp}{\mathrm{comp}}
\condcommand{0}{\size}{\mathrm{size}}
\condcommand{0}{\cossim}{\mathrm{cos\mhyphen sim}}
\condcommand{0}{\sgn}{\mathrm{sgn}}
\condcommand{0}{\disc}{\mathrm{disc}}
\condcommand{0}{\rouge}{\mbox{\scriptsize{ROUGE-1F}}}

\newtheorem{theorem}{Theorem}[section]
\newtheorem{lemma}[theorem]{Lemma}
\newtheorem{corollary}[theorem]{Corollary}
\newtheorem{proposition}[theorem]{Proposition}
\newtheorem{definition}[theorem]{Definition}
\newtheorem{conjecture}[theorem]{Conjecture}

\condcommand{1}{\chaplabel}{\label{chap:#1}}
\condcommand{1}{\chapref}{Chapter~\ref{chap:#1}}
\condcommand{2}{\chapsref}{Chapters~\ref{chap:#1} and~\ref{chap:#2}}

\condcommand{1}{\seclabel}{\label{sec:#1}}
\condcommand{1}{\secref}{Section~\ref{sec:#1}}
\condcommand{2}{\secsref}{Sections~\ref{sec:#1} and~\ref{sec:#2}}

\condcommand{1}{\figlabel}{\label{fig:#1}}
\condcommand{1}{\figref}{Figure~\ref{fig:#1}}
\condcommand{2}{\figsref}{Figures~\ref{fig:#1} and~\ref{fig:#2}}

\condcommand{1}{\tablabel}{\label{tab:#1}}
\condcommand{1}{\tabref}{Table~\ref{tab:#1}}

\condcommand{1}{\eqlabel}{\label{eq:#1}}
\condcommand{1}{\eqref}{Equation~(\ref{eq:#1})}
\condcommand{2}{\eqsref}{Equations~(\ref{eq:#1}) and~(\ref{eq:#2})}

\condcommand{1}{\proplabel}{\label{prop:#1}}
\condcommand{1}{\propref}{Proposition~\ref{prop:#1}}

\condcommand{1}{\conjlabel}{\label{conj:#1}}
\condcommand{1}{\conjref}{Conjecture~\ref{conj:#1}}

\condcommand{1}{\deflabel}{\label{def:#1}}
\condcommand{1}{\defref}{Definition~\ref{def:#1}}

\condcommand{1}{\lemlabel}{\label{lem:#1}}
\condcommand{1}{\lemref}{Lemma~\ref{lem:#1}}

\condcommand{1}{\thmlabel}{\label{thm:#1}}
\condcommand{1}{\thmref}{Theorem~\ref{thm:#1}}

\condcommand{1}{\alglabel}{\label{alg:#1}}
\condcommand{1}{\algref}{Algorithm~\ref{alg:#1}}

\begin{document}

\title{Determinantal point processes for machine learning}
\author{Alex Kulesza \quad Ben Taskar}
\date{}

\maketitle

\begin{abstract}
  Determinantal point processes (DPPs) are elegant probabilistic
  models of repulsion that arise in quantum physics and random matrix
  theory.  In contrast to traditional structured models like Markov
  random fields, which become intractable and hard to approximate in
  the presence of negative correlations, DPPs offer efficient and
  exact algorithms for sampling, marginalization, conditioning, and
  other inference tasks.  We provide a gentle introduction to DPPs,
  focusing on the intuitions, algorithms, and extensions that are most
  relevant to the machine learning community, and show how DPPs can be
  applied to real-world applications like finding diverse sets of
  high-quality search results, building informative summaries by
  selecting diverse sentences from documents, modeling non-overlapping
  human poses in images or video, and automatically building timelines
  of important news stories.
\end{abstract}

\section{Introduction}
\seclabel{intro}

Probabilistic modeling and learning techniques have become
indispensable tools for analyzing data, discovering patterns, and
making predictions in a variety of real-world settings.  In recent
years, the widespread availability of both data and processing
capacity has led to new applications and methods involving more
complex, structured output spaces, where the goal is to simultaneously
make a large number of interrelated decisions.  Unfortunately, the
introduction of structure typically involves a combinatorial explosion
of output possibilities, making inference computationally impractical
without further assumptions.

A popular compromise is to employ graphical models, which are
tractable when the graph encoding local interactions between variables
is a tree.  For loopy graphs, inference can often be approximated in
the special case when the interactions between variables are positive
and neighboring nodes tend to have the same labels.  However, dealing
with global, negative interactions in graphical models remains
intractable, and heuristic methods often fail in practice.

Determinantal point processes (DPPs) offer a promising and
complementary approach.  Arising in quantum physics and random matrix
theory, DPPs are elegant probabilistic models of global, negative
correlations, and offer efficient algorithms for sampling,
marginalization, conditioning, and other inference tasks.  While they
have been studied extensively by mathematicians, giving rise to a
deep and beautiful theory, DPPs are relatively new in machine
learning.  We aim to provide a comprehensible introduction to DPPs,
focusing on the intuitions, algorithms, and extensions that are most
relevant to our community.

\subsection{Diversity}

A DPP is a distribution over subsets of a fixed ground set, for
instance, sets of search results selected from a large database.
Equivalently, a DPP over a ground set of $N$ items can be seen as
modeling a binary characteristic vector of length $N$.  The essential
characteristic of a DPP is that these binary variables are negatively
correlated; that is, the inclusion of one item makes the inclusion of
other items less likely.  The strengths of these negative correlations
are derived from a kernel matrix that defines a global measure of
similarity between pairs of items, so that more similar items are less
likely to co-occur.  As a result, DPPs assign higher probability to
sets of items that are \textit{diverse}; for example, a DPP will
prefer search results that cover multiple distinct aspects of a user's
query, rather than focusing on the most popular or salient one.

This focus on diversity places DPPs alongside a number of recently
developed techniques for working with diverse sets, particularly in
the information retrieval community
\citep{carbonell1998use,zhai2003beyond,chen2006less,yue2008predicting,radlinski2008learning,swaminathan2009essential,raman2012learning}.
However, unlike these methods, DPPs are fully probabilistic, opening
the door to a wider variety of potential applications, without
compromising algorithmic tractability.

The general concept of diversity can take on a number of forms
depending on context and application.  Including multiple kinds of
search results might be seen as \textit{covering} or
\textit{summarizing} relevant interpretations of the query or its
associated topics; see \figref{summarization}.  Alternatively, items
inhabiting a continuous space may exhibit diversity as a result of
\textit{repulsion}, as in \figref{repulsion}.  In fact, certain
repulsive quantum particles are known to be precisely described by a
DPP; however, a DPP can also serve as a model for general repulsive
phenomena, such as the locations of trees in a forest, which appear
diverse due to physical and resource constraints.  Finally, diversity
can be used as a \textit{filtering} prior when multiple selections
must be based on a single detector or scoring metric.  For instance,
in \figref{filtering} a weak pose detector favors large clusters of
poses that are nearly identical, but filtering through a DPP ensures
that the final predictions are well-separated.

\begin{figure}
  \centering
  \includegraphics[width=4in]{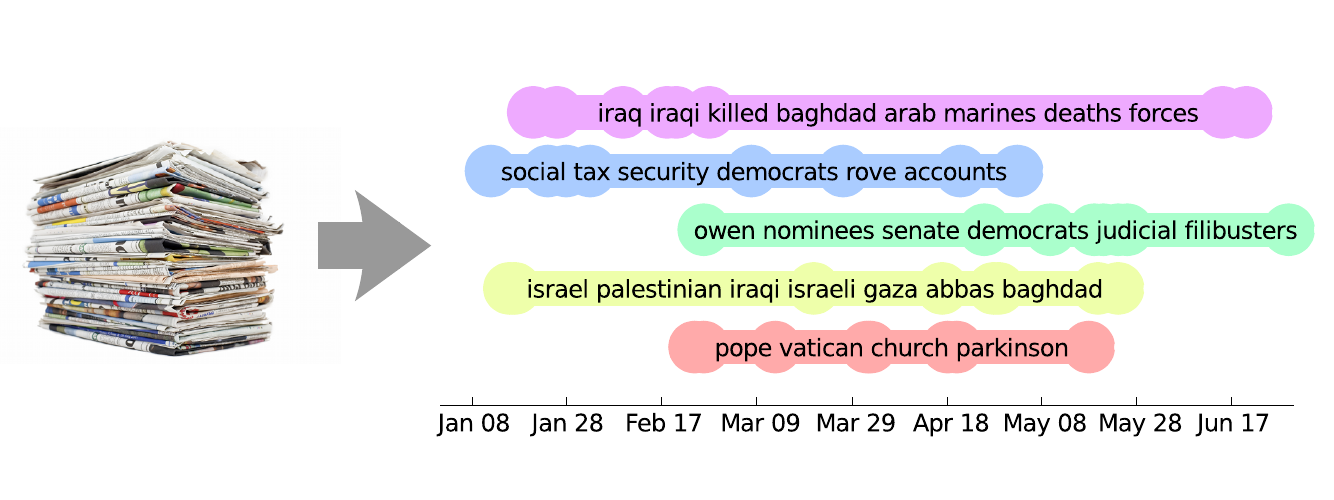}
  \caption{Diversity is used to generate a set of summary timelines
    describing the most important events from a large news corpus.}
  \figlabel{summarization}
\end{figure}

\begin{figure}
  \centering
  \includegraphics[width=4in]{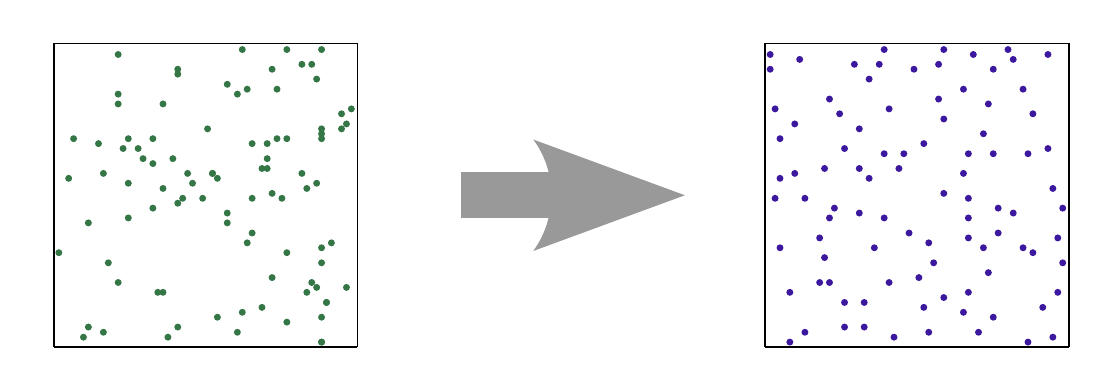}
  \caption{On the left, points are sampled randomly; on the right,
    repulsion between points leads to the selection of a diverse set
    of locations.}  
  \figlabel{repulsion}
\end{figure}

\begin{figure}
  \centering
  \includegraphics[width=4in]{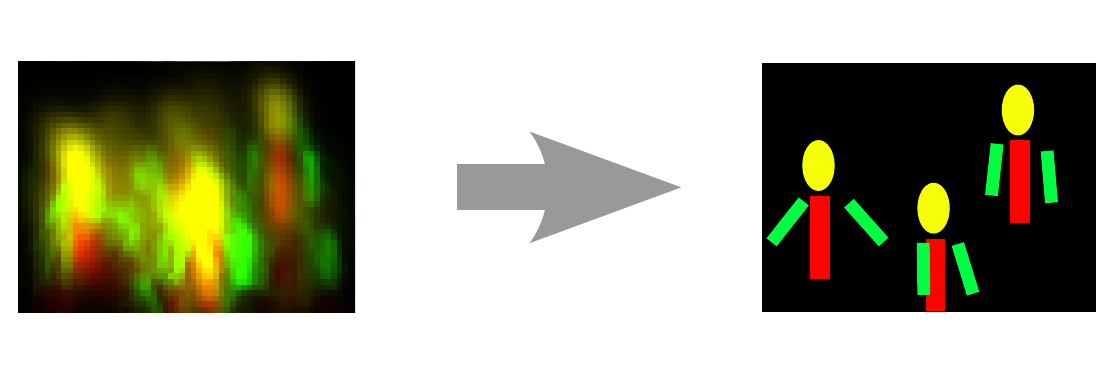}
  \caption{On the left, the output of a human pose detector is noisy
    and uncertain; on the right, applying diversity as a filter leads
    to a clean, separated set of predictions.}  
  \figlabel{filtering}
\end{figure}

Throughout this survey we demonstrate applications for DPPs in a
variety of settings, including:
\begin{itemize}
  \item The DUC 2003/2004 text summarization task, where we form
    extractive summaries of news articles by choosing diverse subsets of
    sentences (\secref{documentsummarization});
  \item An image search task, where we model human judgments of
    diversity for image sets returned by Google Image Search
    (\secref{imagesearch});
  \item A multiple pose estimation task, where we improve the
    detection of human poses in images from television shows by
    incorporating a bias toward non-overlapping predictions
    (\secref{poseestimation});
  \item A news threading task, where we automatically extract
    timelines of important news stories from a large corpus by
    balancing intra-timeline coherence with inter-timeline diversity
    (\secref{newsthreading}).
\end{itemize}


\subsection{Outline}

In this paper we present general mathematical background on DPPs along
with a range of modeling extensions, efficient algorithms, and
theoretical results that aim to enable practical modeling and
learning.  The material is organized as follows.

\paragraph{\secref{background}: \nameref{sec:background}.}
We begin with an introduction to determinantal point processes
tailored to the interests of the machine learning community.  We focus
on discrete DPPs, emphasizing intuitions and including new, simplified
proofs for some theoretical results.  We provide descriptions of known
efficient inference algorithms, and characterize their computational
properties.

\paragraph{\secref{tools}: \nameref{sec:tools}.} 
We describe a decomposition of the DPP that makes explicit its
fundamental tradeoff between quality and diversity.  We compare the
expressive power of DPPs and MRFs, characterizing the tradeoffs in
terms of modeling power and computational efficiency.  We also
introduce a dual representation for DPPs, showing how it can be used
to perform efficient inference over large ground sets.  When the data
are high-dimensional and dual inference is still too slow, we show
that random projections can be used to maintain a provably close
approximation to the original model while greatly reducing
computational requirements.

\paragraph{\secref{learning}: \nameref{sec:learning}.} 
We derive an efficient algorithm for learning the parameters of a
quality model when the diversity model is held fixed.  We employ this
learning algorithm to perform extractive summarization of news text.

\paragraph{\secref{kdpps}: \nameref{sec:kdpps}.} 
We present an extension of DPPs that allows for explicit control over
the number of items selected by the model.  We show not only that this
extension solves an important practical problem, but also that it
increases expressive power: a $k$-DPP can capture distributions that a
standard DPP cannot.  The extension to $k$-DPPs necessitates new
algorithms for efficient inference based on recursions for the
elementary symmetric polynomials.  We validate the new model
experimentally on an image search task.

\paragraph{\secref{sdpps}: \nameref{sec:sdpps}.} 
We extend DPPs to model diverse sets of structured items, such as
sequences or trees, where there are combinatorially many possible
configurations.  In this setting the number of possible subsets is
doubly-exponential, presenting a daunting computational challenge.
However, we show that a factorization of the quality and diversity
models together with the dual representation for DPPs makes efficient
inference possible using second-order message passing.  We demonstrate
structured DPPs on a toy geographical paths problem, a still-image
multiple pose estimation task, and two high-dimensional text threading
tasks.

\section{Determinantal point processes}
\seclabel{background}

Determinantal point processes (DPPs) were first identified as a class
by \citet{macchi1975coincidence}, who called them ``fermion
processes'' because they give the distributions of fermion systems at
thermal equilibrium.  The Pauli exclusion principle states that no two
fermions can occupy the same quantum state; as a consequence fermions
exhibit what is known as the ``anti-bunching'' effect.  This
repulsion is described precisely by a DPP.

In fact, years before Macchi gave them a general treatment, specific
DPPs appeared in major results in random matrix theory
\citep{mehta1960density, dyson1962i, dyson1962ii, dyson1962iii,
  ginibre1965statistical}, where they continue to play an important
role \citep{diaconis2003patterns, johansson2005random}.  Recently,
DPPs have attracted a flurry of attention in the mathematics community
\citep{borodin2000distributions, borodin2003janossy,
  borodin2005eynard, borodin2010adding, burton1993local,
  johansson2002non, johansson2004determinantal, johansson2005arctic,
  okounkov2001infinite, okounkov2003correlation, shirai2000fermion},
and much progress has been made in understanding their formal
combinatorial and probabilistic properties.  The term
``determinantal'' was first used by \citet{borodin2000distributions},
and has since become accepted as standard.  Many good mathematical
surveys are now available \citep{borodin2009determinantal,
  hough2006determinantal, shirai2003randomi, shirai2003randomii,
  lyons2003determinantal, soshnikov2000determinantal,
  tao09determinantal}.


We begin with an overview of the aspects of DPPs most relevant to the
machine learning community, emphasizing intuitions, algorithms, and
computational properties.

\subsection{Definition}

A point process $\P$ on a ground set $\Y$ is a probability measure
over ``point patterns'' or ``point configurations'' of $\Y$, which are
finite subsets of $\Y$.  For instance, $\Y$ could be a continuous time
interval during which a scientist records the output of a brain
electrode, with $\P(\{y_1,y_2,y_3\})$ characterizing the likelihood of
seeing neural spikes at times $y_1$, $y_2$, and $y_3$.  Depending on
the experiment, the spikes might tend to cluster together, or they
might occur independently, or they might tend to spread out in time.
$\P$ captures these correlations.

For the remainder of this paper, we will focus on discrete, finite
point processes, where we assume without loss of generality that $\Y =
\{1,2,\dots,N\}$; in this setting we sometimes refer to elements of
$\Y$ as {\it items}.  Much of our discussion extends to the continuous
case, but the discrete setting is computationally simpler and often
more appropriate for real-world data---e.g., in practice, the
electrode voltage will only be sampled at discrete intervals.  The
distinction will become even more apparent when we apply our methods
to $\Y$ with no natural continuous interpretation, such as the set of
documents in a corpus.

In the discrete case, a point process is simply a probability measure
on $2^\Y$, the set of all subsets of $\Y$.  A sample from $\P$ might
be the empty set, the entirety of $\Y$, or anything in between.  $\P$
is called a {\it determinantal point process} if, when $\bY$ is a
random subset drawn according to $\P$, we have, for every $A \subseteq
\Y$,
\begin{equation}
\P(A \subseteq \bY) = \det(K_A)
\eqlabel{marginal}
\end{equation}
for some real, symmetric $N \times N$ matrix $K$ indexed by the
elements of $\Y$.\footnote{In general, $K$ need not be symmetric.
  However, in the interest of simplicity, we proceed with this
  assumption; it is not a significant limitation for our purposes.}
Here, $K_A \equiv \left[K_{ij}\right]_{i,j\in A}$ denotes the
restriction of $K$ to the entries indexed by elements of $A$, and we
adopt $\det(K_{\emptyset}) = 1$.  Note that normalization is
unnecessary here, since we are defining marginal probabilities that
need not sum to 1.

Since $\P$ is a probability measure, all principal minors $\det(K_A)$
of $K$ must be nonnegative, and thus $K$ itself must be positive
semidefinite.  It is possible to show in the same way that the
eigenvalues of $K$ are bounded above by one using
\eqref{complement_dpp}, which we introduce later.  These requirements
turn out to be sufficient: any $K$, $0 \preceq K \preceq I$, defines a
DPP.  This will be a consequence of \thmref{sampling}.

We refer to $K$ as the {\it marginal kernel} since it contains all the
information needed to compute the probability of any subset $A$ being
included in $\bY$.  A few simple observations follow
from \eqref{marginal}.  If $A = \{i\}$ is a singleton, then we have
\begin{equation}
\P(i \in \bY) = K_{ii}\,.
\eqlabel{1_marginals}
\end{equation}
That is, the diagonal of $K$ gives the marginal probabilities of
inclusion for individual elements of $\Y$.  Diagonal entries close to
1 correspond to elements of $\Y$ that are almost always selected by
the DPP.  Furthermore, if $A=\{i,j\}$ is a two-element set, then
\begin{align}
\P(i,j \in \bY) &= 
\left| \begin{array}{cc}
  K_{ii} & K_{ij}\\
  K_{ji} & K_{jj}
\end{array} \right|\\
&= K_{ii} K_{jj} - K_{ij} K_{ji} \\
&= \P(i \in \bY)\P(j \in \bY)-K^2_{ij}\,.
\eqlabel{2_marginals}
\end{align}
Thus, the off-diagonal elements determine the negative correlations
between pairs of elements: large values of $K_{ij}$ imply that $i$ and
$j$ tend not to co-occur.

\eqref{2_marginals} demonstrates why DPPs are ``diversifying''.  If we
think of the entries of the marginal kernel as measurements of
similarity between pairs of elements in $\Y$, then highly similar
elements are unlikely to appear together.  If $K_{ij} =
\sqrt{K_{ii}K_{jj}}$, then $i$ and $j$ are ``perfectly similar'' and
do not appear together almost surely.  Conversely, when $K$ is
diagonal there are no correlations and the elements appear
independently.  Note that DPPs cannot represent distributions where
elements are {\it more} likely to co-occur than if they were
independent: correlations are always nonpositive.

\begin{figure}
  \centering
  \includegraphics[width=3in]{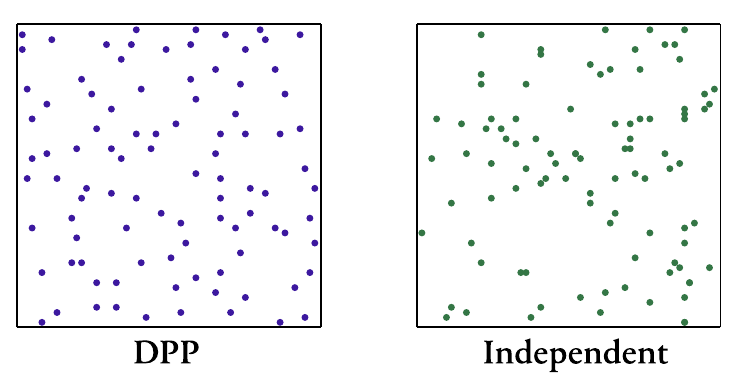}
  \caption[Samples of points in the plane]
    {A set of points in the plane drawn from a DPP (left), and the
    same number of points sampled independently using a Poisson point
    process (right).}  
  \figlabel{planesamples}
\end{figure}

\figref{planesamples} shows the difference between sampling a set of
points in the plane using a DPP (with $K_{ij}$ inversely related to
the distance between points $i$ and $j$), which leads to a relatively
uniformly spread set with good coverage, and sampling points
independently, which results in random clumping.

\subsubsection{Examples}

In this paper, our focus is on using DPPs to model real-world data.
However, many theoretical point processes turn out to be exactly
determinantal, which is one of the main reasons they have received so
much recent attention.  In this section we briefly describe a few
examples; some of them are quite remarkable on their own, and as a
whole they offer some intuition about the types of distributions that
are realizable by DPPs.  Technical details for each example can be
found in the accompanying reference.

\paragraph{Descents in random sequences 
  \textmd{\citep{borodin2010adding}}} 

Given a sequence of $N$ random numbers drawn uniformly and
independently from a finite set (say, the digits 0--9), the locations
in the sequence where the current number is less than the previous
number form a subset of $\{2,3,\dots,N\}$.  This subset is distributed
as a determinantal point process.  Intuitively, if the current number
is less than the previous number, it is probably not too large, thus
it becomes less likely that the next number will be smaller yet.  In
this sense, the positions of decreases repel one another.

\paragraph{Non-intersecting random walks
  \textmd{\citep{johansson2004determinantal}}}

Consider a set of $k$ independent, simple, symmetric random walks of
length $T$ on the integers.  That is, each walk is a sequence
$x_1,x_2,\dots,x_T$ where $x_i-x_{i+1}$ is either -1 or +1 with equal
probability.  If we let the walks begin at positions
$x_1^1,x_1^2,\dots,x_1^k$ and condition on the fact that they end at
positions $x_T^1,x_T^2,\dots,x_T^k$ and do not intersect, then the
positions $x_t^1,x_t^2,\dots,x_t^k$ at any time $t$ are a subset of
$\ints$ and distributed according to a DPP.  Intuitively, if the
random walks do not intersect, then at any time step they are likely
to be far apart.

\paragraph{Edges in random spanning trees 
  \textmd{\citep{burton1993local}}}

Let $G$ be an arbitrary finite graph with $N$ edges, and let $T$ be a
random spanning tree chosen uniformly from the set of all the spanning
trees of $G$.  The edges in $T$ form a subset of the edges of $G$ that
is distributed as a DPP.  The marginal kernel in this case is the
transfer-impedance matrix, whose entry $K_{e_1e_2}$ is the expected
signed number of traversals of edge $e_2$ when a random walk begins at
one endpoint of $e_1$ and ends at the other (the graph edges are first
oriented arbitrarily).  Thus, edges that are in some sense ``nearby''
in $G$ are similar according to $K$, and as a result less likely to
participate in a single uniformly chosen spanning tree.  As this
example demonstrates, some DPPs assign zero probability to sets whose
cardinality is not equal to a particular $k$; in this case, $k$ is the
number of nodes in the graph minus one---the number of edges in any
spanning tree.  We will return to this issue in \secref{kdpps}.

\paragraph{Eigenvalues of random matrices
  \textmd{\citep{ginibre1965statistical, mehta1960density}}}  

Let $M$ be a random matrix obtained by drawing every entry
independently from the complex normal distribution.  This is the
complex Ginibre ensemble.  The eigenvalues of $M$, which form a finite
subset of the complex plane, are distributed according to a DPP.  If a
Hermitian matrix is generated in the corresponding way, drawing each
diagonal entry from the normal distribution and each pair of
off-diagonal entries from the complex normal distribution, then we
obtain the Gaussian unitary ensemble, and the eigenvalues are now a
DPP-distributed subset of the real line.

\paragraph{Aztec diamond tilings
  \textmd{\citep{johansson2005arctic}}}  

The Aztec diamond is a diamond-shaped union of lattice squares, as
depicted in \figref{aztec}.  (Half of the squares have been colored
gray in a checkerboard pattern.)  A domino tiling is a perfect cover
of the Aztec diamond using $2\times 1$ rectangles, as in
\figref{aztec_tiled}.  Suppose that we draw a tiling uniformly at
random from among all possible tilings.  (The number of tilings is
known to be exponential in the width of the diamond.)  We can identify
this tiling with the subset of the squares that are (a) painted gray
in the checkerboard pattern and (b) covered by the left half of a
horizontal tile or the bottom half of a vertical tile (see
\figref{aztec_particles}).  This subset is distributed as a DPP.

\begin{figure}
  \centering
  \subfloat[][Aztec diamond (checkered)]{
    \includegraphics[width=1.6in]{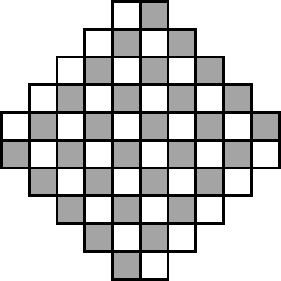}
    \figlabel{aztec}
  }\hspace{0.25in}
  \subfloat[][Domino tiling]{
    \includegraphics[width=1.6in]{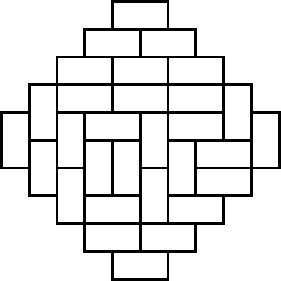}
    \figlabel{aztec_tiled}
  }\hspace{0.25in}
  \subfloat[][DPP]{
    \includegraphics[width=1.6in]{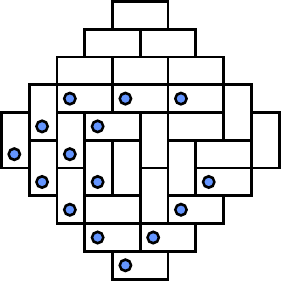}
    \figlabel{aztec_particles}
  }
  \caption[Aztec diamonds]{Aztec diamonds.}
\end{figure}

\subsection{L-ensembles}
\seclabel{L_ensembles}

For the purposes of modeling real data, it is useful to slightly
restrict the class of DPPs by focusing on {\it L-ensembles}.  First
introduced by \citet{borodin2005eynard}, an L-ensemble defines a DPP
not through the marginal kernel $K$, but through a real, symmetric
matrix $L$ indexed by the elements of $\Y$:
\begin{equation}
  \P_L(\bY = Y) \propto \det(L_Y)\,.
  \eqlabel{L_ensemble_pre}
\end{equation}
Whereas \eqref{marginal} gave the marginal probabilities of inclusion
for subsets $A$, \eqref{L_ensemble_pre} directly specifies the atomic
probabilities for every possible instantiation of $\bY$.  As for $K$,
it is easy to see that $L$ must be positive semidefinite.  However,
since \eqref{L_ensemble_pre} is only a statement of proportionality,
the eigenvalues of $L$ need not be less than one; any positive
semidefinite $L$ defines an L-ensemble.  The required normalization
constant can be given in closed form due to the fact that
$\sum_{Y\subseteq \Y} \det(L_Y) = \det(L+I)$, where $I$ is the $N
\times N$ identity matrix.  This is a special case of the following
theorem.
\begin{theorem}
  For any $A \subseteq \Y$,
  \begin{equation}
    \sum_{A \subseteq Y\subseteq\Y} \det(L_Y) = \det(L+I_{\bar A})\,,
    \eqlabel{normalization}
  \end{equation}
  where $I_{\bar A}$ is the diagonal matrix with ones in the diagonal
  positions corresponding to elements of $\bar A = \Y - A$, and zeros
  everywhere else.
  \thmlabel{lsums}
\end{theorem}
\begin{proof}
  Suppose that $A = \Y$; then \eqref{normalization} holds trivially.
  Now suppose inductively that the theorem holds whenever $\bar A$ has
  cardinality less than $k$.  Given $A$ such that $|\bar A| = k > 0$,
  let $i$ be an element of $\Y$ where $i \in \bar A$.  Splitting
  blockwise according to the partition $\Y = \{i\} \cup \Y-\{i\}$, we
  can write
  \begin{equation}
    L + I_{\bar A} = \left(\begin{array}{cc}
      L_{ii} + 1 & L_{i\bar i} \\
      L_{\bar ii} & L_{\Y-\{i\}} + I_{\Y-\{i\}-A}
    \end{array}\right)\,,
  \end{equation}
  where $L_{\bar i i}$ is the subcolumn of the $i$th column of $L$
  whose rows correspond to $\bar i$, and similarly for $L_{i\bar i}$.
  By multilinearity of the determinant, then,
  \begin{align}
    \det(L + I_{\bar A}) &= 
    \left|
    \begin{array}{cc}
      L_{ii} & L_{i\bar i} \\
      L_{\bar ii} & L_{\Y - \{i\}} + I_{\Y-\{i\} - A}
    \end{array}
    \right| + 
    \left|
    \begin{array}{cc}
      1 & \bzero \\
      L_{\bar ii} & L_{\Y - \{i\}} + I_{\Y-\{i\} - A} 
    \end{array}
    \right|\\
    & = \det(L + I_{\overline{A \cup \{i\}}}) + 
    \det(L_{\Y - \{i\}} + I_{\Y-\{i\} - A})\,.
  \end{align}
  We can now apply the inductive hypothesis separately to each term,
  giving
  \begin{align}
    \det(L + I_{\bar A}) 
    &= 
    \sum_{A \cup \{i\} \subseteq Y \subseteq \Y} \det(L_Y) +
    \sum_{A \subseteq Y \subseteq \Y - \{i\}} \det(L_Y)\\
    &= \sum_{A \subseteq Y \subseteq \Y} \det(L_Y)\,,
  \end{align}
  where we observe that every $Y$ either contains $i$ and is included
  only in the first sum, or does not contain $i$ and is included only
  in the second sum.
\end{proof}

\noindent Thus we have
\begin{equation}
  \P_L(\bY = Y) = \frac{\det(L_Y)}{\det(L+I)}\,.
  \eqlabel{L_ensemble}
\end{equation}
As a shorthand, we will write $\P_L(Y)$ instead of $\P_L(\bY = Y)$
when the meaning is clear.

We can write a version of \eqref{2_marginals} for L-ensembles, showing
that if $L$ is a measure of similarity then diversity is preferred:
\begin{equation}
  \P_L(\{i,j\}) \propto \P_L(\{i\})\P_L(\{j\}) - 
  \left(\frac{L_{ij}}{\det(L+I)}\right)^2\,.
\end{equation}
In this case we are reasoning about the full contents of $\bY$ rather
than its marginals, but the intuition is essentially the same.
Furthermore, we have the following result of
\citet{macchi1975coincidence}.

\begin{theorem}
  An L-ensemble is a DPP, and its marginal kernel is
  \begin{equation}
    K = L(L+I)^{-1} = I - (L+I)^{-1}\,.  
    \eqlabel{LtoK}
  \end{equation}
  \thmlabel{LtoK}
\end{theorem}
\begin{proof}
  Using \thmref{lsums}, the marginal probability of a set $A$ is
  \begin{align}
    \P_L(A \subseteq \bY) &= \frac{\sum_{A\subseteq Y\subseteq \Y} \det(L_Y)}
      {\sum_{Y \subseteq \Y} \det(L_Y)}\\
        &= \frac{\det(L+I_{\bar A})}{\det(L+I)}\\
        &= \det\left( (L+I_{\bar A})(L+I)^{-1} \right)\,.
  \end{align}
  We can use the fact that $L(L+I)^{-1} = I - (L+I)^{-1}$ to simplify
  and obtain
  \begin{align}
    \P_L(A \subseteq \bY)
    &= \det\left( I_{\bar A}(L+I)^{-1} + I - (L+I)^{-1} \right)\\
    &= \det\left( I - I_A(L+I)^{-1} \right)\\
    &= \det\left( I_{\bar A} + I_AK \right)\,,
  \end{align}
  where we let $K = I - (L+I)^{-1}$.  Now, we observe that left
  multiplication by $I_A$ zeros out all the rows of a matrix except
  those corresponding to $A$.  Therefore we can split blockwise using
  the partition $\Y = \bar A \cup A$ to get
  \begin{align}
    \det\left( I_{\bar A} + I_AK \right)
    &= \left| \begin{array}{cc}
      I_{|\bar A| \times |\bar A|} & \bzero\\
      K_{A\bar A} & K_A
    \end{array} \right|\\
    &= \det\left(K_A\right)\,.
  \end{align}
\end{proof}

Note that $K$ can be computed from an eigendecomposition of $L =
\sum_{n=1}^N \lambda_n \v_n \v_n^\trans$ by a simple rescaling of
eigenvalues:
\begin{equation}
  K = \sum_{n=1}^N \frac{\lambda_n}{\lambda_n+1} \v_n \v_n^\trans\,.
  \eqlabel{LtoKeig}
\end{equation}
Conversely, we can ask when a DPP with marginal kernel $K$ is also an
L-ensemble.  By inverting \eqref{LtoK}, we have
\begin{equation}
  L = K(I-K)^{-1}\,,
  \eqlabel{KtoL}
\end{equation}
and again the computation can be performed by eigendecomposition.
However, while the inverse in \eqref{LtoK} always exists due to the
positive coefficient on the identity matrix, the inverse in
\eqref{KtoL} may not.  In particular, when any eigenvalue of $K$
achieves the upper bound of 1, the DPP is not an L-ensemble.  We will
see later that the existence of the inverse in \eqref{KtoL} is
equivalent to $\P$ giving nonzero probability to the empty set.  (This
is somewhat analogous to the positive probability assumption in the
Hammersley-Clifford theorem for Markov random fields.)  This is not a
major restriction, for two reasons.  First, when modeling real data we
must typically allocate some nonzero probability for rare or noisy
events, so when cardinality is one of the aspects we wish to model,
the condition is not unreasonable.  Second, we will show in
\secref{kdpps} how to control the cardinality of samples drawn from
the DPP, thus sidestepping the representational limits of L-ensembles.

Modulo the restriction described above, $K$ and $L$ offer alternative
representations of DPPs.  Under both representations, subsets that
have higher diversity, as measured by the corresponding kernel, have
higher likelihood.  However, while $K$ gives marginal probabilities,
L-ensembles directly model the atomic probabilities of observing each
subset of $\Y$, which offers an appealing target for optimization.
Furthermore, $L$ need only be positive semidefinite, while the
eigenvalues of $K$ are bounded above.  For these reasons we will focus
our modeling efforts on DPPs represented as L-ensembles.

\subsubsection{Geometry}
\seclabel{geometry}

Determinants have an intuitive geometric interpretation.  Let $B$ be a
$D$ x $N$ matrix such that $L = B^\trans B$.  (Such a $B$ can always
be found for $D \leq N$ when $L$ is positive semidefinite.)  Denote
the columns of $B$ by $B_i$ for $i = 1,2,\dots,N$.  Then:
\begin{equation}
  \P_L(Y) \propto \det(L_Y) = \vol^2(\{B_i\}_{i\in Y})\,,
  \eqlabel{volumes}
\end{equation}
where the right hand side is the squared $|Y|$-dimensional volume of
the parallelepiped spanned by the columns of $B$ corresponding to
elements in $Y$.

Intuitively, we can think of the columns of $B$ as feature vectors
describing the elements of $\Y$.  Then the kernel $L$ measures
similarity using dot products between feature vectors, and
\eqref{volumes} says that the probability assigned by a DPP to a set
$Y$ is related to the volume spanned by its associated feature
vectors.  This is illustrated in \figref{geometry}.

\begin{figure}
\centering
\includegraphics[width=5in]{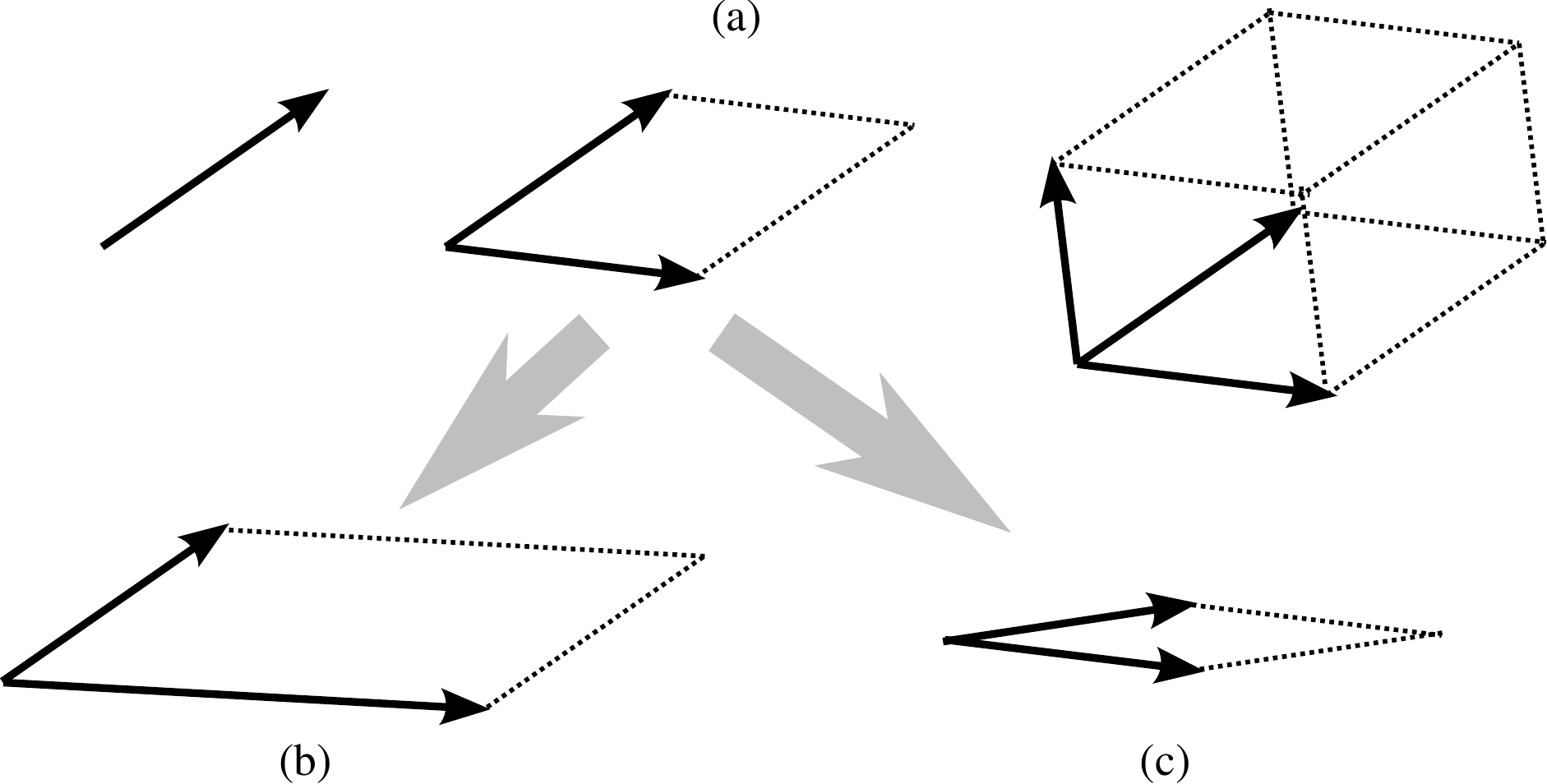}
\caption[Geometry of DPPs]
  {A geometric view of DPPs: each vector corresponds to an
  element of $\Y$.  (a) The probability of a subset $Y$ is the square
  of the volume spanned by its associated feature vectors.  (b) As the
  magnitude of an item's feature vector increases, so do the
  probabilities of sets containing that item.  (c) As the similarity
  between two items increases, the probabilities of sets containing
  both of them decrease.} 
\figlabel{geometry}
\end{figure}

From this intuition we can verify several important DPP properties.
Diverse sets are more probable because their feature vectors are more
orthogonal, and hence span larger volumes.  Items with parallel
feature vectors are selected together with probability zero, since
their feature vectors define a degenerate parallelepiped.  All else
being equal, items with large-magnitude feature vectors are more
likely to appear, because they multiply the spanned volumes for sets
containing them.

We will revisit these intuitions in \secref{decomposition}, where we
decompose the kernel $L$ so as to separately model the direction and
magnitude of the vectors $B_i$.

\subsection{Properties}

In this section we review several useful properties of DPPs.

\paragraph{Restriction}

If $\bY$ is distributed as a DPP with marginal kernel $K$, then $\bY
\cap A$, where $A \subseteq \Y$, is also distributed as a DPP, with
marginal kernel $K_A$.

\paragraph{Complement}

If $\bY$ is distributed as a DPP with marginal kernel $K$, then $\Y -
\bY$ is also distributed as a DPP, with marginal kernel $\bar K = I -
K$.  In particular, we have
\begin{equation}
  \P(A \cap \bY = \emptyset) = \det(\bar K_A) = \det(I - K_A)\,,
  \eqlabel{complement_dpp}
\end{equation}
where $I$ indicates the identity matrix of appropriate size.  It may
seem counterintuitive that the complement of a diversifying process
should also encourage diversity.  However, it is easy to see that
\begin{align}
  \P(i,j \not\in\bY) &= 1 - \P(i\in\bY) - \P(j\in\bY) + \P(i,j\in\bY) \\
  &\leq 1 - \P(i\in\bY) - \P(j\in\bY) + \P(i\in\bY)\P(j\in\bY) \\
  &= \P(i\not\in\bY) + \P(j\not\in\bY) - 1 
  + (1-\P(i\not\in\bY))(1-\P(j\not\in\bY)) \\
  &= \P(i\not\in\bY)\P(j\not\in\bY)\,,
\end{align}
where the inequality follows from \eqref{2_marginals}.

\paragraph{Domination}

If $K \preceq K'$, that is, $K'-K$ is positive semidefinite, then for
all $A \subseteq \Y$ we have
\begin{equation}
  \det(K_A) \leq \det(K'_A)\,.
  \eqlabel{domination}
\end{equation}
In other words, the DPP defined by $K'$ is larger than the one defined
by $K$ in the sense that it assigns higher marginal probabilities to
every set $A$.  An analogous result fails to hold for $L$ due to the
normalization constant.

\paragraph{Scaling}

If $K = \gamma K'$ for some $0 \leq \gamma < 1$, then for all $A
\subseteq \Y$ we have
\begin{equation}
  \det(K_A) = \gamma^{|A|}\det(K'_A)\,.
\end{equation}
It is easy to see that $K$ defines the distribution obtained by taking
a random set distributed according to the DPP with marginal kernel
$K'$, and then independently deleting each of its elements with
probability $1-\gamma$.

\paragraph{Cardinality}

Let $\lambda_1,\lambda_2,\dots,\lambda_N$ be the eigenvalues of $L$.  Then
$|\bY|$ is distributed as the number of successes in $N$ Bernoulli
trials, where trial $n$ succeeds with probability
$\frac{\lambda_n}{\lambda_n+1}$.  This fact follows from
\thmref{sampling}, which we prove in the next section.  One immediate
consequence is that $|\bY|$ cannot be larger than $\rank(L)$.  More
generally, the expected cardinality of $\bY$ is
\begin{equation}
  \E[|\bY|] = \sum_{n=1}^N \frac{\lambda_n}{\lambda_n+1} = \tr(K)\,,
  \eqlabel{card_exp}
\end{equation}
and the variance is
\begin{equation}
  \var(|\bY|) = \sum_{n=1}^N \frac{\lambda_n}{(\lambda_n+1)^2}\,.
  \eqlabel{card_var}
\end{equation}
Note that, by \eqref{LtoK}, $\frac{\lambda_1}{\lambda_1+1},
\frac{\lambda_2}{\lambda_2+1}, \dots, \frac{\lambda_N}{\lambda_N+1}$
are the eigenvalues of $K$.  \figref{lambda_transform} shows a plot of
the function $f(\lambda) = \frac{\lambda}{\lambda+1}$.  It is easy to
see from this why the class of L-ensembles does not include DPPs where
the empty set has probability zero---at least one of the Bernoulli
trials would need to always succeed, and in turn one or more of the
eigenvalues of $L$ would be infinite.
  
In some instances, the sum of Bernoullis may be an appropriate model
for uncertain cardinality in real-world data, for instance when
identifying objects in images where the number of objects is unknown
in advance.  In other situations, it may be more practical to fix the
cardinality of $\bY$ up front, for instance when a set of exactly ten
search results is desired, or to replace the sum of Bernoullis with an
alternative cardinality model.  We show how these goals can be can be
achieved in \secref{kdpps}.

\begin{figure}
\centering
\includegraphics[width=4in]{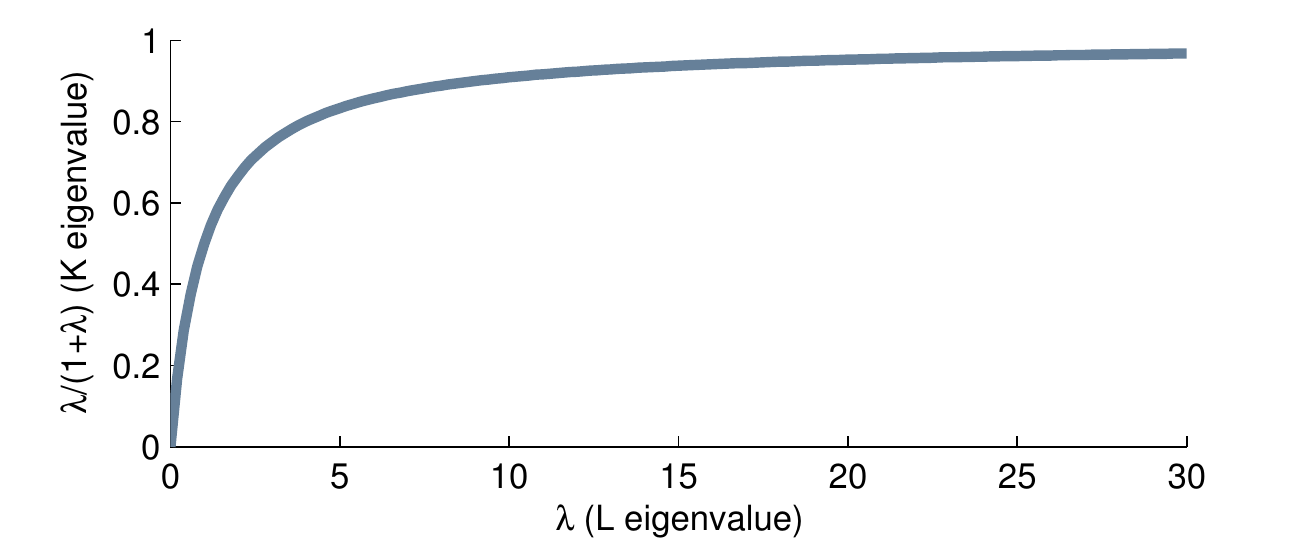}
\caption[Mapping between eigenvalues of $L$ and $K$]
  {The mapping between eigenvalues of $L$ and $K$.}
\figlabel{lambda_transform}
\end{figure}

\subsection{Inference}
\seclabel{dpp_inference}

One of the primary advantages of DPPs is that, although the number of
possible realizations of $\bY$ is exponential in $N$, many types of
inference can be performed in polynomial time.  In this section we
review the inference questions that can (and cannot) be answered
efficiently.  We also discuss the empirical practicality of the
associated computations and algorithms, estimating the largest values
of $N$ that can be handled at interactive speeds (within 2--3 seconds)
as well as under more generous time and memory constraints.  The
reference machine used for estimating real-world performance has eight
Intel Xeon E5450 3Ghz cores and 32GB of memory.

\subsubsection{Normalization}

As we have already seen, the partition function, despite being a sum
over $2^N$ terms, can be written in closed form as $\det(L+I)$.
Determinants of $N\times N$ matrices can be computed through matrix
decomposition in $O(N^3)$ time, or reduced to matrix multiplication
for better asymptotic performance.  The Coppersmith-Winograd
algorithm, for example, can be used to compute determinants in about
$O(N^{2.376})$ time.  Going forward, we will use $\omega$ to denote
the exponent of whatever matrix multiplication algorithm is used.

Practically speaking, modern computers can calculate determinants up
to $N\approx5{,}000$ at interactive speeds, or up to
$N\approx40{,}000$ in about five minutes.  When $N$ grows much larger,
the memory required simply to store the matrix becomes limiting.
(Sparse storage of larger matrices is possible, but computing
determinants remains prohibitively expensive unless the level of
sparsity is extreme.)

\subsubsection{Marginalization}

The marginal probability of any set of items $A$ can be computed using
the marginal kernel as in \eqref{marginal}.  From
\eqref{complement_dpp} we can also compute the marginal probability
that none of the elements in $A$ appear.  (We will see below how
marginal probabilities of arbitrary configurations can be computed
using conditional DPPs.)

If the DPP is specified as an L-ensemble, then the computational
bottleneck for marginalization is the computation of $K$.  The
dominant operation is the matrix inversion, which requires at least
$O(N^{\omega})$ time by reduction to multiplication, or $O(N^3)$ using
Gauss-Jordan elimination or various matrix decompositions, such as the
eigendecomposition method proposed in \secref{L_ensembles}.  Since an
eigendecomposition of the kernel will be central to sampling, the
latter approach is often the most practical when working with DPPs.

Matrices up to $N\approx 2{,}000$ can be inverted at interactive
speeds, and problems up to $N\approx20{,}000$ can be completed in
about ten minutes.

\subsubsection{Conditioning}

The distribution obtained by conditioning a DPP on the event that none
of the elements in a set $A$ appear is easy to compute.  For $B
\subseteq{\Y}$ not intersecting with $A$ we have
\begin{align}
  \P_L(\bY = B \mid A\cap \bY = \emptyset) 
  &= \frac{\P_L(\bY = B)}{\P_L(A\cap \bY = \emptyset)} \\
  &= \frac{\det(L_B)}{\sum_{B' : B' \cap A = \emptyset} \det(L_{B'})} \\
  &= \frac{\det(L_B)}{\det(L_{\bar A} + I)}\,,
  \eqlabel{cond_dpp_excl}
\end{align}
where $L_{\bar A}$ is the restriction of $L$ to the rows and columns
indexed by elements in $\Y - A$.  In other words, the conditional
distribution (over subsets of $\Y-A$) is itself a DPP, and its kernel
$L_{\bar A}$ is obtained by simply dropping the rows and columns of
$L$ that correspond to elements in $A$.

We can also condition a DPP on the event that {\it all} of the
elements in a set $A$ are observed.  For $B$ not intersecting with $A$
we have
\begin{align}
  \P_L(\bY = A \cup B \mid A \subseteq \bY) 
  &= \frac{\P_L(\bY = A \cup B)}{\P_L(A \subseteq \bY)} \\
  &= \frac{\det(L_{A\cup B})}{\sum_{B' : B' \cap A = \emptyset} \det(L_{A\cup B'})} \\
  &= \frac{\det(L_{A\cup B})}{\det(L + I_{\bar A})}\,,
  \eqlabel{cond_dpp_incl}
\end{align}
where $I_{\bar A}$ is the matrix with ones in the diagonal entries
indexed by elements of $\Y - A$ and zeros everywhere else.  Though it
is not immediately obvious, \citet{borodin2005eynard} showed that this
conditional distribution (over subsets of $\Y - A$) is again a DPP,
with a kernel given by
\begin{equation}
  L^A = \left(\left[(L+I_{\bar A})^{-1}\right]_{\bar A}\right)^{-1} - I\,.
  \eqlabel{cond_incl_kernel}
\end{equation}
(Following the $N\times N$ inversion, the matrix is restricted to rows
and columns indexed by elements in $\Y - A$, then inverted again.)  It
is easy to show that the inverses exist if and only if the probability
of $A$ appearing is nonzero.

Combining \eqref{cond_dpp_excl} and \eqref{cond_dpp_incl}, we can
write the conditional DPP given an arbitrary combination of appearing
and non-appearing elements:
\begin{equation}
  \P_L(\bY = A^{\inc} \cup B 
  \mid A^{\inc} \subseteq \bY, A^{\exc}\cap \bY = \emptyset) 
  = \frac{\det(L_{A^{\inc}\cup B})}{\det(L_{\bar A^{\exc}} + I_{\bar A^{\inc}})}\,.
\end{equation}
The corresponding kernel is
\begin{equation}
  L^{A^{\inc}}_{\bar A^{\exc}} 
  = \left(\left[(L_{\bar A^{\exc}}+I_{\bar A^{\inc}})^{-1}
    \right]_{\bar A^{\inc}}\right)^{-1} - I\,.
\end{equation}
Thus, the class of DPPs is closed under most natural conditioning
operations.  

\paragraph{General marginals}

These formulas also allow us to compute arbitrary marginals.  For
example, applying \eqref{LtoK} to \eqref{cond_incl_kernel} yields the
marginal kernel for the conditional DPP given the appearance of $A$:
\begin{equation}
  K^A = I-\left[(L+I_{\bar A})^{-1}\right]_{\bar A}\,.
\end{equation}
Thus we have
\begin{equation}
\P(B \subseteq \bY | A \subseteq \bY) = \det(K^A_B)\,.
\end{equation}
(Note that $K^A$ is indexed by elements of $\Y-A$, so this is only
defined when $A$ and $B$ are disjoint.)  Using \eqref{complement_dpp}
for the complement of a DPP, we can now compute the marginal
probability of any partial assignment, i.e.,
\begin{align}
  \P(A \subseteq \bY, B \cap \bY = \emptyset) 
  &= \P(A \subseteq \bY) \P(B \cap \bY = \emptyset | A \subseteq \bY)\\
  &= \det(K_A)\det(I - K^{A}_B)\,.
\end{align}

Computing conditional DPP kernels in general is asymptotically as
expensive as the dominant matrix inversion, although in some cases
(conditioning only on non-appearance), the inversion is not necessary.
In any case, conditioning is at most a small constant factor more
expensive than marginalization.



\subsubsection{Sampling}
\seclabel{sampling}

\algref{dpp_sampling}, due to \citet{hough2006determinantal}, gives an
efficient algorithm for sampling a configuration $Y$ from a DPP.  The
input to the algorithm is an eigendecomposition of the DPP kernel $L$.
Note that $\e_i$ is the $i$th standard basis $N$-vector, which is all
zeros except for a one in the $i$th position.  We will prove the
following theorem.

\begin{algorithm}[tb]
\begin{algorithmic}
  \STATE {\bfseries Input:} eigendecomposition
    $\{(\v_n,\lambda_n)\}_{n=1}^N$ of $L$\\
  \STATE $J \leftarrow \emptyset$
  \FOR{$n = 1,2,\dots,N$}
  \STATE $J \leftarrow J \cup \{n\}$ with prob. $\frac{\lambda_n}{\lambda_n+1}$
  \ENDFOR\\
  \STATE $V \leftarrow \{\v_n\}_{n\in J}$
  \STATE $Y \leftarrow \emptyset$
  \WHILE{$|V|>0$}
  \STATE Select $i$ from $\Y$ with $\Pr(i) = \frac{1}{|V|}\sum_{\v\in V} 
    (\v^\trans \e_i)^2$
  \STATE $Y \leftarrow Y \cup i$
  \STATE $V \leftarrow V_\bot$, an orthonormal basis for the subspace 
    of $V$ orthogonal to $\e_i$
  \ENDWHILE
  \STATE {\bfseries Output:} $Y$
\end{algorithmic}
\caption{Sampling from a DPP}
\alglabel{dpp_sampling}
\end{algorithm}

\begin{theorem}
  Let $L = \sum_{n=1}^N \lambda_n \v_n \v_n^\trans$ be an orthonormal
  eigendecomposition of a positive semidefinite matrix $L$.  Then
  \algref{dpp_sampling} samples $\bY\sim\P_L$.  
  \thmlabel{sampling}
\end{theorem}

\algref{dpp_sampling} has two main loops, corresponding to two phases
of sampling.  In the first phase, a subset of the eigenvectors is
selected at random, where the probability of selecting each
eigenvector depends on its associated eigenvalue.  In the second
phase, a sample $Y$ is produced based on the selected vectors.  Note
that on each iteration of the second loop, the cardinality of $Y$
increases by one and the dimension of $V$ is reduced by one.  Since
the initial dimension of $V$ is simply the number of selected
eigenvectors ($|J|$), \thmref{sampling} has the previously stated
corollary that the cardinality of a random sample is distributed as a
sum of Bernoulli variables.

To prove \thmref{sampling} we will first show that a DPP can be
expressed as a mixture of simpler, {\it elementary} DPPs.  We will
then show that the first phase chooses an elementary DPP according to
its mixing coefficient, while the second phase samples from the
elementary DPP chosen in phase one.

\begin{definition}
  A DPP is called {\bf elementary} if every eigenvalue of its marginal
  kernel is in $\{0,1\}$.  We write $\P^V$, where $V$ is a set of
  orthonormal vectors, to denote an elementary DPP with marginal
  kernel $K^V = \sum_{\v \in V} \v\v^\trans$.
\end{definition}
We introduce the term ``elementary'' here;
\citet{hough2006determinantal} refer to elementary DPPs as
determinantal {\it projection} processes, since $K^V$ is an
orthonormal projection matrix to the subspace spanned by $V$.  Note
that, due to \eqref{KtoL}, elementary DPPs are not generally
L-ensembles.  We start with a technical lemma.

\begin{lemma}
  Let $W_n$ for $n=1,2,\dots,N$ be an arbitrary sequence of $k \times k$
  rank-one matrices, and let $(W_{n})_i$ denote the $ith$ column of
  $W_n$.  Let $W_J = \sum_{n\in J} W_n$.  Then
  \begin{equation}
    \det(W_J) = \sum_{n_1,n_2,\dots,n_k \in J, \atop \distinct}
    \det([(W_{n_1})_1 (W_{n_2})_2\dots (W_{n_k})_k])\,.
  \end{equation}
  \lemlabel{dettricks}
\end{lemma}
\begin{proof}
  Expanding on the first column of $W_J$ using the multilinearity of the
  determinant,
  \begin{equation}
    \det(W_J) = \sum_{n \in J} \det([(W_{n})_1 (W_{J})_2 \dots (W_{J})_k])\,,
  \end{equation}
  and, applying the same operation inductively to all columns,
  \begin{equation}
    \det(W_J) = \sum_{n_1,n_2,\dots,n_k \in J} 
    \det([(W_{n_1})_1 (W_{n_2})_2 \dots (W_{n_k})_k])\,.
  \end{equation}
  Since $W_n$ has rank one, the determinant of any matrix containing
  two or more columns of $W_n$ is zero; thus the terms in the sum
  vanish unless $n_1,n_2,\dots,n_k$ are distinct.
\end{proof}

\begin{lemma}
  A DPP with kernel $L = \sum_{n=1}^N \lambda_n \v_n \v_n^\trans$ is a
  mixture of elementary DPPs:
  \begin{equation}
    \P_L = \frac{1}{\det(L+I)}
    \sum_{J \subseteq \{1,2,\dots,N\}} \P^{V_J} \prod_{n \in J} \lambda_n\,,
    \eqlabel{elementarymixture}
  \end{equation}
  where $V_J$ denotes the set $\{\v_n\}_{n \in J}$.
  \lemlabel{elementarymixture}
\end{lemma}
\begin{proof}
  Consider an arbitrary set $A$, with $k = |A|$.  Let $W_n =
  [\v_n\v_n^\trans]_A$ for $n = 1,2,\dots,N$; note that all of the $W_n$
  have rank one.  From the definition of $K^{V_J}$, the mixture
  distribution on the right hand side of \eqref{elementarymixture}
  gives the following expression for the marginal probability of $A$:
  \begin{equation}
    \frac{1}{\det(L+I)}\sum_{J \subseteq \{1,2,\dots,N\}}
    \det\left(\sum_{n\in J}W_n\right) \prod_{n \in J} \lambda_n\,.
  \end{equation}
  Applying \lemref{dettricks}, this is equal to
  \begin{align}
    &\frac{1}{\det(L+I)}\sum_{J \subseteq \{1,2,\dots,N\}}
    \sum_{n_1,\dots,n_k \in J, \atop \distinct}
    \det([(W_{n_1})_{1}\dots (W_{n_k})_{k}]) \prod_{n \in J} \lambda_n\\
    &= \frac{1}{\det(L+I)}\sum_{n_1,\dots,n_k =1, \atop \distinct}^N
    \det([(W_{n_1})_{1}\dots (W_{n_k})_{k}]) 
    \sum_{J \supseteq \{n_1,\dots,n_k\}} \prod_{n \in J} \lambda_n\\
    &= \frac{1}{\det(L+I)}\sum_{n_1,\dots,n_k =1, \atop \distinct}^N
    \det([(W_{n_1})_{1}\dots (W_{n_k})_{k}]) \frac{\lambda_{n_1}}{\lambda_{n_1}+1} \cdots
    \frac{\lambda_{n_k}}{\lambda_{n_k}+1}
    \prod_{n=1}^N (\lambda_n+1) \\
    &= \sum_{n_1,\dots,n_k =1, \atop \distinct}^N
    \det\left(\left[\frac{\lambda_{n_1}}{\lambda_{n_1}+1}(W_{n_1})_{1}\dots 
      \frac{\lambda_{n_k}}{\lambda_{n_k}+1}(W_{n_k})_{k}\right]\right)\,,
  \end{align}
  using the fact that $\det(L+I) = \prod_{n=1}^N (\lambda_n + 1)$.
  Applying \lemref{dettricks} in reverse and then the definition of
  $K$ in terms of the eigendecomposition of $L$, we have that the
  marginal probability of $A$ given by the mixture is
  \begin{equation}
    \det\left(\sum_{n=1}^N \frac{\lambda_n}{\lambda_n+1} W_n\right)
    = \det(K_A)\,.
  \end{equation}
  Since the two distributions agree on all marginals, they are
  equal.
\end{proof}

\noindent
Next, we show that elementary DPPs have fixed cardinality.

\begin{lemma}
  If $\bY$ is drawn according to an elementary DPP $\P^V$, then $|\bY|
  = |V|$ with probability one.   
  \lemlabel{elementaryfixed}
\end{lemma}
\begin{proof}
  Since $K^V$ has rank $|V|$, $\P^V(Y \subseteq \bY) = 0$ whenever
  $|Y| > |V|$, so $|\bY| \leq |V|$.  But we also have
  \begin{align}
    E\left[|\bY|\right] &= E\left[\sum_{n=1}^N \I(n \in \bY)\right]\\
    &= \sum_{n=1}^N E\left[\I(n \in \bY)\right]\\
    &= \sum_{n=1}^N K_{nn} = \tr(K) = |V|\,.
  \end{align}
  Thus $|\bY| = |V|$ almost surely.
\end{proof}

\noindent
We can now prove the theorem.

\begin{proof}[Proof of \thmref{sampling}]

  \lemref{elementarymixture} says that the mixture weight of $\P^{V_J}
  $ is given by the product of the eigenvalues $\lambda_{n}$
  corresponding to the eigenvectors $\v_{n}\in V_{J}$, normalized by
  $\det(L+I) = \prod_{n=1}^{N}(\lambda_{n}+1)$.  This shows that the
  first loop of \algref{dpp_sampling} selects an elementary DPP $\P^V$
  with probability equal to its mixture component.  All that remains
  is to show that the second loop samples $\bY \sim \P^V$.
  
  Let $B$ represent the matrix whose rows are the eigenvectors in $V$,
  so that $K^V = B^\trans B$.  Using the geometric interpretation of
  determinants introduced in \secref{geometry}, $\det(K^V_Y)$ is equal
  to the squared volume of the parallelepiped spanned by
  $\{B_i\}_{i\in Y}$.  Note that since $V$ is an orthonormal set,
  $B_i$ is just the projection of $\e_i$ onto the subspace spanned by
  $V$.
  
  Let $k = |V|$.  By \lemref{elementaryfixed} and symmetry, we can
  consider without loss of generality a single $Y = \{1,2,\dots,k\}$.
  Using the fact that any vector both in the span of $V$ and
  perpendicular to $\e_i$ is also perpendicular to the projection of
  $\e_i$ onto the span of $V$, by the base $\times$ height formula for
  the volume of a parallelepiped we have
  \begin{equation}
    \vol\left(\{B_i\}_{i\in Y}\right) = \Vert B_1 \Vert 
    \vol\left(\{\proj_{\perp\e_1} B_i\}_{i=2}^{k}\right)\,,
  \end{equation}
  where $\proj_{\perp\e_1}$ is the projection operator onto the
  subspace orthogonal to $\e_1$.  Proceeding inductively,
  \begin{equation}
    \vol\left(\{B_i\}_{i\in Y}\right) = 
    \Vert B_1 \Vert \Vert \proj_{\perp\e_1} B_2 \Vert 
    \cdots \Vert \proj_{\perp\e_1,\dots,\e_{k-1}} B_k \Vert\,.
    \eqlabel{vol_decomp}
  \end{equation}

  Assume that, as iteration $j$ of the second loop in
  \algref{dpp_sampling} begins, we have already selected $y_1 =
  1,y_2=2,\dots,y_{j-1} = j-1$.  Then $V$ in the algorithm has been
  updated to an orthonormal basis for the subspace of the original $V$
  perpendicular to $\e_1,\dots,\e_{j-1}$, and the probability of
  choosing $y_j = j$ is exactly 
  \begin{equation}
    \frac{1}{|V|}\sum_{\v\in V} (\v^\trans \e_j)^2 = 
    \frac{1}{k-j+1} \Vert
    \proj_{\perp\e_1,\dots,\e_{j-1}} B_j \Vert^2\,.
  \end{equation}
Therefore, the
  probability of selecting the sequence $1,2,\dots,k$ is
  \begin{equation}
    \frac{1}{k!} \Vert B_1 \Vert^2 \Vert \proj_{\perp\e_1} B_2 \Vert^2
    \cdots \Vert \proj_{\perp\e_1,\dots,\e_{k-1}} B_k \Vert^2 =
    \frac{1}{k!}\vol^2\left(\{B_i\}_{i\in Y}\right)\,.
  \end{equation}
  Since volume is symmetric, the argument holds identically for all of
  the $k!$ orderings of $Y$.  Thus the total probability that
  \algref{dpp_sampling} selects $Y$ is $\det(K^V_Y)$.
\end{proof}

\begin{corollary}
  \algref{dpp_sampling} generates $Y$ in uniformly random order.
\end{corollary}

\paragraph{Discussion}

To get a feel for the sampling algorithm, it is useful to visualize
the distributions used to select $i$ at each iteration, and to see
how they are influenced by previously chosen items.
\figref{marginals_line} shows this progression for a simple DPP where
$\Y$ is a finely sampled grid of points in $[0,1]$, and the kernel is
such that points are more similar the closer together they are.
Initially, the eigenvectors $V$ give rise to a fairly uniform
distribution over points in $\Y$, but as each successive point is
selected and $V$ is updated, the distribution shifts to avoid points
near those already chosen.  \figref{marginals_plane} shows a similar
progression for a DPP over points in the unit square.

\begin{figure}
\centering
\subfloat[][Sampling points on an interval] {
\includegraphics[trim=0.25in 0 1in 0,clip,width=5in]{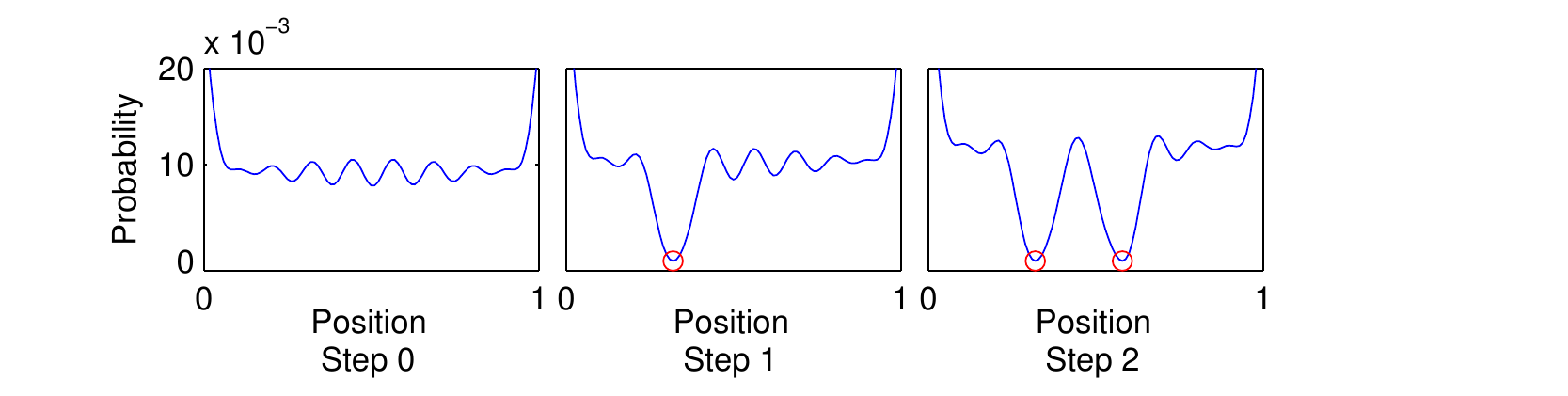}
\figlabel{marginals_line}
}\\
\subfloat[][Sampling points in the plane] {
\includegraphics[trim=0.45in 3.4in 0.6in 3.4in,clip,width=5in]
                {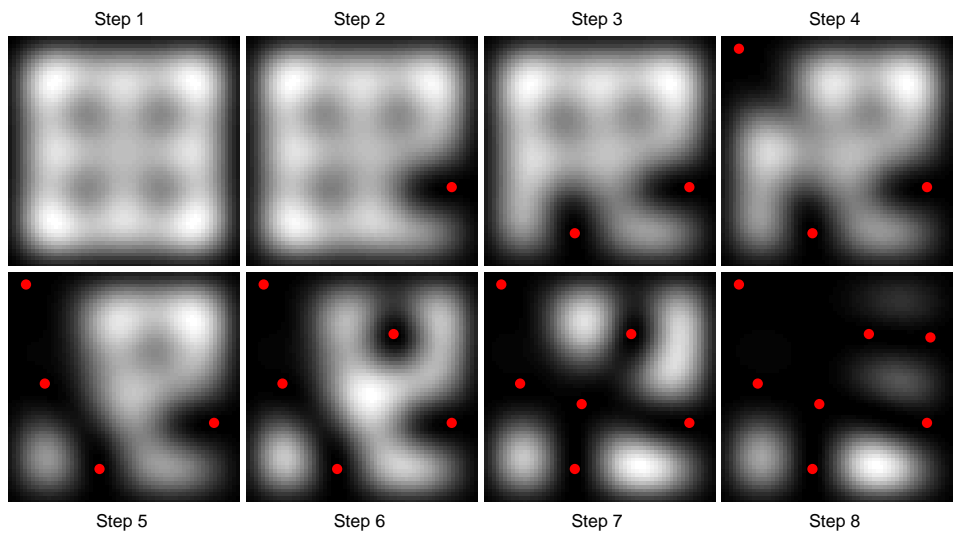}
\figlabel{marginals_plane}
}
\caption[Sampling illustration]
  {Sampling DPP over one-dimensional (top) and two-dimensional
  (bottom) particle positions.  Red circles indicate already selected
  positions. On the bottom, lighter color corresponds to higher
  probability.  The DPP naturally reduces the probabilities for
  positions that are similar to those already selected.}
\end{figure}

The sampling algorithm also offers an interesting analogy to
clustering.  If we think of the eigenvectors of $L$ as representing
soft clusters, and the eigenvalues as representing their
strengths---the way we do for the eigenvectors and eigenvalues of the
Laplacian matrix in spectral clustering---then a DPP can be seen as
performing a clustering of the elements in $\Y$, selecting a random
subset of clusters based on their strength, and then choosing one
element per selected cluster.  Of course, the elements are not chosen
independently and cannot be identified with specific clusters;
instead, the second loop of \algref{dpp_sampling} coordinates the
choices in a particular way, accounting for overlap between the
eigenvectors.

\algref{dpp_sampling} runs in time $O(Nk^3)$, where $k = |V|$ is the
number of eigenvectors selected in phase one.  The most expensive
operation is the $O(Nk^2)$ Gram-Schmidt orthonormalization required to
compute $V_\perp$.  If $k$ is large, this can be reasonably expensive,
but for most applications we do not want high-cardinality DPPs.  (And
if we want very high-cardinality DPPs, we can potentially save time by
using \eqref{complement_dpp} to sample the complement instead.)  In
practice, the initial eigendecomposition of $L$ is often the
computational bottleneck, requiring $O(N^3)$ time.  Modern multi-core
machines can compute eigendecompositions up to $N\approx 1{,}000$ at
interactive speeds of a few seconds, or larger problems up to
$N\approx 10{,}000$ in around ten minutes.  In some instances, it may
be cheaper to compute only the top $k$ eigenvectors; since phase one
tends to choose eigenvectors with large eigenvalues anyway, this can
be a reasonable approximation when the kernel is expected to be low
rank.  Note that when multiple samples are desired, the
eigendecomposition needs to be performed only once.

\citet{deshpande2010efficient} recently proposed a
$(1-\epsilon)$-approximate algorithm for sampling that runs in time
$O(N^2 \log N \frac{k^2}{\epsilon^2} + N \log^\omega N
\frac{k^{2\omega+1}}{\epsilon^{2\omega}} \log(\frac{k}{\epsilon}\log
N))$ when $L$ is already decomposed as a Gram matrix, $L = B^\trans
B$.  When $B$ is known but an eigendecomposition is not (and $N$ is
sufficiently large), this may be significantly faster than the exact
algorithm.

\subsubsection{Finding the mode}
\seclabel{dppmap}

Finding the mode of a DPP---that is, finding the set $Y \subseteq \Y$
that maximizes $\P_L(Y)$---is NP-hard.  In conditional models, this
problem is sometimes referred to as maximum {\it a posteriori} (or
MAP) inference, and it is also NP-hard for most general structured
models such as Markov random fields.  Hardness was first shown for
DPPs by \citet{ko1995exact}, who studied the closely-related maximum
entropy sampling problem: the entropy of a set of jointly Gaussian
random variables is given (up to constants) by the log-determinant of
their covariance matrix; thus finding the maximum entropy subset of
those variables requires finding the principal covariance submatrix
with maximum determinant.  Here, we adapt the argument of
\citet{civril2009selecting}, who studied the problem of finding
maximum-volume submatrices.

\begin{theorem}
  Let {\sc dpp-mode} be the optimization problem of finding, for a
  positive semidefinite $N\times N$ input matrix $L$ indexed by
  elements of $\Y$, the maximum value of $\det(L_Y)$ over all $Y
  \subseteq \Y$.  {\sc dpp-mode} is NP-hard, and furthermore it is
  NP-hard even to approximate {\sc dpp-mode} to a factor of
  $\frac{8}{9}+\epsilon$.
  \thmlabel{argmax}
\end{theorem}
\begin{proof}
  We reduce from {\sc exact 3-cover} (X3C).  An instance of X3C is a
  set $S$ and a collection $C$ of three-element subsets of $S$; the
  problem is to decide whether there is a sub-collection $C' \subseteq
  C$ such that every element of $S$ appears exactly once in $C'$
  (that is, $C'$ is an exact 3-cover).  X3C is known to be
  NP-complete.

  The reduction is as follows.  Let $\Y = \{1,2,\dots,|C|\}$, and let
  $B$ be an $|S| \times |C|$ matrix where $B_{si} = \frac{1}{\sqrt{3}}$
  if $C_i$ contains $s \in S$ and zero otherwise.  Define $L = \gamma
  B^\trans B$, where $1 < \gamma \leq \frac{9}{8}$.  Note that the
  diagonal of $L$ is constant and equal to $\gamma$, and an
  off-diagonal entry $L_{ij}$ is zero if and only if $C_i$ and $C_j$
  do not intersect.  $L$ is positive semidefinite by construction, and
  the reduction requires only polynomial time.  Let $k =
  \frac{|S|}{3}$.  We will show that the maximum value of $\det(L_Y)$
  is greater than $\gamma^{k-1}$ if and only if $C$ contains an exact
  3-cover of $S$.

  $(\leftarrow)$ If $C' \subseteq C$ is an exact 3-cover of $S$, then
  it must contain exactly $k$ 3-sets.  Letting $Y$ be the set of
  indices in $C'$, we have $L_Y = \gamma I$, and thus its determinant
  is $\gamma^k > \gamma^{k-1}$.

  $(\rightarrow)$ Suppose there is no 3-cover of $S$ in $C$.  Let $Y$
  be an arbitrary subset of $\Y$.  If $|Y| < k$, then
  \begin{equation}  
    \det(L_Y) \leq \prod_{i\in Y} L_{ii} = \gamma^{|Y|} \leq \gamma^{k-1}\,.
  \end{equation}
  Now suppose $|Y| \geq k$, and assume without loss of generality that
  $Y = \{1,2,\dots,|Y|\}$.  We have $L_Y = \gamma B_Y^\trans B_Y$, and
  \begin{equation}
    \det(L_Y) = \gamma^{|Y|}\vol^2\left(\{B_i\}_{i\in Y}\right)\,.
  \end{equation}
  By the base $\times$ height formula,
  \begin{equation}
    \vol\left(\{B_i\}_{i\in Y}\right) = 
    \Vert B_1 \Vert \Vert \proj_{\perp B_1} B_2 \Vert 
    \cdots \Vert \proj_{\perp B_1,\dots,B_{|Y|-1}} B_{|Y|} \Vert\,.
  \end{equation}
  Note that, since the columns of $B$ are normalized, each term in the
  product is at most one.  Furthermore, at least $|Y| - k + 1$ of the
  terms must be strictly less than one, because otherwise there would
  be $k$ orthogonal columns, which would correspond to a 3-cover.  By
  the construction of $B$, if two columns $B_i$ and $B_j$ are not
  orthogonal then $C_i$ and $C_j$ overlap in at least one of three
  elements, so we have
  \begin{align}
    \Vert \proj_{\perp B_j} B_i \Vert 
    &= \Vert B_i - (B_i^\trans B_j)B_j \Vert \\
    &\leq \Vert B_i - \frac{1}{3}B_j \Vert \\
    &\leq \sqrt{\frac{8}{9}}\,.
  \end{align}
  Therefore,
  \begin{align}
    \det(L_Y) &\leq \gamma^{|Y|}\left(\frac{8}{9}\right)^{|Y|-k+1}\\
    &\leq \gamma^{k-1}\,,
  \end{align}
  since $\gamma \leq \frac{9}{8}$.

  We have shown that the existence of a 3-cover implies that the optimal
  value of {\sc dpp-mode} is at least $\gamma^k$, while the optimal
  value cannot be more than $\gamma^{k-1}$ if there is no 3-cover.
  Thus any algorithm that can approximate {\sc dpp-mode} to better
  than a factor of $\frac{1}{\gamma}$ can be used to solve X3C in
  polynomial time.  We can choose $\gamma = \frac{9}{8}$ to show that
  an approximation ratio of $\frac{8}{9} + \epsilon$ is NP-hard.
\end{proof}

Since there are only $|C|$ possible cardinalities for $Y$,
\thmref{argmax} shows that {\sc dpp-mode} is NP-hard even under
cardinality constraints.

\citep{ko1995exact} propose an exact, exponential branch-and-bound
algorithm for finding the mode using greedy heuristics to build
candidate sets; they tested their algorithm on problems up to $N=75$,
successfully finding optimal solutions in up to about an hour.  Modern
computers are likely a few orders of magnitude faster; however, this
algorithm is still probably impractical for applications with large
$N$.  \citet{civril2009selecting} propose an efficient greedy
algorithm for finding a set of size $k$, and prove that it achieves an
approximation ratio of $O(\frac{1}{k!})$.  While this guarantee is
relatively poor for all but very small $k$, in practice the results
may be useful nonetheless.

\paragraph{Submodularity}

$\P_L$ is {\it log-submodular}; that is,
\begin{equation}
  \log\P_L(Y \cup \{i\}) - \log\P_L(Y) \geq \log\P_L(Y' \cup \{i\}) -
  \log\P_L(Y')
\end{equation}
whenever $Y \subseteq Y' \subseteq \Y - \{i\}$.  Intuitively, adding
elements to $Y$ yields diminishing returns as $Y$ gets larger.  (This
is easy to show by a volume argument.)  Submodular functions can be
minimized in polynomial time \citep{schrijver2000combinatorial}, and
many results exist for approximately maximizing {\it monotone}
submodular functions, which have the special property that supersets
always have higher function values than their subsets
\citep{nemhauser1978analysis,fisher1978analysis,feige1998threshold}.
In \secref{summ_inf} we will discuss how these kinds of greedy
algorithms can be adapted for DPPs.  However, in general $\P_L$ is
highly non-monotone, since the addition of even a single element can
decrease the probability to zero.

Recently, \citet{feige2007maximizing} showed that even non-monotone
submodular functions can be approximately maximized in polynomial time
using a local search algorithm, and a growing body of research has
focused on extending this result in a variety of ways
\citep{lee2009non,gharan2011submodular,vondrak2011submodular,feldman2011nonmonotone,feldman2011unified,chekuri2011submodular}.
In our recent work we showed how the computational structure of DPPs
gives rise to a particularly efficient variant of these methods
\citep{kulesza2012near}.

\subsection{Related processes}

Historically, a wide variety of point process models have been
proposed and applied to applications involving diverse subsets,
particularly in settings where the items can be seen as points in a
physical space and diversity is taken to mean some sort of
``spreading'' behavior.  However, DPPs are essentially unique among
this class in having efficient and exact algorithms for probabilistic
inference, which is why they are particularly appealing models for
machine learning applications.  In this section we briefly survey the
wider world of point processes and discuss the computational
properties of alternative models; we will focus on point processes
that lead to what is variously described as diversity, repulsion,
(over)dispersion, regularity, order, and inhibition.

\subsubsection{Poisson point processes}

Perhaps the most fundamental point process is the Poisson point
process, which is depicted on the right side of \figref{planesamples}
\citep{daley2003introduction}.  While defined for continuous $\Y$, in
the discrete setting the Poisson point process can be simulated by
flipping a coin independently for each item, and including those items
for which the coin comes up heads.  Formally,
\begin{equation}
  \P(\bY = Y) = \prod_{i\in Y} p_i \prod_{i\not\in Y} (1-p_i)\,,
\end{equation}
where $p_i \in [0,1]$ is the bias of the $i$th coin.  The process is
called \textit{stationary} when $p_i$ does not depend on $i$; in a
spatial setting this means that no region has higher density than any
other region.

A random set $\bY$ distributed as a Poisson point process has the
property that whenever $A$ and $B$ are disjoint subsets of $\Y$, the
random variables $\bY \cap A$ and $\bY \cap B$ are independent; that
is, the points in $\bY$ are not correlated.  It is easy to see that
DPPs generalize Poisson point processes by choosing the marginal
kernel $K$ with $K_{ii} = p_i$ and $K_{ij} = 0, i\neq j$.  This
implies that inference for Poisson point processes is at least as
efficient as for DPPs; in fact, it is more efficient, since for
instance it is easy to compute the most likely configuration.
However, since Poisson point processes do not model correlations
between variables, they are rather uninteresting for most real-world
applications.

Addressing this weakness, various procedural modifications of the
Poisson process have been proposed in order to introduce correlations
between items.  While such constructions can be simple and intuitive,
leading to straightforward sampling algorithms, they tend to make
general statistical inference difficult.

\paragraph{Mat{\'e}rn repulsive processes}

\citet{matern1960spatial,matern1986spatial} proposed a set of
techniques for thinning Poisson point processes in order to induce a
type of repulsion when the items are embedded in a Euclidean space.
The Type I process is obtained from a Poisson set $\bY$ by removing
all items in $\bY$ that lie within some radius of another item in
$\bY$.  That is, if two items are close to each other, they are both
removed; as a result all items in the final process are spaced at
least a fixed distance apart.  The Type II Mat{\'e}rn repulsive
process, designed to achieve the same minimum distance property while
keeping more items, begins by independently assigning each item in
$\bY$ a uniformly random ``time'' in $[0,1]$.  Then, any item within a
given radius of a point having a smaller time value is removed.  Under
this construction, when two items are close to each other only the
later one is removed.  Still, an item may be removed due to its
proximity with an earlier item that was itself removed.  This leads to
the Type III process, which proceeds dynamically, eliminating items in
time order whenever an earlier point which has not been removed lies
within the radius.

Inference for the Mat{\'e}rn processes is computationally daunting.
First and second order moments can be computed for Types I and II, but
in those cases computing the likelihood of a set $Y$ is seemingly
intractable \citep{moller2010perfect}.  Recent work by
\citet{huber2009likelihood} shows that it is possible to compute
likelihood for certain restricted Type III processes, but computing
moments cannot be done in closed form.  In the general case,
likelihood for Type III processes must be estimated using an expensive
Markov chain Monte Carlo algorithm.

The Mat{\'e}rn processes are called ``hard-core'' because they
strictly enforce a minimum radius between selected items.  While this
property leads to one kind of diversity, it is somewhat limited, and
due to the procedural definition it is difficult to characterize the
side effects of the thinning process in a general way.
\citet{stoyan1985one} considered an extension where the radius is
itself chosen randomly, which may be more natural for certain
settings, but it does not alleviate the computational issues.

\paragraph{Random sequential adsorption}

The Mat{\'e}rn repulsive processes are related in spirit to the random
sequential adsorption (RSA) model, which has been used in physics and
chemistry to model particles that bind to two-dimensional surfaces,
e.g., proteins on a cell membrane
\citep{tanemura1979random,finegold1979maximum,feder1980random,swendsen1981dynamics,hinrichsen1986geometry,ramsden1993review}.
RSA is described generatively as follows.  Initially, the surface is
empty; iteratively, particles arrive and bind uniformly at random to a
location from among all locations that are not within a given radius
of any previously bound particle.  When no such locations remain (the
``jamming limit''), the process is complete.

Like the Mat{\'e}rn processes, RSA is a hard-core model, designed
primarily to capture packing distributions, with much of the theoretical
analysis focused on the achievable density.  If the set of locations
is further restricted at each step to those found in an initially
selected Poisson set $\bY$, then it is equivalent to a Mat{\'e}rn Type
III process \citep{huber2009likelihood}; it therefore shares the same
computational burdens.

\subsubsection{Gibbs and Markov point processes}

While manipulating the Poisson process procedurally has some intuitive
appeal, it seems plausible that a more holistically-defined process
might be easier to work with, both analytically and algorithmically.
The Gibbs point process provides such an approach, offering a general
framework for incorporating correlations among selected items
\citep{preston1976random,ripley1977markov,ripley1991statistical,van2000markov,moller2004statistical,moller2007modern,daley2008introduction}.
The Gibbs probability of a set $Y$ is given by
\begin{equation}
  \P(\bY = Y) \propto \exp(-U(Y))\,,
  \eqlabel{markovpp}
\end{equation}
where $U$ is an energy function.  Of course, this definition is fully
general without further constraints on $U$.  A typical assumption is
that $U$ decomposes over subsets of items in $Y$; for instance
\begin{equation}
  \exp(-U(Y)) = \prod_{A \subseteq Y, |A| \leq k} \psi_{|A|}(A)
\end{equation}
for some small constant order $k$ and potential functions $\psi$.  In
practice, the most common case is $k=2$, which is sometimes called a
pairwise interaction point process \citep{diggle1987nonparametric}:
\begin{equation}
  \P(\bY = Y) \propto \prod_{i\in Y} \psi_1(i) 
                      \prod_{i,j \subseteq Y} \psi_2(i,j)\,.
\end{equation}
In spatial settings, a Gibbs point process whose potential functions
are identically 1 whenever their input arguments do not lie within a
ball of fixed radius---that is, whose energy function can be
decomposed into only local terms---is called a Markov point process.
A number of specific Markov point processes have become well-known.

\paragraph{Pairwise Markov processes}

\citet{strauss1975model} introduced a simple pairwise Markov point
process for spatial data in which the potential function $\psi_2(i,j)$
is piecewise constant, taking the value 1 whenever $i$ and $j$ are at
least a fixed radius apart, and the constant value $\gamma$ otherwise.
When $\gamma > 1$, the resulting process prefers clustered items.
(Note that $\gamma > 1$ is only possible in the discrete case; in the
continuous setting the distribution becomes non-integrable.)  We are
more interested in the case $0 < \gamma < 1$, where configurations in
which selected items are near one another are discounted.  When
$\gamma = 0$, the resulting process becomes hard-core, but in general
the Strauss process is ``soft-core'', preferring but not requiring
diversity.

The Strauss process is typical of pairwise Markov processes in that
its potential function $\psi_2(i,j) = \psi(|i-j|)$ depends only on the
distance between its arguments.  A variety of alternative definitions
for $\psi(\cdot)$ have been proposed
\citep{ripley1977markov,ogata1984likelihood}.  For instance,
\begin{align}
  \psi(r) &= 1-\exp(-(r/\sigma)^2)\\
  \psi(r) &= \exp(-(\sigma/r)^n),\quad n > 2\\
  \psi(r) &= \min(r/\sigma,1)
\end{align}
where $\sigma$ controls the degree of repulsion in each case.  Each
definition leads to a point process with a slightly different concept of
diversity.

\paragraph{Area-interaction point processes}

\citet{baddeley1995area} proposed a non-pairwise spatial Markov point
process called the \textit{area-interaction} model, where $U(Y)$ is
given by $\log \gamma$ times the total area contained in the union of
discs of fixed radius centered at all of the items in $Y$.  When
$\gamma > 1$, we have $\log \gamma > 0$ and the process prefers sets
whose discs cover as little area as possible, i.e., whose items are
clustered.  When $0 < \gamma < 1$, $\log \gamma$ becomes negative, so
the process prefers ``diverse'' sets covering as much area as
possible.

If none of the selected items fall within twice the disc radius of
each other, then $\exp(-U(Y))$ can be decomposed into potential
functions over single items, since the total area is simply the sum of
the individual discs.  Similarly, if each disc intersects with at most
one other disc, the area-interaction process can be written as a
pairwise interaction model.  However, in general, an unbounded number
of items might appear in a given disc; as a result the
area-interaction process is an infinite-order Gibbs process.  Since
items only interact when they are near one another, however, local
potential functions are sufficient and the process is Markov.

\paragraph{Computational issues}

Markov point processes have many intuitive properties.  In fact, it
is not difficult to see that, for discrete ground sets $\Y$, the
Markov point process is equivalent to a Markov random field (MRF) on
binary variables corresponding to the elements of $\Y$.  In
\secref{dppsvsmrfs} we will return to this equivalence in order to
discuss the relative expressive possibilities of DPPs and MRFs.  For
now, however, we simply observe that, as for MRFs with negative
correlations, repulsive Markov point processes are computationally
intractable.  Even computing the normalizing constant for
\eqref{markovpp} is NP-hard in the cases outlined above
\citep{daley2003introduction,moller2004statistical}.

On the other hand, quite a bit of attention has been paid to
approximate inference algorithms for Markov point processes, employing
pseudolikelihood
\citep{besag1977some,besag1982point,jensen1991pseudolikelihood,ripley1991statistical},
Markov chain Monte Carlo methods
\citep{ripley1977markov,besag1993spatial,haggstrom1999characterization,berthelsen2006bayesian},
and other approximations
\citep{ogata1985estimation,diggle1994parameter}.  Nonetheless, in
general these methods are slow and/or inexact, and closed-form
expressions for moments and densities rarely exist
\citep{moller2007modern}.  In this sense the DPP is unique.

\subsubsection{Generalizations of determinants}

The determinant of a $k \times k$ matrix $K$ can be written as a
polynomial of degree $k$ in the entries of $K$; in particular,
\begin{equation}
  \det(K) = \sum_{\pi} \sgn(\pi) \prod_{i=1}^k K_{i,\pi(i)}\,,
  \eqlabel{determinant}
\end{equation}
where the sum is over all permutations $\pi$ on $1,2,\dots,k$, and
$\sgn$ is the permutation sign function.  In a DPP, of course, when
$K$ is (a submatrix of) the marginal kernel \eqref{determinant} gives
the appearance probability of the $k$ items indexing $K$.  A natural
question is whether generalizations of this formula give rise to
alternative point processes of interest.

\paragraph{Immanantal point processes}

In fact, \eqref{determinant} is a special case of the more general
\textit{matrix immanant}, where the $\sgn$ function is replaced by
$\chi$, the irreducible representation-theoretic character of the
symmetric group on $k$ items corresponding to a particular partition
of $1,2,\dots,k$.  When the partition has $k$ parts, that is, each
element is in its own part, $\chi(\pi) = \sgn(\pi)$ and we recover the
determinant.  When the partition has a single part, $\chi(\pi) = 1$
and the result is the permanent of $K$.  The associated
\textit{permanental process} was first described alongside DPPs by
\citet{macchi1975coincidence}, who referred to it as the ``boson
process.''  Bosons do not obey the Pauli exclusion principle, and the
permanental process is in some ways the opposite of a DPP, preferring
sets of points that are more tightly clustered, or less diverse, than
if they were independent.  Several recent papers have considered its
properties in some detail
\citep{hough2006determinantal,mccullagh2006permanental}.  Furthermore,
\citet{diaconis2000immanants} considered the point processes induced
by general immanants, showing that they are well defined and in some
sense ``interpolate'' between determinantal and permanental processes.

Computationally, obtaining the permanent of a matrix is \#P-complete
\citep{valiant1979complexity}, making the permanental process
difficult to work with in practice.  Complexity results for immanants
are less definitive, with certain classes of immanants apparently hard
to compute \citep{burgisser2000computational,brylinski2003complexity},
while some upper bounds on complexity are known
\citep{hartmann1985complexity,barvinok1990computational}, and at least
one non-trivial case is efficiently computable
\citep{grone1984algorithm}.  It is not clear whether the latter result
provides enough leverage to perform inference beyond computing
marginals.

\paragraph{$\alpha$-determinantal point processes}

An alternative generalization of \eqref{determinant} is given by the
so-called $\alpha$-determinant, where $\sgn(\pi)$ is replaced by
$\alpha^{k-\nu(\pi)}$, with $\nu(\pi)$ counting the number of cycles
in $\pi$ \citep{vere1997alpha,hough2006determinantal}.  When $\alpha =
-1$ the determinant is recovered, and when $\alpha = +1$ we have
again the permanent.  Relatively little is known for other values of
$\alpha$, although \citet{shirai2003randomi} conjecture that the
associated process exists when $0 \leq \alpha \leq 2$ but not when
$\alpha > 2$.  Whether $\alpha$-determinantal processes have useful
properties for modeling or computational advantages remains an open
question.

\paragraph{Hyperdeterminantal point processes}

A third possible generalization of \eqref{determinant} is the
hyperdeterminant originally proposed by \citet{cayley1843theory} and
discussed in the context of point processes by
\citet{evans2009hyperdeterminantal}.  Whereas the standard determinant
operates on a two-dimensional matrix with entries indexed by pairs of
items, the hyperdeterminant operates on higher-dimensional kernel
matrices indexed by sets of items.  The hyperdeterminant potentially
offers additional modeling power, and
\citet{evans2009hyperdeterminantal} show that some useful properties
of DPPs are preserved in this setting.  However, so far relatively
little is known about these processes.

\subsubsection{Quasirandom processes}

Monte Carlo methods rely on draws of random points in order to
approximate quantities of interest; randomness guarantees that,
regardless of the function being studied, the estimates will be
accurate in expectation and converge in the limit.  However, in
practice we get to observe only a finite set of values drawn from the
random source.  If, by chance, this set is ``bad'', the resulting
estimate may be poor.  This concern has led to the development of
so-called \textit{quasirandom} sets, which are in fact
deterministically generated, but can be substituted for random sets in
some instances to obtain improved convergence guarantees
\citep{niederreiter1992quasi,sobol1998quasi}.

In contrast with pseudorandom generators, which attempt to mimic
randomness by satisfying statistical tests that ensure
unpredictability, quasirandom sets are not designed to appear random,
and their elements are not (even approximately) independent.  Instead,
they are designed to have low \textit{discrepancy}; roughly speaking,
low-discrepancy sets are ``diverse'' in that they cover the sample
space evenly.  Consider a finite subset $Y$ of $[0,1]^D$, with
elements $\x^{(i)} = (x^{(i)}_1,x^{(i)}_2,\dots,x^{(i)}_D)$ for $i =
1,2,\dots,k$.  Let $S_{\x} = [0,x_1) \times [0,x_2) \times \cdots
    \times [0,x_D)$ denote the box defined by the origin and the point
      $\x$.  The discrepancy of $Y$ is defined as follows.
\begin{equation}
  \disc(Y) = \max_{\x \in Y} \left| \frac{|Y \cap S_{\x}|}{k} 
  - \vol(S_{\x}) \right|\,.
\end{equation}
That is, the discrepancy measures how the empirical density $|Y\cap
S_{\x}|/k$ differs from the uniform density $\vol(S_{\x})$ over the
boxes $S_{\x}$.  Quasirandom sets with low discrepancy cover the unit
cube with more uniform density than do pseudorandom sets, analogously
to \figref{planesamples}.

This deterministic uniformity property makes quasirandom sets useful
for Monte Carlo estimation via (among other results) the Koksma-Hlawka
inequality \citep{hlawka1961funktionen,niederreiter1992quasi}.  For a
function $f$ with bounded variation $V(f)$ on the unit cube, the
inequality states that
\begin{equation}
  \left| \frac{1}{k} \sum_{\x\in Y} f(\x) - \int_{[0,1]^D} f(\x) d\x \right|
  \leq V(f)\disc(Y)\,.
\end{equation}
Thus, low-discrepancy sets lead to accurate quasi-Monte Carlo
estimates.  In contrast to typical Monte Carlo guarantees, the
Koksma-Hlawka inequality is deterministic.  Moreover, since the rate
of convergence for standard stochastic Monte Carlo methods is
$k^{-1/2}$, this result is an (asymptotic) improvement when the
discrepancy diminishes faster than $k^{-1/2}$.

In fact, it is possible to construct quasirandom sequences where the
discrepancy of the first $k$ elements is $O((\log k)^D / k)$; the
first such sequence was proposed by \citet{halton1960efficiency}.  The
Sobol sequence \citep{sobol1967distribution}, introduced later, offers
improved uniformity properties and can be generated efficiently
\citep{bratley1988algorithm}.

It seems plausible that, due to their uniformity characteristics,
low-discrepancy sets could be used as computationally efficient but
non-probabilistic tools for working with data exhibiting diversity.
An algorithm generating quasirandom sets could be seen as an efficient
prediction procedure if made to depend somehow on input data and a set
of learned parameters.  However, to our knowledge no work has yet
addressed this possibility.

\section{Representation and algorithms}
\seclabel{tools}

Determinantal point processes come with a deep and beautiful theory,
and, as we have seen, exactly characterize many theoretical processes.
However, they are also promising models for real-world data that
exhibit diversity, and we are interested in making such applications
as intuitive, practical, and computationally efficient as possible.
In this section, we present a variety of fundamental techniques and
algorithms that serve these goals and form the basis of the extensions
we discuss later.

We begin by describing a decomposition of the DPP kernel that offers
an intuitive tradeoff between a unary model of quality over the items
in the ground set and a global model of diversity.  The geometric
intuitions from \secref{background} extend naturally to this
decomposition.  Splitting the model into quality and diversity
components then allows us to make a comparative study of {\it
  expressiveness}---that is, the range of distributions that the model
can describe.  We compare the expressive powers of DPPs
and negative-interaction Markov random fields,
showing that the two models are incomparable in general but exhibit
qualitatively similar characteristics, despite the computational
advantages offered by DPPs.

Next, we turn to the challenges imposed by large data sets, which are
common in practice.  We first address the case where $N$, the number
of items in the ground set, is very large.  In this setting, the
super-linear number of operations required for most DPP inference
algorithms can be prohibitively expensive.  However, by introducing a
dual representation of a DPP we show that efficient DPP inference
remains possible when the kernel is low-rank.  When the kernel is not
low-rank, we prove that a simple approximation based on random
projections dramatically speeds inference while guaranteeing that the
deviation from the original distribution is bounded.  These techniques
will be especially useful in \secref{sdpps}, when we consider
exponentially large $N$.

Finally, we discuss some alternative formulas for the likelihood of a
set $Y$ in terms of the marginal kernel $K$.  Compared to the
L-ensemble formula in \eqref{L_ensemble}, these may be analytically
more convenient, since they do not involve ratios or arbitrary
principal minors.

\subsection{Quality vs. diversity}
\seclabel{decomposition}

An important practical concern for modeling is interpretability; that
is, practitioners should be able to understand the parameters of the
model in an intuitive way.  While the entries of the DPP kernel are
not totally opaque in that they can be seen as measures of
similarity---reflecting our primary qualitative characterization of
DPPs as diversifying processes---in most practical situations we want
diversity to be balanced against some underlying preferences for
different items in $\Y$.  In this section, we propose a decomposition
of the DPP that more directly illustrates the tension between
diversity and a per-item measure of quality.

In \secref{background} we observed that the DPP kernel $L$ can be
written as a Gram matrix, $L = B^\trans B$, where the columns of $B$
are vectors representing items in the set $\Y$.  We now take this one
step further, writing each column $B_i$ as the product of a {\it
  quality} term $q_i \in\reals^+$ and a vector of normalized {\it
  diversity features} $\phi_i \in\reals^D$, $\Vert\phi_i\Vert = 1$.
(While $D = N$ is sufficient to decompose any DPP, we keep $D$
arbitrary since in practice we may wish to use high-dimensional
feature vectors.)  The entries of the kernel can now be written as
\begin{equation}
  L_{ij} = q_i \phi_i^\trans\phi_j q_j\,.
\end{equation}
We can think of $q_i \in \reals^+$ as measuring the intrinsic
``goodness'' of an item $i$, and $\phi_i^\trans\phi_j \in [-1,1]$ as a
signed measure of similarity between items $i$ and $j$.  We use the
following shorthand for similarity:
\begin{equation}
S_{ij} \equiv \phi_i^\trans\phi_j = \frac{L_{ij}}{\sqrt{L_{ii}L_{jj}}}\,.
\eqlabel{computeS}
\end{equation}

This decomposition of $L$ has two main advantages.  First, it
implicitly enforces the constraint that $L$ must be positive
semidefinite, which can potentially simplify learning (see
\secref{learning}).  Second, it allows us to independently model
quality and diversity, and then combine them into a unified model.  In
particular, we have:
\begin{equation}
\P_L(Y) \propto \left(\prod_{i\in Y} q^2_i\right) \det(S_{Y})\,,
\end{equation}
where the first term increases with the quality of the selected items,
and the second term increases with the diversity of the selected
items.  We will refer to $q$ as the {\it quality model} and $S$ or
$\phi$ as the {\it diversity model}.  Without the diversity model, we
would choose high-quality items, but we would tend to choose similar
high-quality items over and over.  Without the quality model, we would
get a very diverse set, but we might fail to include the most
important items in $\Y$, focusing instead on low-quality outliers.  By
combining the two models we can achieve a more balanced result.

Returning to the geometric intuitions from \secref{geometry}, the
determinant of $L_Y$ is equal to the squared volume of the
parallelepiped spanned by the vectors $q_i\phi_i$ for $i \in Y$.  The
magnitude of the vector representing item $i$ is $q_i$, and its
direction is $\phi_i$.  \figref{geometry2} (reproduced from the
previous section) now makes clear how DPPs decomposed in this way
naturally balance the two objectives of high quality and high
diversity.  Going forward, we will nearly always assume that our
models are decomposed into quality and diversity components.  This
provides us not only with a natural and intuitive setup for real-world
applications, but also a useful perspective for comparing DPPs with
existing models, which we turn to next.

\begin{figure}
\centering
\includegraphics[width=5in]{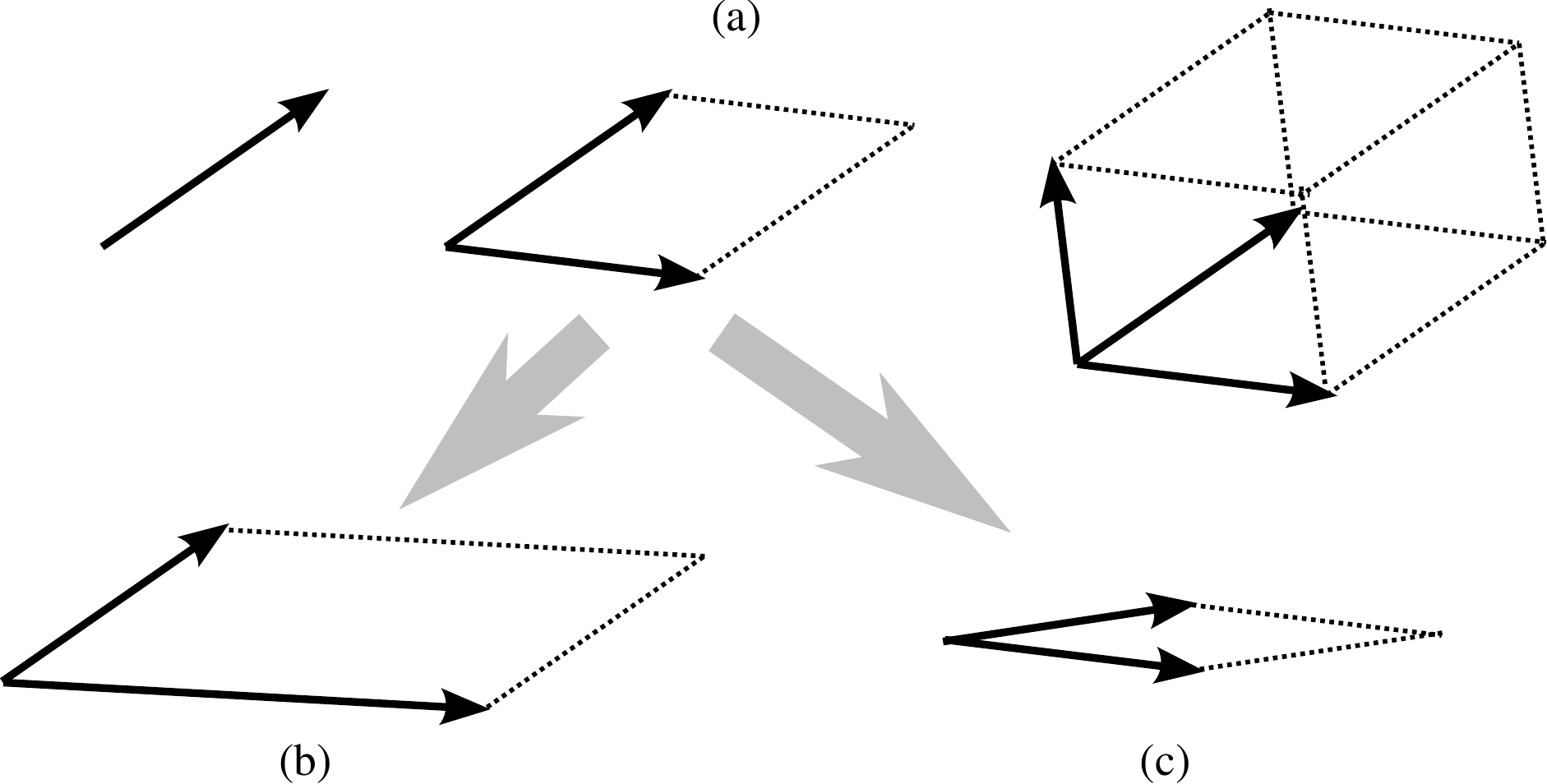}
\caption[Revisiting DPP geometry] 
  {Revisiting DPP geometry: (a) The probability of a subset $Y$ is the
  square of the volume spanned by $q_i\phi_i$ for $i\in Y$.  (b) As
  item $i$'s quality $q_i$ increases, so do the probabilities of sets
  containing item $i$.  (c) As two items $i$ and $j$ become more
  similar, $\phi_i^\trans\phi_j$ increases and the probabilities of
  sets containing both $i$ and $j$ decrease.}  \figlabel{geometry2}
\end{figure}

\subsection{Expressive power}
\seclabel{expressiveness}

Many probabilistic models are known and widely used within the machine
learning community.  A natural question, therefore, is what advantages
DPPs offer that standard models do not.  We have seen already how a
large variety of inference tasks, like sampling and conditioning, can
be performed efficiently for DPPs; however efficiency is essentially a
prerequisite for any practical model.  What makes DPPs particularly
unique is the marriage of these computational advantages with the
ability to express {\it global, negative} interactions between
modeling variables; this repulsive domain is notoriously intractable
using traditional approaches like graphical models
\citep{murphy1999loopy,boros2002pseudo,ishikawa2003exact,taskar2004learning,yanover2002approximate,yanover2006linear,kulesza2008structured}.
In this section we elaborate on the expressive powers of DPPs and
compare them with those of Markov random fields, which we take as
representative graphical models.

\subsubsection{Markov random fields}
\seclabel{mrfs}

A Markov random field (MRF) is an undirected graphical model defined
by a graph $G$ whose nodes $1,2,\dots,N$ represent random variables.
For our purposes, we will consider binary MRFs, where each output
variable takes a value from $\{0,1\}$.  We use $y_i$ to denote a value
of the $i$th output variable, bold $\y_c$ to denote an assignment to a
set of variables $c$, and $\y$ for an assignment to all of the output
variables.  The graph edges $E$ encode direct dependence relationships
between variables; for example, there might be edges between similar
elements $i$ and $j$ to represent the fact that they tend not to
co-occur.  MRFs are often referred to as {\it conditional random
  fields} when they are parameterized to depend on input data, and
especially when $G$ is a chain \citep{lafferty2001conditional}.

An MRF defines a joint probability distribution over the output
variables that factorizes across the cliques $\C$ of $G$:
\begin{equation}
  \P(\y) = \frac{1}{Z} \prod_{c \in \C} \psi_c(\y_c)\,.
  \eqlabel{mrfdist}
\end{equation}
Here each $\psi_c$ is a potential function that assigns a nonnegative
value to every possible assignment $\y_c$ of the clique $c$, and $Z$
is the normalization constant $\sum_{\y'} \prod_{c \in \C}
\psi_c(\y'_c)$.  Note that, for a binary MRF, we can think of $\y$ as
the characteristic vector for a subset $Y$ of $\Y = \{1,2,\dots,N\}$.
Then the MRF is equivalently the distribution of a random subset
$\bY$, where $\P(\bY = Y)$ is equivalent to $\P(\y)$.

The Hammersley-Clifford theorem states that $\P(\y)$ defined in
\eqref{mrfdist} is always Markov with respect to $G$; that is, each
variable is conditionally independent of all other variables given its
neighbors in $G$.  The converse also holds: any distribution that is
Markov with respect to $G$, as long as it is strictly positive, can be
decomposed over the cliques of $G$ as in \eqref{mrfdist}
\citep{grimmett1973theorem}.  MRFs therefore offer an intuitive way to
model problem structure.  Given domain knowledge about the nature of
the ways in which outputs interact, a practitioner can construct a
graph that encodes a precise set of conditional independence
relations.  (Because the number of unique assignments to a clique $c$
is exponential in $|c|$, computational constraints generally limit us
to small cliques.)

For comparison with DPPs, we will focus on {\it pairwise} MRFs, where
the largest cliques with interesting potential functions are the
edges; that is, $\psi_c(\y_c) = 1$ for all cliques $c$ where $|c|>2$.
The pairwise distribution is
\begin{equation}
  \P(\y) = \frac{1}{Z} \prod_{i=1}^N \psi_i(y_i) 
  \prod_{ij \in E} \psi_{ij}(y_i,y_j)\,.
  \eqlabel{pairwisemrfdist}
\end{equation}
We refer to $\psi_i(y_i)$ as node potentials, and $\psi_{ij}(y_i,y_j)$
as edge potentials.

MRFs are very general models---in fact, if the cliques are unbounded
in size, they are fully general---but inference is only tractable in
certain special cases.  \citet{cooper1990computational} showed that
general probabilistic inference (conditioning and marginalization) in
MRFs is NP-hard, and this was later extended by
\citet{dagum1993approximating}, who showed that inference is NP-hard
even to approximate.  \citet{shimony1994finding} proved that the MAP
inference problem (finding the mode of an MRF) is also NP-hard, and
\citet{abdelbar1998approximating} showed that the MAP problem is
likewise hard to approximate.  In contrast, we showed in
\secref{background} that DPPs offer efficient exact probabilistic
inference; furthermore, although the MAP problem for DPPs is NP-hard,
it can be approximated to a constant factor under cardinality
constraints in polynomial time.

The first tractable subclass of MRFs was identified by
\citet{pearl1982reverend}, who showed that belief propagation can be
used to perform inference in polynomial time when $G$ is a tree.  More
recently, certain types of inference in binary MRFs with {\it
  associative} (or submodular) potentials $\psi$ have been shown to be
tractable
\citep{boros2002pseudo,taskar2004learning,kolmogorov2004energy}.
Inference in non-binary associative MRFs is NP-hard, but can be
efficiently approximated to a constant factor depending on the size of
the largest clique \citep{taskar2004learning}.  Intuitively, an edge
potential is called associative if it encourages the endpoint nodes
take the same value (e.g., to be both in or both out of the set $Y$).
More formally, associative potentials are at least one whenever the
variables they depend on are all equal, and exactly one otherwise.  

We can rewrite the pairwise, binary MRF of \eqref{pairwisemrfdist} in
a canonical log-linear form:
\begin{equation}
  \P(\y) \propto \exp\left(\sum_{i} w_{i}y_{i} +\sum_{ij \in E}
  w_{ij}y_{i}y_{j}\right)\,.  
  \eqlabel{canonicalmrf}
\end{equation}
Here we have eliminated redundancies by forcing $\psi_i(0) = 1$,
$\psi_{ij}(0,0) = \psi_{ij}(0,1) = \psi_{ij}(1,0) = 1$, and setting
$w_i = \log\psi_i(1)$, $w_{ij} = \log\psi_{ij}(1,1)$.  This
parameterization is sometimes called the fully visible Boltzmann
machine.  Under this representation, the MRF is associative whenever
$w_{ij} \geq 0$ for all $ij\in E$.

We have seen that inference in MRFs is tractable when we restrict the
graph to a tree or require the potentials to encourage agreement.
However, the repulsive potentials necessary to build MRFs exhibiting
diversity are the opposite of associative potentials (since they imply
$w_{ij} < 0$), and lead to intractable inference for general graphs.
Indeed, such negative potentials can create ``frustrated cycles'',
which have been used both as illustrations of common MRF inference
algorithm failures \citep{kulesza2008structured} and as targets for
improving those algorithms \citep{sontag2007new}.  A wide array of
(informally) approximate inference algorithms have been proposed to
mitigate tractability problems, but none to our knowledge effectively
and reliably handles the case where potentials exhibit strong
repulsion.

\subsubsection{Comparing DPPs and MRFs}
\seclabel{dppsvsmrfs}

Despite the computational issues outlined in the previous section,
MRFs are popular models and, importantly, intuitive for practitioners,
both because they are familiar and because their potential functions
directly model simple, local relationships.  We argue that DPPs have a
similarly intuitive interpretation using the decomposition in
\secref{decomposition}.  Here, we compare the distributions
realizable by DPPs and MRFs to see whether the tractability of DPPs
comes at a large expressive cost.

Consider a DPP over $\Y = \{1,2,\dots,N\}$ with $N \times N$ kernel
matrix $L$ decomposed as in \secref{decomposition}; we have
\begin{equation}
  \P_L(Y) \propto \det(L_Y) = \left(\prod_{i\in Y} q^2_i\right)
  \det(S_{Y})\,.
\end{equation}
The most closely related MRF is a pairwise, complete graph on $N$
binary nodes with negative interaction terms.  We let $y_i = \I(i \in
Y)$ be indicator variables for the set $Y$, and write the MRF in the
log-linear form of \eqref{canonicalmrf}:
\begin{equation}
  \P_\mrf(Y) \propto \exp\left(\sum_{i} w_{i}y_{i} +\sum_{i < j}
  w_{ij}y_{i}y_{j}\right)\,,
  \eqlabel{mrfdiverse}
\end{equation}
where $w_{ij} \leq 0$. 

Both of these models can capture negative correlations between
indicator variables $y_{i}$.  Both models also have $\frac{N(N+1)}{2}$
parameters: the DPP has quality scores $q_i$ and similarity measures
$S_{ij}$, and the MRF has node log-potentials $w_i$ and edge
log-potentials $w_{ij}$.  The key representational difference is that,
while $w_{ij}$ are individually constrained to be nonpositive, the
positive semidefinite constraint on the DPP kernel is global.  One
consequence is that, as a side effect, the MRF can actually capture
certain limited positive correlations; for example, a 3-node MRF with
$w_{12}, w_{13} < 0$ and $w_{23}=0$ induces a positive correlation
between nodes two and three by virtue of their mutual disagreement
with node one.  As we have seen in \secref{background}, the
semidefinite constraint prevents the DPP from forming any positive
correlations.

More generally, semidefiniteness means that the DPP diversity feature
vectors must satisfy the triangle inequality, leading to
\begin{equation}
  \sqrt{1-S_{ij}}+\sqrt{1-S_{jk}} \geq \sqrt{1-S_{ik}}
  \eqlabel{transitivity}
\end{equation}
for all $i,j,k \in \Y$ since $\Vert \phi_i - \phi_j \Vert \propto
\sqrt{1-S_{ij}}$.  The similarity measure therefore obeys a type of
transitivity, with large $S_{ij}$ and $S_{jk}$ implying large
$S_{ik}$.

\eqref{transitivity} is not, by itself, sufficient to guarantee that
$L$ is positive semidefinite, since $S$ must also be realizable using
unit length feature vectors.  However, rather than trying to develop
further intuition algebraically, we turn to visualization.  While it
is difficult to depict the feasible distributions of DPPs and MRFs in
high dimensions, we can get a feel for their differences even with a
small number of elements $N$.

When $N=2$, it is easy to show that the two models are equivalent, in
the sense that they can both represent any distribution with negative
correlations:
\begin{equation}
\P(y_{1}=1)\P(y_{2}=1) \ge \P(y_{1}=1,y_{2}=1)\,.
\end{equation}

When $N=3$, differences start to become apparent.  In this setting
both models have six parameters: for the DPP they are
$(q_{1},q_{2},q_{3},S_{12},S_{13},S_{23})$, and for the MRF they are
$(w_{1},w_{2},w_{3},w_{12},w_{13},w_{23})$.  To place the two models
on equal footing, we represent each as the product of unnormalized
per-item potentials $\psi_{1},\psi_{2},\psi_{3}$ and a single
unnormalized ternary potential $\psi_{123}$.  This representation
corresponds to a factor graph with three nodes and a single, ternary
factor (see \figref{factor}).  The probability of a set $Y$ is then
given by
\begin{equation}
  \P(Y) \propto
  \psi_{1}(y_1)\psi_2(y_2)\psi_3(y_3)\psi_{123}(y_1,y_2,y_3)\,.
\end{equation}
For the DPP, the node potentials are $\psi_{i}^\dpp(y_{i}) =
q_{i}^{2y_{i}}$, and for the MRF they are $\psi_{i}^\mrf(y_{i}) =
e^{w_{i}y_{i}}$.  The ternary factors are
\begin{align}
  \psi_{123}^\dpp(y_1,y_{2},y_{3}) &= 
  \det(S_Y)\,,\\
  \psi_{123}^\mrf(y_1,y_{2},y_{3}) &= 
  \exp\left(\sum_{i<j}w_{ij}y_{i}y_{j}\right)\,.
\end{align}

\begin{figure}
  \centering
  \includegraphics[width=1in]{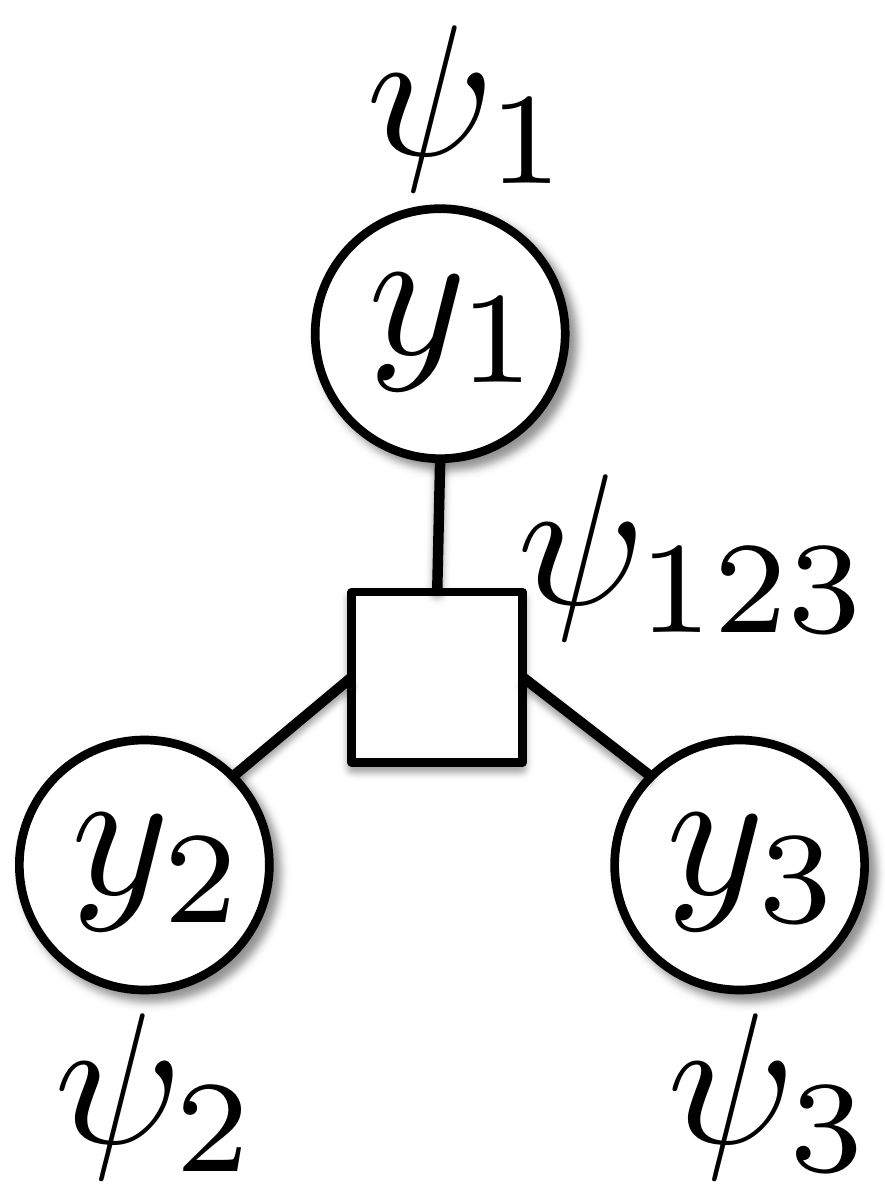}
  \caption[Factor graph representation of 3-item MRF or DPP]
      {A factor graph representation of a 3-item MRF or DPP.} 
  \figlabel{factor}
\end{figure}

\begin{table}
  \centering
  \renewcommand{\arraystretch}{1.2}
  \begin{tabular}{ccccc}
    $Y$ & $y_{1}y_{2}y_{3}$& $\psi_{123}^\mrf$ & $\psi_{123}^\dpp$ \\
    \hline
    \{\} & 000 & 1 & 1\\
    \{1\} & 100 & 1 & 1\\
    \{2\} & 010 & 1 & 1\\
    \{3\} & 001 & 1 & 1\\
    \{1,2\} & 110 & $e^{w_{12}}$ & $1-S_{12}^{2}$ \\
    \{1,3\} & 101 & $e^{w_{13}}$ & $1-S_{13}^{2}$\\
    \{2,3\} & 011 & $e^{w_{23}}$ & $1-S_{23}^{2}$\\
    \{1,2,3\} & 111 & $e^{w_{12}+w_{13}+w_{23}}$ & $1+2S_{12}S_{13}S_{23}- S_{12}^{2}-S_{13}^{2}-S_{23}^{2}$ \\
    \hline
  \end{tabular}
  \caption[Ternary factors for 3-item MRFs and DPPs]
          {Values of ternary factors for 3-item MRFs and DPPs.}
  \tablabel{factor}
\end{table}

Since both models allow setting the node potentials arbitrarily, we
focus now on the ternary factor.  \tabref{factor} shows the values of
$\psi_{123}^\dpp$ and $\psi_{123}^\mrf$ for all subsets $Y \subseteq
\Y$.  The last four entries are determined, respectively, by the three
edge parameters of the MRF and three similarity parameters $S_{ij}$ of
the DPP, so the sets of realizable ternary factors form 3-D manifolds
in 4-D space.  We attempt to visualize these manifolds by showing 2-D
slices in 3-D space for various values of $\psi_{123}(1,1,1)$ (the
last row of \tabref{factor}).

\figref{dpp_slices} depicts four such slices of the realizable DPP
distributions, and \figref{mrf_slices} shows the same slices of the
realizable MRF distributions.  Points closer to the origin (on the
lower left) correspond to ``more repulsive'' distributions, where the
three elements of $\Y$ are less likely to appear together.  When
$\psi_{123}(1,1,1)$ is large (gray surfaces), negative correlations are
weak and the two models give rise to qualitatively similar
distributions.  As the value of the $\psi_{123}(1,1,1)$ shrinks to zero
(red surfaces), the two models become quite different.  MRFs, for
example, can describe distributions where the first item is strongly
anti-correlated with both of the others, but the second and third are
not anti-correlated with each other.  The transitive nature of the DPP
makes this impossible.

To improve visibility, we have constrained $S_{12},S_{13},S_{23} \geq
0$ in \figref{dpp_slices}.  \figref{dpp_slice} shows a single slice
without this constraint; allowing negative similarity makes it
possible to achieve strong three-way repulsion with less pairwise
repulsion, closing the surface away from the origin.  The
corresponding MRF slice is shown in \figref{mrf_slice}, and the two
are overlaid in \figref{overlay} and \figref{overlay2}.  Even though
there are relatively strong interactions in these plots
($\psi_{123}(1,1,1) = 0.1$), the models remain roughly comparable in
terms of the distributions they can express.

\begin{figure}
\centering
\subfloat[DPP slices (partial)]{
\includegraphics[trim=0 0.3in 0 0.3in,clip=true,width=2.5in]{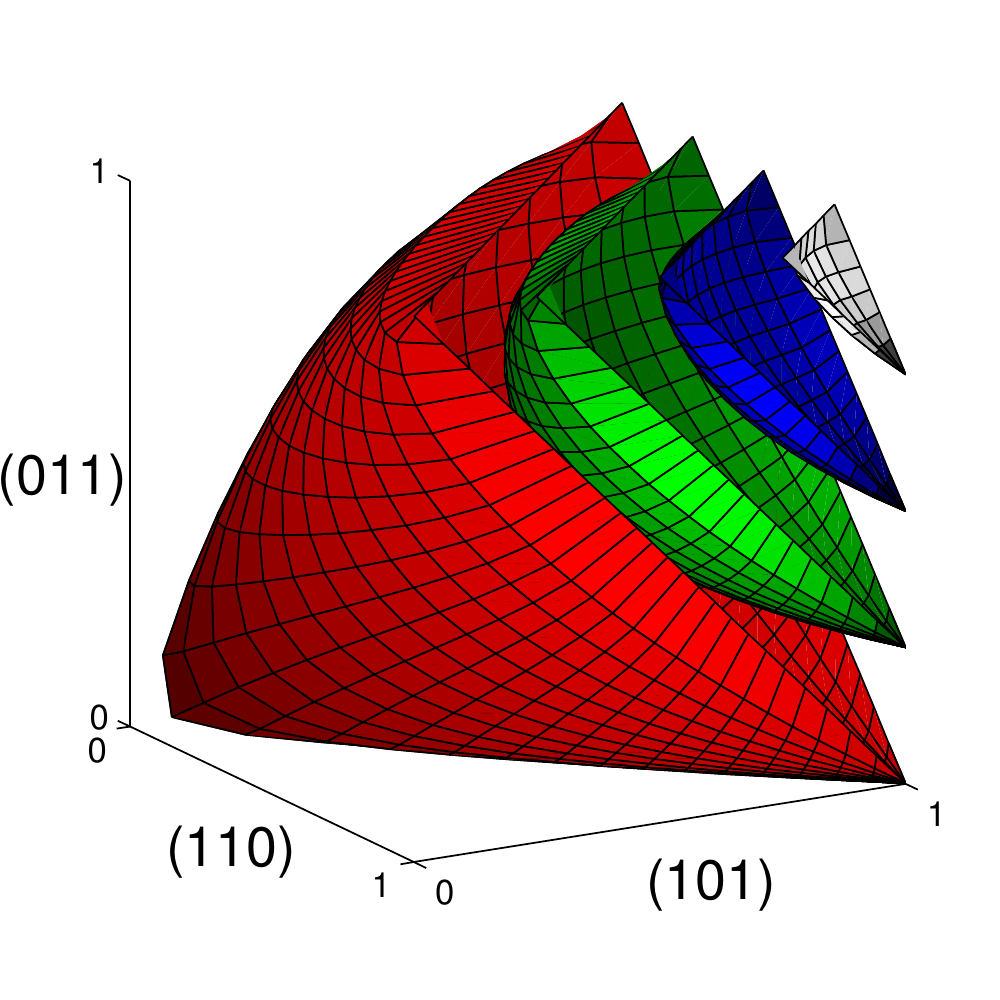}
\figlabel{dpp_slices}}
\subfloat[MRF slices]{
\includegraphics[trim=0 0.3in 0 0.3in,clip=true,width=2.5in]{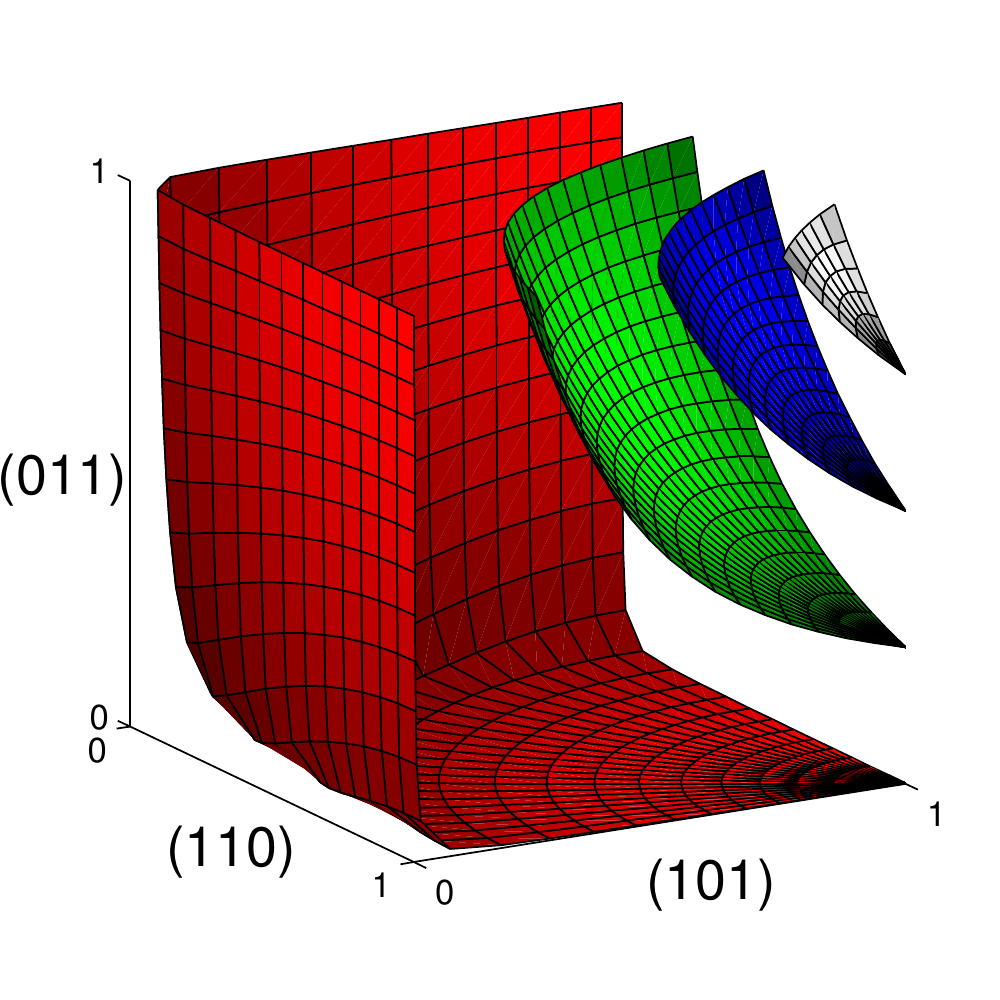}
\figlabel{mrf_slices}}\\
\subfloat[DPP slice (full) for $\psi_{123}(1,1,1) = 0.1$]{
\includegraphics[trim=0 0.3in 0 0.3in,clip=true,width=2.5in]{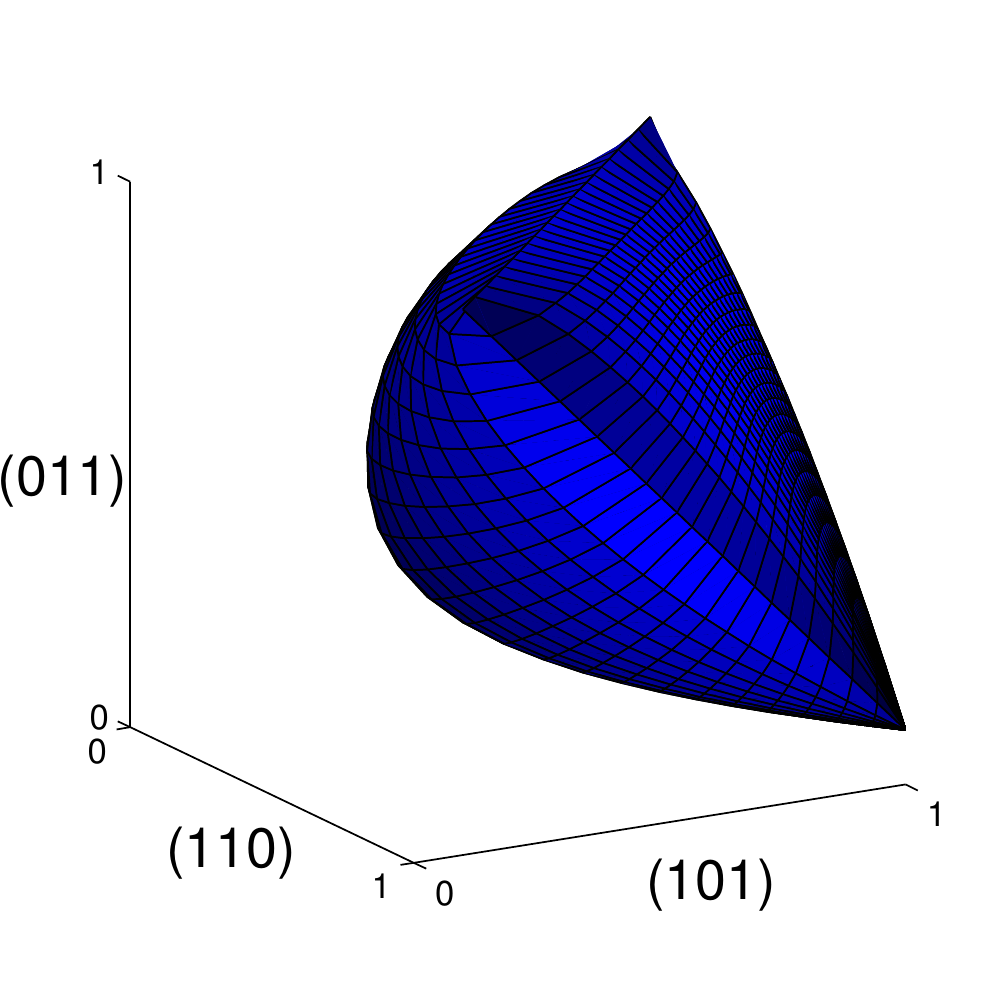}
\figlabel{dpp_slice}}
\subfloat[MRF slice for $\psi_{123}(1,1,1) = 0.1$]{
\includegraphics[trim=0 0.3in 0 0.3in,clip=true,width=2.5in]{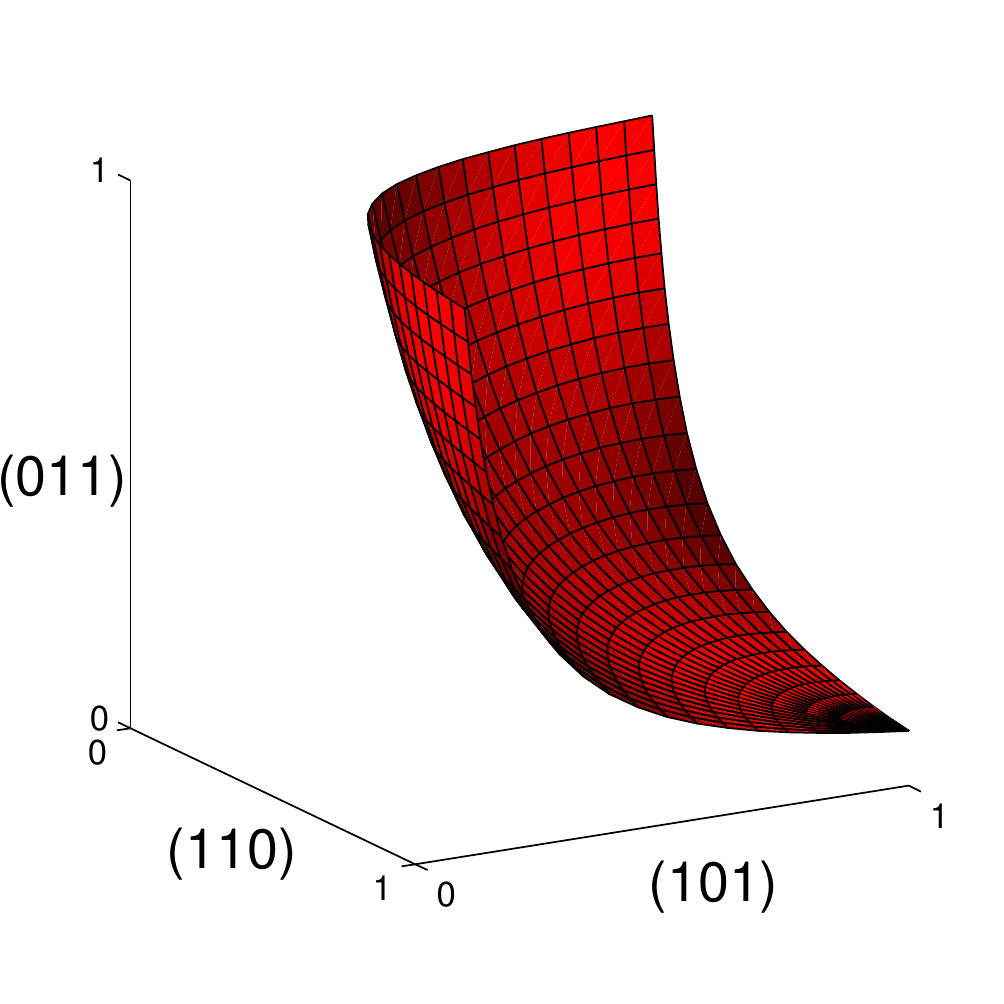}
\figlabel{mrf_slice}}\\
\subfloat[Overlay]{
\includegraphics[trim=0 0.3in 0 0.3in,clip=true,width=2.5in]{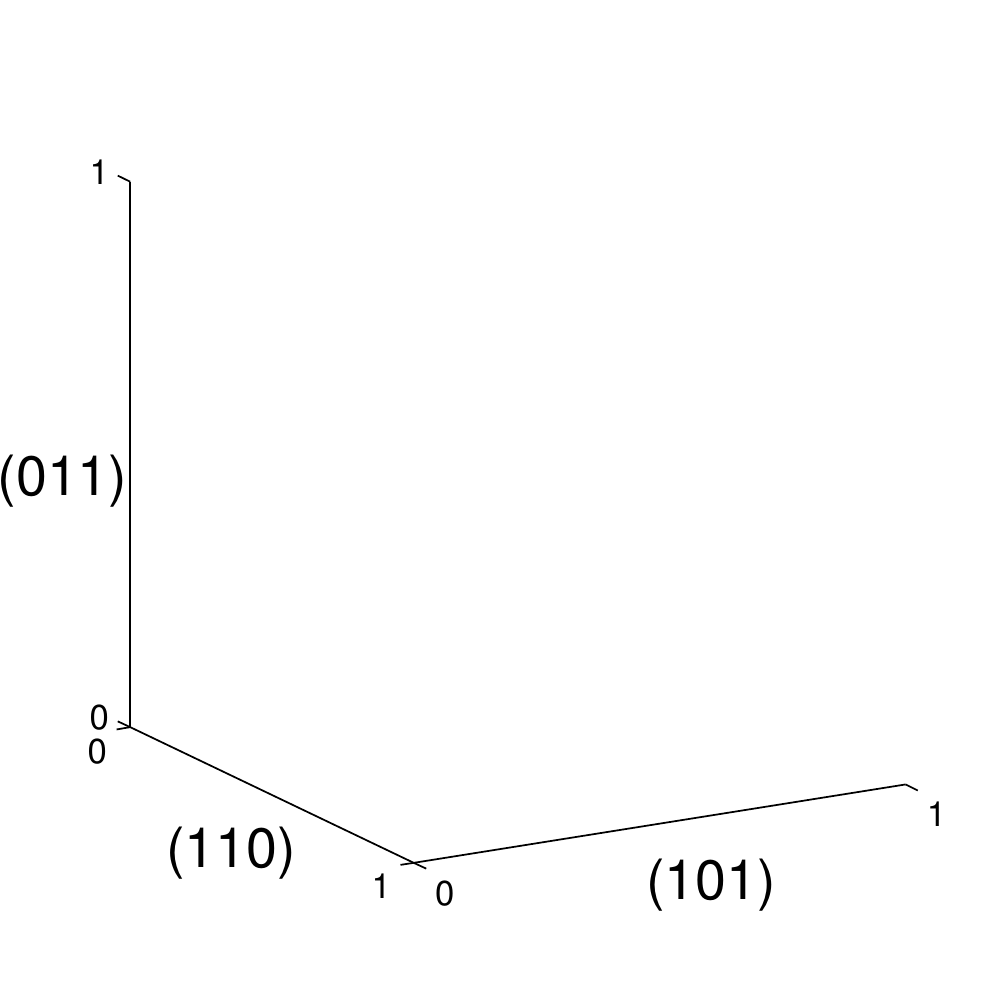}
\hspace{-1.72in}\raisebox{0.5in}{
  \includegraphics[width=1.4in]{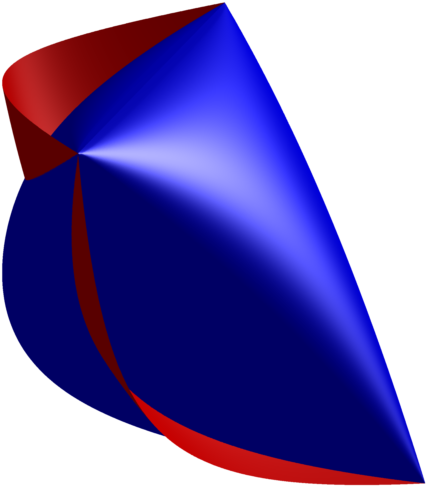}
}
\figlabel{overlay}}
\hspace{0.2in}
\subfloat[Overlay (rotated)]{
\includegraphics[trim=0 0.3in 0 0.3in,clip=true,width=2.5in]{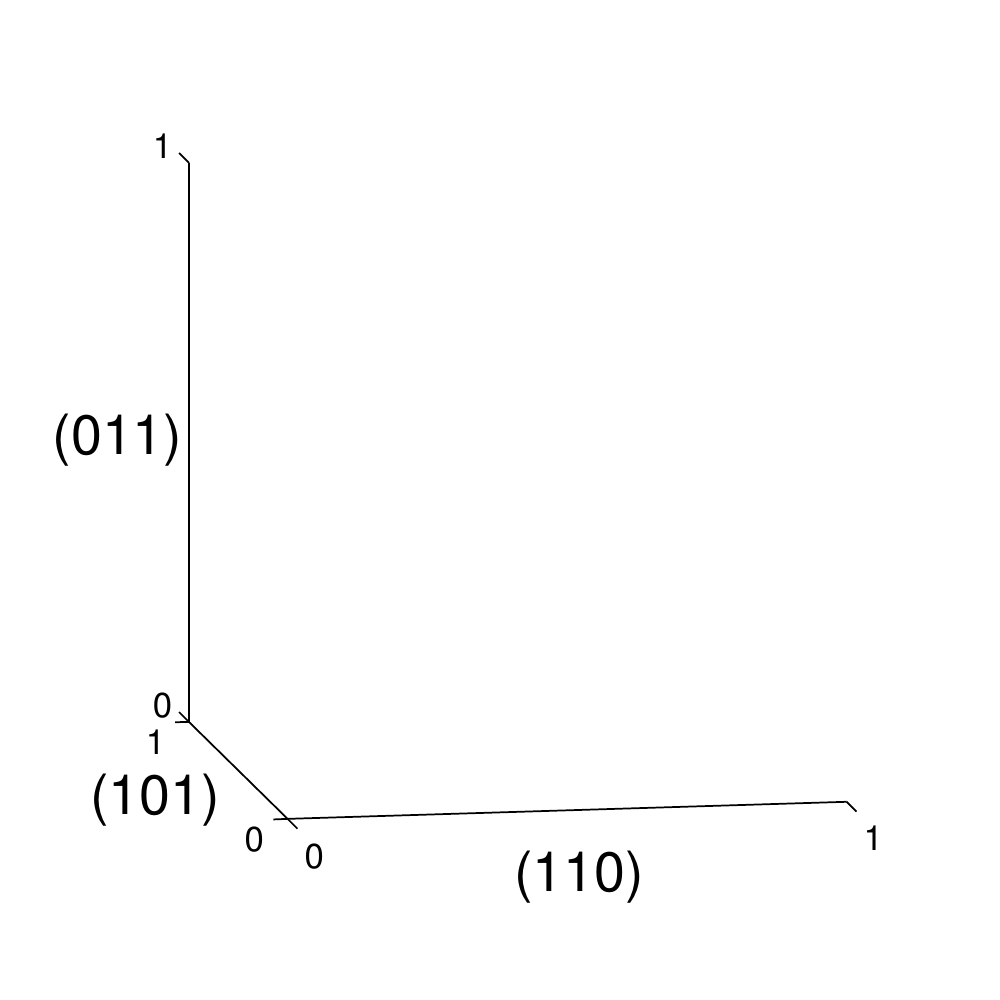}
\hspace{-1.98in}\raisebox{0.65in}{
  \includegraphics[width=1.5in]{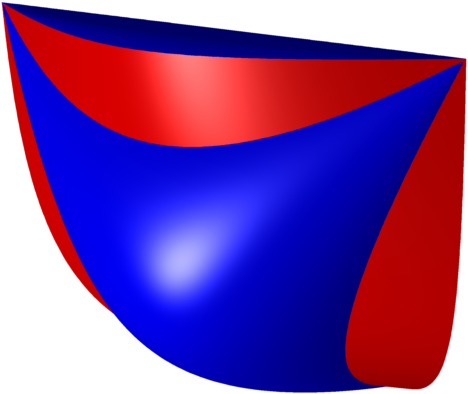}
}
\figlabel{overlay2}}
\hspace{0.3in}
\caption[Realizable distributions for 3-item MRFs and DPPs]
  {(a,b) Realizable values of $\psi_{123}(1,1,0)$,
  $\psi_{123}(1,0,1)$, and $\psi_{123}(0,1,1)$ in a 3-factor when
  $\psi_{123}(1,1,1) = 0.001$ (red), $0.25$ (green), $0.5$ (blue), and
  $0.75$ (grey). (c,d) Surfaces for $\psi_{123}(1,1,1)=0.1$, allowing
  negative similarity for the DPP.  (e,f) DPP (blue) and MRF (red)
  surfaces superimposed.}  \figlabel{cubics}
\end{figure}

As $N$ gets larger, we conjecture that the story is essentially the
same.  DPPs are primarily constrained by a notion of transitivity on
the similarity measure; thus it would be difficult to use a DPP to
model, for example, data where items repel ``distant'' items rather
than similar items---if $i$ is far from $j$ and $j$ is far from $k$ we
cannot necessarily conclude that $i$ is far from $k$.  One way of
looking at this is that repulsion of distant items induces positive
correlations between the selected items, which a DPP cannot represent.

MRFs, on the other hand, are constrained by their local nature and do
not effectively model data that are ``globally'' diverse.  For
instance, a pairwise MRF we cannot exclude a set of three or more
items without excluding some pair of those items.  More generally, an
MRF assumes that repulsion does not depend on (too much) context, so
it cannot express that, say, there can be only a certain number of
selected items overall.  The DPP can naturally implement this kind of
restriction though the rank of the kernel.

\subsection{Dual representation}
\seclabel{dual_dpps}

The standard inference algorithms for DPPs rely on manipulating the
kernel $L$ through inversion, eigendecomposition, and so on.  However,
in situations where $N$ is large we may not be able to work
efficiently with $L$---in some cases we may not even have the memory
to write it down.  In this section, instead, we develop a dual
representation of a DPP that shares many important properties with the
kernel $L$ but is often much smaller.  Afterwards, we will show how
this dual representation can be applied to perform efficient
inference.

Let $B$ be the $D \times N$ matrix whose columns are given by $B_i =
q_i\phi_i$, so that $L=B^\trans B$.  Consider now the matrix
\begin{equation}
  C = BB^\trans\,.
  \eqlabel{dualmatrix}
\end{equation}
By construction, $C$ is symmetric and positive semidefinite.  In
contrast to $L$, which is too expensive to work with when $N$ is
large, $C$ is only $D\times D$, where $D$ is the dimension of the
diversity feature function $\phi$.  In many practical situations, $D$
is fixed by the model designer, while $N$ may grow without bound as
new items become available; for instance, a search engine may
continually add to its database of links.  Furthermore, we have the
following result.

\begin{proposition}
  The nonzero eigenvalues of $C$ and $L$ are identical, and the
  corresponding eigenvectors are related by the matrix $B$.  That is,
  \begin{equation}
    C = \sum_{n=1}^D \lambda_n \cv_n \cv_n^\trans
  \end{equation}
  is an eigendecomposition of $C$ if and only if
  \begin{equation}
    L = \sum_{n=1}^D \lambda_n 
    \left(\frac{1}{\sqrt{\lambda_n}}B^\trans \cv_n\right) 
    \left(\frac{1}{\sqrt{\lambda_n}}B^\trans \cv_n\right)^\trans
  \end{equation}
  is an eigendecomposition of $L$.
  \proplabel{dualkernel}
\end{proposition}
\begin{proof}
  In the forward direction, we assume that
  $\{(\lambda_n,\cv_n)\}_{n=1}^D$ is an eigendecomposition of $C$.  We
  have
  \begin{align}
    \sum_{n=1}^D \lambda_n 
    \left(\frac{1}{\sqrt{\lambda_n}}B^\trans \cv_n\right) 
    \left(\frac{1}{\sqrt{\lambda_n}}B^\trans \cv_n\right)^\trans
    &= B^\trans \left(\sum_{n=1}^D \cv_n \cv_n^\trans\right) B\\
    &= B^\trans B = L\,,
  \end{align}
  since $\cv_n$ are orthonormal by assumption.  Furthermore, for any
  $n$ we have
  \begin{align}
    \left\Vert \frac{1}{\sqrt{\lambda_n}}B^\trans \cv_n \right\Vert^2
    &= \frac{1}{\lambda_n} (B^\trans \cv_n)^\trans(B^\trans \cv_n)\\
    &= \frac{1}{\lambda_n} \cv_n^\trans C \cv_n\\
    &= \frac{1}{\lambda_n} \lambda_n \Vert \cv_n \Vert\\
    &= 1\,,
  \end{align}
  using the fact that $C\cv_n = \lambda_n\cv_n$ since $\cv_n$ is an
  eigenvector of $C$.  Finally, for any distinct $1 \leq a,b \leq D$,
  we have
  \begin{align}
    \left(\frac{1}{\sqrt{\lambda_a}}B^\trans \cv_a\right)^\trans
    \left(\frac{1}{\sqrt{\lambda_b}}B^\trans \cv_b\right)
    &= \frac{1}{\sqrt{\lambda_a \lambda_b}} \cv_a^\trans C \cv_b\\
    &= \frac{\sqrt{\lambda_b}}{\sqrt{\lambda_a}} \cv_a^\trans \cv_b\\
    &= 0\,.
  \end{align}
  Thus $\left\{\left(\lambda_n, \frac{1}{\sqrt{\lambda_n}}B^\trans
  \cv_n\right)\right\}_{n=1}^D$ is an eigendecomposition of $L$.  In
  the other direction, an analogous argument applies once we observe
  that, since $L = B^\trans B$, $L$ has rank at most $D$ and therefore
  at most $D$ nonzero eigenvalues.
\end{proof}

\noindent \propref{dualkernel} shows that $C$ contains quite a bit of
information about $L$.  In fact, $C$ is sufficient to perform nearly
all forms of DPP inference efficiently, including normalization and
marginalization in constant time with respect to $N$, and sampling in
time linear in $N$.

\subsubsection{Normalization}

Recall that the normalization constant for a DPP is given by
$\det(L+I)$.  If $\lambda_1,\lambda_2,\dots,\lambda_N$ are the
eigenvalues of $L$, then the normalization constant is equal to
$\prod_{n=1}^N (\lambda_n + 1)$, since the determinant is the product
of the eigenvalues of its argument.  By \propref{dualkernel}, the
nonzero eigenvalues of $L$ are also the eigenvalues of the dual
representation $C$.  Thus, we have
\begin{equation}
  \det(L+I) = \prod_{n=1}^D (\lambda_n + 1) = \det(C+I)\,.
\end{equation}
Computing the determinant of $C+I$ requires $O(D^{\omega})$ time.

\subsubsection{Marginalization}

Standard DPP marginalization makes use of the marginal kernel $K$,
which is of course as large as $L$.  However, the dual representation
$C$ can be used to compute the entries of $K$.  We first
eigendecompose the dual representation as $C = \sum_{n=1}^D \lambda_n
\cv_n\cv_n^\trans$, which requires $O(D^{\omega})$ time.  Then, we can
use the definition of $K$ in terms of the eigendecomposition of $L$ as
well as \propref{dualkernel} to compute
\begin{align}
  K_{ii} &= \sum_{n=1}^D \frac{\lambda_n}{\lambda_n+1} 
  \left(\frac{1}{\sqrt{\lambda_n}}B_i^\trans \cv_n\right)^2\\
  &= q^2_i\sum_{n=1}^D \frac{1}{\lambda_n+1} 
  (\phi_i^\trans\cv_n)^2 \,.
  \eqlabel{dual_marginal1}
\end{align}
That is, the diagonal entries of $K$ are computable from the dot
products between the diversity features $\phi_i$ and the eigenvectors
of $C$; we can therefore compute the marginal probability of a single
item $i \in \Y$ from an eigendecomposition of $C$ in $O(D^2)$ time.
Similarly, given two items $i$ and $j$ we have
\begin{align}
  K_{ij} &= \sum_{n=1}^D \frac{\lambda_n}{\lambda_n+1} 
  \left(\frac{1}{\sqrt{\lambda_n}}B_i^\trans \cv_n\right)
  \left(\frac{1}{\sqrt{\lambda_n}}B_j^\trans \cv_n\right)\\
  &= q_i q_j \sum_{n=1}^D \frac{1}{\lambda_n+1} 
  (\phi_i^\trans \cv_n) (\phi_j^\trans \cv_n)\,,
  \eqlabel{dual_marginal2}
\end{align}
so we can compute arbitrary entries of $K$ in $O(D^2)$ time.  This
allows us to compute, for example, pairwise marginals $\P(i,j\in
\bY) = K_{ii}K_{jj} - K_{ij}^2$.  More generally, for a set $A \in
\Y$, $|A|=k$, we need to compute $\frac{k(k+1)}{2}$ entries of $K$ to
obtain $K_A$, and taking the determinant then yields $\P(A \subseteq
\bY)$.  The process requires only $O(D^2k^2 + k^\omega)$ time; for
small sets $|A|$ this is just quadratic in the dimension of $\phi$.

\subsubsection{Sampling}
\seclabel{dual_sampling}

Recall the DPP sampling algorithm, which is reproduced for convenience
in \algref{dpp_sampling2}.  We will show that this algorithm can be
implemented in a tractable manner by using the dual representation
$C$.  The main idea is to represent $V$, the orthonormal set of
vectors in $\reals^N$, as a set $\cV$ of vectors in $\reals^D$, with
the mapping
\begin{equation}
  V = \left\{B^\trans \cv \mid \cv \in \cV\right\}\,.
  \eqlabel{CVtoV}
\end{equation}
Note that, when $\cV$ contains eigenvectors of $C$, this is (up to
scale) the relationship established by \propref{dualkernel} between
eigenvectors $\cv$ of $C$ and eigenvectors $\v$ of $L$.

\begin{algorithm}[tb]
\begin{algorithmic}
  \STATE {\bfseries Input:} eigendecomposition
    $\{(\v_n,\lambda_n)\}_{n=1}^N$ of $L$\\
  \STATE $J \leftarrow \emptyset$
  \FOR{$n = 1,2,\dots,N$}
  \STATE $J \leftarrow J \cup \{n\}$ with prob. $\frac{\lambda_n}{\lambda_n+1}$
  \ENDFOR\\
  \STATE $V \leftarrow \{\v_n\}_{n\in J}$
  \STATE $Y \leftarrow \emptyset$
  \WHILE{$|V|>0$}
  \STATE Select $i$ from $\Y$ with $\Pr(i) = \frac{1}{|V|}\sum_{\v\in V} 
    (\v^\trans \e_i)^2$
  \STATE $Y \leftarrow Y \cup i$
  \STATE $V \leftarrow V_\bot$, an orthonormal basis for the subspace 
    of $V$ orthogonal to $\e_i$
  \ENDWHILE
  \STATE {\bfseries Output:} $Y$
\end{algorithmic}
\caption[]{Sampling from a DPP}
\alglabel{dpp_sampling2}
\end{algorithm}

The mapping in \eqref{CVtoV} has several useful properties.  If $\v_1
= B^\trans\cv_1$ and $\v_2 = B^\trans\cv_2$, then $\v_1+\v_2 =
B^\trans(\cv_1+\cv_2)$, and likewise for any arbitrary linear
combination.  In other words, we can perform implicit scaling and
addition of the vectors in $V$ using only their preimages in $\cV$.
Additionally, we have
\begin{align}
  \v_1^\trans\v_2 &= (B^\trans\cv_1)^\trans(B^\trans\cv_2)\\
  &= \cv_1^\trans C \cv_2\,,
\end{align}
so we can compute dot products of vectors in $V$ in $O(D^2)$ time.
This means that, for instance, we can implicitly normalize $\v =
B^\trans\cv$ by updating $\cv \leftarrow \frac{\cv}{\sqrt{\cv^\trans C\cv}}$.

We now show how these operations allow us to efficiently implement key
parts of the sampling algorithm.  Because the nonzero eigenvalues of
$L$ and $C$ are equal, the first loop of the algorithm, where we
choose in index set $J$, remains unchanged.  Rather than using $J$ to
construct orthonormal $V$ directly, however, we instead build the set
$\cV$ by adding $\frac{\cv_n}{\sqrt{\cv_n^\trans C\cv_n}}$ for every $n\in
J$.

In the last phase of the loop, we need to find an orthonormal basis
$V_\bot$ for the subspace of $V$ orthogonal to a given $\e_i$.  This
requires two steps.  In the first, we subtract a multiple of one of
the vectors in $V$ from all of the other vectors so that they are zero
in the $i$th component, leaving us with a set of vectors spanning the
subspace of $V$ orthogonal to $\e_i$.  In order to do this we must be
able to compute the $i$th component of a vector $\v \in V$; since $\v
= B^\trans\cv$, this is easily done by computing the $i$th column of
$B$, and then taking the dot product with $\cv$.  This takes only
$O(D)$ time.  In the second step, we use the Gram-Schmidt process to
convert the resulting vectors into an orthonormal set.  This requires
a series of dot products, sums, and scalings of vectors in $V$;
however, as previously argued all of these operations can be performed
implicitly.  Therefore the mapping in \eqref{CVtoV} allows us to
implement the final line of the second loop using only tractable
computations on vectors in $\cV$.

All that remains, then, is to efficiently choose an item $i$
according to the distribution
\begin{align}
  \Pr(i) &= \frac{1}{|V|}\sum_{\v\in V} (\v^\trans \e_i)^2\\
  &= \frac{1}{|\cV|}\sum_{\cv\in \cV} ((B^\trans\cv)^\trans \e_i)^2
\end{align}
in the first line of the while loop.  Simplifying, we have
\begin{equation}
  \Pr(i) = \frac{1}{|\cV|}\sum_{\cv\in \cV} (\cv^\trans B_i)^2\,.
\end{equation}
Thus the required distribution can be computed in time $O(NDk)$, where
$k = |\cV|$.  The complete dual sampling algorithm is given in
\algref{dual_sampling}; the overall runtime is $O(NDk^2+D^2k^3)$.

\begin{algorithm}[tb]
\begin{algorithmic}
  \STATE {\bfseries Input:} eigendecomposition
    $\{(\cv_n,\lambda_n)\}_{n=1}^N$ of $C$\\
  \STATE $J \leftarrow \emptyset$
  \FOR{$n = 1,2,\dots,N$}
  \STATE $J \leftarrow J \cup \{n\}$ with prob. $\frac{\lambda_n}{\lambda_n+1}$
  \ENDFOR\\
  \STATE $\cV \leftarrow 
  \Bigl\{\frac{\cv_n}{\sqrt{\cv_n^\trans C \cv_n}}\Bigr\}_{n\in J}$
  \STATE $Y \leftarrow \emptyset$
  \WHILE{$|\cV|>0$}
  \STATE Select $i$ from $\Y$ with 
  $\Pr(i) = \frac{1}{|\cV|}\sum_{\cv\in \cV} (\cv^\trans B_i)^2$
  \STATE $Y \leftarrow Y \cup i$
  \STATE Let $\cv_0$ be a vector in $\cV$ with $B_i^\trans \cv_0 \neq 0$
  \STATE Update $\cV \leftarrow \left\{ \cv 
  - \frac{\cv^\trans B_i}{\cv_0^\trans B_i}\cv_0
  \ \vert\ \cv \in\cV - \{\cv_0\} \right\}$
  \STATE Orthonormalize $\cV$ with respect to the dot product 
  $\langle \cv_1,\cv_2 \rangle = \cv_1^\trans C \cv_2$
  \ENDWHILE
  \STATE {\bfseries Output:} $Y$
\end{algorithmic}
\caption{Sampling from a DPP (dual representation)}
\alglabel{dual_sampling}
\end{algorithm}

\subsection{Random projections}
\seclabel{projection}

As we have seen, dual DPPs allow us to deal with settings where $N$ is
too large to work efficiently with $L$ by shifting the computational
focus to the dual kernel $C$, which is only $D \times D$.  This is an
effective approach when $D \ll N$.  Of course, in some cases $D$ might
also be unmanageably large, for instance when the diversity features
are given by word counts in natural language settings, or
high-resolution image features in vision.

To address this problem, we describe a method for reducing the
dimension of the diversity features while maintaining a close
approximation to the original DPP model.  Our approach is based on
applying random projections, an extremely simple technique that
nonetheless provides an array of theoretical guarantees, particularly
with respect to preserving distances between points
\citep{vempala2004random}.  A classic result of
\citet{johnson1984extensions}, for instance, shows that
high-dimensional points can be randomly projected onto a logarithmic
number of dimensions while approximately preserving the distances
between them.  More recently, \citet{magen2008near} extended this idea
to the preservation of volumes spanned by sets of points.  Here, we
apply the connection between DPPs and spanned volumes to show that
random projections allow us to reduce the dimensionality of $\phi$,
dramatically speeding up inference, while maintaining a provably close
approximation to the original, high-dimensional model.  We begin by
stating a variant of Magen and Zouzias' result.

\begin{lemma} (Adapted from \citet{magen2008near})
  Let $X$ be a $D \times N$ matrix.  Fix $k < N$ and $0 <
  \epsilon,\delta < 1/2$, and set the projection dimension
  \begin{equation}
    d = \max\left\{\frac{2k}{\epsilon},\frac{24}{\epsilon^2}
    \left(\frac{\log(3/\delta)}{\log N} + 1\right)
    \left(\log N + 1\right) + k-1\right\}\,.
    \eqlabel{ddef}
  \end{equation}
  Let $G$ be a $d \times D$ random projection matrix whose entries are
  independently sampled from $\N(0,\frac{1}{d})$, and let $X_Y$, where
  $Y \subseteq \{1,2,\dots,N\}$, denote the $D \times |Y|$ matrix
  formed by taking the columns of $X$ corresponding to indices in $Y$.
  Then with probability at least $1 - \delta$ we have, for all $Y$
  with cardinality at most $k$:
  \begin{equation}
    (1-\epsilon)^{|Y|} \leq \frac{\vol(GX_Y)}{\vol(X_Y)} \leq
    (1+\epsilon)^{|Y|}\,,
  \end{equation}
  where $\vol(X_Y)$ is the $k$-dimensional volume of the
  parallelepiped spanned by the columns of $X_Y$.  
  \lemlabel{magen}
\end{lemma}

\noindent
\lemref{magen} says that, with high probability, randomly projecting
to
\begin{equation}
  d = O(\max\{k/\epsilon, (\log(1/\delta) + \log N) / \epsilon^2\})
\end{equation}
dimensions is sufficient to approximately preserve all volumes spanned
by $k$ columns of $X$.  We can use this result to bound the
effectiveness of random projections for DPPs.  

In order to obtain a result that is independent of $D$, we will
restrict ourselves to the portion of the distribution pertaining to
subsets $Y$ with cardinality at most a constant $k$.  This restriction
is intuitively reasonable for any application where we use DPPs to
model sets of relatively small size compared to $N$, which is common
in practice.  However, formally it may seem a bit strange, since it
implies conditioning the DPP on cardinality.  In \secref{kdpps} we
will show that this kind of conditioning is actually very practical
and efficient, and \thmref{dpp_proj}, which we prove shortly, will
apply directly to the $k$-DPPs of \secref{kdpps} without any
additional work.

For now, we will seek to approximate the distribution $\P^{\leq k}(Y)
= \P(\bY = Y\ \vert\ |\bY| \leq k)$, which is simply the original DPP
conditioned on the cardinality of the modeled subset:
\begin{equation}
  \P^{\leq k}(Y) = \frac{ \left(\prod_{i\in Y} q_i^2\right) \det(\phi(Y)^\trans
  \phi(Y)) }
  {\sum_{|Y'| \leq k} \left(\prod_{i\in Y} q_i^2\right) \det(\phi(Y)^\trans
  \phi(Y)) }\,,
\end{equation}
where $\phi(Y)$ denotes the $D \times |Y|$ matrix formed from columns
$\phi_i$ for $i \in Y$.  Our main result follows.

\begin{theorem}
  Let $\P^{\leq k}(Y)$ be the cardinality-conditioned DPP distribution
  defined by quality model $q$ and $D$-dimensional diversity feature
  function $\phi$, and let
  \begin{equation}
    \tilde \P^{\leq k}(Y) \propto \left(\prod_{i \in Y} q_i^2\right)
    \det([G\phi(Y)]^\trans[G\phi(Y)])
  \end{equation}
  be the cardinality-conditioned DPP distribution obtained by
  projecting $\phi$ with $G$.  Then for projection dimension $d$ as in
  \eqref{ddef}, we have
  \begin{equation}
    \Vert \P^{\leq k} - \tilde \P^{\leq k} \Vert_1 \leq e^{6k\epsilon} - 1 
  \end{equation}
  with probability at least $1-\delta$.  Note that $e^{6k\epsilon} - 1
  \approx 6k\epsilon$ when $k\epsilon$ is small.
  \thmlabel{dpp_proj}
\end{theorem}

\noindent
The theorem says that for $d$ logarithmic in $N$ and linear in $k$,
the $L_1$ variational distance between the original DPP and the
randomly projected version is bounded.  In order to use
\lemref{magen}, which bounds volumes of parallelepipeds, to prove this
bound on determinants, we will make use of the following relationship:
\begin{equation}
  \vol(X_Y) = \sqrt{\det(X_Y^\trans X_Y)}\,.
  \eqlabel{detvol}
\end{equation}
In order to handle the conditional DPP normalization constant
\begin{equation}
\sum_{|Y| \leq k} \left(\prod_{i\in Y} q_i^2\right) \det(\phi(Y)^\trans
  \phi(Y))\,,
\end{equation}
we also must adapt \lemref{magen} to \emph{sums} of determinants.
Finally, for technical reasons we will change the symmetry of the
upper and lower bounds from the sign of $\epsilon$ to the sign of the
exponent.  The following lemma gives the details.

\begin{lemma}
  Under the definitions and assumptions of \lemref{magen}, we have,
  with probability at least $1-\delta$,
  \begin{equation}
    (1+2\epsilon)^{-2k} \leq 
    \frac{\sum_{|Y|\leq k} \det((GX_Y)^\trans(GX_Y))}{\sum_{|Y|\leq k} 
      \det(X_Y^\trans X_Y)}
    \leq (1+\epsilon)^{2k}\,.
  \end{equation}
  \lemlabel{normbound}
\end{lemma}
\begin{proof}
  \begin{align}
    \sum_{|Y| \leq k} \det((GX_Y)^\trans(GX_Y))
    &= \sum_{|Y| \leq k} \vol^2(GX_Y)\\
    &\geq \sum_{|Y|\leq k} \left(\vol(X_Y)(1-\epsilon)^{|Y|}\right)^2\\
    &\geq (1-\epsilon)^{2k}\sum_{|Y|\leq k} \vol^2(X_Y)\\
    &\geq (1+2\epsilon)^{-2k} \sum_{|Y|\leq k} \det(X_Y^\trans X_Y)\,,
  \end{align}
  where the first inequality holds with probability at least
  $1-\delta$ by \lemref{magen}, and the third follows from the fact
  that $(1-\epsilon)(1+2\epsilon) \geq 1$ (since $\epsilon < 1/2$),
  thus $(1-\epsilon)^{2k} \geq (1+2\epsilon)^{-2k}$.  The upper bound
  follows directly:
  \begin{align}
    \sum_{|Y|\leq k} \left(\vol(GX_Y)\right)^2
    &\leq \sum_{|Y|\leq k} \left(\vol(X_Y)(1+\epsilon)^{|Y|}\right)^2\\
    &\leq (1+\epsilon)^{2k} \sum_{|Y|\leq k} \det(X_Y^\trans X_Y)\,.
  \end{align}
\end{proof}

\noindent
We can now prove \thmref{dpp_proj}.

\begin{proof}[Proof of \thmref{dpp_proj}]
  Recall the matrix $B$, whose columns are given by $B_i =
  q_i\phi_i$.  We have
  \begin{align}
    \Vert P^{\leq k} - \tilde P^{\leq k} \Vert_1 
    &= \sum_{|Y|\leq k} |P^{\leq k}(Y) - \tilde P^{\leq k}(Y)|\\
    &= \sum_{|Y|\leq k} P^{\leq k}(Y) \left|1 
    - \frac{\tilde P^{\leq k}(Y)}{P^{\leq k}(Y)}\right| \\
    &= \sum_{|Y|\leq k} P^{\leq k}(Y) \left| 1 -
    \frac{\det([GB_Y^\trans][GB_Y])}{\det(B_Y^\trans B_Y)}
    \frac{\sum_{|Y'|\leq k} \det(B_{Y'}^\trans B_{Y'})}
         {\sum_{|Y'|\leq k} \det([GB_{Y'}^\trans][GB_{Y'}])} \right| \nonumber\\
    & \leq  \left| 1 - (1+\epsilon)^{2k}(1+2\epsilon)^{2k} \right| 
         \sum_{|Y|\leq k} P^{\leq k}(Y) \\
    & \leq e^{6k\epsilon} - 1\,,
  \end{align}
  where the first inequality follows from \lemref{magen} and
  \lemref{normbound}, which hold simultaneously with probability at
  least $1-\delta$, and the second follows from $(1 + a)^b \leq
  e^{ab}$ for $a,b \geq 0$.
\end{proof}

By combining the dual representation with random projections, we can
deal simultaneously with very large $N$ and very large $D$.  In fact,
in \secref{sdpps} we will show that $N$ can even be exponentially
large if certain structural assumptions are met.  These techniques
vastly expand the range of problems to which DPPs can be practically
applied.

\subsection{Alternative likelihood formulas}

Recall that, in an L-ensemble DPP, the likelihood of a particular set
$Y \subseteq \Y$ is given by
\begin{equation}
  \P_L(Y) = \frac{\det(L_Y)}{\det(L+I)}\,.
  \eqlabel{orig_likelihood}
\end{equation}
This expression has some nice intuitive properties in terms of
volumes, and, ignoring the normalization in the denominator, takes a
simple and concise form.  However, as a ratio of determinants on
matrices of differing dimension, it may not always be analytically
convenient.  Minors can be difficult to reason about directly, and
ratios complicate calculations like derivatives.  Moreover, we might
want the likelihood in terms of the marginal kernel $K = L(L+I)^{-1} =
I - (L+I)^{-1}$, but simply plugging in these identities yields a
expression that is somewhat unwieldy.

As alternatives, we will derive some additional formulas that,
depending on context, may have useful advantages.  Our starting point
will be the observation, used previously in the proof of
\thmref{LtoK}, that minors can be written in terms of full matrices
and diagonal indicator matrices; specifically, for positive
semidefinite $L$,
\begin{align}
  \det(L_Y) &= \det(I_Y L + I_{\bar Y}) \eqlabel{minor1}\\
  &= (-1)^{|\bar Y|}\det(I_Y L - I_{\bar Y}) \eqlabel{minor2}\\
  &= \left|\det(I_Y L - I_{\bar Y})\right|\,,
\end{align}
where $I_Y$ is the diagonal matrix with ones in the diagonal positions
corresponding to elements of $Y$ and zeros everywhere else, and $\bar
Y = \Y - Y$.  These identities can be easily shown by examining the
matrices blockwise under the partition $\Y = Y \cup \bar Y$, as in the
proof of \thmref{LtoK}.

Applying \eqref{minor1} to \eqref{orig_likelihood}, we get
\begin{align}
  \P_L(Y) &= \frac{\det(I_Y L + I_{\bar Y})}{\det(L+I)}\\
  &= \det((I_Y L + I_{\bar Y})(L+I)^{-1})\\
  &= \det(I_Y L(L+I)^{-1} + I_{\bar Y}(L+I)^{-1})\,.
\end{align}
Already, this expression, which is a single determinant of an $N
\times N$ matrix, is in some ways easier to work with.  We can also
more easily write the likelihood in terms of $K$:
\begin{equation}
  \P_L(Y) = \det(I_Y K + I_{\bar Y}(I-K))\,.
  \eqlabel{altlike1}
\end{equation}
Recall from \eqref{complement_dpp} that $I-K$ is the marginal kernel
of the complement DPP; thus, in an informal sense we can read
\eqref{altlike1} as combining the marginal probability that $Y$ is
selected with the marginal probability that $\bar Y$ is not selected.

We can also make a similar derivation using \eqref{minor2}:
\begin{align}
  \P_L(Y) &= (-1)^{|\bar Y|}\frac{\det(I_Y L - I_{\bar Y})}{\det(L+I)}\\
  &= (-1)^{|\bar Y|}\det((I_Y L - I_{\bar Y})(L+I)^{-1})\\
  &= (-1)^{|\bar Y|}\det(I_Y L(L+I)^{-1} - I_{\bar Y}(L+I)^{-1})\\
  &= (-1)^{|\bar Y|}\det(I_Y K - I_{\bar Y}(I-K))\\
  &= (-1)^{|\bar Y|}\det(K - I_{\bar Y})\\
  &= \left|\det(K - I_{\bar Y})\right|\,.
  \eqlabel{altlike2}
\end{align}
Note that \eqref{altlike1} involves asymmetric matrix products, but
\eqref{altlike2} does not; on the other hand, $K - I_{\bar Y}$ is (in
general) indefinite.


\section{Learning}
\seclabel{learning}

We have seen that determinantal point process offer appealing modeling
intuitions and practical algorithms, capturing geometric notions of
diversity and permitting computationally efficient inference in a
variety of settings.  However, to accurately model real-world data we
must first somehow determine appropriate values of the model
parameters.  While an expert could conceivably design an appropriate
DPP kernel from prior knowledge, in general, especially when dealing
with large data sets, we would like to have an automated method for
learning a DPP.

We first discuss how to parameterize DPPs conditioned on input data.
We then define what we mean by learning, and, using the quality
vs. diversity decomposition introduced in \secref{decomposition}, we
show how a parameterized quality model can be learned efficiently from
a training set.

\subsection{Conditional DPPs}
\seclabel{condDPP}

Suppose we want to use a DPP to model the seats in an auditorium
chosen by students attending a class.  (Perhaps we think students tend
to spread out.)  In this context each meeting of the class is a new
sample from the empirical distribution over subsets of the (fixed)
seats, so we merely need to collect enough samples and we should be
able to fit our model, as desired.

For many problems, however, the notion of a single fixed base set $\Y$
is inadequate.  For instance, consider extractive document
summarization, where the goal is to choose a subset of the sentences
in a news article that together form a good summary of the entire
article.  In this setting $\Y$ is the set of sentences in the news
article being summarized, thus $\Y$ is not fixed in advance but
instead depends on context.  One way to deal with this problem is to
model the summary for each article as its own DPP with a separate
kernel matrix.  This approach certainly affords great flexibility, but
if we have only a single sample summary for each article, there is
little hope of getting good parameter estimates.  Even more
importantly, we have learned nothing that can be applied to generate
summaries of unseen articles at test time, which was presumably our
goal in the first place.

Alternatively, we could let $\Y$ be the set of all sentences appearing
in {\it any} news article; this allows us to learn a single model for
all of our data, but comes with obvious computational issues and does
not address the other concerns, since sentences are rarely repeated.

To solve this problem, we need a DPP that depends parametrically on
the input data; this will enable us to share information across
training examples in a justifiable and effective way.  We first
introduce some notation.  Let $\X$ be the input space; for example,
$\X$ might be the space of news articles.  Let $\Y(X)$ denote the
ground set of items implied by an input $X \in \X$, e.g., the set of
all sentences in news article $X$.  We have the following definition.

\begin{definition}
A {\bf conditional DPP} $\P(\bY=Y|X)$ is a conditional probabilistic
model which assigns a probability to every possible subset $Y
\subseteq \Y(X)$.  The model takes the form of an L-ensemble:
\begin{equation}
  \P(\bY=Y|X) \propto \det(L_Y(X))\,,
\end{equation}
where $L(X)$ is a positive semidefinite $|\Y(X)| \times |\Y(X)|$
kernel matrix that depends on the input.  
\end{definition}

As discussed in \secref{background}, the normalization constant for a
conditional DPP can be computed efficiently and is given by
$\det(L(X)+I)$.  Using the quality/diversity decomposition introduced
in \secref{decomposition}, we have
\begin{equation}
  L_{ij}(X) = q_i(X) \phi_i(X)^\trans\phi_j(X) q_j(X)
  \eqlabel{condL}
\end{equation}
for suitable $q_i(X) \in \reals^+$ and $\phi_i(X) \in \reals^D$,
$\Vert \phi_i(X)\Vert = 1$, which now depend on $X$.

In the following sections we will discuss application-specific
parameterizations of the quality and diversity models $q$ and $\phi$
in terms of the input.  First, however, we review our learning setup.

\subsubsection{Supervised learning}

The basic supervised learning problem is as follows.  We receive a
training data sample $\{(X^{(t)},Y^{(t)})\}_{t=1}^T$ drawn
independently and identically from a distribution $D$ over pairs
$(X,Y) \in \X \times 2^{\Y(X)}$, where $\X$ is an input space and
$\Y(X)$ is the associated ground set for input $X$.  We assume that the
conditional DPP kernel $L(X;\theta)$ is parameterized in terms of a
generic $\theta$, and let
\begin{equation}
  \P_\theta(Y|X) = \frac{\det(L_Y(X;\theta))}{\det(L(X;\theta) + I)}
\end{equation}
denote the conditional probability of an output $Y$ given input $X$
under parameter $\theta$.  The goal of learning is to choose
appropriate $\theta$ based on the training sample so that we can make
accurate predictions on unseen inputs.

While there are a variety of objective functions commonly used for
learning, here we will focus on {\it maximum likelihood} learning (or
maximum likelihood estimation, often abbreviated MLE), where the goal
is to choose $\theta$ to maximize the conditional log-likelihood of
the observed data:
\begin{align}
  \L(\theta) &= \log \prod_{t=1}^T \P_{\theta}(Y^{(t)}|X^{(t)})\\
  &= \sum_{t=1}^T \log \P_{\theta}(Y^{(t)}|X^{(t)})\\
  &= \sum_{t=1}^T \left[ \log\det(L_{Y^{(t)}}(X^{(t)};\theta)) 
    - \log\det(L(X^{(t)};\theta) + I) \right]\,.
\end{align}
Optimizing $\L$ is consistent under mild assumptions; that is, if the
training data are actually drawn from a conditional DPP with
parameter $\theta^*$, then the learned $\theta\to\theta^*$ as $T
\to\infty$.  Of course real data are unlikely to exactly follow any
particular model, but in any case the maximum likelihood approach has
the advantage of calibrating the DPP to produce reasonable probability
estimates, since maximizing $\L$ can be seen as minimizing the
log-loss on the training data.

To optimize the log-likelihood, we will use standard algorithms such
as gradient ascent or L-BFGS \citep{nocedal1980updating}.  These
algorithms depend on the gradient $\nabla\L(\theta)$, which must exist
and be computable, and they converge to the optimum whenever
$\L(\theta)$ is concave in $\theta$.  Thus, our ability to optimize
likelihood efficiently will depend fundamentally on these two
properties.

\subsection{Learning quality}
\seclabel{learning_q}

We begin by showing how to learn a parameterized quality model
$q_i(X;\theta)$ when the diversity feature function $\phi_i(X)$ is
held fixed \citep{kulesza2011learning}.  This setup is somewhat analogous to support vector
machines \citep{vapnik2000nature}, where a kernel is fixed by the
practitioner and then the per-example weights are automatically
learned.  Here, $\phi_i(X)$ can consist of any desired measurements
(and could even be infinite-dimensional, as long as the resulting
similarity matrix $S$ is a proper kernel).  We propose computing the
quality scores using a log-linear model:
\begin{equation}
  q_i(X;\theta) = \exp\left(\frac{1}{2} \theta^\trans \f_i(X)\right)\,,
  \eqlabel{qualitymodel}
\end{equation}
where $\f_i(X) \in \reals^m$ is a feature vector for item $i$ and the
parameter $\theta$ is now concretely an element of $\reals^m$.  Note
that feature vectors $\f_i(X)$ are in general distinct from
$\phi_i(X)$; the former are used for modeling quality, and will be
``interpreted'' by the parameters $\theta$, while the latter define
the diversity model $S$, which is fixed in advance.  We have
\begin{equation}
  \P_\theta(Y|X) = \frac{\prod_{i\in Y}
    \left[\exp\left(\theta^\trans\f_i(X)\right) \right] \det (S_Y(X))}
    { \sum_{Y' \subseteq \Y(X)} \prod_{i\in Y'}
      \left[\exp\left(\theta^\trans\f_i(X)\right)\right]
      \det(S_{Y'}(X))}\,.
\end{equation}

For ease of notation, going forward we will assume that the training
set contains only a single instance $(X,Y)$, and drop the instance
index $t$.  All of the following results extend easily to multiple
training examples.  First, we show that under this parameterization
the log-likelihood function is concave in $\theta$; then we will show
that its gradient can be computed efficiently.  With these results in
hand we will be able to apply standard optimization techniques.

\begin{proposition}
$\L(\theta)$ is concave in $\theta$.
\end{proposition}
\begin{proof} 
We have
\begin{align}
  \L(\theta) &= \log \P_\theta(Y|X)\\
  &= \theta^\trans\sum_{i\in Y} \f_i(X) + \log \det (S_Y(X))\nonumber\\
  &\quad\quad-\log\sum_{Y' \subseteq\Y(X)} 
  \exp\left(\theta^\trans\sum_{i\in Y'} \f_i(X)\right) \det(S_{Y'}(X))\,.
\eqlabel{llbreakdown}
\end{align}
With respect to $\theta$, the first term is linear, the second is
constant, and the third is the composition of a concave function
(negative log-sum-exp) and an affine function, so the overall
expression is concave.
\end{proof}

We now derive the gradient $\nabla\L(\theta)$, using
\eqref{llbreakdown} as a starting point.
\begin{align}
  \nabla \L(\theta)
  &= \sum_{i\in Y} \f_i(X) 
  - \nabla\left[\log\sum_{Y' \subseteq \Y(X)} \exp\left(\theta^\trans\sum_{i\in Y'} 
    \f_i(X)\right) \det(S_{Y'}(X))\right]
  \eqlabel{likemaxent}\\
  &= \sum_{i\in Y} \f_i(X) 
  - \sum_{Y' \subseteq \Y(X)} \frac{\exp\left(\theta^\trans\sum_{i\in Y'} 
    \f_i(X)\right) \det(S_{Y'}(X)) \sum_{i\in Y'}\f_i(X)} 
  {\sum_{Y'} \exp\left(\theta^\trans\sum_{i\in Y'} 
    \f_i(X)\right) \det(S_{Y'}(X))}\\
  &= \sum_{i\in Y} \f_i(X) 
  - \sum_{Y' \subseteq \Y(X)} \P_\theta(Y'|X) \sum_{i\in Y'}\f_i(X)\,.
  \eqlabel{qualgradient}
\end{align}
Thus, as in standard maximum entropy modeling, the gradient of the
log-likelihood can be seen as the difference between the empirical
feature counts and the expected feature counts under the model
distribution.  The difference here, of course, is that $\P_\theta$ is
a DPP, which assigns higher probability to diverse sets.  Compared
with a standard independent model obtained by removing the diversity
term from $\P_\theta$, \eqref{qualgradient} actually emphasizes those
training examples that are {\it not} diverse, since these are the
examples on which the quality model must focus its attention in order
to overcome the bias imposed by the determinant.  In the experiments
that follow we will see that this distinction is important in
practice.

The sum over $Y'$ in \eqref{qualgradient} is exponential in $|\Y(X)|$;
hence we cannot compute it directly.  Instead, we can rewrite it by
switching the order of summation:
\begin{equation}
  \sum_{Y' \subseteq \Y(X)} \P_{\theta}(Y'|X) \sum_{i\in Y'} \f_i(X) = \sum_{i}
  \f_i(X) \sum_{Y' \supseteq \{i\}} \P_{\theta}(Y'|X)\,.
\end{equation}
Note that $\sum_{Y' \supseteq \{i\}} \P_{\theta}(Y'|X)$ is the
marginal probability of item $i$ appearing in a set sampled from the
conditional DPP.  That is, the expected feature counts are computable
directly from the marginal probabilities.  Recall that we can
efficiently marginalize DPPs; in particular, per-item marginal
probabilities are given by the diagonal of $K(X;\theta)$, the marginal
kernel (which now depends on the input and the parameters).  We can
compute $K(X;\theta)$ from the kernel $L(X;\theta)$ using matrix
inversion or eigendecomposition.  \algref{gradient} shows how we can
use these ideas to compute the gradient of $\L(\theta)$ efficiently.

In fact, note that we do not need all of $K(X;\theta)$, but only its
diagonal.  In \algref{gradient} we exploit this in the main loop,
using only $O(N^2)$ multiplications rather than the $O(N^3)$ we would
need to construct the entire marginal kernel.  (In the dual
representation, this can be improved further to $O(ND)$
multiplications.)  Unfortunately, these savings are asymptotically
irrelevant since we still need to eigendecompose $L(X;\theta)$,
requiring about $O(N^3)$ time (or $O(D^3)$ time for the corresponding
eigendecomposition in the dual).  It is conceivable that a faster
algorithm exists for computing the diagonal of $K(X;\theta)$ directly,
along the lines of ideas recently proposed by \citet{tang2011probing}
(which focus on sparse matrices); however, we are not currently aware
of a useful improvement over \algref{gradient}.

\begin{algorithm}[tb]
\begin{algorithmic}
  \STATE {\bfseries Input:} instance $(X,Y)$, parameters $\theta$
  \STATE Compute $L(X;\theta)$ as in \eqref{condL}
  \STATE Eigendecompose $L(X;\theta) = \sum_{n=1}^N \lambda_n \v_n \v_n^\trans$
  \FOR{$i \in \Y(X)$}
  \STATE $K_{ii} \leftarrow \sum_{n=1}^N\frac{\lambda_n}{\lambda_n+1} \v_{ni}^2$
  \ENDFOR\\
  \STATE $\nabla\L(\theta) \leftarrow\sum_{i \in Y}\f_i(X)-\sum_i K_{ii}\f_i(X)$
  \STATE {\bfseries Output:} gradient $\nabla\L(\theta)$
\end{algorithmic}
\caption{Gradient of the log-likelihood}
\alglabel{gradient}
\end{algorithm}

\subsubsection{Experiments: document summarization}
\seclabel{documentsummarization}

We demonstrate learning for the conditional DPP quality model on an
extractive multi-document summarization task using news text.  The
basic goal is to generate a short piece of text that summarizes the
most important information from a news story.  In the {\it extractive}
setting, the summary is constructed by stringing together sentences
found in a cluster of relevant news articles.  This selection problem
is a balancing act: on the one hand, each selected sentence should be
relevant, sharing significant information with the cluster as a whole;
on the other, the selected sentences should be diverse as a group so
that the summary is not repetitive and is as informative as possible
given its length \citep{dang2005overview,nenkova2006compositional}.
DPPs are a natural fit for this task, viewed through the decomposition
of \secref{decomposition} \citep{kulesza2011learning}.

As in \secref{condDPP}, the input $X$ will be a cluster of documents,
and $\Y(X)$ a set of candidate sentences from those documents.  In our
experiments $\Y(X)$ contains all sentences from all articles in the
cluster, although in general preprocessing could also be used to try
to improve the candidate set \citep{conroy2004left}.  We will learn a
DPP to model good summaries $Y$ for a given input $X$.  Because DPPs
model unordered sets while summaries are linear text, we construct a
written summary from $Y$ by placing the sentences it contains in the
same order in which they appeared in the original documents.  This
policy is unlikely to give optimal results, but it is consistent with
prior work \citep{lin2010-submod-sum-nlp} and seems to perform well.
Furthermore, it is at least partially justified by the fact that
modern automatic summary evaluation metrics like ROUGE, which we
describe later, are mostly invariant to sentence order.

We experiment with data from the multi-document summarization task
(Task 2) of the 2003 and 2004 Document Understanding Conference (DUC)
\citep{dang2005overview}.  The article clusters used for these tasks
are taken from the NIST TDT collection.  Each cluster contains
approximately 10 articles drawn from the AP and New York Times
newswires, and covers a single topic over a short time span.  The
clusters have a mean length of approximately 250 sentences and 5800
words.  The 2003 task, which we use for training, contains 30
clusters, and the 2004 task, which is our test set, contains 50
clusters.  Each cluster comes with four reference human summaries
(which are not necessarily formed by sentences from the original
articles) for evaluation purposes.  Summaries are required to be at
most 665 characters in length, including spaces.  \figref{summdata}
depicts a sample cluster from the test set.

\begin{figure}
  \begin{center}
    \includegraphics[width=\textwidth]{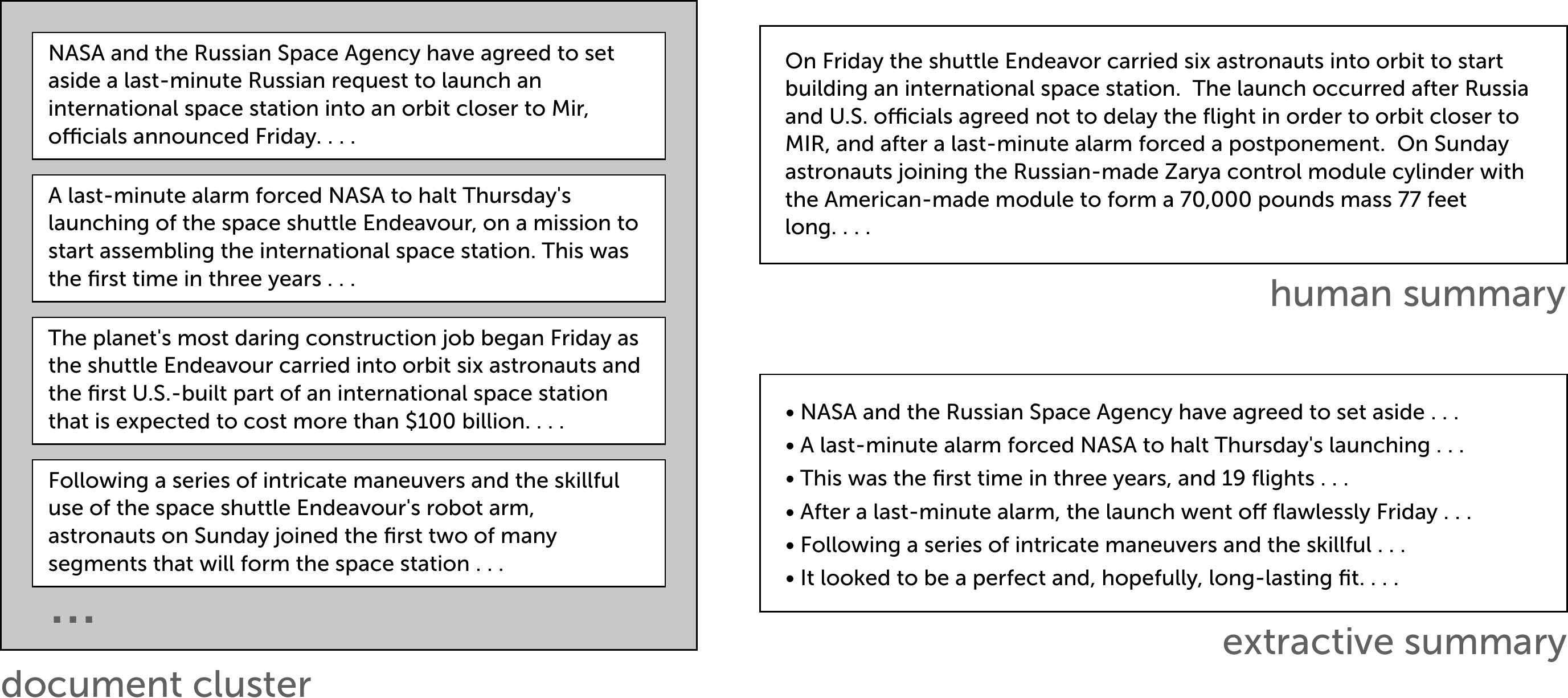}
  \end{center}
  \caption[Sample multi-document summaries]
    {A sample cluster from the DUC 2004 test set, with one of the
    four human reference summaries and an (artificial) extractive
    summary.}  
  \figlabel{summdata}
\end{figure}

To measure performance on this task we follow the original evaluation
and use ROUGE, an automatic evaluation metric for summarization
\citep{lin2004rouge}.  ROUGE measures $n$-gram overlap statistics
between the human references and the summary being scored, and
combines them to produce various sub-metrics.  ROUGE-1, for example,
is a simple unigram recall measure that has been shown to correlate
quite well with human judgments \citep{lin2004rouge}.  Here, we use
ROUGE's unigram F-measure (which combines ROUGE-1 with a measure of
precision) as our primary metric for development.  We refer to this
measure as ROUGE-1F.  We also report ROUGE-1P and ROUGE-1R (precision
and recall, respectively) as well as ROUGE-2F and ROUGE-SU4F, which
include bigram match statistics and have also been shown to correlate
well with human judgments.  Our implementation uses ROUGE version
1.5.5 with stemming turned on, but without stopword removal.  These
settings correspond to those used for the actual DUC competitions
\citep{dang2005overview}; however, we use a more recent version of
ROUGE.

\paragraph{Training data}

Recall that our learning setup requires a training sample of pairs
$(X,Y)$, where $Y \subseteq \Y(X)$.  Unfortunately, while the human
reference summaries provided with the DUC data are of high quality, they
are not extractive, thus they do not serve as examples of summaries
that we can actually model.  To obtain high-quality extractive
``oracle'' summaries from the human summaries, we employ a simple
greedy algorithm (\algref{greedydata}).  On each round the sentence
that achieves maximal unigram F-measure to the human references,
normalized by length, is selected and added to the extractive summary.
Since high F-measure requires high precision as well as recall, we
then update the references by removing the words ``covered'' by the
newly selected sentence, and proceed to the next round.

\begin{algorithm}[tb]
  \begin{algorithmic}
    \STATE {\bfseries Input:} article cluster $X$, human reference 
    word counts $H$, character limit $b$
    \STATE $U \leftarrow \Y(X)$
    \STATE $Y \leftarrow \emptyset$
    \WHILE{$U \neq \emptyset$}
    \STATE $i \leftarrow \argmax_{i' \in U} \left( 
           \frac{\rouge(\words(i'),H)}
                {\sqrt{\length(i')}}\right)$
    \STATE $Y \leftarrow Y \cup \{i\}$
    \STATE $H \leftarrow \max(H - \words(i),0)$
    \STATE  $U \leftarrow U - 
    (\{i\} \cup \{i'|\length(Y) + \length(i') > b\})$
    \ENDWHILE
    \STATE {\bfseries Output:} extractive oracle summary $Y$
  \end{algorithmic}
  \caption{Constructing extractive training data}
  \alglabel{greedydata}
\end{algorithm}

We can measure the success of this approach by calculating ROUGE
scores of our oracle summaries with respect to the human summaries.
\tabref{gold} shows the results for the DUC 2003 training set.  For
reference, the table also includes the ROUGE scores of the best
automatic system from the DUC competition in 2003 (``machine''), as
well as the human references themselves (``human'').  Note that, in
the latter case, the human summary being evaluated is also one of the
four references used to compute ROUGE; hence the scores are probably
significantly higher than a human could achieve in practice.
Furthermore, it has been shown that extractive summaries, even when
generated optimally, are by nature limited in quality compared with
unconstrained summaries \citep{genest2010hextac}.  Thus we believe
that the oracle summaries make strong targets for training.

\begin{table}
\centering 
\begin{tabular}{lccc}
System & {\bf \scriptsize ROUGE-1F} & 
{\bf \scriptsize ROUGE-2F} & {\bf \scriptsize ROUGE-SU4F} \\
\hline
Machine & 35.17 & \, 9.15 & 12.47 \\
Oracle & 46.59 & 16.18 & 19.52 \\
Human & 56.22 & 33.37 & 36.50 \\
\hline
\end{tabular}
\caption[Oracle ROUGE scores]
  {ROUGE scores for the best automatic system from
  DUC 2003, our heuristically-generated oracle extractive summaries, and
  human summaries.}  \tablabel{gold}
\end{table}

\paragraph{Features}
\seclabel{summ_features}

We next describe the feature functions that we use for this task.  For
diversity features $\phi_i(X)$, we generate standard normalized tf-idf
vectors.  We tokenize the input test, remove stop words and
punctuation, and apply a Porter stemmer.\footnote{Code for this
  preprocessing pipeline was provided by Hui Lin and Jeff Bilmes.}
Then, for each word $\word$, the {\it term frequency} $\tf_i(\word)$
of $\word$ in sentence $i$ is defined as the number of times the word
appears in the sentence, and the {\it inverse document frequency}
$\idf(\word)$ is the negative logarithm of the fraction of articles in
the training set where $\word$ appears.  A large value of
$\idf(\word)$ implies that $\word$ is relatively rare.  Finally, the
vector $\phi_i(X)$ has one element per word, and the value of the
entry associated with word $\word$ is proportional to
$\tf_i(\word)\idf(\word)$.  The scale of $\phi_i(X)$ is set such that
$\Vert \phi_i(X) \Vert = 1$.

Under this definition of $\phi$, the similarity $S_{ij}$ between
sentences $i$ and $j$ is known as their {\it cosine similarity}:
\begin{equation}
  S_{ij} = \frac{\sum_\word \tf_i(\word) \tf_j(\word) \idf^2(\word)}
  {\sqrt{\sum_\word \tf^2_i(\word) \idf^2(\word)} 
    \sqrt{\sum_\word \tf^2_j(\word) \idf^2(\word)}}
  \in [0,1]\,.
\end{equation}
Two sentences are cosine similar if they contain many of the same
words, particularly words that are uncommon (and thus more likely to
be salient).

We augment $\phi_i(X)$ with an additional constant feature taking the
value $\rho \geq 0$, which is a hyperparameter.  This has the effect
of making all sentences more similar to one another, increasing
repulsion.  We set $\rho$ to optimize ROUGE-1F score on the training
set; in our experiments, the best choice was $\rho = 0.7$.

We use the very standard cosine distance as our similarity metric
because we need to be confident that it is sensible; it will remain
fixed throughout the experiments.  On the other hand, weights for the
quality features are learned, so we can use a variety of intuitive
measures and rely on training to find an appropriate combination.  The
quality features we use are listed below.  For some of the features,
we make use of cosine distances; these are computed using the same
tf-idf vectors as the diversity features.  When a feature is
intrinsically real-valued, we produce a series of binary features by
binning.  The bin boundaries are determined either globally or
locally.  Global bins are evenly spaced quantiles of the feature
values across all sentences in the training set, while local bins are
quantiles of the feature values in the current cluster only.
\begin{itemize}
\item {\bf Constant}: A constant feature allows the model to bias
  towards summaries with a greater or smaller number of sentences.
\item {\bf Length}: We bin the length of the sentence (in characters) into
  five global bins.
\item {\bf Document position}: We compute the position of the sentence
  in its original document and generate binary features indicating
  positions 1--5, plus a sixth binary feature indicating all other
  positions.  We expect that, for newswire text, sentences that appear
  earlier in an article are more likely to be useful for
  summarization.
\item {\bf Mean cluster similarity}: For each sentence we compute the
  average cosine distance to all other sentences in the cluster.  This
  feature attempts to measure how well the sentence reflects the
  salient words occurring most frequently in the cluster.  We use the
  raw score, five global bins, and ten local bins.
\item {\bf LexRank}: We compute continuous LexRank scores by finding
  the principal eigenvector of the row-normalized cosine similarity
  matrix.  (See \citet{erkan2004lexrank} for details.)  This provides
  an alternative measure of centrality.  We use the raw score, five
  global bins, and five local bins.
\item {\bf Personal pronouns}: We count the number of personal
  pronouns (``he'', ``her'', ``themselves'', etc.) appearing in each
  sentence.  Sentences with many pronouns may be poor for
  summarization since they omit important entity names.
\end{itemize}
In total we have 40 quality features; including $\rho$ our model
has 41 parameters.

\paragraph{Inference}
\seclabel{summ_inf}

At test time, we need to take the learned parameters $\theta$ and use
them to predict a summary $Y$ for a previously unseen document cluster
$X$.  One option is to sample from the conditional distribution, which
can be done exactly and efficiently, as described in
\secref{sampling}.  However, sampling occasionally produces
low-probability predictions.  We obtain better performance on this
task by applying two alternative inference techniques.

\textit{Greedy MAP approximation.}
One common approach to prediction in probabilistic models is maximum
{\it a posteriori} (MAP) decoding, which selects the highest
probability configuration.  For practical reasons, and because the
primary metrics for evaluation were recall-based, the DUC evaluations
imposed a length limit of 665 characters, including spaces, on all
summaries.  In order to compare with prior work we also apply this
limit in our tests.  Thus, our goal is to find the most likely
summary, subject to a budget constraint:
\begin{align}
  Y^{\map} = \argmax_Y\ \ &\P_\theta(Y|X) \nonumber\\
  \st\ \ &\sum_{i\in Y} \length(i) \leq b\,,
\eqlabel{map}
\end{align}
where $\length(i)$ is the number of characters in sentence $i$, and $b
= 665$ is the limit on the total length.  As discussed in
\secref{dppmap}, computing $Y^{\map}$ exactly is NP-hard, but,
recalling that the optimization in \eqref{map} is submodular, we can
approximate it through a simple greedy algorithm (\algref{inference}).

\begin{algorithm}[tb]
  \begin{algorithmic}
    \STATE {\bfseries Input:} document cluster $X$, parameter $\theta$, 
    character limit $b$
    \STATE $U \leftarrow \Y(X)$
    \STATE $Y \leftarrow \emptyset$
    \WHILE{$U \neq \emptyset$}
    \STATE $i \leftarrow \argmax_{i' \in U} \left( 
           \frac{\P_\theta(Y \cup \{i\}|X) - \P_\theta(Y|X)}
                {\length(i)}\right)$
    \STATE $Y \leftarrow Y \cup \{i\}$
    \STATE  $U \leftarrow U - 
    (\{i\} \cup \{i'|\length(Y) + \length(i') > b\})$
    \ENDWHILE
    \STATE {\bfseries Output:} summary $Y$
  \end{algorithmic}
\caption{Approximately computing the MAP summary}
\alglabel{inference}
\end{algorithm}

\algref{inference} is closely related to those given by
\citet{krause2005note} and especially \citet{lin2010-submod-sum-nlp}.
As discussed in \secref{dppmap}, algorithms of this type have formal
approximation guarantees for monotone submodular problems.  Our MAP
problem is not generally monotone; nonetheless, \algref{inference}
seems to work well in practice, and is very fast (see
\tabref{inference_speed}).

\textit{Minimum Bayes risk decoding.}
The second inference technique we consider is minimum Bayes risk (MBR)
decoding.  First proposed by \citet{goel2000minimum} for automatic
speech recognition, MBR decoding has also been used successfully for
word alignment and machine translation
\citep{kumar2002minimum,kumar2004minimum}.  The idea is to choose a
prediction that minimizes a particular application-specific loss
function under uncertainty about the evaluation target.  In our
setting we use ROUGE-1F as a (negative) loss function, so we have
\begin{equation}
  Y^{\mbr} = \argmax_Y\ \E\left[\rouge(Y,\bY^*)\right]\,,
\end{equation}
where the expectation is over realizations of $\bY^*$, the true
summary against which we are evaluated.  Of course, the distribution
of $\bY^*$ is unknown, but we can assume that our trained model
$\P_\theta(\cdot|X)$ gives a reasonable approximation.  Since there
are exponentially many possible summaries, we cannot expect to perform
an exact search for $Y^{\mbr}$; however, we can approximate it through
sampling, which is efficient.

Combining these approximations, we have the following inference rule:
\begin{equation}
  \tilde Y^{\mbr} = \argmax_{Y^{r'},\ r' \in \{1,2,\dots,R\}}\ 
  \frac{1}{R}\sum_{r=1}^R \rouge(Y^{r'},Y^r)\,,
\end{equation}
where $Y^1,Y^2,\dots,Y^R$ are samples drawn from $\P_\theta(\cdot|X)$.
In order to satisfy the length constraint imposed by the evaluation,
we consider only samples with length between 660 and 680 characters
(rejecting those that fall outside this range), and crop $\tilde
Y^{\mbr}$ to the limit of 665 bytes if necessary.  The choice of $R$
is a tradeoff between fast running time and quality of inference.  In
the following section, we report results for $R = 100, 1000,$ and
$5000$; \tabref{inference_speed} shows the average time required to
produce a summary under each setting.  Note that MBR decoding is
easily parallelizable, but the results in \tabref{inference_speed} are
for a single processor.  Since MBR decoding is randomized, we report
all results averaged over 100 trials.
  
\begin{table}
  \centering 
  \begin{tabular}{lr@{.}l}
    System & \multicolumn{2}{c}{Time (s)}\\
    \hline
    \textsc{dpp-greedy} & 0&15 \\
    \textsc{dpp-mbr100} & 1&30 \\
    \textsc{dpp-mbr1000} & 16&91 \\
    \textsc{dpp-mbr5000} & 196&86 \\       
    \hline
  \end{tabular}
  \caption[Inference speed] {The average time required to produce a
    summary for a single cluster from the DUC 2004 test set (without
    parallelization).}  \tablabel{inference_speed}
\end{table}

\paragraph{Results}

We train our model with a standard L-BFGS optimization algorithm.  We
place a zero-mean Gaussian prior on the parameters $\theta$, with
variance set to optimize ROUGE-1F on a development subset of the 2003
data.  We learn parameters $\theta$ on the DUC 2003 corpus, and test
them using DUC 2004 data.  We generate predictions from the trained
DPP using the two inference algorithms described in the previous
section, and compare their performance to a variety of baseline
systems.

Our first and simplest baseline merely returns the first 665 bytes of
the cluster text.  Since the clusters consist of news articles, this
is not an entirely unreasonable summary in many cases.  We refer to
this baseline as \textsc{begin}.

We also compare against an alternative DPP-based model with identical
similarity measure and quality features, but where the quality model
has been trained using standard logistic regression.  To learn this
baseline, each sentence is treated as a unique instance to be
classified as included or not included, with labels derived from our
training oracle.  Thus, it has the advantages of a DPP at test time,
but does not take into account the diversity model while training;
comparing to this baseline allows us to isolate the contribution of
learning the model parameters in context.  Note that MBR inference is
impractical for this model because its training does not properly
calibrate for overall summary length, so nearly all samples are either
too long or too short.  Thus, we report only the results obtained from
greedy inference.  We refer to this model as \textsc{lr+dpp}.

Next, we employ as baselines a range of previously proposed methods
for multi-document summarization.  Perhaps the simplest and most
popular is Maximum Marginal Relevance (MMR), which uses a greedy
selection process \citep{carbonell1998use}.  MMR relies on a
similarity measure between sentences, for which we use the cosine
distance measure $S$, and a measure of relevance for each sentence,
for which we use the same logistic regression-trained quality model as
above.  Sentences are chosen iteratively according to
\begin{equation}
  \argmax_{i \in \Y(X)} \left[ \alpha q_i(X) - 
    (1-\alpha) \max_{j \in Y} S_{ij}\right]\,,
\end{equation}
where $Y$ is the set of sentences already selected (initially empty),
$q_i(X)$ is the learned quality score, and $S_{ij}$ is the cosine
similarity between sentences $i$ and $j$.  The tradeoff $\alpha$ is
optimized on a development set, and sentences are added until the
budget is full.  We refer to this baseline as \textsc{lr+mmr}.

We also compare against the three highest-scoring systems that
actually competed in the DUC 2004 competition---peers 65, 104, and
35---as well as the submodular graph-based approach recently described
by \citet{lin2010-submod-sum-nlp}, which we refer to as
\textsc{submod1}, and the improved submodular learning approach
proposed by \citet{lin2012submodular}, which we denote
\textsc{submod2}.  We produced our own implementation of
\textsc{submod1}, but rely on previously reported numbers for
\textsc{submod2}, which include only ROUGE-1 scores.

\tabref{results} shows the results for all methods on the DUC 2004
test corpus.  Scores for the actual DUC competitors differ slightly
from the originally reported results because we use an updated version
of the ROUGE package.  Bold entries highlight the best performance in
each column; in the case of MBR inference, which is stochastic, the
improvements are significant at 99\% confidence.  The DPP models
outperform the baselines in most cases; furthermore, there is a
significant boost in performance due to the use of DPP maximum
likelihood training in place of logistic regression.  MBR inference
performs best, assuming we take sufficiently many samples; on the
other hand, greedy inference runs more quickly than
\textsc{dpp-mbr100} and produces superior results.  Relative to most
other methods, the DPP model with MBR inference seems to more strongly
emphasize recall.  Note that MBR inference was performed with respect
to ROUGE-1F, but could also be run to optimize other metrics if
desired.

\begin{table}
  \centering
  \begin{tabular}{lccccc}
    System & \bf{\scriptsize ROUGE-1F} & 
    \bf{\scriptsize ROUGE-1P} & \bf{\scriptsize ROUGE-1R} &
    \bf{\scriptsize ROUGE-2F} & \bf{\scriptsize ROUGE-SU4F} 
    \\
    \hline
    \textsc{begin} & 32.08 & 31.53 & 32.69  & 6.52 & 10.37\\
    \textsc{lr+mmr} & 37.58 & 37.15 & 38.05 & 9.05 & 13.06\\
    \textsc{lr+dpp} & 37.96 & 37.67 & 38.31  & 8.88 & 13.13\\
    \textsc{peer 35} & 37.54 & 37.69 & 37.45 & 8.37 & 12.90\\
    \textsc{peer 104} & 37.12 & 36.79 & 37.48 & 8.49 & 12.81\\
    \textsc{peer 65} & 37.87 & 37.58 & 38.20 & 9.13 & 13.19\\
    \textsc{submod1} & 38.73 & 38.40 & 39.11 & 8.86 & 13.11\\
    \textsc{submod2} & 39.78 & 39.16 & 40.43 & - & -\\
    \textsc{dpp-greedy} & 38.96 & 38.82 & 39.15 & \textbf{9.86} & 13.83 \\
    \textsc{dpp-mbr100} & 38.83 & 38.06 & 39.67 & 8.85 & 13.38 \\
    \textsc{dpp-mbr1000} & 39.79 & 38.96 & 40.69 & 9.29 & 13.87 \\
    \textsc{dpp-mbr5000} & \textbf{40.33} & \textbf{39.43} & \textbf{41.31}
    & 9.54 & \textbf{14.13} \\
    \hline
  \end{tabular}
  \caption[ROUGE scores for DUC 2004]
          {ROUGE scores on the DUC 2004 test set.}
  \tablabel{results}
\end{table}

\textit{Feature contributions.}  
In \tabref{ablative} we report the performance of \textsc{dpp-greedy}
when different groups of features from \secref{summ_features} are
removed, in order to estimate their relative contributions.  Length
and position appear to be quite important; however, although
individually similarity and LexRank scores have only a modest impact
on performance, when both are omitted the drop is significant.  This
suggests, intuitively, that these two groups convey similar
information---both are essentially measures of centrality---but that
this information is important to achieving strong performance.

\begin{table}
  \centering
  \begin{tabular}{lccc}
    Features & \bf{\scriptsize ROUGE-1F} & 
    \bf{\scriptsize ROUGE-1P} & \bf{\scriptsize ROUGE-1R}
    \\
    \hline
    All                & 38.96 & 38.82 & 39.15 \\
    All but length     & 37.38 & 37.08 & 37.72 \\
    All but position   & 36.34 & 35.99 & 36.72 \\
    All but similarity & 38.14 & 37.97 & 38.35 \\
    All but LexRank    & 38.10 & 37.92 & 38.34 \\
    All but pronouns   & 38.80 & 38.67 & 38.98 \\
    All but similarity, LexRank        & 36.06 & 35.84 & 36.32 \\
    \hline
  \end{tabular}
  \caption[ROUGE scores with features removed]
    {ROUGE scores for \textsc{dpp-greedy} with features removed.}
  \tablabel{ablative}
\end{table}

\section{$k$-DPPs}
\seclabel{kdpps}

A determinantal point process assigns a probability to every subset of
the ground set $\Y$.  This means that, with some probability, a sample
from the process will be empty; with some probability, it will be all
of $\Y$.  In many cases this is not desirable.  For instance, we might
want to use a DPP to model the positions of basketball players on a
court, under the assumption that a team tends to spread out for better
coverage.  In this setting, we know that with very high probability
each team will have exactly five players on the court.  Thus, if our
model gives some probability of seeing zero or fifty players, it is
not likely to be a good fit.

We showed in \secref{sampling} that there exist elementary DPPs having
fixed cardinality $k$; however, this is achieved only by focusing
exclusively (and equally) on $k$ specific ``aspects'' of the data, as
represented by eigenvectors of the kernel.  Thus, for DPPs, the
notions of {\it size} and {\it content} are fundamentally intertwined.
We cannot change one without affecting the other.  This is a serious
limitation on the types of distributions than can be expressed; for
instance, a DPP cannot even capture the uniform distribution over sets
of cardinality $k$.

More generally, even for applications where the number of items is
unknown, the size model imposed by a DPP may not be a good fit.  We
have seen that the cardinality of a DPP sample has a simple
distribution: it is the number of successes in a series of Bernoulli
trials.  But while this distribution characterizes certain types of
data, other cases might look very different.  For example, picnickers
may tend to stake out diverse positions in a park, but on warm weekend
days there might be hundreds of people, and on a rainy Tuesday night
there are likely to be none.  This bimodal distribution is quite
unlike the sum of Bernoulli variables imposed by DPPs.

Perhaps most importantly, in some cases we do not even want to model
cardinality at all, but instead offer it as a parameter.  For example,
a search engine might need to deliver ten diverse results to its
desktop users, but only five to its mobile users.  This ability to
control the size of a DPP ``on the fly'' can be crucial in real-world
applications.

In this section we introduce $k$-DPPs, which address the issues
described above by conditioning a DPP on the cardinality of the random
set $\bY$.  This simple change effectively divorces the DPP content
model, with its intuitive diversifying properties, from the DPP size
model, which is not always appropriate.  We can then use the DPP
content model with a size model of our choosing, or simply set the
desired size based on context.  The result is a significantly more
expressive modeling approach (which can even have limited positive
correlations) and increased control.

We begin by defining $k$-DPPs.  The conditionalization they require,
though simple in theory, necessitates new algorithms for inference
problems like normalization and sampling.  Naively, these tasks
require exponential time, but we show that through recursions for
computing elementary symmetric polynomials we can solve them exactly
in polynomial time.  Finally, we demonstrate the use of $k$-DPPs on an
image search problem, where the goal is to show users diverse sets of
images that correspond to their query.

\subsection{Definition}

A $k$-DPP on a discrete set $\Y = \{1,2,\dots,N\}$ is a distribution
over all subsets $Y \subseteq \Y$ with cardinality $k$ \citep{kulesza2011kdpps}.  In contrast
to the standard DPP, which models both the size and content of a
random subset $\bY$, a $k$-DPP is concerned only with the content of a
random $k$-set.  Thus, a $k$-DPP is obtained by conditioning a
standard DPP on the event that the set $\bY$ has cardinality $k$.
Formally, the $k$-DPP $\P^k_L$ gives probabilities
\begin{equation}
  P^k_L(Y) = \frac{\det(L_Y)}{\sum_{|Y'| = k} \det(L_{Y'})}\,,
  \eqlabel{kdpp}
\end{equation}
where $|Y| = k$ and $L$ is a positive semidefinite kernel.  Compared
to the standard DPP, the only changes are the restriction on $Y$ and
the normalization constant.  While in a DPP every $k$-set $Y$ competes
with all other subsets of $\Y$, in a $k$-DPP it competes only with
sets of the same cardinality.  This subtle change has significant
implications.

For instance, consider the seemingly simple distribution that is
uniform over all sets $Y \subseteq \Y$ with cardinality $k$.  If we
attempt to build a DPP capturing this distribution we quickly run into
difficulties.  In particular, the marginal probability of any single
item is $\frac{k}{N}$, so the marginal kernel $K$, if it exists, must
have $\frac{k}{N}$ on the diagonal.  Likewise, the marginal
probability of any pair of items is $\frac{k(k-1)}{N(N-1)}$, and so by
symmetry the off diagonal entries of $K$ must be equal to a constant.
As a result, any valid marginal kernel has to be the sum of a constant
matrix and a multiple of the identity matrix.  Since a constant matrix
has at most one nonzero eigenvalue and the identity matrix is full
rank, it is easy to show that, except in the special cases
$k=0,1,N-1$, the resulting kernel has full rank.  But we know that a
full rank kernel implies that the probability of seeing all $N$ items
together is nonzero.  Thus the desired process cannot be a DPP unless
$k=0,1,N-1,$ or $N$.  On the other hand, a $k$-DPP with the identity
matrix as its kernel gives the distribution we are looking for.  This
improved expressive power can be quite valuable in practice.

\subsubsection{Alternative models of size}

Since a $k$-DPP is conditioned on cardinality, $k$ must come from
somewhere outside of the model.  In many cases, $k$ may be fixed
according to application needs, or perhaps changed on the fly by users
or depending on context.  This flexibility and control is one of the
major practical advantages of $k$-DPPs.  Alternatively, in situations
where we wish to model size as well as content, a $k$-DPP can be
combined with a size model $\P_\size$ that assigns a probability to
every possible $k \in \{1,2,\dots,N\}$:
\begin{equation}
  \P(Y) = \P_\size(|Y|)\P_L^{|Y|}(Y)\,.
\end{equation}
Since the $k$-DPP is a proper conditional model, the distribution $\P$
is well-defined.  By choosing $\P_\size$ appropriate to the task at
hand, we can effectively take advantage of the diversifying properties
of DPPs in situations where the DPP size model is a poor fit.

As a side effect, this approach actually enables us to use $k$-DPPs to
build models with both negative and positive correlations.  For
instance, if $\P_\size$ indicates that there are likely to be either
hundreds of picnickers in the park (on a nice day) or, otherwise, just
a few, then knowing that there are fifty picnickers today implies that
there are likely to be even more.  Thus, $k$-DPPs can yield more
expressive models than DPPs in this sense as well.

\subsection{Inference}

Of course, increasing the expressive power of the DPP causes us to
wonder whether, in doing so, we might have lost some of the convenient
computational properties that made DPPs useful in the first place.
Naively, this seems to be the case; for instance, while the
normalizing constant for a DPP can be written in closed form, the sum
in \eqref{kdpp} is exponential and seems hard to simplify.  In this
section, we will show how $k$-DPP inference can in fact be performed
efficiently, using recursions for computing the elementary symmetric
polynomials.

\subsubsection{Normalization}

Recall that the $k$th elementary symmetric polynomial on
$\lambda_1,\lambda_2\dots,\lambda_N$ is given by
\begin{equation}
  e_k(\lambda_1,\lambda_2,\dots,\lambda_N) = 
  \sum_{J \subseteq \{1,2,\dots,N\}\atop|J|=k} \prod_{n \in J} \lambda_n\,. 
\end{equation}
For instance,
\begin{align}
  e_1(\lambda_1,\lambda_2,\lambda_3) &= 
  \lambda_1 + \lambda_2 + \lambda_3\\
  e_2(\lambda_1,\lambda_2,\lambda_3) &= 
  \lambda_1\lambda_2 + \lambda_1\lambda_3 + \lambda_2\lambda_3\\
  e_3(\lambda_1,\lambda_2,\lambda_3) &= 
  \lambda_1\lambda_2\lambda_3\,.
\end{align}

\begin{proposition}
  The normalization constant for a $k$-DPP is
  \begin{equation}
    Z_k = \sum_{|Y'| = k} \det(L_{Y'}) =
    e_k(\lambda_1,\lambda_2,\dots,\lambda_N)\,,
  \end{equation}
  where $\lambda_1,\lambda_2,\dots,\lambda_N$ are the eigenvalues of
  $L$.  \proplabel{kdppnorm}
\end{proposition}
\begin{proof}
  One way to see this is to examine the characteristic polynomial of
  $L$, $\det(L-\lambda I)$ \citep{gelfand1989lectures}.  We can also
  show it directly using properties of DPPs.  Recalling that
  \begin{equation}
    \sum_{Y\subseteq\Y} \det(L_Y) = \det(L+I)\,,
  \end{equation}
  we have
  \begin{equation}
    \sum_{|Y'| = k} \det(L_{Y'}) = \det(L+I) \sum_{|Y'| = k} \P_L(Y')\,,
  \end{equation}
  where $\P_L$ is the DPP with kernel $L$.  Applying
  \lemref{elementarymixture}, which expresses any DPP as a mixture of
  elementary DPPs, we have
  \begin{align}
    \det(L+I) \sum_{|Y'| = k} \P_L(Y') 
    &= \sum_{|Y'| = k} \sum_{J \subseteq \{1,2,\dots,N\}} 
    \P^{V_J}(Y') \prod_{n\in J} \lambda_n\\
    &= \sum_{|J| = k} \sum_{|Y'| = k} \P^{V_J}(Y') \prod_{n\in J} \lambda_n \\
    &= \sum_{|J| = k} \prod_{n\in J} \lambda_n\eqlabel{sympoly}\,,
  \end{align}
  where we use \lemref{elementaryfixed} in the last two steps to
  conclude that $\P^{V_J}(Y') = 0$ unless $|J| = |Y'|$.  (Recall that
  $V_J$ is the set of eigenvectors of $L$ associated with $\lambda_n$
  for $n \in J$.)
\end{proof}

To compute the $k$th elementary symmetric polynomial, we can use the
recursive algorithm given in \algref{summation}, which is based on the
observation that every set of $k$ eigenvalues either omits
$\lambda_N$, in which case we must choose $k$ of the remaining
eigenvalues, or includes $\lambda_N$, in which case we get a factor of
$\lambda_N$ and choose only $k-1$ of the remaining eigenvalues.
Formally, letting $e_k^N$ be a shorthand for
$e_k(\lambda_1,\lambda_2,\dots,\lambda_N)$, we have
\begin{equation}
  e_k^N = e_k^{N-1} + \lambda_N e_{k-1}^{N-1}\,.
\end{equation}
Note that a variety of recursions for computing elementary symmetric
polynomials exist, including Newton's identities, the Difference
Algorithm, and the Summation Algorithm \citep{baker1996computing}.
\algref{summation} is essentially the Summation Algorithm, which is
both asymptotically faster and numerically more stable than the other
two, since it uses only sums and does not rely on precise cancellation
of large numbers.

\begin{algorithm}[tb]
\begin{algorithmic}
  \STATE {\bfseries Input:} $k$, 
    eigenvalues $\lambda_1,\lambda_2,\dots\lambda_N$
  \STATE $e_0^n \leftarrow 1\quad\forall\ n \in \{0,1,2,\dots,N\}$
  \STATE $e_l^0 \leftarrow 0\quad\forall\ l \in \{1,2,\dots,k\}$
  \FOR{$l = 1,2,\dots k$}
  \FOR{$n = 1,2,\dots,N$}
  \STATE $e_l^n \leftarrow e_l^{n-1} + \lambda_n e_{l-1}^{n-1}$
  \ENDFOR
  \ENDFOR
  \STATE {\bfseries Output:} $e_k(\lambda_1,\lambda_2,\dots,\lambda_N) = e_k^N$
\end{algorithmic}
\caption{Computing the elementary symmetric polynomials}
\alglabel{summation}
\end{algorithm}

\algref{summation} runs in time $O(Nk)$.  Strictly speaking, the inner
loop need only iterate up to $N-k+l$ in order to obtain $e_k^N$ at the
end; however, by going up to $N$ we compute all of the preceding
elementary symmetric polynomials $e_l^N$ along the way.  Thus, by
running \algref{summation} with $k=N$ we can compute the normalizers
for $k$-DPPs of every size in time $O(N^2)$.  This can be useful when
$k$ is not known in advance.

\subsubsection{Sampling}
\seclabel{kdpp_sampling}

Since a $k$-DPP is just a DPP conditioned on size, we could sample a
$k$-DPP by repeatedly sampling the corresponding DPP and rejecting the
samples until we obtain one of size $k$.  To make this more efficient,
recall from \secref{sampling} that the standard DPP sampling algorithm
proceeds in two phases.  First, a subset $V$ of the eigenvectors of
$L$ is selected at random, and then a set of cardinality $|V|$ is
sampled based on those eigenvectors.  Since the size of a sample is
fixed in the first phase, we could reject the samples before the
second phase even begins, waiting until we have $|V|=k$.  However,
rejection sampling is likely to be slow.  It would be better to
directly sample a set $V$ conditioned on the fact that its cardinality
is $k$.  In this section we show how sampling $k$ eigenvectors can be
done efficiently, yielding a sampling algorithm for $k$-DPPs that is
asymptotically as fast as sampling standard DPPs.

We can formalize the intuition above by rewriting the $k$-DPP
distribution in terms of the corresponding DPP:
\begin{equation}
  \P^k_L(Y) = \frac{1}{e_k^N}\det(L+I)\P_L(Y)
\end{equation}
whenever $|Y| = k$, where we replace the DPP normalization constant
with the $k$-DPP normalization constant using \propref{kdppnorm}.
Applying \lemref{elementarymixture} and \lemref{elementaryfixed} to
decompose the DPP into elementary parts yields
\begin{equation}
  \P^k_L(Y) = \frac{1}{e_k^N} \sum_{|J| = k} \P^{V_J}(Y) \prod_{n \in J} \lambda_n\,.
  \eqlabel{kdppelem}
\end{equation}
Therefore, a $k$-DPP is also a mixture of elementary DPPs, but it only
gives nonzero weight to those of dimension $k$.  Since the second
phase of DPP sampling provides a means for sampling from any given
elementary DPP, we can sample from a $k$-DPP if we can sample index
sets $J$ according to the corresponding mixture components.  Like
normalization, this is naively an exponential task, but we can do it
efficiently using the recursive properties of elementary symmetric
polynomials.

\begin{algorithm}[tb]
  \caption{Sampling $k$ eigenvectors}
  \alglabel{sampleJ}
  \begin{algorithmic}
    \STATE {\bfseries Input:} $k$, eigenvalues 
    $\lambda_1,\lambda_2,\dots,\lambda_N$
    \STATE compute $e_l^n$ for $l
    = 0,1,\dots,k$ and $n = 0,1,\dots,N$ (\algref{summation})
    \STATE $J \leftarrow \emptyset$
    \STATE $l \leftarrow k$
    \FOR{$n = N,\dots,2,1$}
    \IF{$l = 0$}
    \STATE {\bf break}
    \ENDIF  
    \IF{$u \sim U[0,1] < \lambda_n\frac{e_{l-1}^{n-1}}{e_l^n}$}
    \STATE $J \leftarrow J \cup \{n\}$
    \STATE $l \leftarrow l-1$
    \ENDIF
    \ENDFOR
    \STATE {\bfseries Output:} $J$
  \end{algorithmic}
\end{algorithm}

\begin{theorem}
  Let $\bJ$ be the desired random variable, so that $\Pr(\bJ = J) =
  \frac{1}{e_k^N} \prod_{n\in J} \lambda_n$ when $|J| = k$, and zero
  otherwise.  Then \algref{sampleJ} yields a sample for $\bJ$.
\end{theorem}
\begin{proof}
  If $k=0$, then \algref{sampleJ} returns immediately at the first
  iteration of the loop with $J=\emptyset$, which is the only possible
  value of $\bJ$.

  If $N=1$ and $k=1$, then $\bJ$ must contain the single index $1$.
  We have $e_1^1 = \lambda_1$ and $e_0^0 = 1$, thus $\lambda_1
  \frac{e_0^0}{e_1^1} = 1$, and \algref{sampleJ} returns $J = \{1\}$
  with probability 1.

  We proceed by induction and compute the probability that
  \algref{sampleJ} returns $J$ for $N>1$ and $1 \leq k \leq N$.  By
  inductive hypothesis, if an iteration of the loop in
  \algref{sampleJ} begins with $n < N$ and $0 \leq l \leq n$, then the
  remainder of the algorithm adds to $J$ a set of elements $J'$ with
  probability
  \begin{equation}
    \frac{1}{e_{l}^{n}} \prod_{n'\in J'} \lambda_{n'}
  \end{equation}
  if $|J'| = l$, and zero otherwise.

  Now suppose that $J$ contains $N$, $J = J' \cup \{N\}$.  Then $N$
  must be added to $J$ in the first iteration of the loop, which
  occurs with probability $\lambda_N \frac{e_{k-1}^{N-1}}{e_k^N}$.
  The second iteration then begins with $n = N-1$ and $l = k-1$.  If
  $l$ is zero, we have the immediate base case; otherwise we have $1
  \leq l \leq n$.  By the inductive hypothesis, the remainder of the
  algorithm selects $J'$ with probability
  \begin{equation}
    \frac{1}{e_{k-1}^{N-1}} \prod_{n\in J'} \lambda_n
  \end{equation}
  if $|J'| = k-1$, and zero otherwise.  Thus \algref{sampleJ} returns
  $J$ with probability
  \begin{equation}
    \left(\lambda_N \frac{e_{k-1}^{N-1}}{e_k^N}\right)
    \frac{1}{e_{k-1}^{N-1}} \prod_{n\in J'} \lambda_n = 
    \frac{1}{e_k^N} \prod_{n\in J} \lambda_n
  \end{equation}
  if $|J| = k$, and zero otherwise.

  On the other hand, if $J$ does not contain $N$, then the first
  iteration must add nothing to $J$; this happens with probability
  \begin{equation}
    1-\lambda_N \frac{e_{k-1}^{N-1}}{e_k^N} = \frac{e_{k}^{N-1}}{e_k^N}\,,
    \eqlabel{probnoN}
  \end{equation}
  where we use the fact that $e_k^N - \lambda_N e_{k-1}^{N-1} =
  e_k^{N-1}$.  The second iteration then begins with $n = N-1$ and
  $l=k$.  We observe that if $N-1 < k$, then \eqref{probnoN} is equal
  to zero, since $e_l^n = 0$ whenever $l > n$.  Thus almost surely the
  second iteration begins with $k \leq n$, and we can apply the
  inductive hypothesis.  This guarantees that the remainder of the
  algorithm chooses $J$ with probability
  \begin{equation}
    \frac{1}{e_{k}^{N-1}} \prod_{n\in J} \lambda_n
  \end{equation}
  whenever $|J| = k$.  The overall probability that \algref{sampleJ}
  returns $J$ is therefore
  \begin{equation}
    \left(\frac{e_{k}^{N-1}}{e_k^N}\right)
    \frac{1}{e_{k}^{N-1}} \prod_{n\in J} \lambda_n = 
    \frac{1}{e_k^N} \prod_{n\in J} \lambda_n
  \end{equation}
  if $|J| = k$, and zero otherwise.
\end{proof}

\algref{sampleJ} precomputes the values of $e_1^1,\dots,e_k^N$, which
requires $O(Nk)$ time using \algref{summation}.  The loop then
iterates at most $N$ times and requires only a constant number of
operations, so \algref{sampleJ} runs in $O(Nk)$ time overall.  By
\eqref{kdppelem}, selecting $J$ with \algref{sampleJ} and then
sampling from the elementary DPP $\P^{V_J}$ generates a sample from
the $k$-DPP.  As shown in \secref{sampling}, sampling an elementary
DPP can be done in $O(Nk^3)$ time (see the second loop of
\algref{dpp_sampling}), so sampling $k$-DPPs is $O(Nk^3)$ overall,
assuming we have an eigendecomposition of the kernel in advance.  This
is no more expensive than sampling a standard DPP.

\subsubsection{Marginalization}

Since $k$-DPPs are not DPPs, they do not in general have marginal
kernels.  However, we can still use their connection to DPPs to
compute the marginal probability of a set $A$, $|A| \leq k$:
\begin{align}
  \P^k_L(A \subseteq \bY) 
  &= \sum_{|Y'| = k-|A| \atop A \cap Y' = \emptyset} \P^k_L(Y' \cup A)\\
  &= \frac{\det(L+I)}{Z_k} \sum_{|Y'| = k-|A| \atop A \cap Y' = \emptyset} 
  \P_L(Y' \cup A)\\
  &= \frac{\det(L+I)}{Z_k} \sum_{|Y'| = k-|A| \atop A \cap Y' = \emptyset} 
  \P_L(\bY = Y' \cup A | A \subseteq \bY)\P_L(A \subseteq \bY)\\
  &= \frac{Z_{k-|A|}^{A}}{Z_k} \frac{\det(L+I)}{\det(L^A+I)} 
  \P_L(A \subseteq \bY)\,,
  \eqlabel{kdppmarg}
\end{align}
where $L^A$ is the kernel, given in \eqref{cond_incl_kernel}, of the
DPP conditioned on the inclusion of $A$, and
\begin{align}
  Z_{k - |A|}^{A} &= \det(L^A+I)\sum_{|Y'| = k-|A| \atop A \cap Y' =
    \emptyset} \P_L(\bY = Y' \cup A | A \subseteq \bY)\\
  &= \sum_{|Y'| = k-|A| \atop A \cap Y' =
    \emptyset} \det(L^A_{Y'})
\end{align}
is the normalization constant for the $(k-|A|)$-DPP with kernel $L^A$.
That is, the marginal probabilities for a $k$-DPP are just the
marginal probabilities for a DPP with the same kernel, but with an
appropriate change of normalizing constants.  We can simplify
\eqref{kdppmarg} by observing that
\begin{equation}
  \frac{\det(L_A)}{\det(L+I)} = 
  \frac{\P_L(A \subseteq\bY)}{\det(L^A+I)}\,,
\end{equation}
since the left hand side is the probability (under the DPP with kernel
$L$) that $A$ occurs by itself, and the right hand side is the
marginal probability of $A$ multiplied by the probability of observing
nothing else conditioned on observing $A$: $1/\det(L^A+I)$.  Thus we
have
\begin{equation}
  \P^k_L(A \subseteq \bY) = \frac{Z_{k-|A|}^{A}}{Z_k} \det(L_A)
  = Z_{k-|A|}^A \P^k_L(A)\,.
  \eqlabel{kdppmarg2}
\end{equation}
That is, the marginal probability of $A$ is the probability of
observing exactly $A$ times the normalization constant when
conditioning on $A$.  Note that a version of this formula also holds
for standard DPPs, but there it can be rewritten in terms of the
marginal kernel.

\paragraph{Singleton marginals}

\eqsref{kdppmarg}{kdppmarg2} are general but require computing large
determinants and elementary symmetric polynomials, regardless of the
size of $A$.  Moreover, those quantities (for example, $\det(L^A+I)$)
must be recomputed for each unique $A$ whose marginal probability is
desired.  Thus, finding the marginal probabilities of many small sets
is expensive compared to a standard DPP, where we need only small
minors of $K$.  However, we can derive a more efficient approach in
the special but useful case where we want to know all of the singleton
marginals for a $k$-DPP---for instance, in order to implement quality
learning as described in \secref{learning_q}.

We start by using \eqref{kdppelem} to write the marginal probability
of an item $i$ in terms of a combination of elementary DPPs:
\begin{equation}
  \P^k_L(i\in \bY) = \frac{1}{e^N_k}\sum_{|J|=k}\P^{V_J}(i \in \bY)
  \prod_{n'\in J} \lambda_{n'}\,.
\end{equation}
Because the marginal kernel of the elementary DPP $\P^{V_J}$ is given
by $\sum_{n\in J} \v_n\v_n^\trans$, we have
\begin{align}
  \P^k_L(i\in \bY) &= \frac{1}{e^N_k}\sum_{|J|=k}
  \left(\sum_{n\in J} (\v_n^\trans \e_i)^2\right) 
  \prod_{n'\in J} \lambda_{n'}\\
  &= \frac{1}{e^N_k}\sum_{n=1}^N  (\v_n^\trans \e_i)^2
  \sum_{J\supseteq\{n\},|J|=k} \prod_{n'\in J} \lambda_{n'}\\
  &= \sum_{n=1}^N  (\v_n^\trans \e_i)^2 \lambda_n \frac{e_{k-1}^{-n}}{e^N_k}\,,
\end{align}
where $e^{-n}_{k-1} =
e_{k-1}(\lambda_1,\lambda_2,\dots,\lambda_{n-1},\lambda_{n+1},
\dots,\lambda_N)$ denotes the $(k-1)$-order elementary symmetric
polynomial for all eigenvalues of $L$ except $\lambda_n$.  Note that
$\lambda_n e_{k-1}^{-n}/e^N_k$ is exactly the marginal probability
that $n \in J$ when $J$ is chosen using \algref{sampleJ}; in other
words, the marginal probability of item $i$ is the sum of the
contributions $(\v_n^\trans \e_i)^2$ made by each eigenvector scaled
by the respective probabilities that the eigenvectors are selected.
The contributions are easily computed from the eigendecomposition of
$L$, thus we need only $e^N_k$ and $e^{-n}_{k-1}$ for each value of
$n$ in order to calculate the marginals for all items in $O(N^2)$
time, or $O(ND)$ time if the rank of $L$ is $D < N$.

\algref{summation} computes $e^{N-1}_{k-1} = e^{-N}_{k-1}$ in the
process of obtaining $e^N_k$, so naively we could run
\algref{summation} $N$ times, repeatedly reordering the eigenvectors
so that each takes a turn at the last position.  To compute all of the
required polynomials in this fashion would require $O(N^2k)$ time.
However, we can improve this (for small $k$) to $O(N\log(N) k^2)$; to
do so we will make use of a binary tree on $N$ leaves.  Each node of
the tree corresponds to a set of eigenvalues of $L$; the leaves
represent single eigenvalues, and an interior node of the tree
represents the set of eigenvalues corresponding to its descendant
leaves.  (See \figref{esp_tree}.)  We will associate with each node
the set of elementary symmetric polynomials
$e_1(\Lambda),e_2(\Lambda),\dots,e_k(\Lambda)$, where $\Lambda$ is the
set of eigenvalues represented by the node.

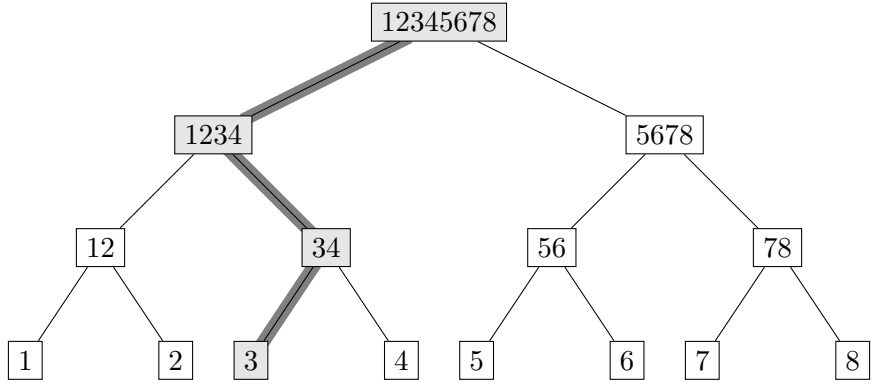
\begin{figure}
\centering
\begin{tikzpicture}[auto,level/.style={sibling distance=60mm/#1}]
\node 
    [draw,fill=black!10] (a1) {12345678}
    child {node [draw,fill=black!10] (a2) {1234}
      child {node [draw] {12}
        child {node [draw] {1}}
        child {node [draw] {2}}
      }
      child {node [draw,fill=black!10] (a3) {34}
        child {node [draw,fill=black!10] (a4) {3}}
        child {node [draw] {4}}
      }
    }
    child {node [draw] {5678}
      child {node [draw] {56}
        child {node [draw] {5}}
        child {node [draw] {6}}
      }
      child {node [draw] {78}
        child {node [draw] {7}}
        child {node [draw] {8}}
      }  
    };
    \begin{pgfonlayer}{background}
      \draw [line width=5pt,black!50,
        shorten <=-4pt,shorten >=-4pt] (a1) to (a2);
      \draw [line width=5pt,black!50,
        shorten <=-4pt,shorten >=-4pt] (a2) to (a3);
      \draw [line width=5pt,black!50,
        shorten <=-4pt,shorten >=-4pt] (a3) to (a4);
    \end{pgfonlayer}
\end{tikzpicture}
\caption[Binary tree for singleton $k$-DPP marginals]
  {Binary tree with $N=8$ leaves; interior nodes represent their
  descendant leaves.  Removing a path from leaf $n$ to the root leaves
  $\log N$ subtrees that can be combined to compute $e^{-n}_{k-1}$.
  \figlabel{esp_tree}}
\end{figure}

These polynomials can be computed directly for leaf nodes in constant
time, and the polynomials of an interior node can be computed given
those of its children using a simple recursion:
\begin{equation}
  e_k(\Lambda_1 \cup \Lambda_2) = 
  \sum_{l=0}^k e_l(\Lambda_1)e_{k-l}(\Lambda_2)\,.
  \eqlabel{esp_recursion}
\end{equation}
Thus, we can compute the polynomials for the entire tree in
$O(N\log(N) k^2)$ time; this is sufficient to obtain $e^N_k$ at the
root node.

However, if we now remove a leaf node corresponding to eigenvalue $n$,
we invalidate the polynomials along the path from the leaf to the
root; see \figref{esp_tree}.  This leaves $\log N$ disjoint subtrees
which together represent all of the eigenvalues of $L$, leaving out
$\lambda_n$.  We can now apply \eqref{esp_recursion} $\log N$ times to
the roots of these trees in order to obtain $e^{-n}_{k-1}$ in
$O(\log(N)k^2)$ time.  If we do this for each value of $n$, the total
additional time required is $O(N\log(N) k^2)$.

The algorithm described above thus takes $O(N\log(N)k^2)$ time to
produce the necessary elementary symmetric polynomials, which in turn
allow us to compute all of the singleton marginals.  This is a
dramatic improvement over applying \eqref{kdppmarg} to each item
separately.

\subsubsection{Conditioning}

Suppose we want to condition a $k$-DPP on the inclusion of a
particular set $A$.  For $|A| + |B| = k$ we have
\begin{align}
  \P^k_L(\bY = A \cup B | A \subseteq \bY) 
  &\propto \P^k_L(\bY = A \cup B)\\
  &\propto \P_L(\bY = A \cup B)\\
  &\propto \P_L(\bY = A \cup B | A \subseteq \bY)\\
  &\propto \det(L^A_B)\,.
\end{align}
Thus the conditional $k$-DPP is a $k-|A|$-DPP whose kernel is the same
as that of the associated conditional DPP.  The normalization constant
is $Z_{k-|A|}^A$.  We can condition on excluding $A$ in the same
manner.

\subsubsection{Finding the mode}

Unfortunately, although $k$-DPPs offer the efficient versions of DPP
inference algorithms presented above, finding the most likely set $Y$
remains intractable.  It is easy to see that the reduction from
\secref{dppmap} still applies, since the cardinality of the $Y$
corresponding to an exact 3-cover, if it exists, is known.  In
practice we can utilize greedy approximations, like we did for
standard DPPs in \secref{documentsummarization}.

\subsection{Experiments: image search}
\seclabel{imagesearch}

We demonstrate the use of $k$-DPPs on an image search task \citep{kulesza2011kdpps}.  The
motivation is as follows.  Suppose that we run an image search engine,
where our primary goal is to deliver the most relevant possible images
to our users.  Unfortunately, the query strings those users provide
are often ambiguous.  For instance, a user searching for
``philadelphia'' might be looking for pictures of the city skyline,
street-level shots of buildings, or perhaps iconic sights like the
Liberty Bell or the Love sculpture.  Furthermore, even if we know the
user is looking for a skyline photograph, he or she might specifically
want a daytime or nighttime shot, a particular angle, and so on.  In
general, we cannot expect users to provide enough information in a
textual query to identify the best image with any certainty.  

For this reason search engines typically provide a small array of
results, and we argue that, to maximize the probability of the user
being happy with at least one image, the results should be relevant to
the query but also diverse with respect to one another.  That is, if
we want to maximize the proportion of users searching ``philadelphia''
who are satisfied by our response, each image we return should satisfy
a large but distinct subset of those users, thus maximizing our
overall coverage.  Since we want diverse results but also require
control over the number of results we provide, a $k$-DPP is a natural
fit.

\subsubsection{Learning setup}

Of course, we do not actually run a search engine and do not have real
users.  Thus, in order to be able to evaluate our model using real
human feedback, we define the task in a manner that allows us to
obtain inexpensive human supervision via Amazon Mechanical Turk.  We
do this by establishing a simple binary decision problem, where the
goal is to choose, given two possible sets of image search results,
the set that is more diverse.  Formally, our labeled training data
comprises comparative pairs of image sets $\{(Y^+_t,Y^-_t)\}_{t=1}^T$,
where set $Y_t^+$ is preferred over set $Y_t^-$, $|Y_t^+| = |Y_t^-| =
k$.  We can measure performance on this classification task using the
zero-one loss, which is zero whenever we choose the correct set from a
given pair, and one otherwise.


For this task we employ a simple method for
learning a combination of $k$-DPPs that is convex and seems to work
well in practice.  Given a set $L_1,L_2,\dots,L_D$ of ``expert''
kernel matrices, which are fixed in advance, define the combination
model
\begin{equation}
  \P^k_\theta = \sum_{l=1}^D \theta_l \P^k_{L_l}\,,
\end{equation}
where $\sum_{l=1}^D \theta_l = 1$.  Note that this is a combination of
distributions, rather than a combination of kernels.  We will learn
$\theta$ to optimize a logistic loss measure on the binary task:
\begin{align}
  \min_\theta\quad & \L(\theta) = 
  \sum_{t=1}^T \log\left(1 + e^{-\gamma\left[\P^k_\theta(Y_t^+) - 
      \P^k_\theta(Y_t^-)\right]}\right)\nonumber\\
  \st\quad & \sum_{l=1}^D \theta_l = 1\,,\eqlabel{optim}
\end{align}
where $\gamma$ is a hyperparameter that controls how aggressively we
penalize mistakes.  Intuitively, the idea is to find a combination of
$k$-DPPs where the positive sets $Y^+_t$ receive higher probability
than the corresponding negative sets $Y^-_t$.  By using the logistic
loss (\figref{logistic}), which acts like a smooth hinge loss, we
focus on making fewer mistakes.

\begin{figure}
  \centering
  \includegraphics[width=3in]{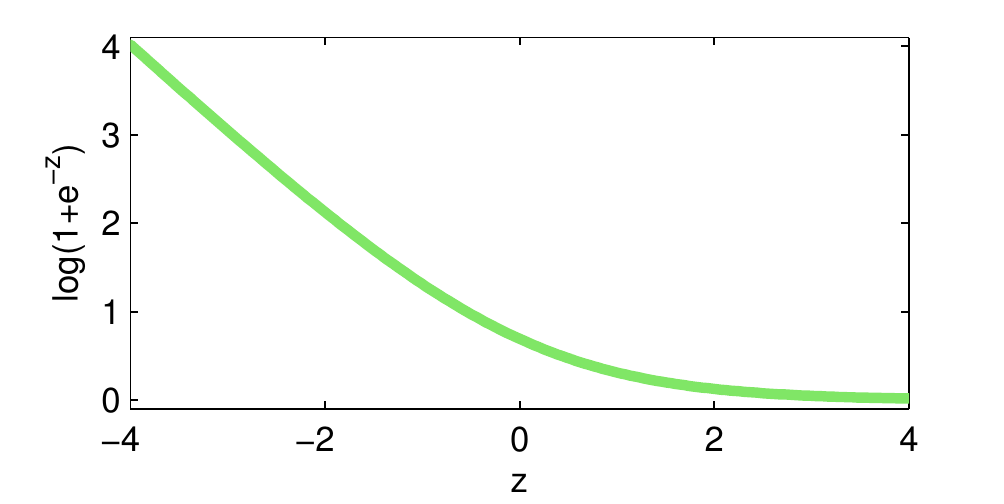}
  \caption[Logistic loss function]
          {The logistic loss function.}
  \figlabel{logistic}
\end{figure}

Because \eqref{optim} is convex in $\theta$ (it is the composition of
the convex logistic loss function with a linear function of $\theta$),
we can optimize it efficiently using projected gradient descent, where
we alternate taking gradient steps and projecting on the constraint
$\sum_{l=1}^D \theta_l = 1$.  The gradient is given by
\begin{equation}
  \nabla\L = \sum_{t=1}^T \frac{e^{\theta^\trans\delta^t}}
           {1+e^{\theta^\trans\delta^t}} \delta^t\,,
\end{equation}
where $\delta^t$ is a vector with entries
\begin{equation}
  \delta^t_l = -\gamma\left[\P^k_{L_l}(Y_t^+) - \P^k_{L_l}(Y_t^-)\right]\,.
\end{equation}
Projection onto the simplex is achieved using standard
algorithms \citep{bertsekas99nonlinear}.

\subsubsection{Data}

We create datasets for three broad image search categories, using
8--12 hand-selected queries for each category.  (See
\tabref{queries}.)  For each query, we retrieve the top 64 results
from Google Image Search, restricting the search to JPEG files that
pass the strictest level of Safe Search filtering.  Of those 64
results, we eliminate any that are no longer available for download.
On average this leaves us with 63.0 images per query, with a range of
59--64.

\begin{table}
  \centering
  \begin{tabular}{ccc}
    {\sc cars} & {\sc cities} & {\sc dogs}\\
    \hline
    chrysler&     baltimore&       beagle\\       
    ford&         barcelona&       bernese\\      
    honda&        london&          blue heeler\\  
    mercedes&     los angeles&     cocker spaniel\\
    mitsubishi&   miami&           collie\\       
    nissan&       new york city&   great dane\\   
    porsche&      paris&           labrador\\     
    toyota&       philadelphia&    pomeranian\\   
    &             san francisco&   poodle\\       
    &             shanghai&        pug\\          
    &             tokyo&           schnauzer\\    
        &             toronto&         shih tzu\\  
    \hline
  \end{tabular}
  \caption[Image search queries]
      {Queries used for data collection.}
  \tablabel{queries}
\end{table}

We then use the downloaded images to generate 960 training instances
for each category, spread evenly across the different queries.  In
order to compare $k$-DPPs directly against baseline heuristic methods
that do not model probabilities of full sets, we generate only
instances where $Y_t^+$ and $Y_t^-$ differ by a single element.  That
is, the classification problem is effectively to choose which of two
candidate images $i_t^+,i_t^i$ is a less redundant addition to a given
partial result set $Y_t$:
\begin{align}
  Y_t^+ &= Y_t \cup \{i_t^+\} & Y_t^- &= Y_t \cup \{i_t^-\}\,.
\end{align}
In our experiments $Y_t$ contains five images, so $k = |Y_t^+| =
|Y_t^-| = 6$.  We sample partial result sets using a $k$-DPP with a
SIFT-based kernel (details below) to encourage diversity.  The
candidates are then selected uniformly at random from the remaining
images, except for 10\% of instances that are reserved for measuring
the performance of our human judges.  For those instances, one of the
candidates is a duplicate image chosen uniformly at random from the
partial result set, making it the obviously more redundant choice.
The other candidate is chosen as usual.

In order to decide which candidate actually results in the more
diverse set, we collect human diversity judgments using Amazon's
Mechanical Turk.  Annotators are drawn from the general pool of Turk
workers, and are able to label as many instances as they wish.
Annotators are paid \$0.01 USD for each instance that they label.  For
practical reasons, we present the images to the annotators at reduced
scale; the larger dimension of an image is always 250 pixels.  The
annotators are instructed to choose the candidate that they feel is
``less similar'' to the images in the partial result set.  We do not
offer any specific guidance on how to judge similarity, since dealing
with uncertainty in human users is central to the task.  The candidate
images are presented in random order.  \figref{labeling_examples}
shows a sample instance from each category.

\begin{figure}
\centering
\includegraphics[clip=true,trim=0 0.15in 0 0,width=\textwidth]
                {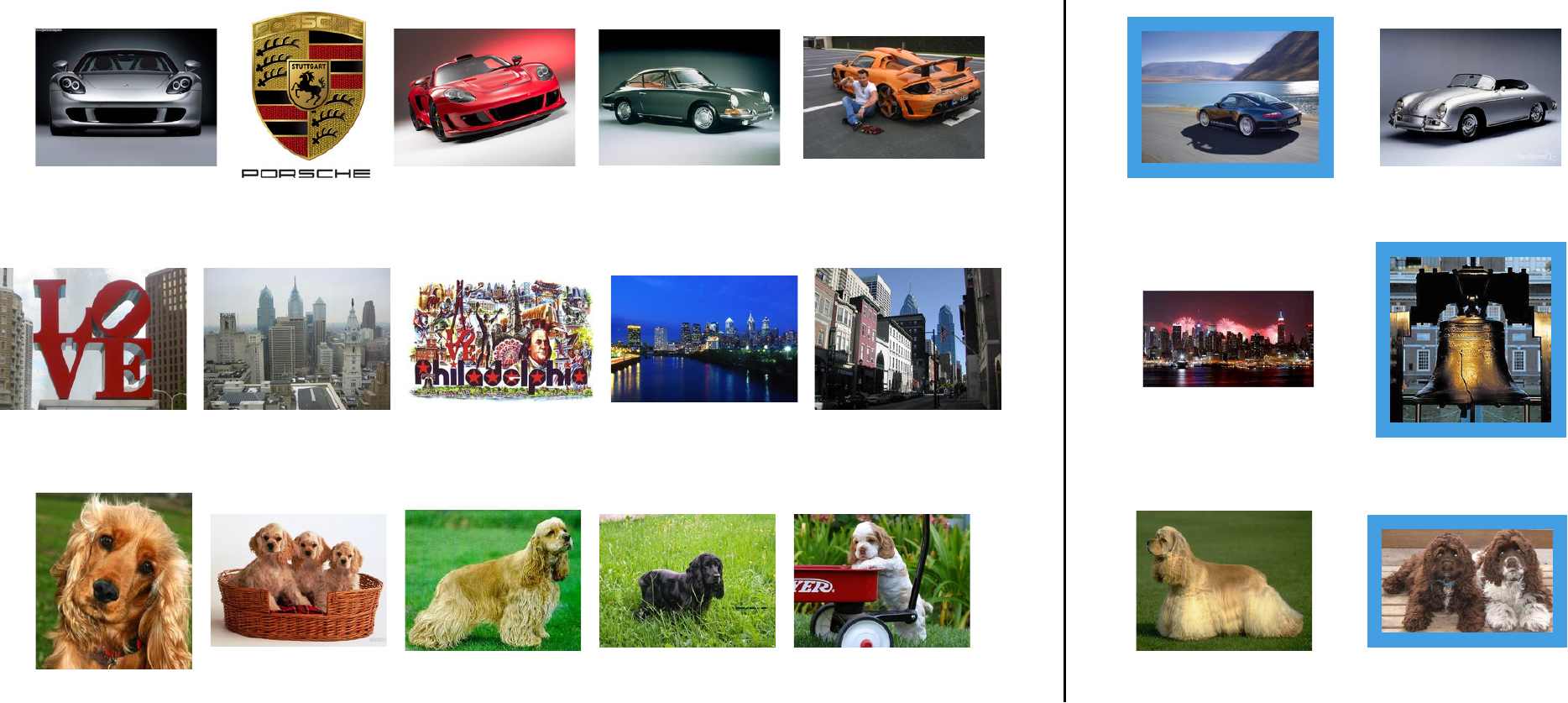}
\caption[Sample image search instances]
  {Sample labeling instances from each search category.  The
  five images on the left form the partial result set, and the two
  candidates are shown on the right.  The candidate receiving the
  majority of annotator votes has a blue border.}
\figlabel{labeling_examples}
\end{figure}

Overall, we find that workers choose the correct image for 80.8\% of
the calibration instances (that is, they choose the one not belonging
to the partial result set).  This suggests only moderate levels of
noise due to misunderstanding, inattention or robot workers.  However,
for non-calibration instances the task is inherently difficult and
subjective.  To keep noise in check, we have each instance labeled by
five independent judges, and keep only those instances where four or
more judges agree.  In the end this leaves us with 408--482 labeled
instances per category, or about half of the original instances.

\subsubsection{Kernels}

We define a set of 55 ``expert'' similarity kernels for the collected
images, which form the building blocks of our combination model and
baseline methods.  Each kernel $L^{\f}$ is the Gram matrix of some
feature function $\f$; that is, $L^{\f}_{ij} = \f(i)\cdot \f(j)$ for
images $i$ and $j$.  We therefore specify the kernels through the
feature functions used to generate them.  All of our feature functions
are normalized so that $\Vert \f(i) \Vert^2 = 1$ for all $i$; this
ensures that no image is {\it a priori} more likely than any other.
Implicitly, thinking in terms of the decomposition in
\secref{decomposition}, we are assuming that all of the images in our
set are equally {\it relevant} in order to isolate the modeling of
diversity.  This assumption is at least partly justified by the fact
that our images come from actual Google searches, and are thus
presumably relevant to the query.

We use the following feature functions, which derive from standard
image processing and feature extraction methods:
\begin{itemize}
\item {\bf Color} (2 variants): Each pixel is assigned a coordinate in
  three-dimensional Lab color space.  The colors are then sorted into
  axis-aligned bins, producing a histogram of either 8 or 64
  dimensions.
\item {\bf SIFT} (2 variants): The images are processed with the {\tt
  vlfeat} toolbox to obtain sets of 128-dimensional SIFT
  descriptors \citep{sift,vedaldi08vlfeat}.  The descriptors for a
  given category are combined, subsampled to a set of 25,000, and then
  clustered using $k$-means into either 256 or 512 clusters.  The
  feature vector for an image is the normalized histogram of the
  nearest clusters to the descriptors in the image.
\item {\bf GIST}: The images are processed using code from
  \citet{gist} to yield 960-dimensional GIST feature vectors
  characterizing properties like ``openness,'' ``roughness,''
  ``naturalness,'' and so on.
\end{itemize}
In addition to the five feature functions described above, we include
another five that are identical but focus only on the center of the
image, defined as the centered rectangle with dimensions half those of
the original image.  This gives our first ten kernels.  We then create
45 pairwise combination kernels by concatenating every possible pair
of the 10 basic feature vectors.  This technique produces kernels that
synthesize more than one source of information, offering greater
flexibility.

Finally, we augment our kernels by adding a constant hyperparameter
$\rho$ to each entry.  $\rho$ acts a knob for controlling the overall
preference for diversity; as $\rho$ increases, all images appear more
similar, thus increasing repulsion.  In our experiments, $\rho$ is
chosen independently for each method and each category to optimize
performance on the training set.

\subsubsection{Methods}

We test four different methods.  Two use $k$-DPPs, and two are derived
from Maximum Marginal Relevance (MMR) \citep{carbonell1998use}.  For
each approach, we test both the single best expert kernel on the
training data and a learned combination of kernels.  All
methods were tuned separately for each of the three query categories.
On each run a random 25\% of the labeled examples are reserved for
testing, and the remaining 75\% form the training set used for
setting hyperparameters and training.  Recall that $Y_t$ is the
five-image partial result set for instance $t$, and let $C_t =
\{i^+_t,i^-_t\}$ denote the set of two candidates images, where
$i^+_t$ is the candidate preferred by the human judges.

\paragraph{Best $k$-DPP}

Given a single kernel $L$, the $k$-DPP prediction is
\begin{equation}
  k\mathrm{DPP}_t = \argmax_{i \in C_t}\ \P^6_L(Y_t \cup \{i\})\,.
\end{equation}
We select the kernel with the best zero-one accuracy on the training
set, and apply it to the test set.

\paragraph{Mixture of $k$-DPPs}

We apply our learning method to the full set of 55 kernels, optimizing
\eqref{optim} on the training set to obtain a 55-dimensional mixture
vector $\theta$.  We set $\gamma$ to minimize the zero-one training
loss.  We then take the learned $\theta$ and apply it to making
predictions on the test set:
\begin{equation}
  k\mathrm{DPPmix}_t = \argmax_{i \in C_t}\ 
  \sum_{l=1}^{55} \theta_l\P^6_{L_l}(Y_t \cup \{i\})\,.
\end{equation}

\paragraph{Best MMR} 

Recall that MMR is a standard, heuristic technique for generating
diverse sets of search results.  The idea is to build a set
iteratively by adding on each round a result that maximizes a weighted
combination of relevance (with respect to the query) and diversity,
measured as the maximum similarity to any of the previously selected
results.  (See \secref{documentsummarization} for more details about
MMR.)  For our experiments, we assume relevance is uniform; hence we
merely need to decide which of the two candidates has the smaller
maximum similarity to the partial result set.  Thus, for a given
kernel $L$, the MMR prediction is
\begin{equation}
  \mathrm{MMR}_t = \argmin_{i \in C_t} \left[ \max_{j \in Y_t} L_{ij} \right]\,.
\end{equation}
As for the $k$-DPP, we select the single best kernel on the training
set, and apply it to the test set.

\paragraph{Mixture MMR} 

We can also attempt to learn a mixture of similarity kernels for MMR.
We use the same training approach as for $k$-DPPs, but replace the
probability score $P^k_\theta(Y_y \cup \{i\})$ with the negative cost
\begin{equation}
  -c_\theta(Y_t,i) = -\max_{j \in Y_t} \sum_{l=1}^D \theta_l [L_l]_{ij}\,,
\end{equation}
which is just the negative similarity of item $i$ to the set $Y_t$
under the combined kernel metric.  Significantly, this substitution
makes the optimization non-smooth and non-convex, unlike the $k$-DPP
optimization.  In practice this means that the global optimum is not easily
found.  However, even a local optimum may provide advantages over the
single best kernel.  In our experiments we use the local optimum found
by projected gradient descent starting from the uniform kernel
combination.

\subsubsection{Results}

\tabref{imresults} shows the mean zero-one accuracy of each method for
each query category, averaged over 100 random train/test splits.
Statistical significance is computed by bootstrapping.  Regardless of
whether we learn a mixture, $k$-DPPs outperform MMR on two of the
three categories, significant at 99\% confidence.  In all cases, the
learned mixture of $k$-DPPs achieves the best performance.  Note that,
because the decision being made for each instance is binary, 50\% is
equivalent to random performance.  Thus the numbers in
\tabref{imresults} suggest that this is a rather difficult task, a
conclusion supported by the rates of noise exhibited by the human
judges.  However, the changes in performance due to learning and the
use of $k$-DPPs are more obviously significant when measured as
improvements above this baseline level.  For example, in the cars
category our mixture of $k$-DPPs performs 14.58 percentage points
better than random, versus 9.59 points for MMR with a mixture of
kernels.  \figref{kdppsamples} shows some actual samples drawn using
the $k$-DPP sampling algorithm.

\begin{table}
  \centering
  \begin{tabular}{lcccc}
    & Best & Best & Mixture & Mixture\\
    Category        & MMR   & $k$-DPP & MMR   & $k$-DPP \\     
    \hline
    {\sc cars}        & 55.95 & 57.98   & 59.59 & {\bf 64.58} \\
    {\sc cities}      & 56.48 & 56.31   & 60.99 &      61.29  \\
    {\sc dogs}        & 56.23 & 57.70   & 57.39 & {\bf 59.84} \\
    \hline
  \end{tabular}
  \caption[Image search results]
    {Percentage of real-world image search examples judged the
    same way as the majority of human annotators.  Bold results are
    significantly higher than others in the same row with 99\%
    confidence.}  
  \tablabel{imresults}
\end{table}

\begin{figure}
\centering
\includegraphics[width=\textwidth]
                {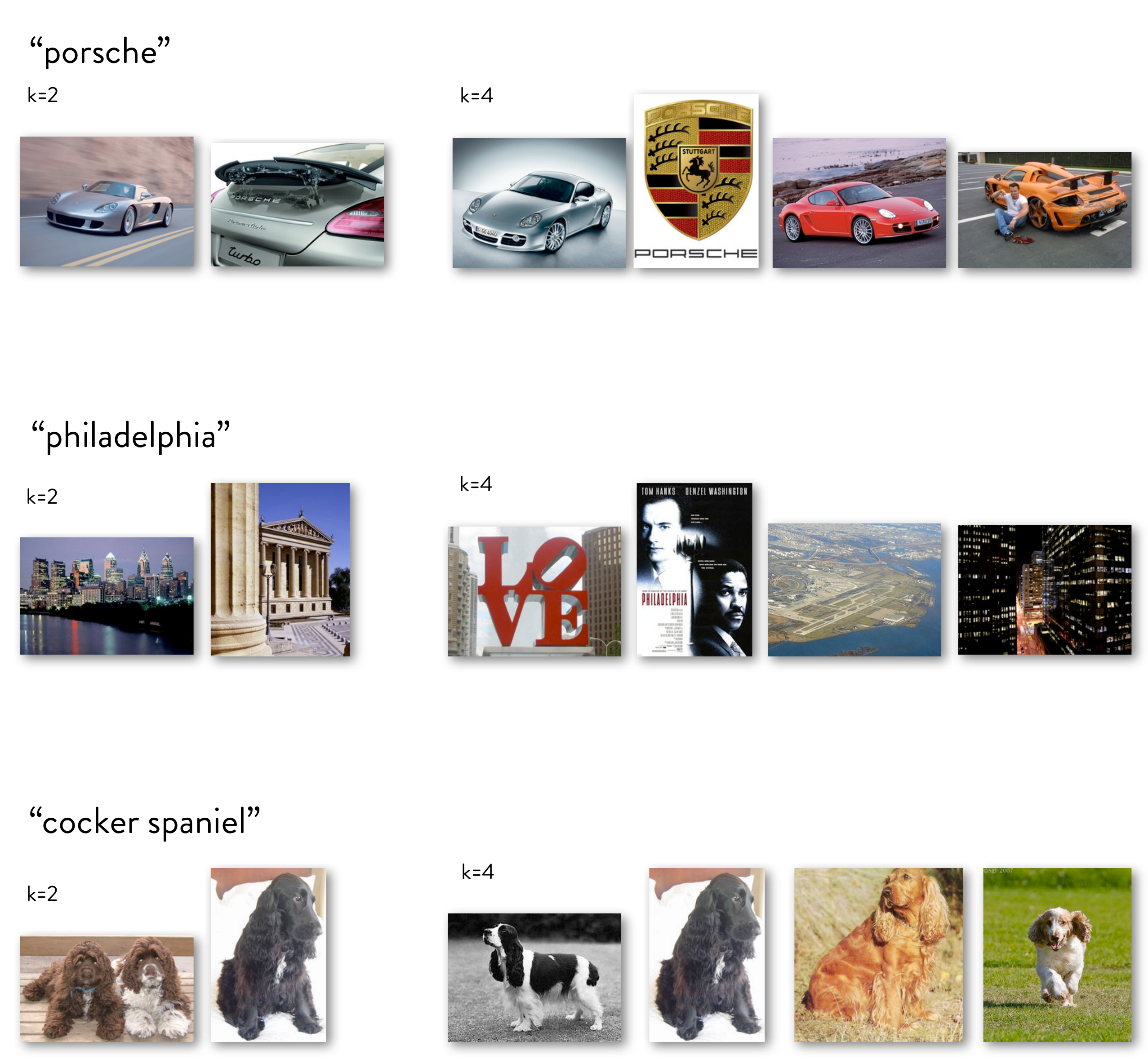}
\caption[Samples from $k$-DPP mixture model]
  {Samples from the $k$-DPP mixture model.}
\figlabel{kdppsamples}
\end{figure}

\tabref{topfeatures} shows, for the $k$-DPP mixture model, the kernels
receiving the highest weights for each search category (on average
over 100 train/test splits).  Combined-feature kernels appear to be
useful, and the three categories exhibit significant differences in
what annotators deem diverse, as we might expect.

\begin{table}
  \centering
  \begin{tabular}{lcr}
    \hline
    & color-8-center \& sift-256 &(0.13) \\
    {\sc cars} & color-8-center \& sift-512 &(0.11)\\
    & color-8-center &(0.07)\\
    \hline
    & sift-512-center &(0.85)\\
    {\sc cities} & gist &(0.08)\\
    & color-8-center \& gist &(0.03)\\
    \hline
    & color-8-center &(0.39)\\
    {\sc dogs} & color-8-center \& sift-512 &(0.21)\\
    & color-8-center \& sift-256 &(0.20)\\
    \hline
  \end{tabular}
  \caption[Highest-weighted kernels]
    {Kernels receiving the highest average weights for each
    category (shown in parentheses).  Ampersands indicate kernels
    generated from pairs of feature functions.}
  \tablabel{topfeatures}
\end{table}

We can also return to our original motivation and try to measure how
well each method ``covers'' the space of likely user intentions.
Since we do not have access to real users who are searching for the
queries in our dataset, we instead simulate them by imagining that
each is looking for a particular target image drawn randomly from the
images in our collection.  For instance, given the query
``philadelphia'' we might draw a target image of the Love sculpture,
and then evaluate each method on whether it selects an image of the
Love sculpture, i.e., whether it satisfies that virtual user.  More
generally, we will simply record the maximum similarity of any image
in the result set to the target image.  We expect better methods to
show higher similarity when averaged over a large number of such
users.

\begin{table}
  \centering
  \begin{tabular}{lccc}
    & Single kernel & Uniform & MMR \\
    Category & (average) & mixture & mixture \\
    \hline
    {\sc cars  } & 57.58 & 68.31 & 58.15 \\
    {\sc cities} & 59.00 & 64.76 & 62.32 \\
    {\sc dogs  } & 57.78 & 62.12 & 57.86 \\
    \hline
  \end{tabular}
  \caption[Coverage results]
    {The percentage of virtual users whose desired image is more
    similar to the $k$-DPP results than the MMR results.  Above 50
    indicates better $k$-DPP performance; below 50 indicates better
    MMR performance.  The results for the 55 individual expert kernels
    are averaged in the first column.}  \tablabel{coverage}
\end{table}

We consider only the mixture models here, since they perform best.
For each virtual user, we sample a ten-image result set $Y_\dpp$ using
the mixture $k$-DPP, and select a second ten-image result set $Y_\mmr$
using the mixture MMR.  For MMR, the first image is selected uniformly
at random, since they are assumed to be uniformly relevant.
Subsequent selections are deterministic.  Given a target image $i$
drawn uniformly at random, we then compute similarities
\begin{align}
  s_\dpp(i) &= \max_{j \in Y_\dpp} L_{ij} &   s_\mmr(i) &= \max_{j \in Y_\mmr} L_{ij}
\end{align}
for a particular similarity kernel $L$.  We report the fraction of the
time that $s_\dpp(i) > s_\mmr(i)$; that is, the fraction of the time
that our virtual user would be better served by the $k$-DPP model.
Because we have no gold standard kernel $L$ for measuring similarity,
we try several possibilities, including all 55 expert kernels, a
uniform combination of the expert kernels, and the combination learned
by MMR.  (Note that the mixture $k$-DPP does not learn a kernel
combination, hence there is no corresponding mixture to try here.)
\tabref{coverage} shows the results, averaged across all of the
virtual users (i.e., all the images in our collection).  Even when
using the mixture learned to optimize MMR itself, the $k$-DPP does a
better job of covering the space of possible user intentions.  All
results in \tabref{coverage} are significantly higher than 50\% at
99\% confidence.

\section{Structured DPPs}
\seclabel{sdpps}

We have seen in the preceding sections that DPPs offer polynomial-time
inference and learning with respect to $N$, the number of items in the
ground set $\Y$.  This is important since DPPs model an exponential
number of subsets $Y \subseteq \Y$, so naive algorithms would be
intractable.  And yet, we can imagine DPP applications for which even
linear time is too slow.  For example, suppose that after modeling the
positions of basketball players, as proposed in the previous section,
we wanted to take our analysis one step further.  An obvious extension
is to realize that a player does not simply occupy a single position,
but instead moves around the court over time.  Thus, we might want to
model not just diverse sets of positions on the court, but diverse
sets of {\it paths} around the court during a game.  While we could
reasonably discretize the possible court positions to a manageable
number $\M$, the number of paths over, say, 100 time steps would be
$\M^{100}$, making it almost certainly impossible to enumerate them
all, let alone build an $\M^{100} \times \M^{100}$ kernel matrix.

However, in this combinatorial setting we can take advantage of the
fact that, even though there are exponentially many paths, they are
{\it structured}; that is, every path is built from a small number of
the same basic components.  This kind of structure has frequently been
exploited in machine learning, for example, to find the best
translation of a sentence, or to compute the marginals of a Markov
random field.  In such cases structure allows us to factor
computations over exponentially many possibilities in an efficient
way.  And yet, the situation for structured DPPs is even worse: when
the number of items in $\Y$ is exponential, we are actually modeling a
distribution over the {\it doubly exponential} number of subsets of an
exponential $\Y$.  If there are $\M^{100}$ possible paths, there are
$2^{\M^{100}}$ subsets of paths, and a DPP assigns a probability to
every one.  This poses an extreme computational challenge.

In order to develop efficient structured DPPs (SDPPs), we will
therefore need to combine the dynamic programming techniques used for
standard structured prediction with the algorithms that make DPP
inference efficient.  We will show how this can be done by applying
the dual DPP representation from \secref{dual_dpps}, which shares
spectral properties with the kernel $L$ but is manageable in size, and
the use of second-order message passing, where the usual sum-product
or min-sum semiring is replaced with a special structure that computes
quadratic quantities over a factor graph \citep{li2009first}.  In the
end, we will demonstrate that it is possible to normalize and sample
from an SDPP in polynomial time.

Structured DPPs open up a large variety of new possibilities for
applications; they allow us to model diverse sets of essentially any
structured objects.  For instance, we could find not only the best
translation but a diverse set of high-quality translations for a
sentence, perhaps to aid a human translator.  Or, we could study the
distinct proteins coded by a gene under alternative RNA splicings,
using the diversifying properties of DPPs to cover the large space of
possibilities with a small representative set.  Later, we will apply
SDPPs to three real-world tasks: identifying multiple human poses in
images, where there are combinatorially many possible poses, and we
assume that the poses are diverse in that they tend not to overlap;
identifying salient lines of research in a corpus of computer science
publications, where the structures are citation chains of important
papers, and we want to find a small number of chains that covers the
major topic in the corpus; and building threads from news text, where
the goal is to extract from a large corpus of articles the most
significant news stories, and for each story present a sequence of
articles covering the major developments of that story through time.

We begin by defining SDPPs and stating the structural assumptions that
are necessary to make inference efficient; we then show how these
assumptions give rise to polynomial-time algorithms using second order
message passing.  We discuss how sometimes even these polynomial
algorithms can be too slow in practice, but demonstrate that by
applying the technique of random projections (\secref{projection}) we
can dramatically speed up computation and reduce memory use while
maintaining a close approximation to the original model
\citep{kulesza2010sdpps}.  Finally, we show how SDPPs can be applied
to the experimental settings described above, yielding improved
results compared with a variety of standard and heuristic baseline
approaches.

\subsection{Factorization}
\seclabel{sdpp_model}

In \secref{dpp_inference} we saw that DPPs remain tractable on modern
computers for $N$ up to around 10,000.  This is no small feat, given
that the number of subsets of 10,000 items is roughly the number of
particles in the observable universe to the 40th power.  Of course,
this is not magic but simply a consequence of a certain type of {\it
  structure}; that is, we can perform inference with DPPs because the
probabilities of these subsets are expressed as combinations of only a
relatively small set of $O(N^2)$ parameters.  In order to make the jump
now to ground sets $\Y$ that are exponentially large, we will need to
make an similar assumption about the structure of $\Y$ itself.  Thus,
a structured DPP (SDPP) is a DPP in which the ground set $\Y$ is given
implicitly by combinations of a set of {\it parts}.  For instance, the
parts could be positions on the court, and an element of $\Y$ a
sequence of those positions.  Or the parts could be rules of a
context-free grammar, and then an element of $\Y$ might be a complete
parse of a sentence.  This assumption of structure will give us the
algorithmic leverage we need to efficiently work with a distribution
over a doubly exponential number of possibilities.

Because elements of $\Y$ are now structures, we will no longer think
of $\Y = \{1,2,\dots,N\}$; instead, each element $\y \in \Y$ is a
structure given by a sequence of $R$ parts
$(y_{1},y_{2},\dots,y_{R})$, each of which takes a value from a finite
set of $\M$ possibilities.  For example, if $\y$ is the path of a
basketball player, then $R$ is the number of time steps at which the
player's position is recorded, and $y_{r}$ is the player's discretized
position at time $r$.  We will use $\y_i$ to denote the $i$th structure
in $\Y$ under an arbitrary ordering; thus $\Y =
\{\y_1,\y_2,\dots,\y_N\}$, where $N = \M^R$.  The parts of $\y_i$ are
denoted $y_{ir}$.

An immediate challenge is that the kernel $L$, which has $N^2$
entries, can no longer be written down explicitly.  We therefore
define its entries using the quality/diversity decomposition presented
in \secref{decomposition}.  Recall that this decomposition gives the
entries of $L$ as follows:
\begin{equation}
  L_{ij} = q(\y_i) \phi(\y_i)^\trans \phi(\y_j) q(\y_j)\,, 
\end{equation}
where $q(\y_i)$ is a nonnegative measure of the quality of structure
$\y_i$, and $\phi(\y_i)$ is a $D$-dimensional vector of diversity
features so that $\phi(\y_i)^\trans\phi(\y_j)$ is a measure of the
similarity between structures $\y_i$ and $\y_j$.  We cannot afford to
specify $q$ and $\phi$ for every possible structure, but we can use
the assumption that structures are built from parts to define a
factorization, analogous to the factorization over cliques that gives
rise to Markov random fields.

Specifically, we assume that the model decomposes over a set of {\it
  factors} $F$, where a factor $\factor \in F$ is a small subset of the
parts of a structure.  (Keeping the factors small will ensure that the
model is tractable.)  We denote by $\y_{\factor}$ the collection of
parts of $\y$ that are included in factor $\factor$; then the
factorization assumption is that the quality score decomposes
multiplicatively over parts, and the diversity features decompose
additively:
\begin{align}
  q(\y) &= \prod_{\factor \in F} q_\factor(\y_{\factor})\\
  \phi(\y) &= \sum_{\factor \in F} \phi_\factor(\y_{\factor})\,.
  \eqlabel{factorization}
\end{align}
We argue that these are quite natural factorizations.  For instance,
in our player tracking example we might have a positional factor for
each time $r$, allowing the quality model to prefer paths that go
through certain high-traffic areas, and a transitional factor for each
pair of times $(r-1,r)$, allowing the quality model to enforce the
smoothness of a path over time.  More generally, if the parts
correspond to cliques in a graph, then the quality scores can be given
by a standard log-linear Markov random field (MRF), which defines a
multiplicative distribution over structures that give labelings of the
graph.  Thus, while in \secref{expressiveness} we compared DPPs and
MRFs as alternative models for the same binary labeling problems,
SDPPs can also be seen as an extension to MRFs, allowing us to take a
model of individual structures and use it as a quality measure for
modeling diverse sets of structures.

Diversity features, on the other hand, decompose additively, so we
can think of them as global feature functions defined by summing local
features, again as done in standard structured prediction.  For
example, $\phi_r(y_r)$ could track the coarse-level position of a
player at time $r$, so that paths passing through similar positions at
similar times are less likely to co-occur.  Note that, in contrast to
the unstructured case, we do not generally have $\Vert\phi(\y)\Vert =
1$, since there is no way to enforce such a constraint under the
factorization in \eqref{factorization}.  Instead, we simply set the
factor features $\phi_\factor(\y_{\factor})$ to have unit norm for all
$\factor$ and all possible values of $\y_\factor$.  This slightly
biases the model towards structures that have the same (or similar)
features at every factor, since such structures maximize
$\Vert\phi\Vert$.  However, the effect of this bias seems to be minor
in practice.

As for unstructured DPPs, the quality and diversity models combine to
produce balanced, high-quality, diverse results.  However, in the
structured case the contribution of the diversity model can be
especially significant due to the combinatorial nature of the items in
$\Y$.  For instance, imagine taking a particular high-quality path and
perturbing it slightly, say by shifting the position at each time step
by a small random amount.  This process results in a new and distinct
path, but is unlikely to have a significant effect on the overall
quality: the path remains smooth and goes through roughly the same
positions.  Of course, this is not unique to the structured case; we
can have similar high-quality items in any DPP.  What makes the
problem especially serious here is that there is a combinatorial
number of such slightly perturbed paths; the introduction of structure
dramatically increases not only the number of items in $\Y$, but also
the number of subtle variations that we might want to suppress.
Furthermore, factored distributions over structures are often very
peaked due to the geometric combination of quality scores across many
factors, so variations of the most likely structure can be much more
probable than any real alternative.  For these reasons independent
samples from an MRF can often look nearly identical; a sample from an
SDPP, on the other hand, is much more likely to contain a truly
diverse set of structures.

\subsubsection{Synthetic example: particle tracking}
\seclabel{tracking}

Before describing the technical details needed to make SDPPs
computationally efficient, we first develop some intuition by studying
the results of the model as applied to a synthetic motion tracking
task, where the goal is to follow a collection of particles as they
travel in a one-dimensional space over time.  This is essentially a
simplified version of our player tracking example, but with the motion
restricted to a line.  We will assume that a path $\y$ has 50 parts,
where each part $y_{r} \in \{1,2,\dots,50\}$ is the particle's position
at time step $r$ discretized into one of 50 locations.  The total
number of possible trajectories in this setting is $50^{50}$, and we
will be modeling $2^{50^{50}}$ possible sets of trajectories.  We
define positional and transitional factors
\begin{equation}
  F = \{\{r\}\ |\ r = 1,2,\dots,50\} \cup 
  \{\{r-1,r\}\ |\ r = 2,3,\dots,50\}\,.
\end{equation}

While a real tracking problem would involve quality scores $q(\y)$ that
depend on some observations---for example, measurements over time from
a set of physical sensors, or perhaps a video feed from a basketball
game---for simplicity we determine the quality of a trajectory here
using only its starting position and a measure of smoothness over
time.  Specifically, we have
\begin{equation}
  q(\y) = q_1(y_{1})\prod_{r=2}^{50}q(y_{r-1},y_{r})\,,
\end{equation}
where the initial quality score $q_1(y_{1})$ is given by a smooth
trimodal function with a primary mode at position 25 and secondary
modes at positions 10 and 40, depicted by the blue curves on the left
side of \figref{trajectory_samples}, and the quality scores for all
other positional factors are fixed to one and have no effect.  The
transition quality is the same at all time steps, and given by
$q(y_{r-1},y_{r}) = f_\N(y_{r-1}-y_{r})$, where $f_\N$ is the density
function of the normal distribution; that is, the quality of a
transition is maximized when the particle does not change location,
and decreases as the particle moves further and further from its
previous location.  In essence, high quality paths start near the
central position and move smoothly through time.

We want trajectories to be considered similar if they travel through
similar positions, so we define a 50-dimensional diversity feature
vector as follows:
\begin{align}
  \phi(\y) &= \sum_{r=1}^{50} \phi_r(y_{r})\\
  \phi_{rl}(y_{r}) &\propto f_\N(l - y_{r}),\quad l=1,2,\dots,50\,.
\end{align}
Intuitively, feature $l$ is activated when the trajectory passes near
position $l$, so trajectories passing through nearby positions will
activate the same features and thus appear similar in the diversity
model.  Note that for simplicity, the time at which a particle reaches
a given position has no effect on the diversity features.  The
diversity features for the transitional factors are zero and have no
effect.

We use the quality and diversity models specified above to define our
SDPP.  In order to obtain good results for visualization, we scale the
kernel so that the expected number of trajectories in a sample from
the SDPP is five.  We then apply the algorithms developed later to
draw samples from the model.  The first row of
\figref{trajectory_samples} shows the results, and for comparison each
corresponding panel on the second row shows an equal number of
trajectories sampled independently, with probabilities proportional to
their quality scores.  As evident from the figure, trajectories
sampled independently tend to cluster in the middle region due to the
strong preference for this starting position.  The SDPP samples,
however, are more diverse, tending to cover more of the space while
still respecting the quality scores---they are still smooth, and still
tend to start near the center.

\begin{figure}
  \centering
  \includegraphics[width=0.605in]{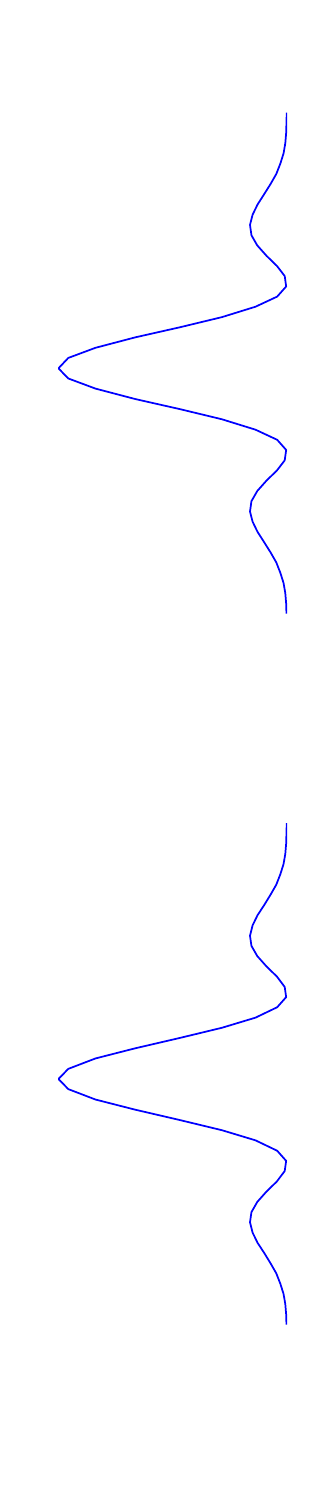}
  \includegraphics[clip,trim=1in 0 0.8in 0,width=5.1in]
                  {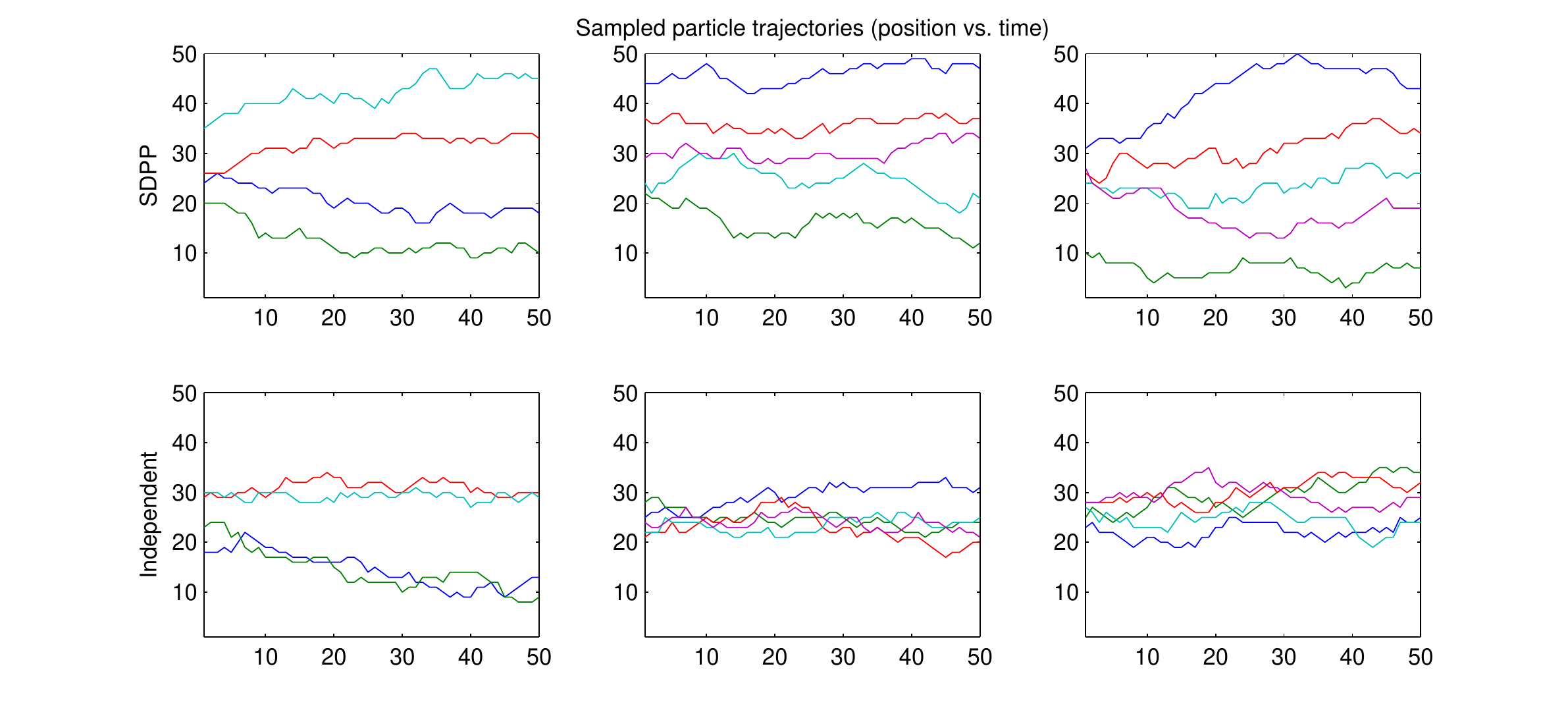}
  \caption[Particle trajectories from SDPP]
    {Sets of particle trajectories sampled from an SDPP (top row)
    and independently using only quality scores (bottom row). The curves
    to the left indicate quality scores for the initial positions of the
    particles.}
  \figlabel{trajectory_samples}
\end{figure}

\subsection{Second-order message passing}

The central computational challenge for SDPPs is the fact that $N =
M^R$ is exponentially large, making the usual inference algorithms
intractable.  However, we showed in \secref{dual_dpps} that DPP
inference can be recast in terms of a smaller dual representation $C$;
recall that, if $B$ is the $D \times N$ matrix whose columns are given
by $B_{\y_i} = q(\y_i)\phi(\y_i)$, then $L = B^\trans B$ and
\begin{align}
  C &= BB^\trans\\
  &= \sum_{\y \in \Y} q^2(\y) \phi(\y)\phi(\y)^\trans\,.
  \eqlabel{Cexpanded}
\end{align}

Of course, for the dual representation to be of any use we must be
able to efficiently compute $C$.  If we think of
$q^2_\factor(\y_{\factor})$ as the factor potentials of a graphical
model $p(\y)\propto\prod_{\factor\in F} q^2_\factor(\y_{\factor})$,
then computing $C$ is equivalent to computing second moments of the
diversity features under $p$ (up to normalization).  Since the
diversity features factor additively, $C$ is quadratic in the local
diversity features $\phi_\factor(\y_{\factor})$.  Thus, we could
naively calculate $C$ by computing the pairwise marginals
$p(\y_{\factor},\y_{\factor'})$ for all realizations of the factors
$\factor,\factor'$ and, by linearity of expectations, adding up their
contributions:
\begin{equation}
  C \propto \sum_{\factor,\factor'}\sum_{\y_{\factor},\y_{\factor'}}
  p(\y_{\factor},\y_{\factor'})
  \phi_\factor(\y_{\factor})\phi_{\factor'}(\y_{\factor'})^\trans\,,
\end{equation}
where the proportionality is due to the normalizing constant of
$p(\y)$.  However, this sum is quadratic in the number of factors and
their possible realizations, and can therefore be expensive when
structures are large.

Instead, we can substitute the factorization from
\eqref{factorization} into \eqref{Cexpanded} to obtain
\begin{equation}
  C = \sum_{\y\in\Y} 
  \left(\prod_{\factor\in F} q^2_\factor(\y_{\factor})\right)
  \left(\sum_{\factor\in F} \phi_\factor(\y_{\factor})\right)
  \left(\sum_{\factor\in F} \phi_\factor(\y_{\factor})\right)^\trans\,.
  \eqlabel{sdpp_C}
\end{equation}
It turns out that this expression is computable in linear time using a
second-order message passing algorithm.


Second-order message passing was first introduced by
\citet{li2009first}.  The main idea is to compute second-order
statistics over a graphical model by using the standard belief
propagation message passing algorithm, but with a special semiring in
place of the usual sum-product or max-product.  This substitution
makes it possible to compute quantities of the form
\begin{equation}
  \sum_{\y\in\Y}
  \left(\prod_{\factor\in F} p_\factor(\y_{\factor})\right)
  \left(\sum_{\factor\in F} a_\factor(\y_{\factor})\right) 
  \left(\sum_{\factor\in F} b_\factor(\y_{\factor})\right)\,,
  \eqlabel{semiringbp}
\end{equation}
where $p_\factor$ are nonnegative and $a_\factor$ and $b_\factor$ are
arbitrary functions.  Note that we can think of $p_\factor$ as
defining a multiplicatively decomposed function
\begin{equation}
  p(\y) = \prod_{\factor\in F} p_\factor(\y_{\factor})\,,
\end{equation}
and $a_\factor$ and $b_\factor$ as defining corresponding additively
decomposed functions $a$ and $b$.

We begin by defining the notion of a factor graph, which provides the
structure for all message passing algorithms.  We then describe
standard belief propagation on factor graphs, and show how it can be
defined in a general way using semirings.  Finally we demonstrate that
belief propagation using the semiring proposed by \citet{li2009first}
computes quantities of the form in \eqref{semiringbp}.

\subsubsection{Factor graphs}
\seclabel{factor_graphs}

Message passing operates on {\it factor graphs}.  A factor graph is an
undirected bipartite graph with two types of vertices: {\it variable}
nodes and {\it factor} nodes.  Variable nodes correspond to the parts
of the structure being modeled; for the SDPP setup described above, a
factor graph contains $R$ variable nodes, each associated with a
distinct part $r$.  Similarly, each factor node corresponds to a
distinct factor $\factor \in F$.  Every edge in the graph connects a
variable node to a factor node, and an edge exists between variable
node $r$ and factor node $\factor$ if and only if $r \in \factor$.
Thus, the factor graph encodes the relationships between parts and
factors.  \figref{factor_graph} shows an example factor graph for the
tracking problem from \secref{tracking}.

\begin{figure}
  \centering
  \begin{tikzpicture}[auto,level/.style={sibling distance=60mm/#1}]
    \matrix [row sep=0.5cm,column sep=1cm] {
      \node (f1) [draw,minimum size=0.5cm] {}; &&
      \node (f2) [draw,minimum size=0.5cm] {}; &&
      \node (f3) [draw,minimum size=0.5cm] {}; &&
      \node (fR) [draw,minimum size=0.5cm] {};
      \\
      \node (y1) [circle,draw] {$y_1$}; &
      \node (f12) [draw,minimum size=0.5cm] {}; &
      \node (y2) [circle,draw] {$y_2$}; &
      \node (f23) [draw,minimum size=0.5cm] {}; &
      \node (y3) [circle,draw] {$y_3$}; &
      \node (dots) {$\quad\cdots\quad$}; &
      \node (yR) [circle,draw] {$y_R$};
      \\
    };
    \path {
      (f1) edge (y1)
      (f2) edge (y2)
      (f3) edge (y3)
      (fR) edge (yR)
      (f12) edge (y1)
      (f12) edge (y2)
      (f23) edge (y2)
      (f23) edge (y3)
      (dots) edge (y3)
      (dots) edge (yR)
    };
  \end{tikzpicture}
  \caption[Factor graphs]
    {A sample factor graph for the tracking problem.  Variable
    nodes are circular, and factor nodes are square.  Positional
    factors that depend only on a single part appear in the top row;
    binary transitional factors appear between parts in the second
    row.}  
  \figlabel{factor_graph}
\end{figure}
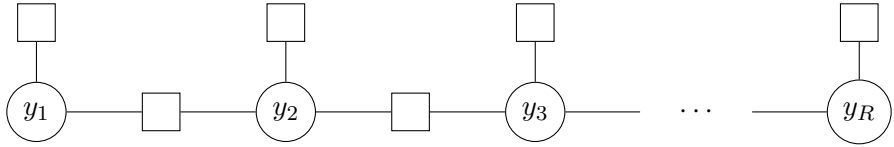

It is obvious that the computation of \eqref{semiringbp} cannot be
efficient when factors are allowed to be arbitrary, since in the limit
a factor could contain all parts and we could assign arbitrary values
to every configuration $\y$.  Thus we will assume that the degree of
the factor nodes is bounded by a constant $c$.  (In
\figref{factor_graph}, as well as all of the experiments we run, we
have $c=2$.)  Furthermore, message passing algorithms are efficient
whenever the factor graph has low treewidth, or, roughly, when only
small sets of nodes need to be merged to obtain a tree.  Going forward
we will assume that the factor graph is a tree, since any
low-treewidth factor graph can be converted into an equivalent factor
tree with bounded factors using the junction tree algorithm
\citep{lauritzen1988local}.

\subsubsection{Belief propagation}

We now describe the basic belief propagation algorithm, first
introduced by \citet{pearl1982reverend}.  Suppose each factor has an
associated real-valued weight function $w_\factor(\y_\factor)$, giving
rise to the multiplicatively decomposed global weight function
\begin{equation}
  w(\y) = \prod_{\factor \in F} w_\factor(\y_\factor)\,.
\end{equation}
Then the goal of belief propagation is to efficiently compute sums of
$w(\y)$ over combinatorially large sets of structures $\y$.

We will refer to a structure $\y$ as an {\it assignment} to the
variable nodes of the factor graph, since it defines a value $y_{r}$
for every part.  Likewise we can think of $\y_{\factor}$ as an
assignment to the variable nodes adjacent to $\factor$, and $y_{r}$ as
an assignment to a single variable node $r$.  We use the notation
$\y_{\factor} \sim y_{r}$ to indicate that $\y_{\factor}$ is consistent
with $y_{r}$, in the sense that it assigns the same value to variable
node $r$.  Finally, denote by $F(r)$ the set of factors in which
variable $r$ participates.

The belief propagation algorithm defines recursive message functions
$m$ to be passed along edges of the factor graph; the formula for the
message depends on whether it is traveling from a variable node to a
factor node, or vice versa:
\begin{itemize}
\item From a variable $r$ to a factor $\factor$:
  \begin{equation}
  m_{r\to\factor}(y_r) = 
  \prod_{\factor' \in F(r) - \{\factor\}}m_{\factor'\to r}(y_r)
  \end{equation}
\item From a factor $\factor$ to a variable $r$:
  \begin{equation}
  m_{\factor\to r}(y_r) = 
  \sum_{\y_\factor \sim y_r} \left[ w_\factor(\y_\factor) 
    \prod_{r' \in \factor - \{r\}} m_{r'\to\factor}(y_{r'}) \right]
  \end{equation}
\end{itemize}
Intuitively, an outgoing message summarizes all of the messages
arriving at the source node, excluding the one coming from the target
node.  Messages from factor nodes additionally incorporate information
about the local weight function.

Belief propagation passes these messages in two phases based on an
arbitrary orientation of the factor tree.  In the first phase, called
the forward pass, messages are passed upwards from the leaves to the
root.  In the second phase, or backward pass, the messages are passed
downward, from the root to the leaves.  Upon completion of the second
phase one message has been passed in each direction along every edge in
the factor graph, and it is possible to prove using an inductive
argument that, for every $y_r$,
\begin{equation}
  \prod_{\factor \in F(r)} m_{\factor \to t}(y_r) = 
  \sum_{\y \sim y_r} \prod_{\factor \in F} w_\factor(\y_\factor)\,.
  \eqlabel{beliefs}
\end{equation}
If we think of the $w_\factor$ as potential functions, then
\eqref{beliefs} gives the (unnormalized) marginal probability of the
assignment $y_r$ under a Markov random field.  

Note that the algorithm passes two messages per edge in the factor
graph, and each message requires considering at most $M^c$
assignments, therefore its running time is $O(\M^c R)$.  The sum on
the right-hand side of \eqref{beliefs}, however, is exponential in the
number of parts.  Thus belief propagation offers an efficient means of
computing certain combinatorial quantities that would naively require
exponential time.

\subsubsection{Semirings}

In fact, the belief propagation algorithm can be easily generalized to
operate over an arbitrary semiring, thus allowing the same basic
algorithm to perform a variety of useful computations.  Recall that a
semiring $\langle W,\oplus,\otimes,\bzero,\bone\rangle$ comprises a
set of elements $W$, an addition operator $\oplus$, a multiplication
operator $\otimes$, an additive identity $\bzero$, and a
multiplicative identity $\bone$ satisfying the following requirements
for all $a,b,c\in W$:
\begin{itemize}
  \item Addition is associative and commutative, with identity
    $\bzero$:
    \begin{align}
      a\oplus (b\oplus c) &= (a\oplus b)\oplus c\\
      a\oplus b &= b\oplus a \\
      a\oplus \bzero &= a
    \end{align}
  \item Multiplication is associative, with identity $\bone$:
    \begin{align}
      a\otimes (b\otimes c) &= (a\otimes b)\otimes c\\
      a\otimes \bone &= \bone\otimes a = a
    \end{align}
  \item Multiplication distributes over addition:
    \begin{align}
      a\otimes(b\oplus c) &= (a\otimes b) \oplus (a\otimes c)\\
      (a\oplus b)\otimes c &= (a\otimes c) \oplus (b\otimes c)
    \end{align}
  \item $\bzero$ is absorbing under multiplication:
    \begin{align}
      a\otimes \bzero &= \bzero\otimes a = \bzero
    \end{align}
\end{itemize}
Obviously these requirements are met when $W = \reals$ and
multiplication and addition are the usual arithmetic operations; this
is the standard sum-product semiring.  We also have, for example, the
max-product semiring, where $W = [0,\infty)$, addition is given by the
maximum operator with identity element 0, and multiplication is as
before.

We can rewrite the messages defined by belief propagation in terms of
these more general operations.  For $w_\factor(\y_\factor) \in W$, we
have
\begin{align}
  m_{r\to\factor}(y_r) &= 
  \bigotimes_{\factor' \in F(r) - \{\factor\}}m_{\factor'\to r}(y_r)
  \eqlabel{messages1}\\
  m_{\factor\to r}(y_r) &= 
  \bigoplus_{\y_\factor \sim y_r} \left[ w_\factor(\y_\factor) \otimes
    \bigotimes_{r' \in \factor - \{r\}} m_{r'\to\factor}(y_{r'}) \right]\,.
  \eqlabel{messages2}
\end{align}

As before, we can pass messages forward and then backward through the
factor tree.  Because the properties of semirings are sufficient to
preserve the inductive argument, we then have the following analog of
\eqref{beliefs}:
\begin{equation}
  \bigotimes_{\factor \in F(r)} m_{\factor \to r}(y_r) = 
  \bigoplus_{\y \sim y_r} \bigotimes_{\factor\in F} w_\factor(\y_\factor)\,.
  \eqlabel{semibeliefs}
\end{equation}
We have seen that \eqref{semibeliefs} computes marginal probabilities
under the sum-product semiring, but other semirings give rise to
useful results as well.  Under the max-product semiring, for instance,
\eqref{semibeliefs} is the so-called max-marginal---the maximum
unnormalized probability of any single assignment $\y$ consistent with
$y_r$.  In the next section we take this one step further, and show
how a carefully designed semiring will allow us to sum second-order
quantities across exponentially many structures $\y$.

\subsubsection{Second-order semiring}

\citet{li2009first} proposed the following second-order semiring over
four-tuples $(q,\phi,\psi,c) \in W = \reals^4$:
\begin{align}
  (q_1,\phi_1,\psi_1,c_1) \oplus (q_2,\phi_2,\psi_2,c_2) 
  &= (q_1+q_2,\ \phi_1+\phi_2,\ \psi_1+\psi_2,\ c_1+c_2) \\
  (q_1,\phi_1,\psi_1,c_1) \otimes (q_2,\phi_2,\psi_2,c_2) 
  &= (q_1q_2,\ q_1\phi_2+q_2\phi_1,\ q_1\psi_2+q_2\psi_1,\nonumber\\
  &\quad\quad q_1c_2 + q_2c_1+\phi_1\psi_2+\phi_2\psi_1)\\
  \bzero &= (0,0,0,0)\\
  \bone &= (1,0,0,0)
\end{align}
It is easy to verify that the semiring properties hold for these
operations.  Now, suppose that the weight function for a factor
$\factor$ is given by
\begin{equation}
  w_\factor(\y_\factor) =
  (p_\factor(\y_\factor),\ p_\factor(\y_\factor)a_\factor(\y_\factor),
  \ p_\factor(\y_\factor)b_\factor(\y_\factor),
  \ p_\factor(\y_\factor)a_\factor(\y_\factor)b_\factor(\y_\factor))\,,
\end{equation}
where $p_\factor$, $a_\factor$, and $b_\factor$ are as before.  Then
$w_\factor(\y_\factor) \in W$, and we can get some intuition about the
multiplication operator by observing that the fourth component of
$w_\factor(\y_\factor) \otimes w_{\factor'}(\y_{\factor'})$ is
\begin{align}
  p_\factor(\y_{\factor})&
  \left[p_{\factor'}(\y_{\factor'})a_{\factor'}(\y_{\factor'})
    b_{\factor'}(\y_{\factor'})\right]
  + p_{\factor'}(\y_{\factor'})
  \left[p_\factor(\y_{\factor})a_\factor(\y_{\factor})b_\factor(\y_{\factor})\right]
  \nonumber\\
  &\quad+ \left[p_\factor(\y_{\factor})a_\factor(\y_{\factor})\right]
  \left[p_{\factor'}(\y_{\factor'})b_{\factor'}(\y_{\factor'})\right] 
  + \left[p_{\factor'}(\y_{\factor'})a_{\factor'}(\y_{\factor'})\right]
  \left[p_\factor(\y_{\factor})b_\factor(\y_{\factor})\right] \\
  &= p_\factor(\y_{\factor})p_{\factor'}(\y_{\factor'})
  \left[a_\factor(\y_{\factor}) + a_{\factor'}(\y_{\factor'})\right]
  \left[b_\factor(\y_{\factor}) + b_{\factor'}(\y_{\factor'})\right]\,.
\end{align}
In other words, multiplication in the second-order semiring combines
the values of $p$ multiplicatively and the values of $a$ and $b$
additively, leaving the result in the fourth component.  It is not
hard to extend this argument inductively and show that the fourth
component of $\bigotimes_{\factor \in F} w_\factor(\y_\factor)$ is
given in general by
\begin{equation}
  \left( \prod_{\factor\in F} p_\factor(\y_\factor) \right)
  \left( \sum_{\factor\in F} a_\factor(\y_\factor) \right)
  \left( \sum_{\factor\in F} b_\factor(\y_\factor) \right)\,.
\end{equation}
Thus, by \eqref{semibeliefs} and the definition of $\oplus$, belief
propagation with the second-order semiring yields messages that
satisfy
\begin{equation}
  \left[\bigotimes_{\factor \in F(r)} m_{\factor \to r}(y_r)\right]_4 = 
  \sum_{\y\sim y_r} \left( \prod_{\factor\in F} p_\factor(\y_\factor) \right)
  \left( \sum_{\factor\in F} a_\factor(\y_\factor) \right)
  \left( \sum_{\factor\in F} b_\factor(\y_\factor) \right)\,.
  \eqlabel{itworks}
\end{equation}
Note that multiplication and addition remain constant-time operations
in the second-order semiring, thus belief propagation can still be
performed in time linear in the number of factors.  In the following
section we will show that the dual representation $C$, as well as
related quantities needed to perform inference in SDPPs, takes the
form of \eqref{itworks}; thus second-order message passing will be an
important tool for efficient SDPP inference.

\subsection{Inference}
\seclabel{sdpp_inference}

The factorization proposed in \eqref{factorization} gives a concise
definition of a structured DPP for an exponentially large $\Y$;
remarkably, under suitable conditions it also gives rise to tractable
algorithms for normalizing the SDPP, computing marginals, and
sampling.  The only restrictions necessary for efficiency are the ones
we inherit from belief propagation: the factors must be of bounded size
so that we can enumerate all of their possible configurations, and
together they must form a low-treewidth graph on the parts of the
structure.  These are precisely the same conditions needed for
efficient graphical model inference \citep{koller2009probabilistic},
which is generalized by inference in SDPPs.

\subsubsection{Computing $C$}
\seclabel{sdpp_C}

As we saw in \secref{dual_dpps}, the dual representation $C$ is
sufficient to normalize and marginalize an SDPP in time constant in
$N$.  Recall from \eqref{sdpp_C} that the dual representation of an
SDPP can be written as
\begin{equation}
  C = \sum_{\y\in\Y} 
  \left(\prod_{\factor\in F} q^2_\factor(\y_{\factor})\right)
  \left(\sum_{\factor\in F} \phi_\factor(\y_{\factor})\right)
  \left(\sum_{\factor\in F} \phi_\factor(\y_{\factor})\right)^\trans\,,
\end{equation}
which is of the form required to apply second-order message passing.
Specifically, we can compute for each pair of diversity features
$(a,b)$ the value of
\begin{equation}
  \sum_{\y\in\Y} 
  \left(\prod_{\factor\in F} q^2_\factor(\y_{\factor})\right)
  \left(\sum_{\factor\in F} \phi_{\factor a}(\y_{\factor})\right)
  \left(\sum_{\factor\in F} \phi_{\factor b}(\y_{\factor})\right)
  \eqlabel{Ccomponent}
\end{equation}
by summing \eqref{itworks} over the possible assignments $y_r$, and
then simply assemble the results into the matrix $C$.  Since there are
$\frac{D(D+1)}{2}$ unique entries in $C$ and message passing runs in
time $O(\M^c R)$, computing $C$ in this fashion requires $O(D^2\M^c
R)$ time.

We can make several practical optimizations to this algorithm, though
they will not affect the asymptotic performance.  First, we note that
the full set of messages at {\it any} variable node $r$ is sufficient
to compute \eqref{Ccomponent}.  Thus, during message passing we need
only perform the forward pass; at that point, the messages at the root
node are complete and we can obtain the quantity we need.  This speeds
up the algorithm by a factor of two.  Second, rather than running
message passing $D^2$ times, we can run it only once using a
vectorized second-order semiring.  This has no effect on the total
number of operations, but can result in significantly faster
performance due to vector optimizations in modern processors.  The
vectorized second-order semiring is over four-tuples $(q,\phi,\psi,C)$
where $q\in \reals$, $\phi,\psi \in \reals^D$, and $C \in \reals^{D
  \times D}$, and uses the following operations:
\begin{align}
  (q_1,\phi_1,\psi_1,C_1) \oplus (q_2,\phi_2,\psi_2,C_2) 
  &= (q_1+q_2,\ \phi_1+\phi_2,\ \psi_1+\psi_2,\ C_1+C_2) \\
  (q_1,\phi_1,\psi_1,C_1) \otimes (q_2,\phi_2,\psi_2,C_2) 
  &= (q_1q_2,\ q_1\phi_2+q_2\phi_1,\ q_1\psi_2+q_2\psi_1,\nonumber\\
  &\quad\quad q_1C_2 + q_2C_1+\phi_1\psi_2^\trans+\phi_2\psi_1^\trans)\\
  \bzero &= (0,\bzero,\bzero,\bzero)\\
  \bone &= (1,\bzero,\bzero,\bzero)\,.
\end{align}
It is easy to verify that computations in this vectorized semiring are
identical to those obtained by repeated use of the scalar semiring.

Given $C$, we can now normalize and compute marginals for an SDPP
using the formulas in \secref{dual_dpps}; for instance
\begin{align}
  K_{ii} &= \sum_{n=1}^D \frac{\lambda_n}{\lambda_n+1} 
  \left(\frac{1}{\sqrt{\lambda_n}}B_i^\trans \cv_n\right)^2\\
  &= q^2(\y_i)\sum_{n=1}^D \frac{1}{\lambda_n+1} 
  (\phi(\y_i)^\trans\cv_n)^2 \,,
  \eqlabel{sdpp_marginal1}
\end{align}
where $C = \sum_{n=1}^D \lambda_n \cv_n \cv_n^\trans$ is an
eigendecomposition of $C$.

\paragraph{Part marginals}

The introduction of structure offers an alternative type of marginal
probability, this time not of structures $\y \in \Y$ but of single
part assignments.  More precisely, we can ask how many of the
structures in a sample from the SDPP can be expected to make the
assignment $\hat y_r$ to part $r$:
\begin{align}
  \mu_r(\hat y_r) 
  &= \E \left[\sum_{\y \in \Y} 
    \I(\y \in \bY \wedge y_{r} = \hat y_r)\right]\\
  &= \sum_{\y \sim \hat y_r} \P_L(\y \in \bY)\,.
\end{align}
The sum is exponential, but we can compute it efficiently using
second-order message passing.  We apply \eqref{sdpp_marginal1} to get
\begin{align}
  \sum_{\y \sim \hat y_r}\P_L(\y \in \bY)
  &= \sum_{\y \sim \hat y_r}
  q^2(\y)\sum_{n=1}^D \frac{1}{\lambda_n+1} (\phi(\y)^\trans\cv_n)^2\\
  &= \sum_{n=1}^D \frac{1}{\lambda_n+1} 
  \sum_{\y \sim \hat y_r} q^2(\y)(\phi(\y)^\trans\cv_n)^2\\
  &= \sum_{n=1}^D \frac{1}{\lambda_n+1} 
  \sum_{\y \sim \hat y_r} 
  \left(\prod_{\factor \in F} q_\factor^2(\y_{\factor})\right)
  \left(\sum_{\factor \in F} \phi_\factor(\y_{\factor})^\trans\cv_n\right)^2\,.
\end{align}
The result is a sum of $D$ terms, each of which takes the form of
\eqref{itworks}, and therefore is efficiently computable by message
passing.  The desired part marginal probability simply requires $D$
separate applications of belief propagation, one per eigenvector
$\cv_n$, for a total runtime of $O(D^2\M^c R)$.  (It is also possible
to vectorize this computation and use a single run of belief
propagation.)  Note that if we require the marginal for only a single
part $\mu_r(\hat y_r)$, we can run just the forward pass if we root
the factor tree at part node $r$.  However, by running both passes we
obtain everything we need to compute the part marginals for any $r$
and $\hat y_r$; the asymptotic time required to compute all part
marginals is the same as the time required to compute just one.

\subsubsection{Sampling}
\seclabel{sdpp_sampling}

While the dual representation provides useful algorithms for
normalization and marginals, the dual sampling algorithm is linear in
$N$; for SDPPs, this is too slow to be useful.  In order to make SDPP
sampling practical, we need to be able to efficiently choose a
structure $\y_i$ according to the distribution
\begin{align}
  \Pr(\y_i) &= \frac{1}{|\cV|}\sum_{\cv\in \cV} (\cv^\trans B_i)^2
\end{align}
in the first line of the while loop in \algref{dual_sampling}.  We can
use the definition of $B$ to obtain
\begin{align}
  \Pr(\y_i) &= \frac{1}{|\cV|}\sum_{\cv\in \cV} q^2(\y_i)(\cv^\trans \phi(\y_i))^2\\
  &= \frac{1}{|\cV|}\sum_{\cv\in \cV} 
  \left( \prod_{\factor\in F} q_\factor^2(\y_{i\factor}) \right)
  \left( \sum_{\factor\in F} \cv^\trans \phi_\factor(\y_{i\factor}) \right)^2\,. 
  \eqlabel{singledist}
\end{align}
Thus, the desired distribution has the familiar form of
\eqref{itworks}.  For instance, the marginal probability of part $r$
taking the assignment $\hat y_r$ is given by
\begin{align}
  \frac{1}{|\cV|}\sum_{\cv\in \cV} \sum_{\y \sim \hat y_r}
  \left( \prod_{\factor\in F} q_\factor^2(\y_{\factor}) \right)
  \left( \sum_{\factor\in F} \cv^\trans \phi_\factor(\y_{\factor}) \right)^2\,,
\end{align}
which we can compute with $k = |\cV|$ runs of belief propagation (or a
single vectorized run), taking only $O(D\M^c R k)$ time.  More
generally, the message-passing computation of these marginals offers
an efficient algorithm for sampling individual full structures $\y_i$.
We will first show a naive method based on iterated computation of
conditional marginals, and then use it to derive a more efficient
algorithm by integrating the sampling of parts into the
message-passing process.

\paragraph{Single structure sampling}

Returning to the factor graph used for belief propagation (see
\secref{factor_graphs}), we can force a part $r'$ to take a certain
assignment $y_{r'}$ by adding a new singleton factor containing only
$r'$, and setting its weight function to $\bone$ for $y_{r'}$ and
$\bzero$ otherwise.  (In practice, we do not need to actually create a
new factor; we can simply set outgoing messages from variable $r'$ to
$\bzero$ for all but the desired assignment $y_{r'}$.)  It is easy to
see that \eqref{semibeliefs} becomes
\begin{equation}
  \bigotimes_{\factor \in F(r)} m_{\factor \to r}(y_r) = 
  \bigoplus_{\y \sim y_r,y_{r'}} \bigotimes_{\factor\in F} w_\factor(\y_\factor)\,,
  \eqlabel{condbeliefs}
\end{equation}
where the sum is now doubly constrained, since any assignment $\y$ that
is not consistent with $y_{r'}$ introduces a $\bzero$ into the
product.  If $\bigotimes_{\factor\in F} w_\factor(\y_\factor)$ gives
rise to a probability measure over structures $\y$, then
\eqref{condbeliefs} can be seen as the unnormalized {\it conditional}
marginal probability of the assignment $y_r$ given $y_{r'}$.  For
example, using the second-order semiring with $p = q^2$ and $a = b =
\cv^\trans\phi$, we have
\begin{align}
  \left[\bigotimes_{\factor \in F(r)} m_{\factor \to r}(y_r)\right]_4 = 
  \sum_{\y\sim y_r,y_{r'}} \left( \prod_{\factor\in F} q^2_\factor(\y_\factor) \right)
  \left( \sum_{\factor\in F} \cv^\trans\phi_\factor(\y_\factor) \right)^2\,.
  \eqlabel{condsingle}
\end{align}
Summing these values for all $\cv\in\cV$ and normalizing the result
yields the conditional distribution of $y_r$ given fixed assignment
$y_{r'}$ under \eqref{singledist}.  Going forward we will assume for
simplicity that $\cV$ contains a single vector $\cv$; however, the
general case is easily handled by maintaining $|\cV|$ messages in
parallel, or by vectorizing the computation.

The observation that we can compute conditional probabilities with
certain assignments held fixed gives rise to a naive algorithm for
sampling a structure according to $\Pr(\y_i)$ in \eqref{singledist},
shown in \algref{single_sampling}.  While polynomial,
\algref{single_sampling} requires running belief propagation $R$
times, which might be prohibitively expensive for large structures.
We can do better by weaving the sampling steps into a single run of
belief propagation.  We discuss first how this can be done for linear
factor graphs, where the intuition is simpler, and then extend it to
general factor trees.

\begin{algorithm}[tb]
\begin{algorithmic}
  \STATE {\bfseries Input:} factored $q$ and $\phi$, $\cv$
  \STATE $S \leftarrow \emptyset$
  \FOR{$r = 1,2,\dots,R$}
  \STATE Run second-order belief propagation with:
  \STATE \quad $\bullet$\; $p=q^2$
  \STATE \quad $\bullet$\; $a=b=\cv^\trans \phi$
  \STATE \quad $\bullet$\; assignments in $S$ held fixed
  \STATE Sample $y_r$ according to $\Pr(y_r|S) \propto
  \left[\bigotimes_{\factor \in F(r)} m_{\factor \to r}(y_r)\right]_4$
  \STATE $S \leftarrow S \cup \{y_r\}$
  \ENDFOR
  \STATE {\bfseries Output:} $\y$ constructed from $S$
\end{algorithmic}
\caption{Sampling a structure (naive)}
\alglabel{single_sampling}
\end{algorithm}

\paragraph{Linear graphs}

Suppose that the factor graph is a linear chain arranged from left to
right.  Each node in the graph has at most two neighbors---one to the
left, and one to the right.  Assume the belief propagation forward
pass proceeds from left to right, and the backward pass from right to
left.  To send a message to the right, a node needs only to receive
its message from the left.  Conversely, to send a message to the left,
only the message from the right is needed.  Thus, the forward and
backward passes can be performed independently.

Consider now the execution of \algref{single_sampling} on this factor
graph.  Assume the variable nodes are numbered in decreasing order
from left to right, so the variable sampled in the first iteration is
the rightmost variable node.  Observe that on iteration $r$, we do not
actually need to run belief propagation to completion; we need only
the messages incoming to variable node $r$, since those suffice to
compute the (conditional) marginals for part $r$.  To obtain those
messages, we must compute all of the forward messages sent from the
left of variable $r$, and the backward messages from the right.  Call
this set of messages $\m(r)$.  

Note that $\m(1)$ is just a full, unconstrained forward pass, which
can be computed in time $O(D\M^c R)$.  Now compare $\m(r)$ to
$\m(r-1)$.  Between iteration $r-1$ and $r$, the only change to $S$ is
that variable $r-1$, to the right of variable $r$, has been assigned.
Therefore the forward messages in $\m(r)$, which come from the left,
do not need to be recomputed, as they are a subset of the forward
messages in $\m(r-1)$.  Likewise, the backward messages sent from the
right of variable $r-1$ are unchanged, so they do not need to be
recomputed.  The only new messages in $\m(r)$ are those backward
messages traveling from $r-1$ to $r$.  These can be computed, using
$\m(r-1)$ and the sampled assignment $y_{r-1}$, in constant time.  See
\figref{linearsampling} for an illustration of this process.

\begin{figure}
  \centering
  \begin{tikzpicture}
    [auto,scale=0.8,
      Mr/.style={thick,dashed,blue,shorten <=4pt,shorten >=4pt},
      both/.style={thick,dashed,shorten <=4pt,shorten >=4pt}
    ]

    \node [yshift=2.25cm] {$\m(r-1)$:};
    \node [yshift=-0.9cm] {$\m(r)$:};
    \matrix [row sep=1cm,column sep=1.1cm] {
      \\
      \node (f12a) [minimum size=0.75cm] {$\cdots$}; &
      \node (y2a) [circle,draw,minimum size=1cm] {$y_{r+1}$}; &
      \node (f23a) [draw,minimum size=0.75cm] {}; &
      \node (y3a) [circle,draw,minimum size=1cm] {$y_r$}; &
      \node (f34a) [draw,minimum size=0.75cm] {}; &
      \node (y4a) [circle,draw,minimum size=1cm] {$y_{r-1}$}; &
      \node (f45a) [draw,minimum size=0.75cm] {}; &
      \node (y5a) [minimum size=0.75cm] {$\cdots$};
      \\\\
      \node (f12b) [minimum size=0.75cm] {$\cdots$}; &
      \node (y2b) [circle,draw,minimum size=1cm] {$y_{r+1}$}; &
      \node (f23b) [draw,minimum size=0.75cm] {}; &
      \node (y3b) [circle,draw,minimum size=1cm] {$y_r$}; &
      \node (f34b) [draw,minimum size=0.75cm] {}; &
      \node (y4b) [circle,draw,minimum size=1cm] {$y_{r-1}$}; 
      \node (y4b_) [circle,draw,minimum size=1cm] {}; &
      \node (f45b) [draw,minimum size=0.75cm] {}; &
      \node (y5b) [minimum size=0.75cm] {$\cdots$};
      \\
    };

    \path {
      (f12a) edge (y2a)
      (f23a) edge (y2a)
      (f23a) edge (y3a)
      (f34a) edge (y3a)
      (f34a) edge (y4a)
      (f45a) edge (y4a)
      (f45a) edge (y5a)
    };

    \path {
      (f12b) edge (y2b)
      (f23b) edge (y2b)
      (f23b) edge (y3b)
      (f34b) edge (y3b)
      (f34b) edge (y4b)
      (f45b) edge (y4b)
      (f45b) edge (y5b)
    };

    \draw[both] [->] (f12a) to [bend left=40]  (y2a);
    \draw[both] [->] (y2a) to [bend left=40]  (f23a);
    \draw[both] [->] (f23a) to [bend left=40]  (y3a);
    \draw[both] [->] (y3a) to [bend left=40]  (f34a);
    \draw[both] [->] (f34a) to [bend left=40]  (y4a);
    \draw[both] [<-] (y4a) to [bend right=40,swap]  (f45a);
    \draw[both] [<-] (f45a) to [bend right=40,swap]  (y5a);

    \draw[both] [->] (f12b) to [bend left=40]  (y2b);
    \draw[both] [->] (y2b) to [bend left=40]  (f23b);
    \draw[both] [->] (f23b) to [bend left=40]  (y3b);
    \draw[Mr]   [<-] (y3b) to [bend right=40,swap] node {*}  (f34b);
    \draw[Mr]   [<-] (f34b) to [bend right=40,swap] node {*} (y4b);
    \draw[both] [<-] (y4b) to [bend right=40,swap]  (f45b);
    \draw[both] [<-] (f45b) to [bend right=40,swap]  (y5b);

  \end{tikzpicture}
  \caption[Messages on a linear chain]
    {Messages on a linear chain.  Only the starred messages need
    to be computed to obtain $\m(r)$ from $\m(r-1)$.  The double
    circle indicates that assignment $y_{r-1}$ has been fixed for
    computing $\m(r)$.}  \figlabel{linearsampling}
\end{figure}
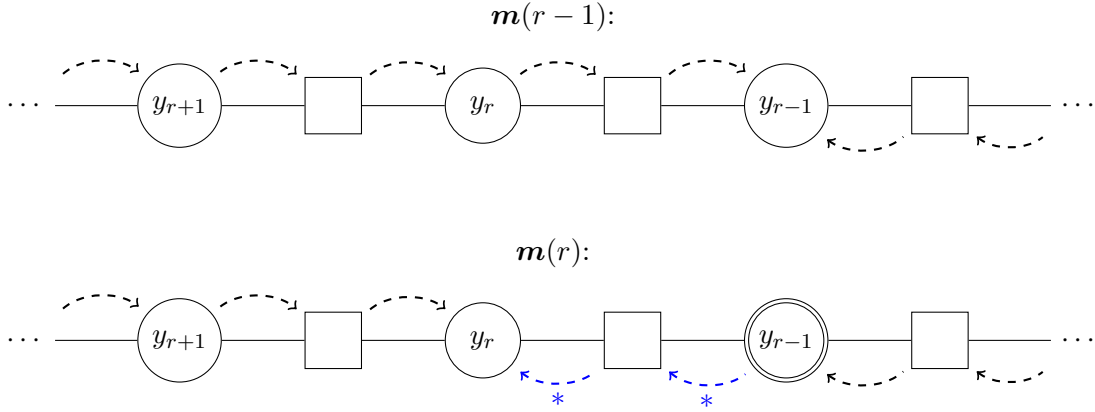

Thus, rather than restarting belief propagation on each loop of
\algref{single_sampling}, we have shown that we need only compute a
small number of additional messages.  In essence we have threaded the
sampling of parts $r$ into the backward pass.  After completing the
forward pass, we sample $y_1$; we then compute the backward messages
from $y_1$ to $y_2$, sample $y_2$, and so on.  When the backward pass
is complete, we sample the final assignment $y_R$ and are finished.
Since the initial forward pass takes $O(D\M^c R)$ time and each of the
$O(R)$ subsequent iterations takes at most $O(D\M^c)$ time, we can
sample from $\Pr(\y_i)$ over a linear graph in $O(D\M^c R)$ time.

\paragraph{Trees}

The algorithm described above for linear graphs can be generalized to
arbitrary factor trees.  For standard graphical model sampling using
the sum-product semiring, the generalization is straightforward---we
can simply pass messages up to the root and then sample on the
backward pass from the root to the leaves.  However, for arbitrary
semirings this is algorithm is incorrect, since an assignment to one
node can affect the messages arriving at its siblings even when the
parent's assignment is fixed.

Let $m_{b\to a}(\cdot|S)$ be the message function sent from node $b$
to node $a$ during a run of belief propagation where the assignments
in $S$ have been held fixed.  Imagine that we re-root the factor tree
with $a$ as the root; then define $T_a(b)$ to be the subtree rooted at
$b$ (see \figref{subtree}).  Several useful observations follow.

\begin{figure}
\centering
\begin{tikzpicture}[auto,level/.style={sibling distance=60mm/#1}]
\node 
    [draw] (a) {$a$}
    child {node [circle,draw] {$d$}
      child {node {$\vdots$}}
      child {node {$\vdots$}}
    }
    child {node [circle,draw] (b) {$b$}
      child {node [draw] (c1) {$c_1$}
        child {node [circle,draw] (ll) {}}
        child {node [circle,draw] {}}
      }
      child {node [draw] (c2) {$c_2$}
        child {node [circle,draw] {}}
        child {node [circle,draw] (lr) {}}
      }  
    };
\draw [->,thick,dashed,shorten <=4pt,shorten >=4pt] (b) to 
      [bend right=30,swap] 
      node {$m_{b\to a}(\cdot|S)$} (a);
\begin{pgfonlayer}{background}
  \node [inner sep=10pt,fill=black!20,fit=(b)(ll)(lr)] (subtree) {};
  \node [anchor=north east,inner sep=5pt] at (subtree.north east) {$T_a(b)$};
\end{pgfonlayer}
\end{tikzpicture}
\caption[Factor tree notation]
  {Notation for factor trees, including $m_{b\to a}(\cdot|S)$
  and $T_a(b)$ when $a$ is a (square) factor node and $b$ is a (round)
  variable node.  The same definitions apply when $a$ is a variable
  and $b$ is a factor.  \figlabel{subtree}}
\end{figure}
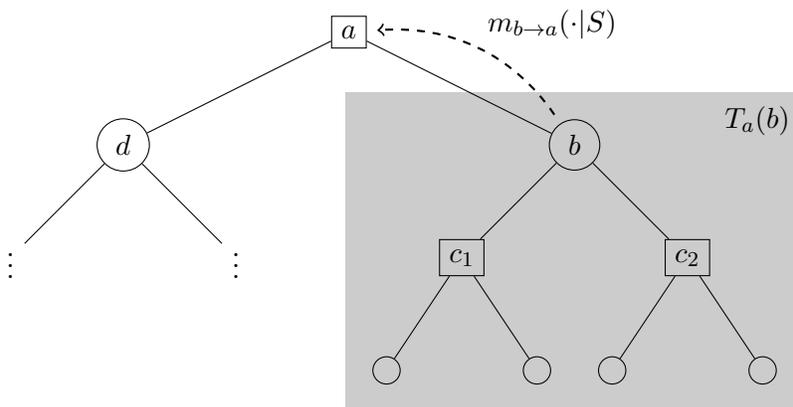

\begin{lemma}
  If $b_1$ and $b_2$ are distinct neighbors of $a$, then $T_a(b_1)$
  and $T_a(b_2)$ are disjoint.  
  \lemlabel{disjointsubtrees}
\end{lemma}
\begin{proof}
The claim is immediate, since the underlying graph is a tree.
\end{proof}

\begin{lemma}
  $m_{b\to a}(\cdot|S)$ can be computed given only the messages
  $m_{c\to b}(\cdot|S)$ for all neighbors $c \neq a$ of $b$ and either
  the weight function $w_b$ (if $b$ is a factor node) or the
  assignment to $b$ in $S$ (if $b$ is a variable node and such an
  assignment exists).  
  \lemlabel{recursivemessages}
\end{lemma}
\begin{proof}
  Follows from the message definitions in
  \eqsref{messages1}{messages2}.
\end{proof}

\begin{lemma}
  $m_{b\to a}(\cdot|S)$ depends only on the assignments in $S$ that
  give values to variables in $T_a(b)$.  
  \lemlabel{localdependence}
\end{lemma}
\begin{proof}
  If $b$ is a leaf (that is, its only neighbor is $a$), the lemma
  holds trivially.  If $b$ is not a leaf, then assume inductively that
  incoming messages $m_{c\to b}(\cdot|S)$, $c\neq a$, depend only on
  assignments to variables in $T_b(c)$.  By
  \lemref{recursivemessages}, the message $m_{b\to a}(\cdot|S)$
  depends only on those messages and (possibly) the assignment to $b$
  in $S$.  Since $b$ and $T_b(c)$ are subgraphs of $T_a(b)$, the claim
  follows.
\end{proof}

\noindent
To sample a structure, we begin by initializing $S_0=\emptyset$ and
setting messages $\hat m_{b\to a} = m_{b\to a}(\cdot|S_0)$ for all
neighbor pairs $(a,b)$.  This can be done in $O(D\M^c R)$ time via
belief propagation.

Now we walk the graph, sampling assignments and updating the current
messages $\hat m_{b\to a}$ as we go.  Step $t$ from node $b$ to $a$
proceeds in three parts as follows:
\begin{enumerate}
  \item Check whether $b$ is a variable node without an assignment in
    $S_{t-1}$.  If so, sample an assignment $y_b$ using the current
    incoming messages $\hat m_{c\to b}$, and set $S_t = S_{t-1} \cup
    \{y_b\}$.  Otherwise set $S_t = S_{t-1}$.
  \item Recompute and update $\hat m_{b\to a}$ using the current
    messages and \eqsref{messages1}{messages2}, taking into account any
    assignment to $b$ in $S_t$.
  \item Advance to node $a$.
\end{enumerate}
This simple algorithm has the following useful invariant.
\begin{theorem}
  Following step $t$ from $b$ to $a$, for every neighbor $d$ of $a$ we
  have
  \begin{equation}
    \hat m_{d\to a} = m_{d\to a}(\cdot|S_t)\,.
  \end{equation}
  \thmlabel{treesample}
\end{theorem}
\begin{proof}
  By design, the theorem holds at the outset of the walk.  Suppose
  inductively that the claim is true for steps $1,2,\dots,t-1$.  Let
  $t'$ be the most recent step prior to $t$ at which we visited $a$,
  or $0$ if step $t$ was our first visit to $a$.  Since the graph is a
  tree, we know that between steps $t'$ and $t$ the walk remained
  entirely within $T_a(b)$.  Hence the only assignments in $S_t -
  S_{t'}$ are to variables in $T_a(b)$.  As a result, for all
  neighbors $d \neq b$ of $a$ we have $\hat m_{d\to a} = m_{d\to
    a}(\cdot|S_{t'}) = m_{d\to a}(\cdot|S_t)$ by the inductive
  hypothesis, \lemref{disjointsubtrees}, and \lemref{localdependence}.
  
  It remains to show that $\hat m_{b\to a} = m_{b\to a}(\cdot|S_i)$.
  For all neighbors $c\neq a$ of $b$, we know that $\hat m_{c\to b} =
  m_{c\to b}(\cdot|S_{i-1}) = m_{c\to b}(\cdot|S_t)$ due to the
  inductive hypothesis and \lemref{localdependence} (since $b$ is not
  in $T_b(c)$).  By \lemref{recursivemessages}, then, we have $\hat
  m_{b\to a} = m_{b\to a}(\cdot|S_t)$.
\end{proof}

\noindent
\thmref{treesample} guarantees that whenever we sample an assignment
for the current variable node in the first part of step $t$, we sample
from the conditional marginal distribution $\Pr(y_b|S_{t-1})$.
Therefore, we can sample a complete structure from the distribution
$\Pr(\y)$ if we walk the entire tree.  This can be done, for example,
by starting at the root and proceeding in depth-first order.  Such a
walk takes $O(R)$ steps, and each step requires computing only a
single message.  Thus, allowing now for $k = |\cV| > 1$, we can sample
a structure in time $O(D\M^c Rk)$, a significant improvement over
\algref{single_sampling}.  The procedure is summarized in
\algref{single_sampling2}.

\begin{algorithm}[tb]
\begin{algorithmic}
  \STATE {\bfseries Input:} factored $q$ and $\phi$, $\cv$
  \STATE $S \leftarrow \emptyset$
  \STATE Initialize $\hat m_{a\to b}$ using second-order belief propagation 
  with $p=q^2$, $a=b=\cv^\trans\phi$
  \STATE Let $a_1,a_2,\dots,a_T$ be a traversal of the factor tree
  \FOR{$t = 1,2,\dots,T$}
  \IF{$a_t$ is a variable node $r$ with no assignment in $S$}
  \STATE Sample $y_r$ according to 
  $\Pr(y_r) \propto 
  \left[\bigotimes_{\factor \in F(r)} \hat m_{\factor \to r}(y_r)\right]_4$
  \STATE $S \leftarrow S \cup \{y_r\}$
  \ENDIF
  \IF{$t < T$}
  \STATE Update $\hat m_{a_t\to a_{t+1}}$ using \eqsref{messages1}{messages2}, 
  fixing assignments in $S$
  \ENDIF
  \ENDFOR
  \STATE {\bfseries Output:} $\y$ constructed from $S$
\end{algorithmic}
\caption{Sampling a structure}
\alglabel{single_sampling2}
\end{algorithm}

\algref{single_sampling2} is the final piece of machinery needed to
replicate the DPP sampling algorithm using the dual representation.
The full SDPP sampling process is given in \algref{sdpp_sampling} and
runs in time $O(D^2k^3 + D\M^c Rk^2)$, where $k$ is the number of
eigenvectors selected in the first loop.  As in standard DPP sampling,
the asymptotically most expensive operation is the orthonormalization;
here we require $O(D^2)$ time to compute each of the $O(k^2)$ dot
products.

\begin{algorithm}[tb]
\begin{algorithmic}
  \STATE {\bfseries Input:} eigendecomposition
    $\{(\cv_n,\lambda_n)\}_{n=1}^D$ of $C$
  \STATE $J \leftarrow \emptyset$
  \FOR{$n = 1,2,\dots,N$}
  \STATE $J \leftarrow J \cup \{n\}$ with prob. $\frac{\lambda_n}{\lambda_n+1}$
  \ENDFOR
  \STATE $\cV \leftarrow 
  \left\{\frac{\cv_n}{\sqrt{\cv_n^\trans C \cv_n}}\right\}_{n\in J}$
  \STATE $Y \leftarrow \emptyset$
  \WHILE{$|\cV|>0$}
  \STATE Select $\y_i$ from $\Y$ with 
  $\Pr(\y_i) = \frac{1}{|\cV|}\sum_{\cv\in \cV} 
  ((B^\trans\cv)^\trans \e_i)^2$ (\algref{single_sampling2})
  \STATE $Y \leftarrow Y \cup \y_i$
  \STATE $\cV \leftarrow \cV_\bot$, where 
  $\{B^\trans \cv \mid \cv\in \cV_\bot\}$ is an orthonormal basis 
  for the subspace of $V$ 
  \STATE\quad\quad\quad\quad orthogonal to $\e_i$
  \ENDWHILE
  \STATE {\bfseries Output:} $Y$
\end{algorithmic}
\caption{Sampling from an SDPP}
\alglabel{sdpp_sampling}
\end{algorithm}

\subsection{Experiments: pose estimation}
\seclabel{poseestimation}

To demonstrate that SDPPs effectively model characteristics of
real-world data, we apply them to a multiple-person pose estimation
task \citep{kulesza2010sdpps}.  Our input will be a still image depicting multiple people, and
our goal is to simultaneously identify the poses---the positions of
the torsos, heads, and left and right arms---of all the people in the
image.  A pose $\y$ is therefore a structure with four parts, in this
case literally body parts.  To form a complete structure, each part
$r$ is assigned a position/orientation pair $y_r$.  Our quality model
will be based on ``part detectors'' trained to estimate the likelihood
of a particular body part at a particular location and orientation;
thus we will focus on identifying poses that correspond well to the
image itself.  Our similarity model, on the other hand, will focus on
the location of a pose within the image.  Since the part detectors
often have uncertainty about the precise location of a part, there may
be many variations of a single pose that outscore the poses of all the
other, less detectable people.  An independent model would thus be
likely to choose many similar poses.  By encouraging the model to
choose a spatially diverse set of poses, we hope to improve the chance
that the model predicts a single pose for each person.

Our dataset consists of 73 still frames taken from various TV shows,
each approximately 720 by 540 pixels in size
\citep{sapp2010}\footnote{The images and code were obtained from {\tt
    http://www.vision.grasp.upenn.edu/video}}.  As much as possible,
the selected frames contain three or more people at similar scale, all
facing the camera and without serious occlusions.  Sample images from
the dataset are shown in \figref{marginals_ps}.  Each person in each
image is annotated by hand; each of the four parts (head, torso, right
arm, and left arm) is labeled with the pixel location of a reference
point (e.g., the shoulder) and an orientation selected from among 24
discretized angles.

\subsubsection{Factorized model}

There are approximately 75,000 possible values for each part, so there
are about $4^{75,000}$ possible poses, and thus we cannot reasonably
use a standard DPP for this problem.  Instead, we build a factorized
SDPP.  Our factors are given by the standard pictorial structure model
\citep{felzenszwalb2005pictorial,fischler1973ps}, treating each pose
as a two-level tree with the torso as the root and the head and arms
as leaves.  Each node (body part) has a singleton factor, and each
edge has a corresponding pairwise factor.

Our quality function derives from the model proposed by
\citet{sapp2010}, and is given by
\begin{equation}
  q(\y) = \gamma\left(\prod_{r = 1}^R q_r(y_r)
         \prod_{(r,r') \in E}q_{r,r'}(y_r,y_{r'})\right)^\beta\,,
\end{equation}
where $E$ is the set of edges in the part tree, $\gamma$ is a scale
parameter that will control the expected number of poses in an SDPP
sample, and $\beta$ is a sharpness parameter that controls the dynamic
range of the quality scores.  We set the values of the hyperparameters
$\gamma$ and $\beta$ using a held-out training set, as discussed
below.  The per-part quality scores $q_r(y_r)$ are provided by the
customized part detectors trained by \citet{sapp2010} on similar
images; they assign a value to every proposed location and orientation
$y_r$ of part $r$.  The pairwise quality scores $q_{r,r'}(y_r,y_{r'})$
are defined according to a Gaussian ``spring'' that encourages, for
example, the left arm to begin near the left shoulder of the torso.
Full details of the model are provided in \citet{sapp2010}.

In order to encourage the model not to choose overlapping poses, 
our diversity features reflect the locations of the constituent parts:
\begin{equation}
  \phi(\y) = \sum_{r = 1}^R \phi_r(y_r)\,,
\end{equation}
where each $\phi_r(y_r) \in \reals^{32}$.  There are no diversity
features on the edge factors.  The local features are based on a $8
\times 4$ grid of reference points $x_1,x_2,\dots,x_{32}$ spaced evenly
across the image; the $l$th feature is
\begin{equation}
  \phi_{rl}(y_r) \propto f_\N\left(\frac{\dist(y_r,x_l)}{\sigma}\right)\,.
\end{equation}
Here $f_\N$ is again the standard normal density function, and
$\dist(y_r,x_l)$ is the Euclidean distance between the position of
part $r$ (ignoring orientation) and the reference point $x_l$.  Poses
that occupy the same part of the image will be near the same reference
points, and thus their feature vectors $\phi$ will be more closely
aligned.  The parameter $\sigma$ controls the width of the kernel;
larger values of $\sigma$ make poses at a given distance appear more
similar  We set $\sigma$ on a held-out training set.

\subsubsection{Methods}

We compare samples from the SDPP defined above to those from two
baseline methods.  The first, which we call the independent model,
draws poses independently according to the distribution obtained by
normalizing the quality scores, which is essentially the graphical
model used by \citet{sapp2010}.  For this model the number of poses to
be sampled must be supplied by the user, so to create a level playing
field we choose the number of poses in an SDPP sample $Y$.  Since this
approach does not incorporate a notion of diversity (or any
correlations between selected poses whatsoever), we expect that we
will frequently see multiple poses that correspond to the same person.

The second baseline is a simple non-maximum suppression model
\citep{canny1986computational}, which incorporates a heuristic for
encouraging diversity.  The first pose is drawn from the normalized
quality model in the same manner as for the independent method.
Subsequent poses, however, are constrained so that they cannot overlap
with the previously selected poses, but otherwise drawn according to
the quality model.  We consider poses overlapping if they cover any of
the same pixels when rendered.  Again, the number of poses must be
provided as an argument, so we use the number of poses from a sample
of the SDPP.  While the non-max approach can no longer generate
multiple poses in the same location, it achieves this using a hard,
heuristic constraint.  Thus, we might expect to perform poorly when
multiple people actually do overlap in the image, for example if one
stands behind the other.

The SDPP, on the other hand, generates samples that prefer, but do not
require poses to be spatially diverse.  That is, strong visual
information in the image can override our prior assumptions about the
separation of distinct poses.  We split our data randomly into a
training set of 13 images and a test set of 60 images.  Using the
training set, we select values for $\gamma$, $\beta$, and $\sigma$
that optimize overall $F_1$ score at radius 100 (see below), as well
as distinct optimal values of $\beta$ for the baselines.  ($\gamma$
and $\sigma$ are irrelevant for the baselines.)  We then use each
model to sample 10 sets of poses for each test image, for a total of
600 samples per model.

\subsubsection{Results}

For each sample from each of the three tested methods, we compute
measures of precision and recall as well as an $F_1$ score.  In our
tests, precision is measured as the fraction of predicted parts for
which both endpoints are within a given radius of the endpoints of an
expert-labeled part of the same type (head, left arm, and so on).  We
report results across a range of radii.  Correspondingly, recall is
the fraction of expert-labeled parts with endpoints within a given
radius of a predicted part of the same type.  Since the SDPP model
encourages diversity, we expect to see improvements in recall at the
expense of precision, compared to the independent model.  $F_1$ score
is the harmonic mean of precision and recall.  We compute all metrics
separately for each sample, and then average the results across
samples and images in the test set.

The results are shown in \figref{f1_radius}.  At tight tolerances,
when the radius of acceptance is small, the SDPP performs comparably
to the independent and non-max samples, perhaps because the quality
scores are simply unreliable at this resolution, thus diversity has
little effect.  As the radius increases, however, the SDPP obtains
better results, significantly outperforming both baselines.
\figref{arm_radius} shows the curves for just the arm parts, which
tend to be more difficult to locate accurately and exhibit greater
variance in orientation.  \figref{pr_radius} shows the
precision/recall obtained by each model.  As expected, the SDPP model
achieves its improved $F_1$ score by increasing recall at the cost of
precision.

\begin{figure}
  \centering
  \subfloat[][]{
    \includegraphics[width=1.9in]{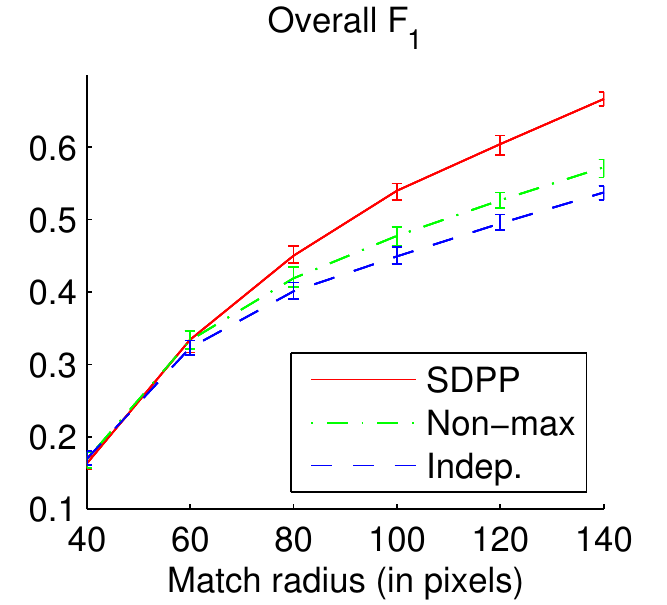}
    \figlabel{f1_radius}
  }
  \subfloat[][]{
    \includegraphics[width=1.9in]{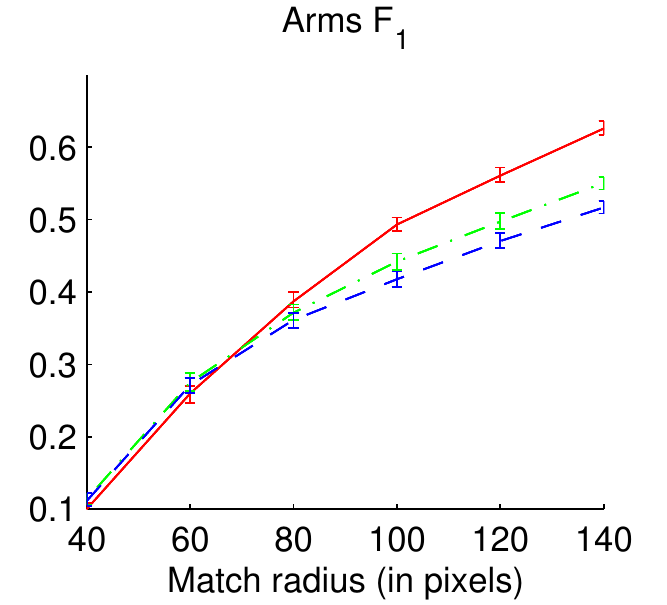}
    \figlabel{arm_radius}
  }
  \subfloat[][]{
    \includegraphics[width=1.9in]{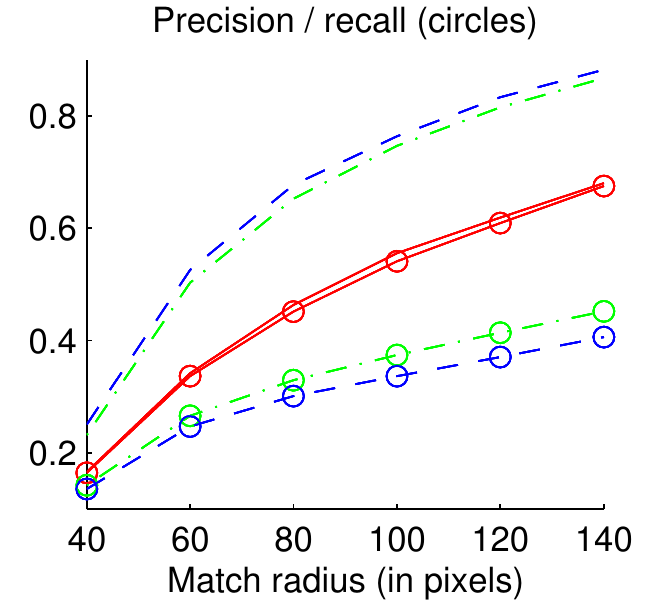}
    \figlabel{pr_radius}
  }
  \caption[Pose estimation results]
    {Results for pose estimation.  The horizontal axis gives the
    acceptance radius used to determine whether two parts are
    successfully matched.  95\% confidence intervals are shown.  (a)
    Overall $F_1$ scores. (b) Arm $F_1$ scores. (c) Overall
    precision/recall curves (recall is identified by circles).}
\end{figure}

For illustration, we show the SDPP sampling process for some sample
images from the test set in \figref{marginals_ps}.  The SDPP part
marginals are visualized as a ``cloud'', where brighter colors
correspond to higher probability.  From left to right, we can see how
the marginals change as poses are selected during the main loop of
\algref{sdpp_sampling}.  As we saw for simple synthetic examples in
\figref{marginals_line}, the SDPP discounts but does not entirely
preclude poses that are similar to those already selected.


\begin{figure}
  \centering
  \includegraphics[width=\linewidth]{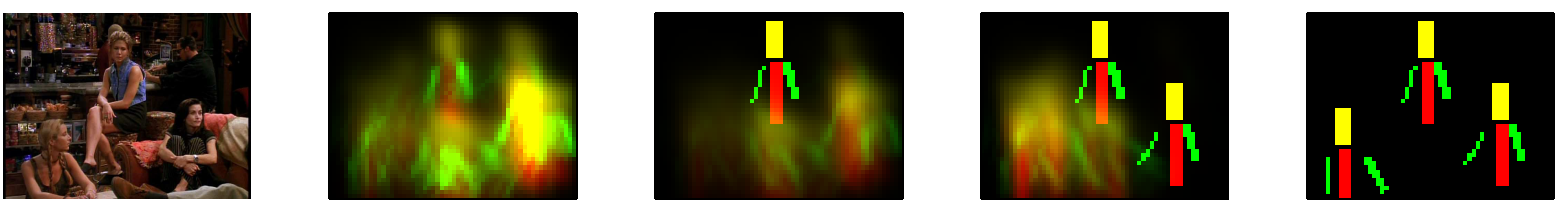}\\
  \includegraphics[width=\linewidth]{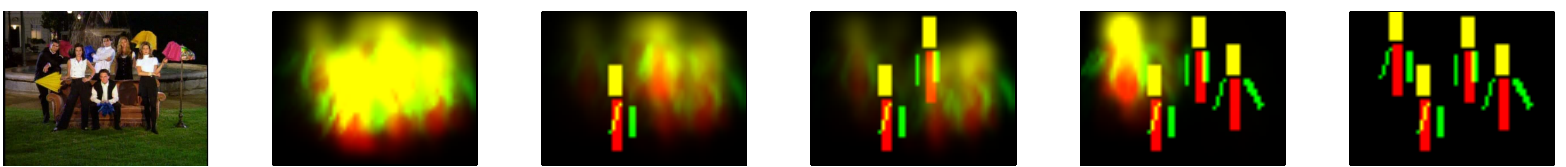}\\
  \includegraphics[width=\linewidth]{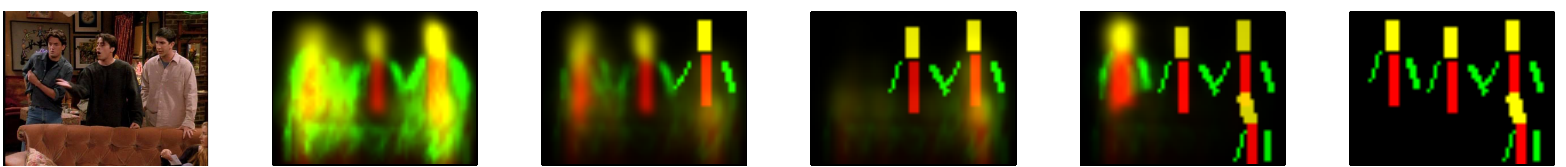}\\
  \includegraphics[width=\linewidth]{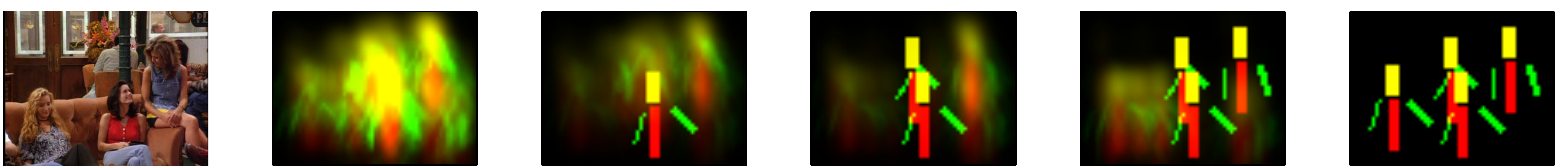}\\
  \includegraphics[width=\linewidth]{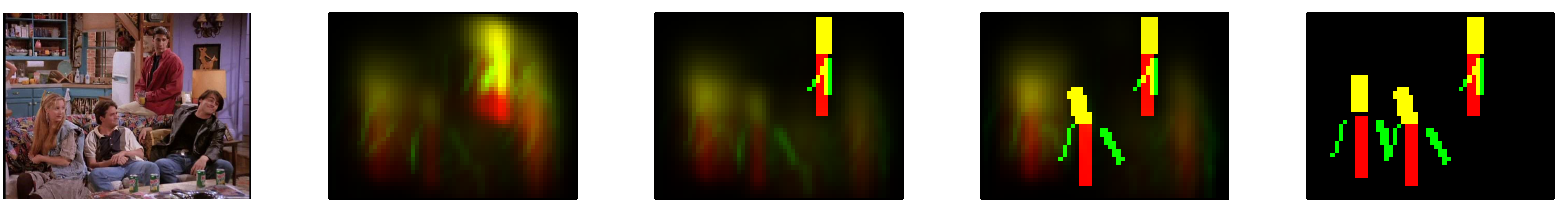}\\
  \includegraphics[width=\linewidth]{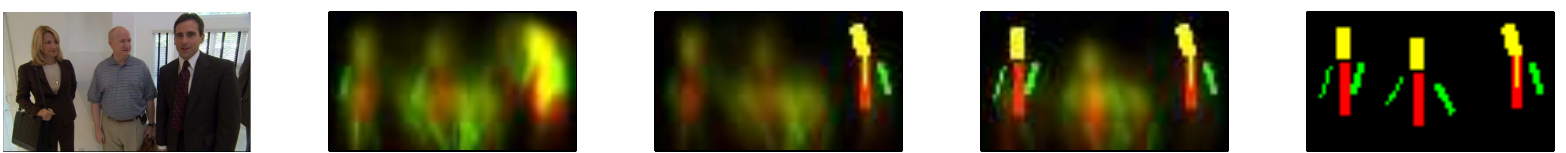}\\
  \includegraphics[width=\linewidth]{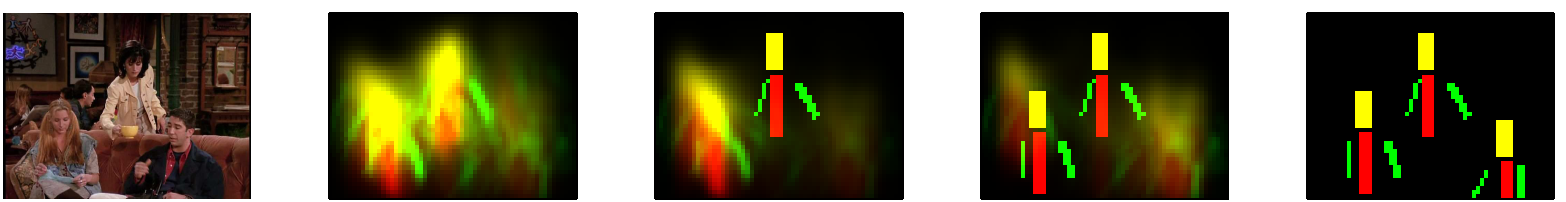}\\
  \includegraphics[width=\linewidth]{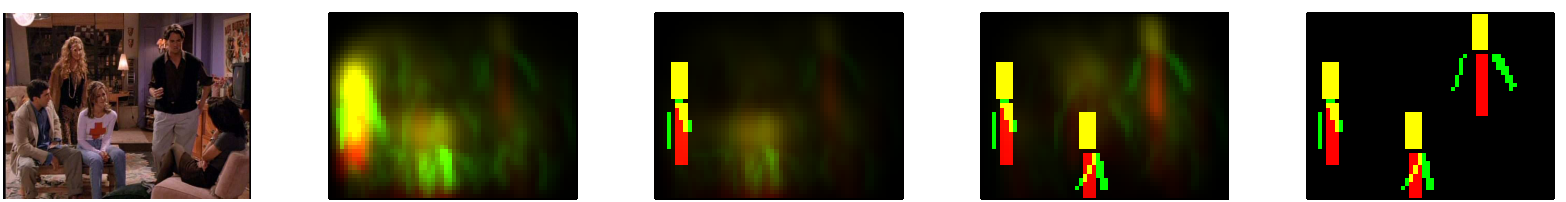}\\
  \includegraphics[width=\linewidth]{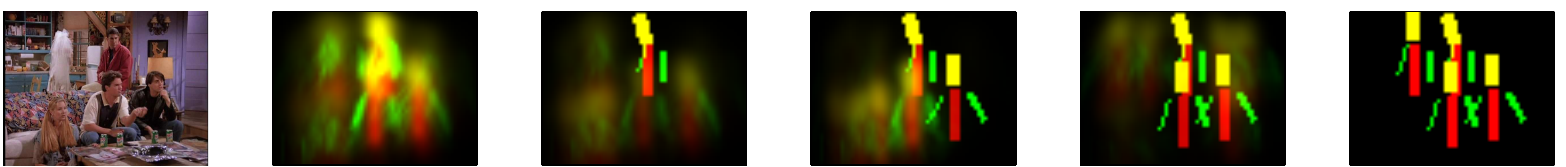}\\
  \includegraphics[width=\linewidth]{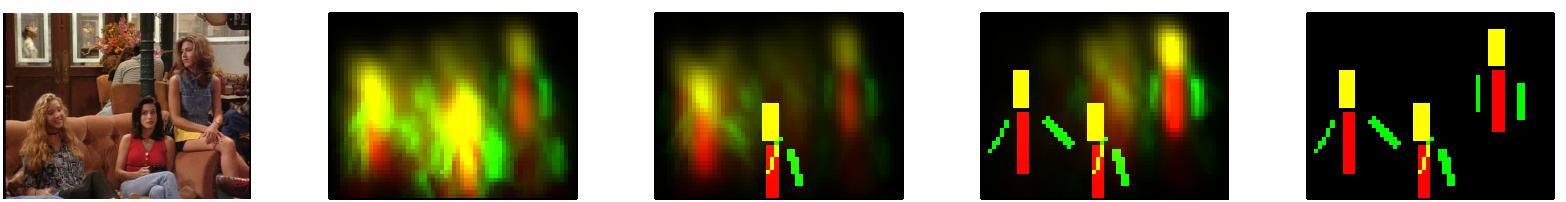}\\
  \caption[Pose estimation marginals]
    {Structured marginals for the pose estimation task,
    visualized as clouds, on successive steps of the sampling
    algorithm.  Already selected poses are superimposed.  Input images
    are shown on the left.}  \figlabel{marginals_ps}
\end{figure}

\subsection{Random projections for SDPPs}
\seclabel{sdppproj}

It is quite remarkable that we can perform polynomial-time inference
for SDPPs given their extreme combinatorial nature.  Even so, in some
cases the algorithms presented in \secref{sdpp_inference} may not be
fast enough.  Eigendecomposing the dual representation $C$, for
instance, requires $O(D^3)$ time, while normalization,
marginalization, and sampling, even when an eigendecomposition has
been precomputed, scale quadratically in $D$, both in terms of time
and memory.  In practice, this limits us to relatively low-dimensional
diversity features $\phi$; for example, in our pose estimation
experiments we built $\phi$ from a fairly coarse grid of 32 points
mainly for reasons of efficiency.  As we move to textual data, this
will become an even bigger problem, since there is no natural
low-dimensional analog for feature vectors based on, say, word
occurrences.  In the following section we will see data where natural
vectors $\phi$ have dimension $D \approx$ 30,000; without dimensionality
reduction, storing even a single belief propagation message would
require over 200 terabytes of memory.

To address this problem, we will make use of the random projection
technique described in \secref{projection}, reducing the dimension of
the diversity features without sacrificing the accuracy of the model.
Because \thmref{dpp_proj} depends on a cardinality condition, we will
focus on $k$-SDPPs.  As described in \secref{kdpps}, a $k$-DPP is
simply a DPP conditioned on the cardinality of the modeled subset
$\bY$:
\begin{equation}
  \P^k(Y) = \frac{ \left(\prod_{\y\in Y} q^2(\y)\right) \det(\phi(Y)^\trans
  \phi(Y)) }
  {\sum_{|Y'| = k} \left(\prod_{\y\in Y} q^2(\y)\right) \det(\phi(Y)^\trans
  \phi(Y)) }\,,
\end{equation}
where $\phi(Y)$ denotes the $D \times |Y|$ matrix formed from columns
$\phi(\y)$ for $\y \in Y$.  When $q$ and $\phi$ factor over parts of a
structure, as in \secref{sdpp_model}, we will refer to this
distribution as a $k$-SDPP.  We note in passing that the algorithms
for normalization and sampling in \secref{kdpps} apply equally well
to $k$-SDPPs, since they depend mainly on the eigenvalues of $L$,
which we can obtain from $C$.

Recall that \thmref{dpp_proj} requires projection dimension
\begin{equation}
  d = O(\max\{k/\epsilon, (\log(1/\delta) + \log N) / \epsilon^2\})\,.
\end{equation}
In the structured setting, $N = M^R$, thus $d$ must be logarithmic in
the number of labels and linear in the number of parts.  Under this
condition, we have, with probability at least $1-\delta$,
\begin{equation}
  \Vert \P^k - \tilde \P^k \Vert_1 \leq e^{6k\epsilon} - 1\,,
\end{equation}
where $\tilde\P^k(Y)$ is the projected $k$-SDPP.

\subsubsection{Toy example: geographical paths}

In order to empirically study the effects of random projections, we
test them on a simple toy application where $D$ is small enough that
the exact model is tractable.  The goal is to identify diverse,
high-quality sets of travel routes between U.S. cities, where
diversity is with respect to geographical location, and quality is
optimized by short paths visiting the most populous or most touristy
cities.  Such sets of paths could be used, for example, by a
politician to plan campaign routes, or by a traveler organizing a
series of vacations.

We model the problem as a $k$-SDPP over path structures having $R=4$
parts, where each part is a stop along the path and can take any of
$M=200$ city values.  The quality and diversity functions are
factored, with a singleton factor for every individual stop and
pairwise factors for consecutive pairs of stops.  The quality of a
singleton factor is based on the Google hit count for the assigned
city, so that paths stopping in popular cities are preferred.  The
quality of a pair of consecutive stops is based on the distance
between the assigned cities, so that short paths are preferred.  In
order to disallow paths that travel back and forth between the same
cities, we augment the stop assignments to include arrival direction,
and assign a quality score of zero to paths that return in the
direction from which they came.  The diversity features are only
defined on the singleton factors; for a given city assignment $y_r$,
$\phi_r(y_r)$ is just the vector of inverse distances between $y_r$
and all of the 200 cities.  As a result, paths passing through the
same or nearby cities appear similar, and the model prefers paths that
travel through different regions of the country.  We have $D=200$.

\figref{maps} shows sets of paths sampled from the $k$-SDPP for
various values of $k$.  For $k=2$, the model tends to choose one path
along the east coast and another along the west coast.  As $k$
increases, a variety of configurations emerge; however, they continue
to emphasize popular cities and the different paths remain
geographically diverse.

\begin{figure}
  \centering
  \includegraphics[width=5.5in]{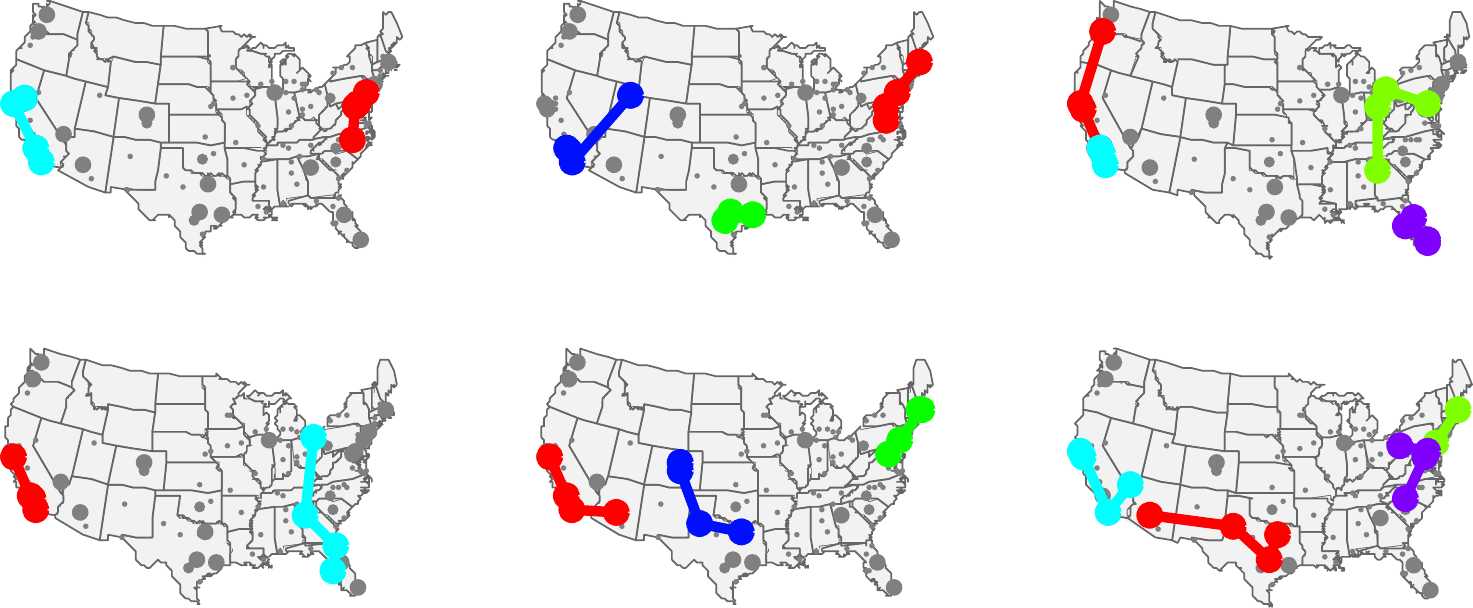}
  \caption[Geographical path samples]
    {Each column shows two samples drawn from a $k$-SDPP; from
    left to right, $k=2,3,4$.  Circle size corresponds to city
    quality.}
  \figlabel{maps}
\end{figure}

We can now investigate the effects of random projections on this
model.  \figref{projdim} shows the $L_1$ variational distance between
the original model and the projected model (estimated by sampling), as
well as the memory required to sample a set of paths for a variety of
projection dimensions $d$.  As predicted by \thmref{dpp_proj}, only a
relatively small number of projection dimensions are needed to obtain
a close approximation to the original model.  Past $d\approx 25$, the
rate of improvement due to increased dimension falls off dramatically;
meanwhile, the required memory and running time start to become
significant.  \figref{projdim} suggests that aggressive use of random
projections, like those we employ in the following section, is not
only theoretically but also empirically justified.

\begin{figure}
  \centering
  \includegraphics[trim=2in 4in 2in 4in,clip=true,width=3.5in]{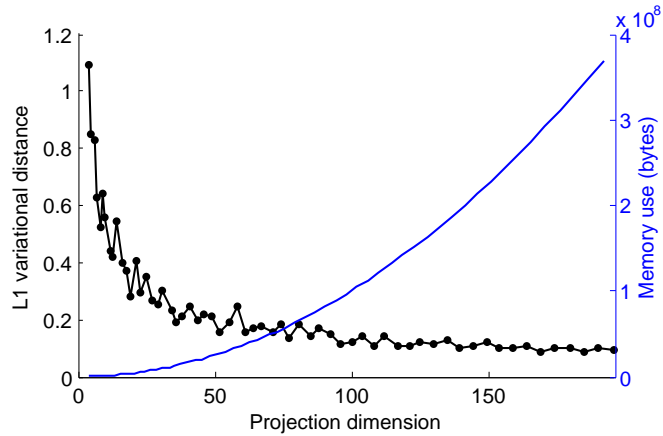}
  \caption[Random projection tests]
    {The effect of random projections.  In black, on the left,
    we estimate the $L_1$ variational distance between the original
    and projected models.  In blue, on the right, we plot the memory
    required for sampling, which is also proportional to running
    time.}  \figlabel{projdim}
\end{figure}

\subsection{Experiments: threading graphs}

In this section we put together many of the techniques introduced in
this paper in order to complete a novel task that we refer to as {\it
  graph threading} \citep{kulesza2012discovering}.  The goal is to extract from a large directed
graph a set of diverse, salient {\it threads}, or singly-connected
chains of nodes.  Depending on the construction of the graph, such
threads can have various semantics.  For example, given a corpus of
academic literature, high-quality threads in the citation graph might
correspond to chronological chains of important papers, each building
on the work of the last.  Thus, graph threading could be used to
identify a set of significant lines of research.  Or, given a
collection of news articles from a certain time period, where each
article is a node connected to previous, related articles, we might
want to display the most significant news stories from that period,
and for each story provide a thread that contains a timeline of its
major events.  We experiment on data from these two domains in the
following sections.  Other possibilities might include discovering
trends on social media sites, for example, where users can post image
or video responses to each other, or mining blog entries for important
conversations through trackback links.  \figref{threading_overview}
gives an overview of the graph threading task for document
collections.

Generally speaking, graph threading offers a means of gleaning
insights from collections of interrelated objects---for instance,
people, documents, images, events, locations, and so on---that are too
large and noisy for manual examination.  In contrast to tools like
search, which require the user to specify a query based on prior
knowledge, a set of threads provides an immediate, concise, high-level
summary of the collection, not just identifying a set of important
objects but also conveying the relationships between them.  As the
availability of such datasets continues to grow, this kind of
automated analysis will be key in helping us to efficiently and
effectively navigate and understand the information they contain.

\begin{figure}
  \centering
  \includegraphics[width=5.5in]{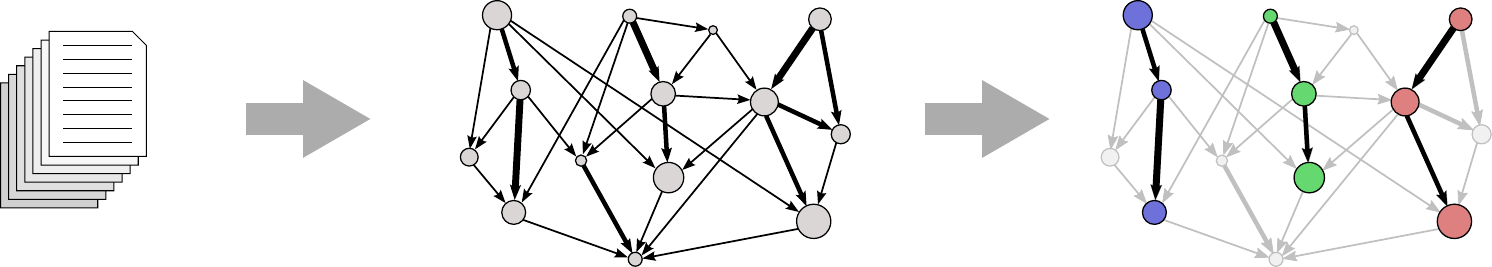}
  \caption[Graph threading]
    {An illustration of graph threading applied to a document
    collection.  We first build a graph from the collection, using
    measures of importance and relatedness to weight nodes (documents)
    and build edges (relationships).  Then, from this graph, we
    extract a diverse, salient set of threads to represent the
    collection.}  
  \figlabel{threading_overview}
\end{figure}

\subsubsection{Related work}

Research from to the Topic Detection and Tracking (TDT) program
\citep{wayne2000multilingual} has led to useful methods for tasks like
link detection, topic detection, and topic tracking that can be seen
as subroutines for graph threading on text collections.  Graph
threading with $k$-SDPPs, however, addresses these tasks jointly,
using a global probabilistic model with a tractable inference
algorithm.

Other work in the topic tracking literature has addressed related
tasks \citep{mei2005discovering,blei2006dynamic,leskovec2009meme}.  In
particular, \citet{blei2006dynamic} proposed dynamic topic models
(DTMs), which, given a division of text documents into time slices,
attempt to fit a generative model where topics evolve over time, and
documents are drawn from the topics available at the time slice during
which they were published.  The evolving topics found by a DTM can be
seen as threads of a sort, but in contrast to graph threading they are
not composed of actual items in the dataset (in this case, documents).
In \secref{newsthreading} we will return to this distinction when we
compare $k$-SDPP threading to a DTM baseline.

The information retrieval community has produced other methods for
extracting temporal information from document collections.
\citet{swan2000timemines} proposed a system for finding temporally
clustered named entities in news text and presenting them on a
timeline.  \citet{allan2001temporal} introduced the task of
\textit{temporal summarization}, which takes as input a stream of news
articles related to a particular topic, and then seeks to extract
sentences describing important events as they occur.
\citet{yan2011evolutionary} evaluated methods for choosing sentences
from temporally clustered documents that are relevant to a query.  In
contrast, graph threading seeks not to extract grouped entities or
sentences, but instead to organize a subset of the objects (documents)
themselves into threads, with topic identification as a side effect.

Some prior work has also focused more directly on threading.
\citet{shahaf2010connecting} and \citet{chieu2004query} proposed
methods for selecting individual threads, while
\citet{shahaf2012trains} recently proposed \textit{metro maps} as
alternative structured representations of related news stories.  Metro
maps are effectively sets of non-chronological threads that are
encouraged to intersect and, in doing so, generate a map of events and
topics.  However, these approaches assume some prior knowledge about
content.  \citet{shahaf2010connecting}, for example, assume the thread
endpoints are specified, and \citet{chieu2004query} require a set of
query words.  Likewise, because they build metro maps individually,
\citet{shahaf2012trains} implicitly assume that the collection is
filtered to a single topic, perhaps from a user query.  These inputs
make it possible to quickly pare down the document graph.  In
contrast, we will apply graph threading to very large graphs, and
consider all possible threads.

\subsubsection{Setup}

In order to be as useful as possible, the threads we extract from a
data graph need to be both high quality, reflecting the most important
parts of the collection, and diverse, so that they cover distinct
aspects of the data.  In addition, we would like to be able to
directly control both the length and the number of threads that we
return, since different contexts might necessitate different settings.
Finally, to be practical our method must be efficient in both time and
memory use.  $k$-SDPPs with random projections allow us to
simultaneously achieve all of these goals.

Given a directed graph on $M$ vertices with edge set $E$ and a
real-valued weight function $w(\cdot)$ on nodes and edges, define the
weight of a thread $\y = (y_1,y_2,\dots,y_R),$ $(y_r,y_{r+1}) \in E$
by
\begin{equation}
  w(\y) = \sum_{r=1}^{R} w(y_{r}) + \sum_{r=2}^{R}
  w(y_{r-1},y_{r})\,.
  \eqlabel{pathweight}
\end{equation}
We can use $w$ to define a simple log-linear quality model for our
$k$-SDPP:
\begin{align}
  q(\y) &= \exp(\beta w(\y))\\ 
  &= \left(\prod_{r=1}^R \exp(w(y_r))
  \prod_{r=2}^R \exp(w(y_{r-1},y_r)) \right)^\beta\,,
\end{align}
where $\beta$ is a hyperparameter controlling the dynamic range of the
quality scores.  We fix the value of $\beta$ on a validation set in
our experiments.

Likewise, let $\phi$ be a feature function from nodes in the graph to
$\reals^D$; then the diversity feature function on threads is
\begin{equation}
  \phi(\y) = \sum_{r=1}^R \phi(y_r)\,.
  \eqlabel{pathsim}
\end{equation}
In some cases it might also be convenient to have diversity features
on edges of the graph as well as nodes.  If so, they can be
accommodated without much difficulty; however, for simplicity we
proceed with the setup above.

We assume that $R$, $k$, and the projection dimension $d$ are
provided; the first two depend on application context, and the third,
as discussed in \secref{sdppproj}, is a tradeoff between
computational efficiency and faithfulness to the original model.  To
generate diverse thread samples, we first project the diversity
features $\phi$ by a random $d \times D$ matrix $G$ whose entries are
drawn independently and identically from $\N(0,\frac{1}{d})$.  We then
apply second-order message passing to compute the dual representation
$C$, as in \secref{sdpp_C}.  After eigendecomposing $C$,
which is only $d \times d$ due to the projection, we can run the first
phase of the $k$-DPP sampling algorithm from \secref{kdpp_sampling} to
choose a set $\cV$ of eigenvectors, and finally complete the SDPP
sampling algorithm in \secref{sdpp_sampling} to obtain a set of $k$
threads $Y$.  We now apply this model to two datasets; one is a
citation graph of computer science papers, and the other is a large
corpus of news text.

\subsubsection{Academic citation data}
\seclabel{cora}

The Cora dataset comprises a large collection of approximately 200,000
academic papers on computer science topics, including citation
information \citep{mccallum2000irj}.  We construct a directed graph
with papers as nodes and citations as edges, and then remove papers
with missing metadata or zero outgoing citations, leaving us with
28,155 papers.  The average out-degree is $3.26$ citations per paper,
and $0.011$\% of the total possible edges are present in the graph.

To obtain useful threads, we set edge weights to reflect the degree of
textual similarity between the citing and the cited paper, and node
weights to correspond with a measure of paper ``importance''.
Specifically, the weight of edge $(a,b)$ is given by the cosine
similarity metric, which for two documents $a$ and $b$ is the dot
product of their normalized tf-idf vectors, as defined in
\secref{documentsummarization}:
\begin{equation}
  \cossim(a,b) = \frac{\sum_{\word \in \allwords} \tf_a(\word) 
    \tf_b(\word) idf^2(\word)}
         {\sqrt{\sum_{\word \in \allwords} \tf^2_a(\word) idf^2(\word)}
           \sqrt{\sum_{\word \in \allwords} \tf^2_b(\word) idf^2(\word)}}\,,
\end{equation}
Here $\allwords$ is a subset of the words found in the documents.  We
select $\allwords$ by filtering according to document frequency; that
is, we remove words that are too common, appearing in more than 10\%
of papers, or too rare, appearing in only one paper.  After filtering,
there are 50,912 unique words.

The node weights are given by the LexRank score of each paper
\citep{erkan2004lexrank}.  The LexRank score is the stationary
distribution of the thresholded, binarized, row-normalized matrix of
cosine similarities, plus a damping term, which we fix to $0.15$.
LexRank is a measure of centrality, so papers that are closely related
to many other papers will receive a higher score.


Finally, we design the diversity feature function $\phi$ to encourage
topical diversity.  Here we apply cosine similarity again,
representing a document by the 1,000 documents to which it is most
similar.  This results in binary $\phi$ of dimension $D = M = 28,155$
with exactly 1,000 non-zeros; $\phi_l(y_r) = 1$ implies that $l$ is
one of the 1,000 most similar documents to $y_r$.  Correspondingly,
the dot product between the diversity features of two documents is
proportional to the fraction of top-1,000 documents they have in
common.  In order to make $k$-SDPP inference efficient, we project
$\phi$ down to $d = 50$ dimensions.

\figref{exthreads} illustrates the behavior of the model when we set
$k=4$ and $R=5$.  Samples from the model, like the one presented in
the figure, not only offer some immediate intuition about the types of
papers contained in the collection, but also, upon examining individual
threads, provide a succinct illustration of the content and
development of each area.  Furthermore, the sampled threads cover
distinct topics, standing apart visually in \figref{exthreads} and
exhibiting diverse salient terms.

\begin{figure}
  \centering
  \includegraphics[width=3.7in,trim=1.9in 3.75in 2in 3.25in,
    clip=true]{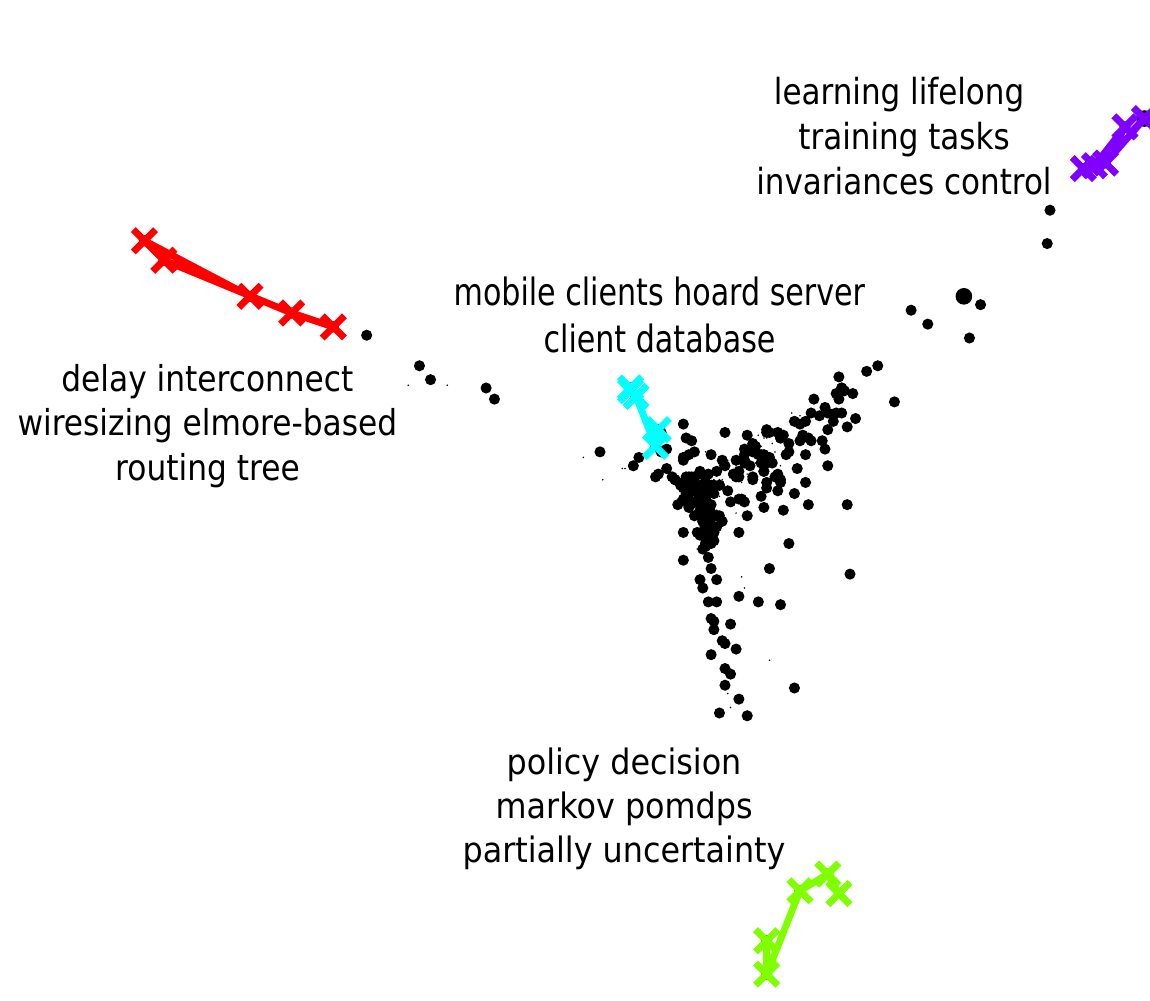}
  \\\vspace{0.4cm}
  \fbox{\parbox{5.75in}{
      \begin{center}
        Thread: {\bf learning lifelong training tasks invariances control}
      \end{center}
      \begin{enumerate}
        \setlength{\itemsep}{-4pt}
      \item Locally Weighted Learning for Control
      \item Discovering Structure in Multiple Learning Tasks: The TC
        Algorithm
      \item Learning One More Thing
      \item Explanation Based Learning for Mobile Robot Perception
      \item Learning Analytically and Inductively
      \end{enumerate}
  }}
  \\\vspace{0.4cm}
  \fbox{\parbox{5.75in}{
      \begin{center}
        Thread: {\bf mobile clients hoard server client database}
      \end{center}
      \begin{enumerate}
        \setlength{\itemsep}{-4pt}
      \item A Database Architecture for Handling Mobile Clients
      \item An Architecture for Mobile Databases
      \item Database Server Organization for Handling Mobile Clients
      \item Mobile Wireless Computing: Solutions and Challenges in
        Data Management
      \item Energy Efficient Query Optimization
      \end{enumerate}
  }}
  \caption[Cora threads]
    {Sampled threads from a $4$-SDPP with thread length $R = 5$
    on the Cora dataset.  Above, we plot a subset of the Cora papers,
    projecting their tf-idf vectors to two dimensions by running PCA
    on the centroids of the threads, and then highlight the thread
    selections in color.  Displayed beside each thread are the words
    in the thread with highest tf-idf score.  Below, we show the
    titles of the papers in two of the threads.}  \figlabel{exthreads}
\end{figure}

\subsubsection{News articles}
\seclabel{newsthreading}

Our news dataset comprises over 200,000 articles from the New York
Times, collected from 2005-2007 as part of the English Gigaword corpus
\citep{graff2009ldc}.  We split the articles into six groups, with six
months' worth of articles in each group.  Because the corpus contains
a significant amount of noise in the form of articles that are short
snippets, lists of numbers, and so on, we filter the results by
discarding articles more than two standard deviations longer than the
mean article, articles less than 400 words, and articles whose
fraction of non-alphabetic words is more than two standard deviations
above the mean.  On average, for each six-month period we are left
with 34,504 articles.

For each time period, we generate a graph with articles as nodes.  As
for the citations dataset, we use cosine similarity to define edge
weights.  The subset of words $\allwords$ used to compute cosine
similarity contains all words that appear in at least 20 articles and
at most 15\% of the articles.  Across the six time periods, this
results in an average of 36,356 unique words.  We include in our graph
only those edges with cosine similarity of at least 0.1; furthermore,
we require that edges go forward in time to enforce the chronological
ordering of threads.  The resulting graphs have an average of $0.32$\%
of the total possible edges, and an average degree of 107.  As before,
we use LexRank for node weights, and the top-1000 similar documents to
define a binary feature function $\phi$.  We add a constant feature
$\rho$ to $\phi$, which controls the overall degree of repulsion;
large values of $\rho$ make all documents more similar to one another.
We set $\rho$ and the quality model hyperparameters to maximize a
cosine similarity evaluation metric (see \secref{eval}), using the
data from the first half of 2005 as a development set.  Finally, we
randomly project the diversity features from $D \approx 34,500$ to
$d=50$ dimensions.  For all of the following experiments, we use
$k=10$ and $R=8$.  All evaluation metrics we report are averaged over
$100$ random samples from the model.

\paragraph{Graph visualizations}

In order to convey the scale and content of the graphs built from news
data, we provide some high-resolution renderings.  \figref{newsgraph}
shows the graph neighborhood of a single article node from our
development set.  Each node represents an article and is annotated
with the corresponding headline; the size of each node reflects its
weight, as does the thickness of an edge.  The horizontal position of
a node corresponds to the time at which the article was published,
from left to right; the vertical positions are optimized for
readability.  In the digital version of this paper,
\figref{newsgraph} can be zoomed in order to read the headlines; in
hardcopy, however, it is likely to be illegible.  As an alternative,
an online, zoomable version of the figure is available at
\url{http://zoom.it/GUCR}.

\begin{figure}
  \centering
  \includegraphics[width=5.5in,trim=0in 2.75in 0in 2.75in,clip=true]
                  {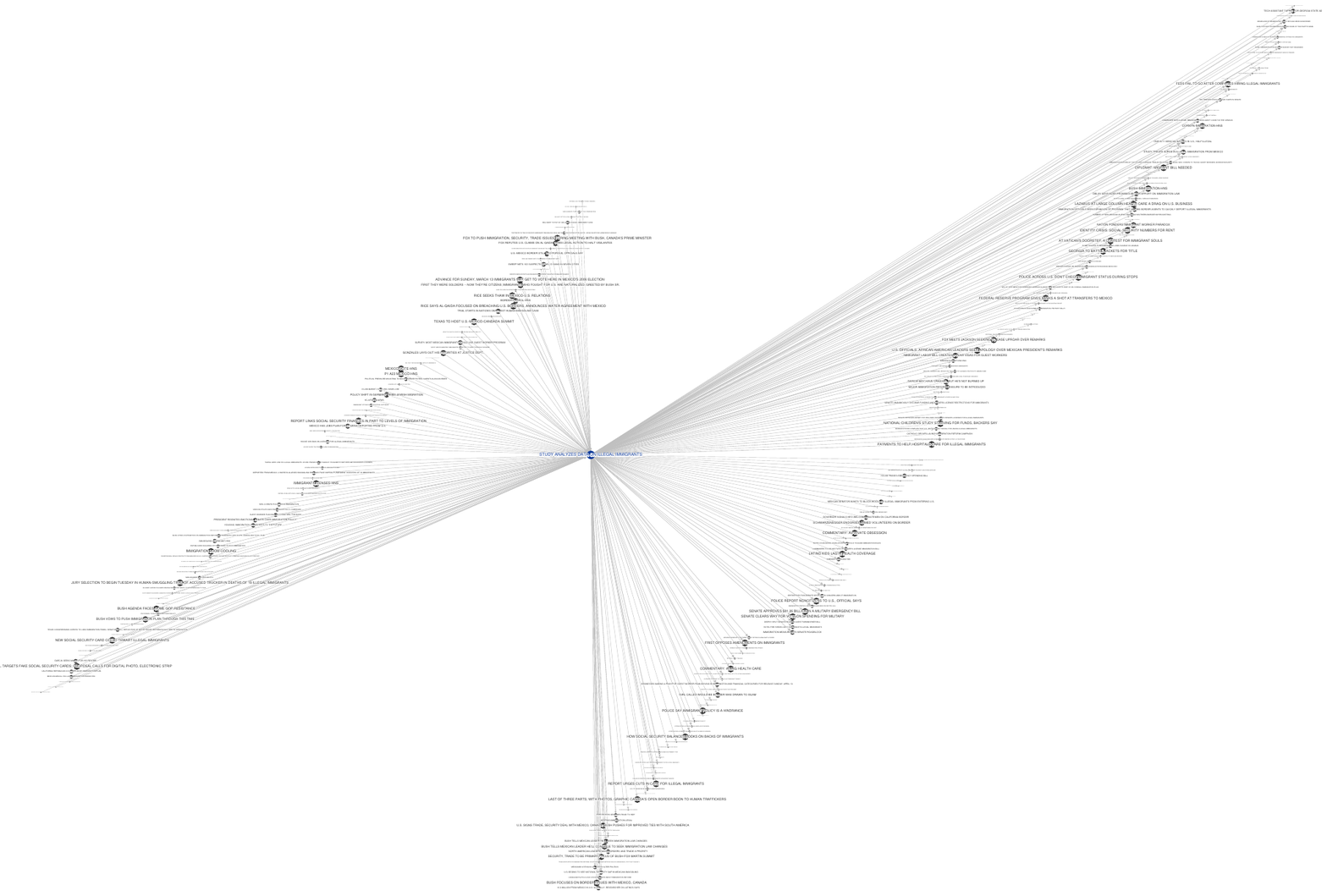}
  \caption[News graph visualization]
    {Visualization of a single article node and all of its
    neighboring article nodes.}  
  \figlabel{newsgraph}
\end{figure}

Visualizing the entire graph is quite challenging since it contains
tens of thousands of nodes and millions of edges; placing such a
figure in the paper would be impractical since the computational
demands of rendering it and the zooming depth required to explore it
would exceed the abilities of modern document viewers.  Instead, we
provide an online, zoomable version based upon a high-resolution (540
megapixel) rendering, available at \url{http://zoom.it/jOKV}.  Even at
this level of detail, only 1\% of the edges are displayed; otherwise
they become visually indistinct.  As in \figref{newsgraph}, each node
represents an article and is sized according to its weight and
overlaid with its headline.  The horizontal position corresponds to
time, ranging from January 2005 (on the left) to June 2005 (on the
right).  The vertical positions are determined by similarity to a set
of threads sampled from the $k$-SDPP, which are rendered in color.



\paragraph{Baselines}

We will compare the $k$-SDPP model to two natural baselines.

\textit{$k$-means baseline.} 
A simple method for this task is to split each six-month period of
articles into $R$ equal-sized time slices, and then apply $k$-means
clustering to each slice, using cosine similarity at the clustering
metric.  We can then select the most central article from each cluster
to form the basis of a set of threads.  The $k$ articles chosen from
time slice $r$ are matched one-to-one with those from slice $r-1$ by
computing the pairing that maximizes the average cosine similarity of
the pairs---that is, the coherence of the threads.  Repeating this
process for all $r$ yields a set of $k$ threads of length $R$, where
no two threads will contain the same article.  However, because
clustering is performed independently for each time slice, it is
likely that the threads will sometimes exhibit discontinuities when
the articles chosen at successive time slices do not naturally align.



\textit{DTM baseline.}
A natural extension, then, is the dynamic topic model (DTM) of
\citet{blei2006dynamic}, which explicitly attempts to find topics that
are smooth through time.  We use publicly available code\footnote{\tt
  http://code.google.com/p/princeton-statistical-learning/} to fit
DTMs with the number of topics set to $k$ and with the data split into
$R$ equal time slices.  We set the hyperparameters to maximize the
cosine similarity metric (see \secref{eval}) on our development set.
We then choose, for each topic at each time step, the document with
the highest per-word probability of being generated by that topic.
Documents from the same topic form a single thread.

\paragraph{}
\figref{dppthreads} shows some of the threads sampled randomly from
the $k$-SDPP for our development set, and \figref{dtmthreads} shows
the same for threads produced by the DTM baseline.  An obvious
distinction is that topic model threads always span nearly the entire
time period, selecting one article per time slice as required by the
form of the model, while the DPP can select threads covering only the
relevant span.  Furthermore, the headlines in the figures suggest that
the $k$-SDPP produces more tightly focused, narrative threads due to
its use of the data graph, while the DTM threads, though topically
related, tend not to describe a single continuous news story.  This
distinction, which results from the fact that topic models are not
designed with threading in mind, and so do not take advantage of the
explicit relation information given by the graph, means that $k$-SDPP
threads often form a significantly more coherent representation of the
news collection.

\begin{figure}
  \centering
  \includegraphics[trim=1.15in 4in 0.75in 4.2in,clip=true,
    width=5in]{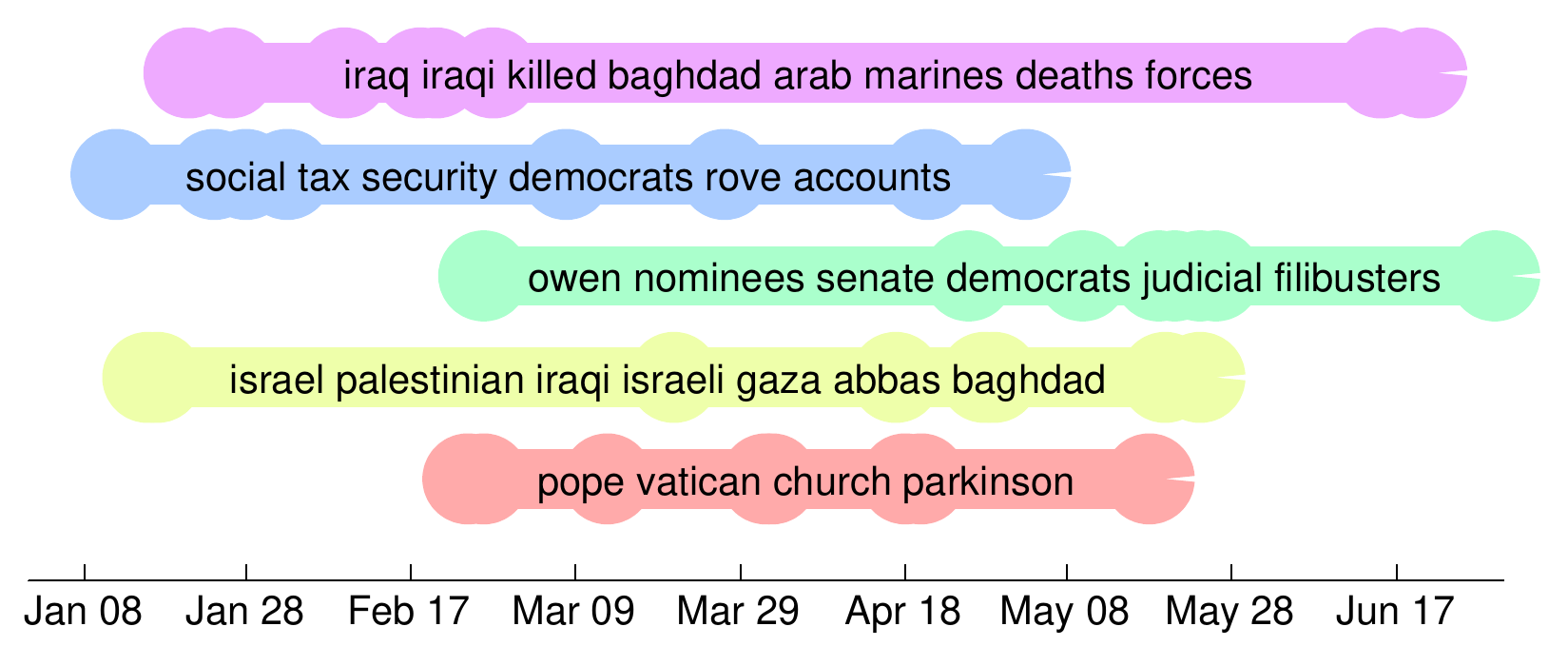}
  \\\vspace{0.5cm}
  \fbox{\parbox{5.75in}{
      \begin{center}
        Thread: {\bf pope vatican church parkinson}
      \end{center}
      \centering
      \begin{tabular}{@{ }r@{: }p{4.5in}}
        Feb 24 & Parkinson's Disease Increases Risks to Pope \\
        Feb 26 & Pope's Health Raises Questions About His Ability to Lead \\
        Mar 13 & Pope Returns Home After 18 Days at Hospital \\
        Apr 01 & Pope's Condition Worsens as World Prepares for End of Papacy \\
        Apr 02 & Pope, Though Gravely Ill, Utters Thanks for Prayers \\
        Apr 18 & Europeans Fast Falling Away from Church \\
        Apr 20 & In Developing World, Choice [of Pope] Met with Skepticism \\
        May 18 & Pope Sends Message with Choice of Name \\
      \end{tabular}
      \\\vspace{0.15in}
  }}
  \caption[$k$-SDPP news threads]
    {A set of five news threads randomly sampled from a $k$-SDPP
    for the first half of 2005.  Above, the threads are shown on a
    timeline with the most salient words superimposed; below, the
    dates and headlines from a single thread are listed.}
  \figlabel{dppthreads}
\end{figure}

\begin{figure}
  \centering
  \includegraphics[trim=1.15in 4in 0.85in 4.2in,clip=true,
    width=5in]{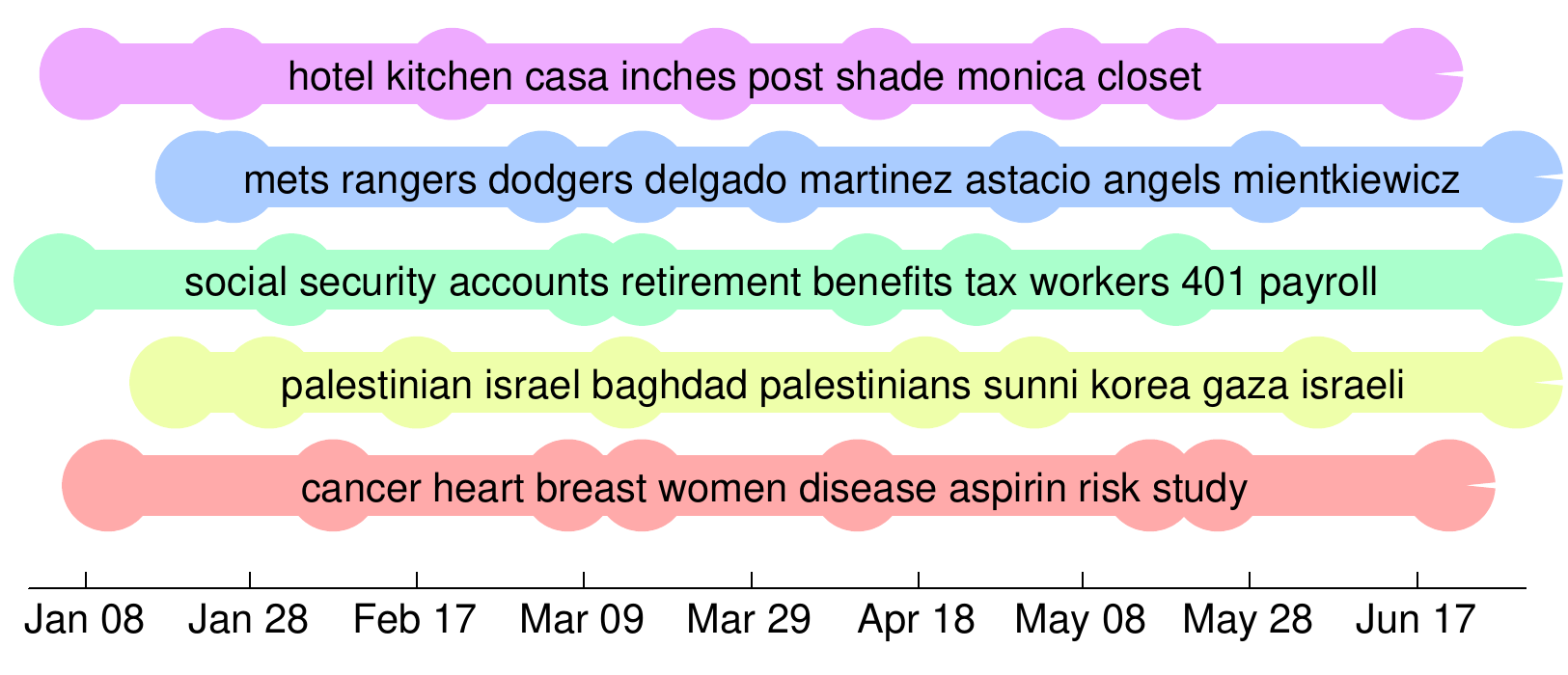}
  \\\vspace{0.5cm}
  \fbox{\parbox{5.75in}{
      \begin{center}
        Thread: {\bf cancer heart breast women disease aspirin risk study}
      \end{center}
      \centering
      \begin{tabular}{@{ }r@{: }p{4.5in}}
        Jan 11 & Study Backs Meat, Colon Tumor Link \\
        Feb 07 & Patients---and Many Doctors---Still Don't Know How Often Women Get Heart Disease \\
        Mar 07 & Aspirin Therapy Benefits Women, but Not in the Way It Aids Men \\
        Mar 16 & Study Shows Radiation Therapy Doesn't Increase Heart Disease Risk for Breast Cancer Patients \\
        Apr 11 & Personal Health: Women Struggle for Parity of the Heart \\
        May 16 & Black Women More Likely to Die from Breast Cancer \\
        May 24 & Studies Bolster Diet, Exercise for Breast Cancer Patients \\
        Jun 21 & Another Reason Fish is Good for You \\
      \end{tabular}
      \\\vspace{0.15in}
  }}
  \caption[DTM news threads]
    {A set of five news threads generated by the dynamic topic
    model for the first half of 2005.  Above, the threads are shown on
    a timeline with the most salient words superimposed; below, the
    dates and headlines from a single thread are
    listed.}  
  \figlabel{dtmthreads}
\end{figure}

\paragraph{Comparison to human summaries}
\seclabel{eval}

We provide a quantitative evaluation of the threads generated by our
baselines and sampled from the $k$-SDPP by comparing them to a set of
human-generated news summaries.  The human summaries are not threaded;
they are flat, approximately daily news summaries found in the Agence
France-Presse portion of the Gigaword corpus, distinguished by their
``multi'' type tag.  The summaries generally cover world news, which
is only a subset of the contents of our dataset.  Nonetheless, they
allow us to provide an extrinsic evaluation for this novel task
without generating gold standard timelines manually, which is a
difficult task given the size of the corpus.  We compute four metrics:
\begin{itemize}
\item {\bf Cosine similarity.} We concatenate the human summaries over
  each six-month period to obtain a target tf-idf vector, concatenate
  the set of threads to be evaluated to obtain a predicted tf-idf
  vector, and then compute the cosine similarity (in percent) between
  the target and predicted vectors.  All hyperparameters are chosen to
  optimize this metric on a validation set.
\item {\bf ROUGE-1, 2, and SU4.} As described in
  \secref{documentsummarization}, ROUGE is an automatic evaluation
  metric for text summarization based on $n$-gram overlap statistics
  \citep{lin2004rouge}.  We report three standard variants.
\end{itemize}

\tabref{news_results} shows the results of these comparisons, averaged
over all six half-year intervals.  Under each metric, the $k$-SDPP produces
threads that more closely resemble human summaries.

\begin{table}
  \centering
  \begin{tabular}{cccc@{/}ccc@{/}ccc@{/}c}
    &
    & 
    \multicolumn{3}{c}{\bf\scriptsize ROUGE-1} & 
    \multicolumn{3}{c}{\bf\scriptsize ROUGE-2} & 
    \multicolumn{3}{c}{\bf\scriptsize ROUGE-SU4} \\
    System
    &
    Cos-sim
    & 
    F & Prec & Rec &
    F & Prec & Rec &
    F & Prec & Rec \\
    \hline
    
    $k$-means &
    29.9 &
    16.5 &  17.3 & 15.8 &
    0.695 & 0.73 & 0.67 &
    3.76 & 3.94 & 3.60 \\
    
    DTM &
    27.0 &
    14.7 & 15.5 & 14.0 &
    0.750 & 0.81 & 0.70 &
    3.44 & 3.63 & 3.28 \\
    
    $k$-SDPP &
    \bf 33.2 &
    \bf 17.2 & \bf 17.7 & \bf 16.7 &
    \bf 0.892 & \bf 0.92 & \bf 0.87 &
    \bf 3.98 & \bf 4.11 & \bf 3.87 \\
    \hline
  \end{tabular}
  \caption[News timeline results]
    {Similarity of automatically generated timelines to human
    summaries.  Bold entries are significantly higher than others in
    the column at 99\% confidence, verified using bootstrapping.}
   \tablabel{news_results}
\end{table}

\paragraph{Mechanical Turk evaluation}

\begin{table}
  \centering
  \begin{tabular}{ccc}
    System & Rating & Interlopers \\
    \hline
    $k$-means & 2.73 & 0.71 \\ 
    DTM & 3.19 & 1.10 \\
    $k$-SDPP & {\bf 3.31} & 1.15 \\   
    \hline
  \end{tabular}
  \caption[News timeline Turk results]
    {Rating: average coherence score from 1 (worst) to 5 (best).
    Interlopers: average number of interloper articles identified (out
    of 2).  Bold entries are significantly higher with 95\% confidence.}
  \tablabel{turk}
\end{table}

An important distinction between the baselines and the $k$-SDPP is
that the former are \textit{topic}-oriented, choosing articles that
relate to broad subject areas, while the $k$-SDPP approach is
\textit{story}-oriented, chaining together articles with direct
individual relationships.  An example of this distinction can be seen
in \figsref{dppthreads}{dtmthreads}.

To obtain a large-scale evaluation of this type of thread coherence,
we employ Mechanical Turk, on online marketplace for inexpensively and
efficiently completing tasks requiring human judgment.  We asked
Turkers to read the headlines and first few sentences of each article
in a timeline and then rate the overall narrative coherence of the
timeline on a scale of 1 (``the articles are totally unrelated'') to 5
(``the articles tell a single clear story'').  Five separate Turkers
rated each timeline.  The average ratings are shown in the left column
of \tabref{turk}; the $k$-SDPP timelines are rated as significantly
more coherent, while $k$-means does poorly since it has no way to
ensure that clusters are similar between time slices.

In addition, we asked Turkers to evaluate threads implicitly by
performing a second task.  (This also had the side benefit of ensuring
that Turkers were engaged in the rating task and did not enter random
decisions.)  We displayed timelines into which two additional
``interloper'' articles selected at random had been inserted, and
asked users to remove the two articles that they thought should be
removed to improve the flow of the timeline.  A screenshot of the task
is provided in \figref{turktask}.  Intuitively, the true interlopers
should be selected more often when the original timeline is coherent.
The average number of interloper articles correctly identified is
shown in the right column of \tabref{turk}.

\begin{figure}
  \centering
  \includegraphics[width=5.5in]{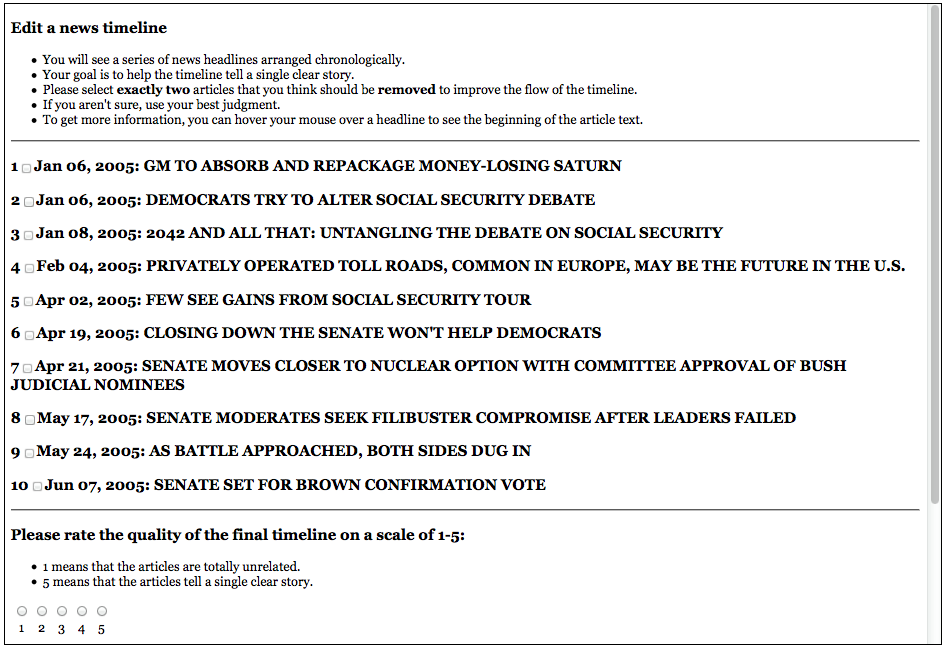}
  \caption[Mechanical Turk task]
    {A screenshot of the Mechanical Turk task presented to
    annotators.}  
  \figlabel{turktask}
\end{figure}

\paragraph{Runtime}

Finally, assuming that tf-idf and feature values have been computed in
advance (this process requires approximately 160 seconds), we report
in \tabref{runtime} the time required to produce a set of threads on
the development set.  This measurement includes clustering for the
$k$-means baseline, model fitting for the DTM baseline, and random
projections, computation of the covariance matrix, and sampling for
the $k$-SDPP.  The tests were run on a machine with eight Intel Xeon
E5450 cores and 32G of memory.  Thanks to the use of random
projections, the $k$-SDPP is not only the most faithful to human news
summaries, but also the fastest by a large margin.

\begin{table}
\centering
\begin{tabular}{lr@{.}l}
System & \multicolumn{2}{c}{Runtime (s)} \\\hline
$k$-means & 625 & 63 \\
DTM & 19,433 & 80 \\
$k$-SDPP & \textbf{252} & \textbf{38} \\
\hline
\end{tabular}
\caption[News threads running time]
    {Running time for the tested methods.}
\tablabel{runtime}
\end{table}

\section{Conclusion}
\seclabel{conclusion}

We believe that DPPs offer exciting new possibilities for a wide range
of practical applications.  Unlike heuristic diversification
techniques, DPPs provide coherent probabilistic semantics, and yet
they do not suffer from the computational issues that plague existing
models when negative correlations arise.  Before concluding, we
briefly mention two open technical questions, as well as some
possible directions for future research.

\subsection{Open question: concavity of entropy}

The Shannon entropy of the DPP with marginal kernel $K$ is given by
\begin{equation}
  H(K) = - \sum_{Y \subseteq \Y} \P(Y) \log \P(Y)\,.
\end{equation}
\begin{conjecture}[\citet{lyons2003determinantal}]
  $H(K)$ is concave in $K$.
\end{conjecture}
While numerical simulation strongly suggests that the conjecture is
true, to our knowledge no proof currently exists.

\subsection{Open question: higher-order sums}

In order to calculate, for example, the Hellinger distance between a
pair of DPPs, it would be useful to be able to compute quantities of
the form
\begin{equation}
  \sum_{Y\subseteq\Y} \det(L_Y)^p
\end{equation}
for $p > 1$.  To our knowledge it is not currently known whether it is
possible to compute these quantities efficiently.

\subsection{Research directions}

A variety of interesting machine learning questions remain for future
research.

\begin{itemize}
  \item Would DPPs based on Hermitian or asymmetric kernels offer
    worthwhile modeling advantages?
  \item Is there a simple characterization of the conditional
    independence relations encoded by a DPP?
  \item Can we perform DPP inference under more complicated
    constraints on allowable sets?  (For instance, if the items
    correspond to edges in a graph, we might only consider sets that
    comprise a valid matching.)
  \item How can we learn the similarity kernel for a DPP (in addition
    to the quality model) from labeled training data?
  \item How can we efficiently (perhaps approximately) work with SDPPs
    over loopy factor graphs?
  \item Can SDPPs be used to diversify $n$-best lists and improve
    reranking performance, for instance in parsing or machine
    translation?
\end{itemize}


\bibliographystyle{plainnat}
\bibliography{paper}

\end{document}